\documentclass{article}

\usepackage{microtype}
\usepackage{graphicx}
\usepackage{booktabs}
\usepackage[table]{xcolor}

\definecolor{findingsbg}{HTML}{EDDDAB}
\colorlet{findingslabelbg}{findingsbg!85!black}
\colorlet{findingsbar}{findingsbg!45!black}
\newcommand{\findingbox}[1]{%
  \par\addvspace{5pt}%
  \noindent\begingroup
  \setlength{\fboxsep}{5pt}%
  \hbox to \linewidth{%
    {\color{findingsbar}\vrule width 2.5pt}%
    \colorbox{findingsbg}{%
      \parbox{\dimexpr\linewidth-2.5pt-2\fboxsep\relax}{%
        \normalsize
        {\setlength{\fboxsep}{1pt}%
         \colorbox{findingslabelbg}{%
           \textcolor{findingsbar}{\bfseries\MakeUppercase{Findings}}}}%
        \enspace #1%
      }%
    }%
    \hss%
  }%
  \endgroup\par\addvspace{5pt}%
}
\usepackage{hyperref}

\usepackage{triattention}

\usepackage{amsmath}
\usepackage{amssymb}
\usepackage{amsfonts}
\usepackage{multirow}
\usepackage{adjustbox}
\usepackage{xspace}
\usepackage{enumitem}
\usepackage{fancyvrb}
\usepackage{wrapfig}
\usepackage{capt-of}
\usepackage{multicol}
\usepackage{url}
\usepackage{tikz}
\definecolor{abovegreen}{RGB}{118,176,83}
\definecolor{belowblue}{RGB}{68,141,192}
\definecolor{heatneutral}{gray}{0.94}
\newcommand{\ds}[1]{%
  \tikz[baseline=(X.base)]
  \node[draw,circle,inner sep=0.6pt,minimum size=1em]
  (X) {\scriptsize #1};%
}

\newcommand{\modelname}{\textit{TriAttention}\xspace}

\preprinttitlerunning{\modelname: Unified Sparse Mixed-Precision Attention}

\begin{document}

\twocolumn[
\preprinttitle{\modelname: Unified Sparse Mixed-Precision Attention for Long-Context LLM Inference}

\begin{preprintauthorlist}
\preprintauthor{Hanzhi Zhang}{1}
\preprintauthor{Qiao Zhang}{2}
\preprintauthor{Qinglei Cao}{2}
\preprintauthor{Heng Fan}{1}
\preprintauthor{Yan Huang}{1}
\preprintauthor{Kewei Sha}{3}
\preprintauthor{Yunhe Feng}{1}
\end{preprintauthorlist}

\hypersetup{pdftitle={TriAttention: Unified Sparse Mixed-Precision Attention for Long-Context LLM Inference},pdfauthor={Hanzhi Zhang, Qiao Zhang, Qinglei Cao, Heng Fan, Yan Huang, Kewei Sha, Yunhe Feng},pdfsubject={},pdfkeywords={Efficient inference, sparse attention, quantization}}

\vskip 0.3in

\begin{abstract}
Sparsity and quantization reduce the cost of attention by skipping interactions and lowering the precision of those retained.
However, useful interactions differ in their precision requirements: some tolerate low-precision scores while remaining sensitive to removal.
We therefore introduce \modelname, which assigns query-key (QK) tiles to skipping, 4-bit, or 8-bit computation, using 4-bit to retain quantization-tolerant interactions at lower cost.
To represent compactly, we encode each head's computation regions with two diagonal-distance boundaries, keeping policy storage independent of context length.
Offline calibration then evaluates candidate boundary pairs using normalized projected-output errors and complete-tile execution costs.
Using these measurements, a model-wide allocator selects each head's boundaries to minimize calibration error under a shared computation budget.
During prefill, the selected boundaries generate precision-grouped traversals for a fused kernel that skips omitted tiles and combines contributions from both QK precisions into one normalized attention output.
The quantized keys and values prepared during prefill are then reused in a compressed key-value (KV) cache for decoding, reducing repeated reads of prompt data.
Experiments show that \modelname achieves end-to-end generation speedups of up to $7.88\times$ over the Torch attention and $2.65\times$ over the BF16 baseline.
Meanwhile, \modelname maintains quality comparable to the original models on general-purpose and long-context tasks.
These results show how a compact retention-and-precision policy translates local sensitivities into attention computation and repeated KV-cache access.
The implementation is \url{https://github.com/HanzhiZhang-Ulrica/TriAttention}.
\end{abstract}

]

\preprintaffiliations{%
\textsuperscript{1}LLaVi Lab, Computer Science \& Engineering, University of North Texas.
\textsuperscript{2}Computer Science, Saint Louis University.
\textsuperscript{3}Data Science, University of North Texas.\par
\texttt{\{hanzhi.zhang,\allowbreak heng.fan,\allowbreak yan.huang,\allowbreak kewei.sha,\allowbreak yunhe.feng\}@unt.edu}\\
\texttt{\{qiao.zhang,\allowbreak qinglei.cao\}@slu.edu}.}

\section{Introduction}
\label{sec:introduction}

\begin{figure*}[t]
    \centering
    \includegraphics[width=\textwidth]{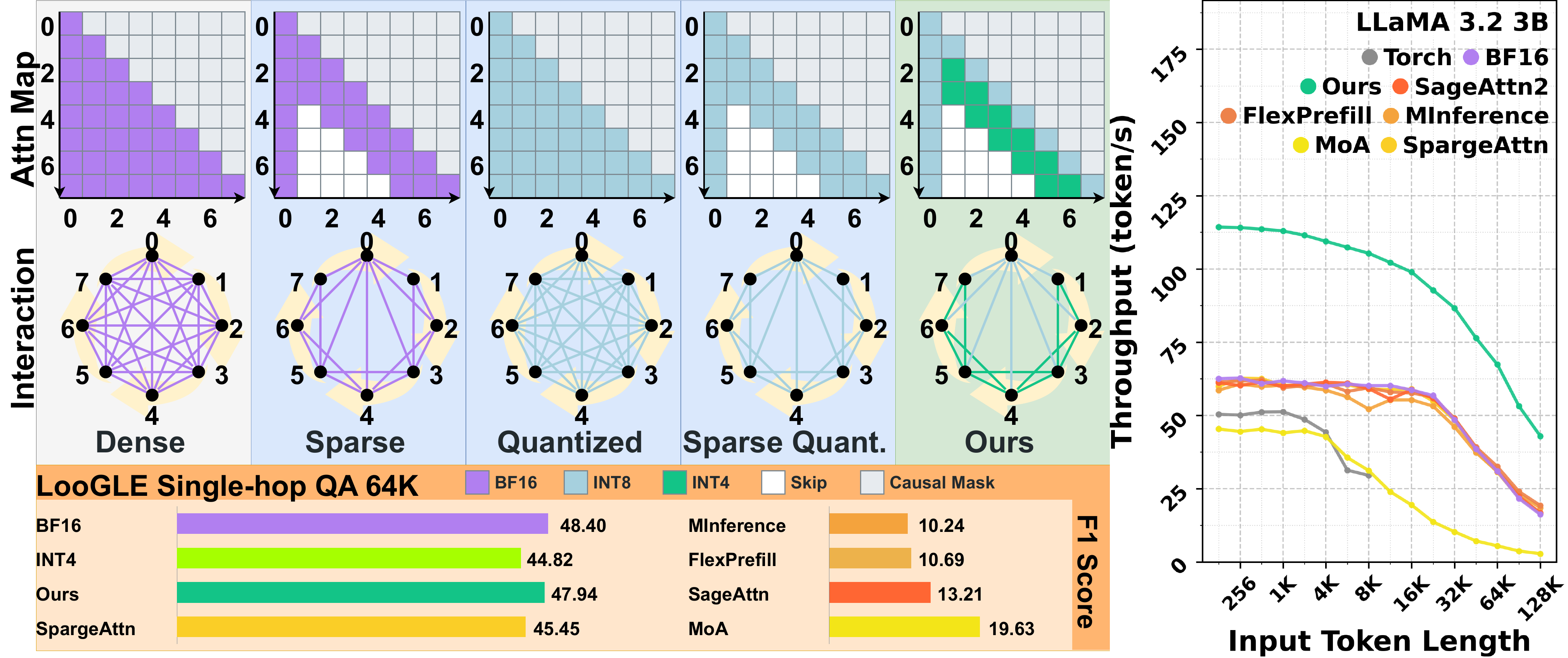}
    \vspace{-20pt}
    \caption{Attention computation choices and inference performance. Top left: attention maps and token-interaction graphs compare dense, sparse, quantized, and sparse-quantized designs with \modelname{}, which jointly assigns skipping, 4-bit, and 8-bit. Bottom left: LooGLE single-hop QA F1 at 64K. Right: throughput across input lengths. Both evaluations use Llama-3.2-3B-Instruct.}
    \label{fig:intro}
    \vspace{-12pt}
\end{figure*}

Efficient deployment of large language models (LLMs) requires reducing inference latency while preserving model quality across conversational assistance, code generation, reasoning, and document summarization \citep{grattafiori2024llama,bai2024longbench}.
For attention, this involves both matrix computation during prompt processing (prefill) and repeated key-value (KV) cache reads during token generation (decode) \citep{dao2022flashattention,yang2025lserve}.
Sparsity reduces attention computation by skipping selected interactions, while quantization accelerates query-key (QK) multiplication through lower-precision arithmetic \citep{jiang2024minference,zhang2025sageattention}.
However, attention interactions differ in their sensitivity to skipping and reduced precision: some can be removed with little effect on the output, while other useful contributions tolerate lower QK precision but remain sensitive to removal.
Model quality therefore depends on both which interactions are retained and the precision used to compute their contributions.

Figure~\ref{fig:intro} contrasts efficient attention strategies and illustrates their long-context QA and inference-throughput differences.
Dense attention kernels reduce memory transfers and improve GPU utilization, yet still evaluate a quadratic number of causal interactions \citep{dao2022flashattention,dao2024flashattention}.
To reduce the interaction count, structured sparse methods select attention patterns or spans, but retained regions still incur floating-point computation. Therefore, further savings from sparsity require excluding more context, and discarding useful interactions degrades model quality \citep{lai2025flexprefill,jiang2024minference,fu2024moa,zhu2025sampleattention}.
Quantized attention increases QK multiplication throughput with lower-bit arithmetic. However, using a uniformly higher precision protects sensitive regions but spends unnecessary computation on regions that tolerate lower precision, while reducing precision to suit tolerant regions introduces excessive error in sensitive ones and degrades model quality \citep{zhang2025sageattention,zhang2024sageattention2}.
Sparse quantized attention combines skipping with low-bit QK computation, yet retained regions still share one precision even though some require higher score accuracy and others tolerate further quantization \citep{zhang2025spargeattention}.
This leads to our central question: \textbf{how can attention jointly allocate retention and precision for efficient GPU execution?}

Our key insight is that \textit{attention regions differ in their sensitivity to skipping and reduced score precision}.
Exploiting these differences lets us preserve useful context at lower cost while reserving higher precision for sensitive regions.
However, translating this flexibility into efficient inference requires resolving two coupled challenges.
For allocation, comparable reductions in computation produce different output errors across layers and heads, so a compression level tolerated by some regions introduces excessive errors in less tolerant regions and degrades model quality.
For execution, an explicit tile-action map grows quadratically with context length. Query-block traversal metadata also competes with staged K/V operands for shared memory.

To address these challenges, we introduce \modelname, a sparse mixed-precision attention method that jointly assigns skipping, 4-bit, and 8-bit QK computation within each head.
To control policy-metadata growth, we define each layer--head configuration using two diagonal-distance boundaries that describe entire computation regions with storage independent of context length.
To compare sensitivity across layers and heads, offline calibration measures each candidate boundary pair's error after output projection and normalizes it by the layer's reference attention-output energy.
We pair these errors with complete-tile costs, so allocation accounts for the full computation performed on retained tiles.
Using these comparisons, a global allocator selects a boundary pair under a shared budget, preserving more computation in sensitive regions and concentrating savings in more tolerant ones.
During prefill, the selected boundaries generate only the current query block's retained tile indices, keeping traversal metadata local to the work staged in shared memory.
The fused kernel processes these indices in precision groups to reuse query operands, skips omitted tiles, and combines retained contributions into one normalized attention output.
At prompt completion, the selected ranges determine which prepared quantized keys and values are packed into the decode cache, extending the allocation's savings from prefill to repeated prompt-data reads.

Across the evaluated workloads, \modelname{} achieves end-to-end generation speedups of up to $7.88\times$ over models using Torch attention and $2.65\times$ over the BF16 baseline.
These efficiency gains are achieved while maintaining quality comparable to the original models on general-purpose tasks and long-context question answering.

\noindent Our contributions are:
\begin{enumerate}[nosep,left=0pt]
    \item \textbf{Three-Level QK Computation.}
    We develop a tile-level organization of skipping, 4-bit, and 8-bit QK computation that matches graded sensitivity with graded computational cost.
    The shared tile unit makes retention and precision jointly selectable within each head.

    \item \textbf{Compact Distance Encoding.}
    We represent each head's computation regions with two diagonal-distance boundaries, providing a common representation for offline calibration and local, precision-grouped GPU traversal while keeping policy storage independent of context length.

    \item \textbf{Sensitivity-Guided Global Allocation.}
    We develop an offline allocation rule that compares normalized output errors and execution costs of alternative three-level assignments.
    A model-wide budget coordinates these choices across layers and heads, prioritizing computation where it most reduces projected-output error.

    \item \textbf{Cross-Phase Execution.}
    Fused prefill executes the selected computation regions, while decoding reuses the prepared quantized keys and values in a compressed KV cache to reduce repeated prompt-data reads.
\end{enumerate}

\section{Related Work}
\label{sec:related_work}

Efficient attention connects GPU execution, interaction selection, and numerical precision.

\textit{Hardware-efficient attention execution} improves data reuse, work partitioning, and kernel specialization. Tiled fusion keeps score computation, online softmax, and value aggregation on chip, reducing intermediate transfers while dense query-key computation remains quadratic \citep{dao2022flashattention,dao2024flashattention,ye2025flashinfer}. Workload-aware partitioning of query blocks or KV segments across thread blocks improves occupancy and load balance, although KV splitting adds coordination and partial-output reduction \citep{dao2024flashattention,sanovar2025leanattention,ye2025flashinfer}. Asynchronous designs overlap transfers, matrix multiplication, and reduce partial-softmax synchronization to improve pipeline parallelism\citep{shah2024flashattention,hong2024flashdecoding++}. Meanwhile, programmable systems specialize fused kernels for score transformations, masks, and KV-cache layouts, supporting attention variants while leaving interaction selection to applications \citep{dong2025flexattention,ye2025flashinfer}. 

\textit{Structured sparse attention} selects tokens, spans, heads, or interaction blocks. Streaming windows retain initial attention sinks and recent tokens, bounding KV working sets but denying windowed heads direct access to excluded distant evidence \citep{xiao2024efficient,xiao2025duoattention}. For heterogeneous heads, calibration selects per-head spans or full-context retrieval and streaming roles, leaving retained QK precision fixed \citep{fu2024moa,xiao2025duoattention}. Input-dependent methods use sampled-query attention and lightweight block-importance estimates to select columns, diagonals, or blocks, adding online scoring and sparse-index construction \citep{jiang2024minference,lai2025flexprefill,zhu2025sampleattention,xu2025xattention}. Estimated cumulative attention mass adapts retention per input but omits output errors from quantizing retained scores \citep{lai2025flexprefill,zhu2025sampleattention,xu2025xattention}. Learning-based methods distill block-importance predictors from attention maps, requiring extra predictor training changes \citep{gao2024seerattention,yuan2025native}. 

\textit{Quantized and mixed-precision attention} controls arithmetic precision and operand representations. Blockwise scales and integer kernels accelerate 8-bit QK, but outlier-dominated scales reduce resolution for smaller values \citep{zhang2025sageattention,kang2025turboattention}. Smoothing and orthogonal transforms reduce outlier error, while finer groups, probability rescaling, and accumulation control enable lower-bit integer and floating-point attention with uniform QK precision within heads despite regional sensitivity differences \citep{zhang2024sageattention2,shah2024flashattention,zhang2026sageattention3}. Beyond scale selection, spatial mixed precision gives diagonal neighborhoods higher precision, yet computes all unmasked interactions, including low-impact ones \citep{ding2026diagonal,sharratt2026thriftattention}. KV-cache quantization tailors scale axes, quantization levels, and outlier handling to key and value distributions, reducing bytes per entry while cached tokens and per-step reads grow with context \citep{liu2024kivi,hooper2024kvquant}. Joint sparse-quantized systems add block skipping to low-bit QK, but retained regions share precision\citep{zhang2025spargeattention}. Other systems combine selective KV access or eviction with cache quantization, reducing traffic but requiring online page or token scoring during generation \citep{yang2025lserve,zhang2024q}.

\vspace{-5pt}
\section{Preliminaries}
\label{sec:preliminaries}
\vspace{-5pt}

Prefill computes attention for all prompt positions, allowing many queries to reuse each key and value. Decode processes one new query per request and rereads the accumulated KV cache each step. For a context of length $L$, dense attention evaluates $\mathcal{O}(L^2)$ causal interactions during prefill and $\mathcal{O}(L)$ per decode step~\citep{yang2025lserve}.

Both phases use the same attention operation. Let $Q\in\mathbb{R}^{L_q\times d}$ contain the queries and $K,V\in\mathbb{R}^{L_k\times d}$ the keys and values, where $d$ is the head dimension. The output is
{
\setlength{\abovedisplayskip}{3pt}
\setlength{\belowdisplayskip}{3pt}
\setlength{\abovedisplayshortskip}{2pt}
\setlength{\belowdisplayshortskip}{3pt}
\begin{equation}
P=\operatorname{softmax}\!\left(\frac{QK^{\top}}{\sqrt{d}}+M\right),\qquad O=PV,
\label{eq:ta-attention}
\end{equation}
}
where $M$ applies the causal mask and softmax operates. QK multiplication produces attention scores, and PV multiplication aggregates values using the attention probabilities in $P$. Grouped-query attention (GQA) allows several query heads to share the same $K$ and $V$~\citep{yang2025lserve}.

\begin{figure*}[t]
\centering
\includegraphics[width=\textwidth]{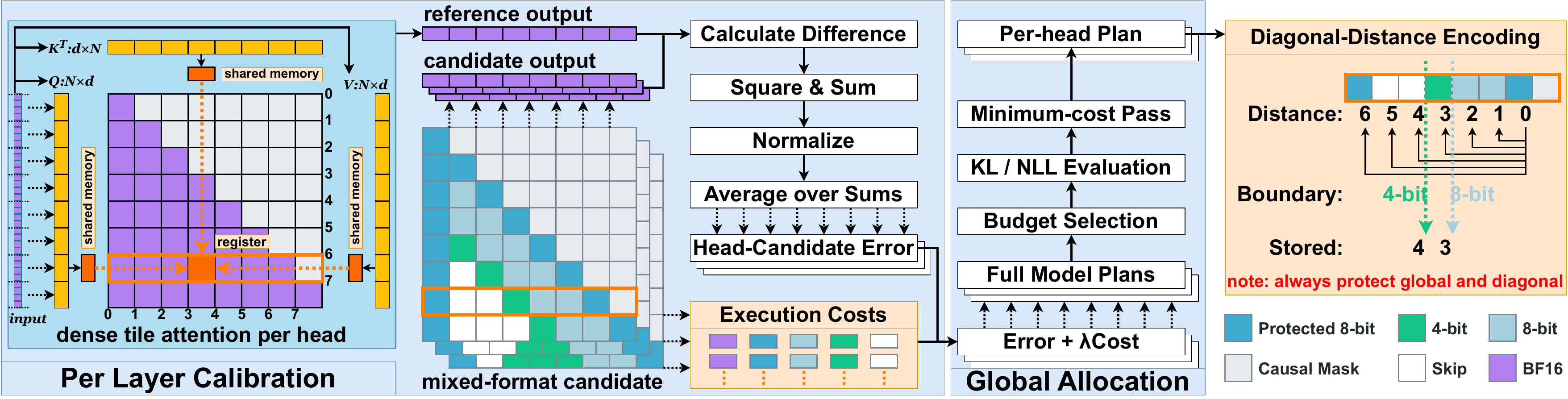}
\vspace{-20pt}
\caption{Offline calibration and compact distance encoding. Left: candidate boundary pairs are evaluated by normalized projected-output error. Center: errors and complete-tile execution costs guide allocation under a shared computation budget. Full-model prediction checks select the final boundary table. Right: two diagonal-distance boundaries per layer and query head encode 8-bit, 4-bit, and skipped QK regions, with the first key block and diagonal tiles protected.}
\label{fig:ta-framework}
\vspace{-10pt}
\end{figure*}

\textbf{Integer operands and scales.}
Integer quantization represents a group of values $X$ through integer codes $\widehat X$ and a scale $s_X$, with $X\approx s_X\widehat X$~\citep{zhang2025sageattention}. The corresponding operand scales restore the numerical magnitude of integer products. Products from different groups or precisions can then be accumulated in common units.

\textbf{Tiled attention and partial aggregation.}
Fused attention kernels pair query and key blocks into tiles, stage operands in shared memory, and keep intermediate scores and probabilities on chip~\citep{dao2022flashattention}. For one query row, a subset of keys produces a partial state $(m_r,l_r,o_r)$: its maximum score, the sum of exponentiated scores relative to that maximum, and the correspondingly weighted value sum. Partial states over disjoint key subsets are combined by aligning their score maxima:
{
\setlength{\abovedisplayskip}{3pt}
\setlength{\belowdisplayskip}{3pt}
\setlength{\abovedisplayshortskip}{2pt}
\setlength{\belowdisplayshortskip}{3pt}
\setlength{\jot}{1pt}
\begin{equation}
\begin{aligned}
m&=\max_r m_r,\\
l&=\sum_r \exp(m_r-m)l_r,\\
o&=\sum_r \exp(m_r-m)o_r,\qquad O=o/l.
\end{aligned}
\label{eq:ta-partial-merge}
\end{equation}
}
This rule supports both online accumulation across successive tiles and parallel reduction across key segments. The partials use common score and value units so that the denominator and numerator describe the same attention weights.

\vspace{-5pt}
\section{TriAttention}
\label{sec:method}
\vspace{-5pt}

\modelname uses regional sensitivity to assign attention tiles to skipping, 4-bit, or 8-bit QK computation. The diagonal-distance boundaries define each layer--head configuration for both offline candidate evaluation and precision-grouped traversal. Offline calibration evaluates candidate boundary pairs, and a model-wide allocator uses their errors and costs to select a boundary table under a shared computation budget (Figure~\ref{fig:ta-framework}). During inference, this boundary table guides fused prefill and compressed KV-cache for decode.

\vspace{-5pt}
\subsection{Compact Distance Encoding}
\label{sec:method-representation}

\vspace{-5pt}

Long contexts enlarge the attention tile map, while each GPU thread block has a limited shared-memory budget for traversal metadata and staged operands. We organize the three QK actions by distance from the attention diagonal. Each boundary pair describes a query head's retained context and QK precision regions.

For query block $i$ and key block $j$, let $\Delta=i-j$ be their distance in blocks. A configuration $c=(w_8,w_4)$ specifies the extent of the 8-bit QK region and the overall retained region, with $1\leq w_8\leq w_4$. We protect the first key block and the diagonal tile with 8-bit QK, preserving access to the prompt beginning and the query's own block. For the remaining tiles, the configuration assigns
{
\setlength{\abovedisplayskip}{3pt}
\setlength{\belowdisplayskip}{3pt}
\setlength{\abovedisplayshortskip}{2pt}
\setlength{\belowdisplayshortskip}{3pt}
\renewcommand{\arraystretch}{0.85}
\begin{equation}
a_c(\Delta)=
\begin{cases}
8\text{-bit}, & 1\leq\Delta<w_8,\\
4\text{-bit}, & w_8\leq\Delta<w_4,\\
\mathrm{skip}, & \Delta\geq w_4.
\end{cases}
\label{eq:ta-policy}
\end{equation}
}
Extending $w_4$ retains additional context at 4-bit precision, while increasing $w_8$ assigns 8-bit computation. We select a separate pair for every layer and query head from a candidate set $\mathcal{C}$ that includes full-context configurations.

\begin{figure}[thb]
\centering
\vspace{-10pt}
\includegraphics[width=\linewidth]{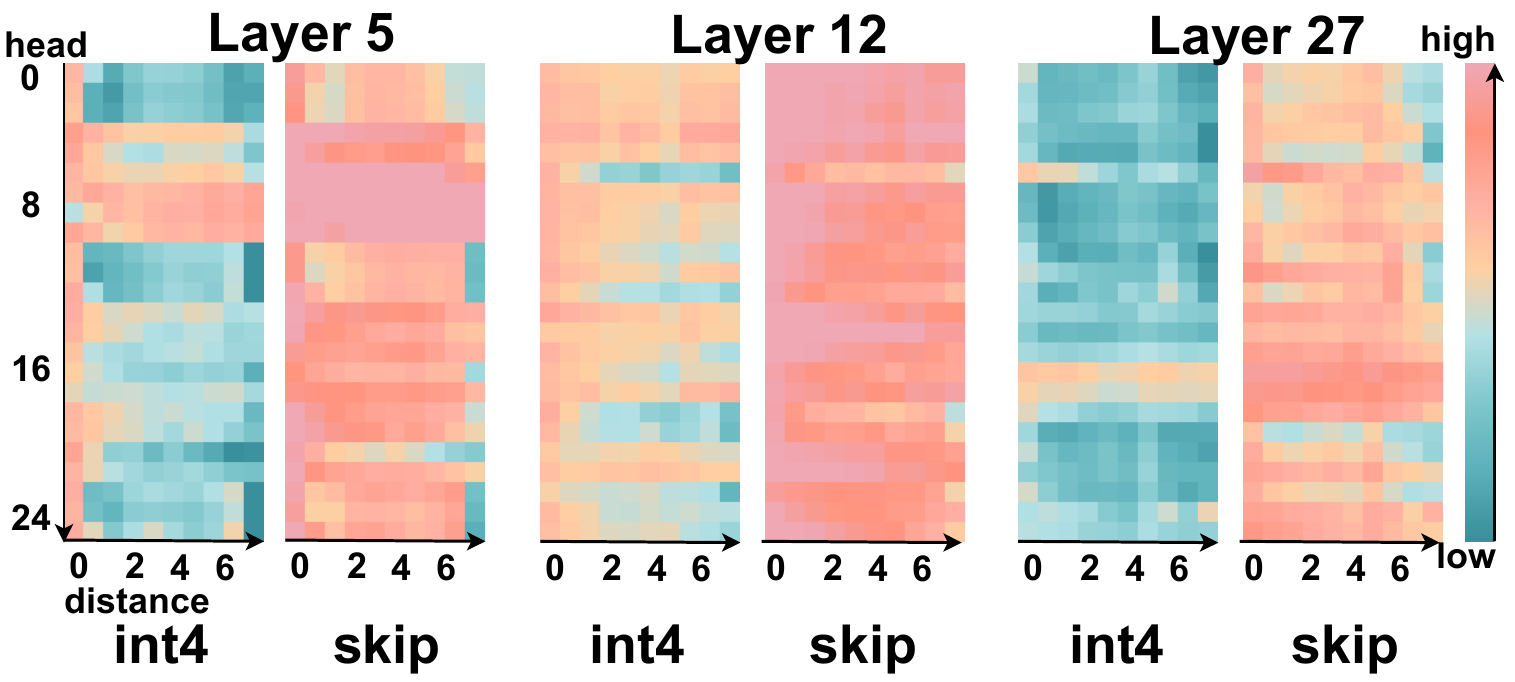}
\vspace{-23pt}
\caption{Per-layer QK sensitivity in Llama-3.2-3B-Instruct. Rows are query heads; columns are diagonal-distance bands ordered near to far. Colors show layer-normalized projected-output error from 4-bit QK or skipping, with nearer bands kept in 8-bit.}
\label{fig:per-layer-sens}

\includegraphics[width=\linewidth]{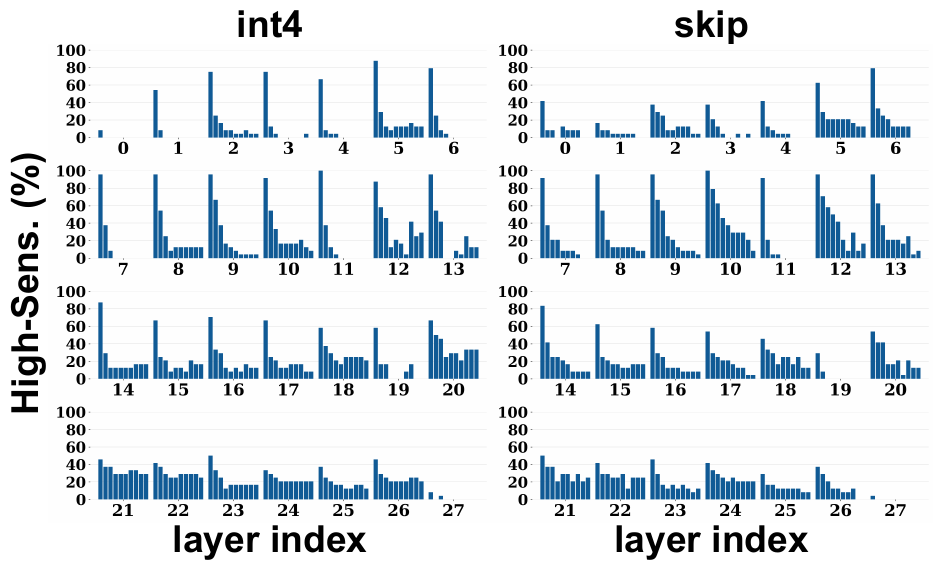}
\vspace{-23pt}
\caption{Distribution of high-sensitivity QK bands in Llama-3.2-3B-Instruct. Ten bars per layer. Each bar reports the percentage of heads whose error lies in the model-wide top 20\% of layer--head--band errors, ranked separately for 4-bit and skipping.}
\label{fig:global-sens}
\vspace{-15pt}
\end{figure}

The compact distance encoding in Figure~\ref{fig:ta-framework} stores two integers per layer and query head, keeping policy storage independent of context length. At execution time, each thread block uses these boundaries to generate only the retained key-block indices for its query block. The indices are grouped by precision, so their group boundaries specify the arithmetic paths. Calibration evaluates these boundary pairs, and allocation selects one pair per layer and query head to form the boundary table.

\begin{figure*}[ht]
\centering
\includegraphics[width=\textwidth]{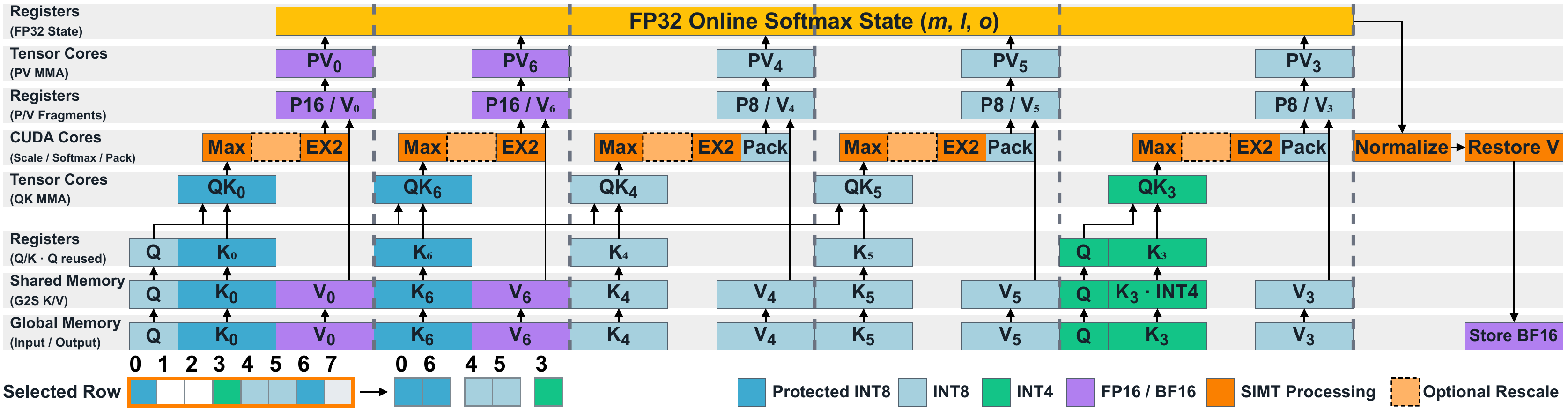}
\vspace{-20pt}
\caption{Fused prefill with compact, precision-grouped traversal. The illustrated query block visits protected tiles 0 and 6, 8-bit QK tiles 4 and 5, and 4-bit QK tile 3. Protected tiles use 16-bit PV, while other retained tiles use 8-bit PV. Shared-memory staging and query reuse feed one register-resident attention state, followed by normalization, value restoration, and 16-bit output.}
\label{fig:ta-prefill}
\vspace{-10pt}
\end{figure*}

\vspace{-5pt}
\subsection{Sensitivity-Guided Global Allocation}
\label{sec:method-allocation}
\vspace{-5pt}

Figures~\ref{fig:per-layer-sens} and~\ref{fig:global-sens} show that output errors from skipping and 4-bit QK vary across layers, heads, and diagonal-distance bands.
We therefore evaluate candidate boundary pairs for each layer and query head and select a boundary table under a shared computation budget.

\textbf{Measuring the effect on the layer output.}
To compare these configurations, we measure how each one changes a head's contribution to the representation passed through the model by the attention output projection. So we evaluate its error after this projection. We normalize each head's projected-output error by the layer's reference attention-output energy, making errors comparable across layers.

For calibration input $x$, let $O_{\ell h}^{c}(x)$ and $O_{\ell h}^{\mathrm{ref}}(x)$ denote the candidate and dense 16-bit reference outputs of query head $h$ in layer $\ell$ at the scored calibration positions. Let $W_{\ell h}^{O}$ be the block of the output projection that maps this head into the model's hidden dimension, and let $Y_{\ell}^{\mathrm{ref}}=\sum_h O_{\ell h}^{\mathrm{ref}}W_{\ell h}^{O}$ be the complete reference attention output. Over calibration prompts $\mathcal{D}$, we score the candidate by
{
\setlength{\abovedisplayskip}{3pt}
\setlength{\belowdisplayskip}{3pt}
\setlength{\abovedisplayshortskip}{2pt}
\setlength{\belowdisplayshortskip}{3pt}
\begin{equation}
e_{\ell h}(c)=\frac{1}{|\mathcal{D}|}\sum_{x\in\mathcal{D}}
\frac{\left\|\bigl(O_{\ell h}^{c}(x)-O_{\ell h}^{\mathrm{ref}}(x)\bigr)W_{\ell h}^{O}\right\|_F^2}
{\left\|Y_{\ell}^{\mathrm{ref}}(x)\right\|_F^2}.
\label{eq:ta-error}
\end{equation}
}
A lower score identifies a configuration that preserves the head's projected contribution more closely. All candidates' reference output continues to the next layer. Their scores thus compare the combined effects of context retention and quantized computation under common inputs.

\textbf{Allocating a shared computation budget.}
Each error score is paired with the computation needed to produce that output. At a common target context length, let $n_8(c)$ and $n_4(c)$ count the candidate's retained 8-bit and 4-bit QK tiles, excluding the protected tiles shared by all candidates. Profiling complete tiles, including QK, softmax, and PV, gives the relative cost $\gamma$ of a tile using 4-bit QK compared with one using 8-bit QK. We estimate normalized cost as $C(c)=[n_8(c)+\gamma n_4(c)]/n_{\mathrm{all8}}$, where $n_{\mathrm{all8}}$ is the corresponding count when all eligible tiles use 8-bit QK. This cost prices both context extension and increased precision.

For each layer and head, we choose $c_{\ell h}$ to minimize summed error under a shared normalized-cost budget $B$:
{
\setlength{\abovedisplayskip}{3pt}
\setlength{\belowdisplayskip}{3pt}
\setlength{\abovedisplayshortskip}{2pt}
\setlength{\belowdisplayshortskip}{3pt}
\begin{equation}
\min_{\{c_{\ell h}\in\mathcal{C}\}}\sum_{\ell,h}e_{\ell h}(c_{\ell h})
\;\text{s.t.}\;
\sum_{\ell,h}C(c_{\ell h})\leq B.
\label{eq:ta-allocation}
\end{equation}
}
We generate candidate boundary tables by sweeping a shared cost weight $\lambda\geq0$, with each head selecting $c_{\ell h}(\lambda)=\arg\min_{c\in\mathcal{C}}[e_{\ell h}(c)+\lambda C(c)]$. The shared weight assigns the same price to computation in every head, giving wider context or higher precision priority where it yields greater error reduction per added cost. Per budget, we retain the feasible sampled table with the lowest summed error.

The resulting tables are evaluated in the complete model on separate selection prompts. We choose the lowest-cost table that satisfies the prescribed thresholds for mean next-token KL divergence from the reference model and the mean increase in teacher-forced response negative log-likelihood within every context-length group. Appendix~\ref{app:calibration} specifies the calibration prompts, candidate grid, cost-weight sweep, and selection thresholds.

\vspace{-5pt}
\subsection{Fused Prefill with Compact Traversal}
\label{sec:method-prefill}
\vspace{-5pt}

Given the selected boundary table and a layer's $Q$, $K$, and $V$, prefill prepares the required operand formats and constructs a compact traversal for each query block. The traversal determines which K/V tiles enter shared memory and which arithmetic path processes them. The kernel accumulates their contributions into one attention output. Figure~\ref{fig:ta-prefill} follows the selected tiles through this execution.

\begin{figure*}[t]
\centering
\includegraphics[width=\textwidth]{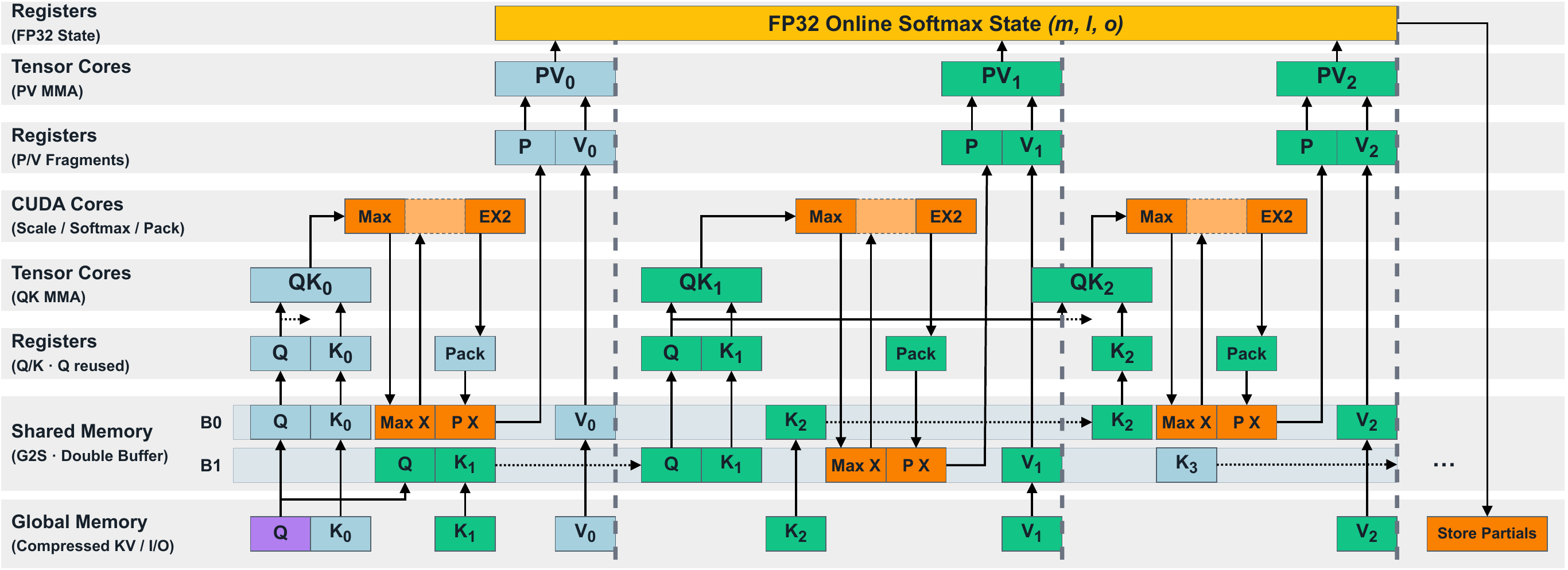}
\vspace{-20pt}
\caption{Mixed-precision decode over the compressed prompt KV cache. Alternating shared-memory buffers B0/B1 stage retained KV tiles for 8-bit (blue) or 4-bit (green) computation. The kernel reuses query fragments and maintains attention state in registers across tiles. Each key segment writes partial states for subsequent merging with other segments and the 16-bit generated-token tail.}
\label{fig:ta-decode}
\vspace{-10pt}
\end{figure*}

\textbf{Preparing reusable operands.}
After rotary position embedding, the runtime centers keys and applies a common orthogonal signed Hadamard transform to queries and keys to control their dynamic range before quantization. It prepares the 8-bit and 4-bit Q/K formats for the selected table, with their scales. Value formats share channelwise centering and scaling statistics, allowing reconstruction after aggregation. Preparation produces 16-bit and 8-bit values for prefill and 4-bit values for the compressed prompt KV cache. GPU memory holds the packed operands and statistics.

\textbf{Constructing the shared-memory traversal.}
Each thread block reads its head's two boundaries and combines them with its query-block index to generate the retained key-block indices. Threads write these indices cooperatively in shared memory, with consecutive segments for protected tiles, 8-bit QK tiles, and 4-bit QK tiles. Segment boundaries specify the arithmetic path, so each list entry carries only a key-block index. The kernel consumes this list to stage the corresponding packed K/V tiles and reuses the query representation within each pass. Skipping removes tiles from the list and their associated QK, softmax, and PV work. This organization places the current query block's traversal data alongside its operand buffers, linking the compact distance policy directly to on-chip execution.

\textbf{Executing the precision groups.}
Protected tiles use 8-bit QK with 16-bit PV.
Both remaining QK precisions use 8-bit probabilities and 8-bit values for PV. Tile-local probability scaling preserves resolution within each retained tile. The integer path uses the same probability codes for its normalization sum and PV product, and their scale places both contributions in the shared state. All groups update the same attention state in registers, keeping QK, probability formation, and PV accumulation fused across the traversal. After the final tile, the kernel normalizes and restores the value coordinates to produce $O$ for the layer's output projection. Quantization scales, probability coding, and shared-state updates are detailed in Appendix~\ref{app:numerics}. Prepared prompt K/V and statistics then supply the cache.

\vspace{-5pt}
\subsection{Decoding from the Compressed Cache}
\label{sec:method-decode}
\vspace{-5pt}

To reduce repeated prompt-data reads, we use the selected boundaries to compact the prepared quantized K/V into a compressed prompt KV cache. Cache construction applies each head's boundaries to the final prompt query block, identifies the retained prompt tiles, and packs this selection once for reuse throughout generation. Each new query attends to this compressed prompt KV cache and a 16-bit tail containing generated-token keys and values. Figure~\ref{fig:ta-decode} shows the computation over the retained prompt data.

\textbf{Mapping head decisions to shared storage.}
The policy specifies context per query head, while GQA shares physical K/V storage across a group of query heads. We retain the union of the group's requested tiles and store each key tile once at the highest requested precision. A head mask identifies which query heads retain the tile, and participating heads compute QK at its stored precision. Compact buffers store retained K/V tiles with indices, masks, and reconstruction statistics. Values use 8-bit storage for the first prompt block and a recent window at the prompt's end, with 4-bit storage for the other retained blocks. This combines shared storage with each head's selected context and reduces prompt-data reads at each generation step. 

\textbf{Sharing reads and parallelizing the retained context.}
The incoming query receives the same rotation and quantization used in prefill, allowing it to consume the stored prompt keys directly. Queries sharing a KV head form rows of one matrix operation, so each loaded KV tile serves the participating heads in the group. Since decode supplies one query position per request, the kernel splits the retained key tiles across GPU thread blocks to expose parallelism along the context. Each block processes its assigned segment using the stored key precision and probability codes matched to the value precision, returning $(m_r,l_r,o_r)$ for merging.

\textbf{Producing the output and extending the context.}
Newly generated keys and values are appended to the 16-bit tail. The merge kernel computes attention over this tail using the prompt's key-centering offset. It combines the prompt partials and restores their value coordinates with the stored statistics. With both sources in common score and value units, the kernel merges their contributions and normalizes the result to produce the current query's output. 
Appendix~\ref{app:execution} details compact traversal, shared cache indexing, and prompt--tail merging.

\begin{table*}[t]
\centering
\caption{End-to-end inference efficiency for \textbf{Llama-3.2-3B-Instruct} on A100-SXM4-40GB, with 65,536 input tokens, 128 output tokens, and batch size one. Peak VRAM covers generation including prefill. Cell colors are relative to BF16 under each metric's direction: green denotes equal or better results, blue worse results, and gray BF16. Bold and underlined values mark the best and worst results among Ours and Others; displayed ties share the mark.}
\label{tab:efficiency}
\small
\setlength{\tabcolsep}{3pt}
\begin{adjustbox}{max width=\textwidth}
\begin{tabular*}{\textwidth}{@{\extracolsep{\fill}}llrrrrrrr@{}}
\toprule
\multirow{2}{*}{\textbf{Metric}} & \multirow{2}{*}{\textbf{Unit}} & \multicolumn{1}{c}{\textbf{Upper Ref.}} & \multicolumn{1}{c}{\textbf{Ours}} & \multicolumn{5}{c}{\textbf{Others}}\\
\cmidrule(lr){3-3}\cmidrule(lr){4-4}\cmidrule(lr){5-9}
 & & \textbf{BF16} & \textbf{\modelname} & \textbf{MInference} & \textbf{FlexPrefill} & \textbf{SageAttn} & \textbf{MoA} & \textbf{SpargeAttn}\\
\midrule
TPOT $\downarrow$ & ms/token & \cellcolor{heatneutral}26.64 & \cellcolor{abovegreen!70}\textbf{10.06} & \cellcolor{belowblue!20}26.91 & \cellcolor{belowblue!20}26.79 & \cellcolor{belowblue!20}26.93 & \cellcolor{belowblue!80}\underline{176.64} & \cellcolor{belowblue!20}26.92\\
Throughput $\uparrow$ & tok/s & \cellcolor{heatneutral}14.06 & \cellcolor{abovegreen!70}\textbf{22.80} & \cellcolor{abovegreen!20}14.23 & \cellcolor{abovegreen!50}17.39 & \cellcolor{abovegreen!20}14.59 & \cellcolor{belowblue!70}\underline{4.64} & \cellcolor{belowblue!20}13.97\\
Peak VRAM $\downarrow$ & GiB & \cellcolor{heatneutral}15.65 & \cellcolor{abovegreen!50}\textbf{11.76} & \cellcolor{belowblue!20}15.77 & \cellcolor{belowblue!20}16.40 & \cellcolor{abovegreen!20}15.65 & \cellcolor{belowblue!20}15.77 & \cellcolor{belowblue!20}\underline{16.82}\\
Energy $\downarrow$ & J/1K tokens & \cellcolor{heatneutral}24788.26 & \cellcolor{abovegreen!55}\textbf{15737.62} & \cellcolor{abovegreen!30}21536.48 & \cellcolor{abovegreen!50}18358.31 & \cellcolor{abovegreen!20}23459.16 & \cellcolor{belowblue!80}\underline{71430.99} & \cellcolor{abovegreen!20}23365.47\\
\bottomrule
\end{tabular*}
\end{adjustbox}
\vspace{-10pt}
\end{table*}

\begin{figure*}[!tbp]
\centering
\includegraphics[width=0.27650431\textwidth]{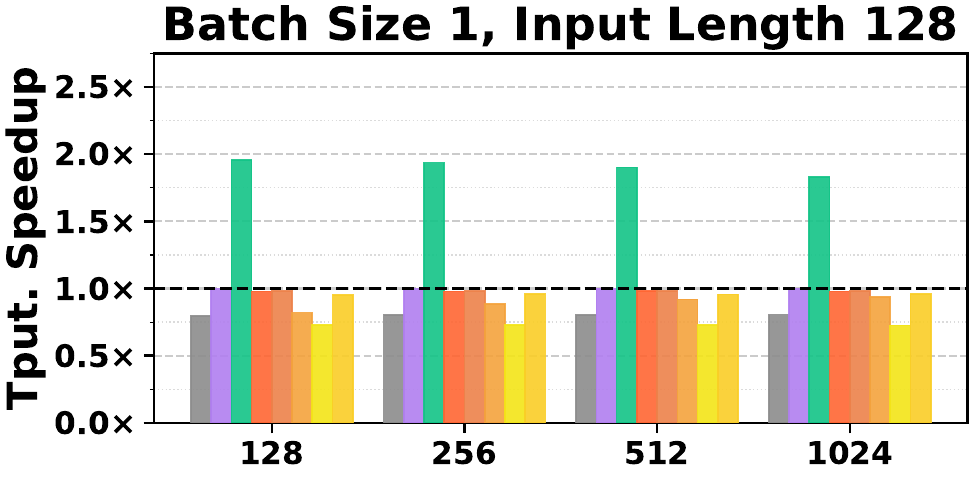}\hfill
\includegraphics[width=0.23449856\textwidth]{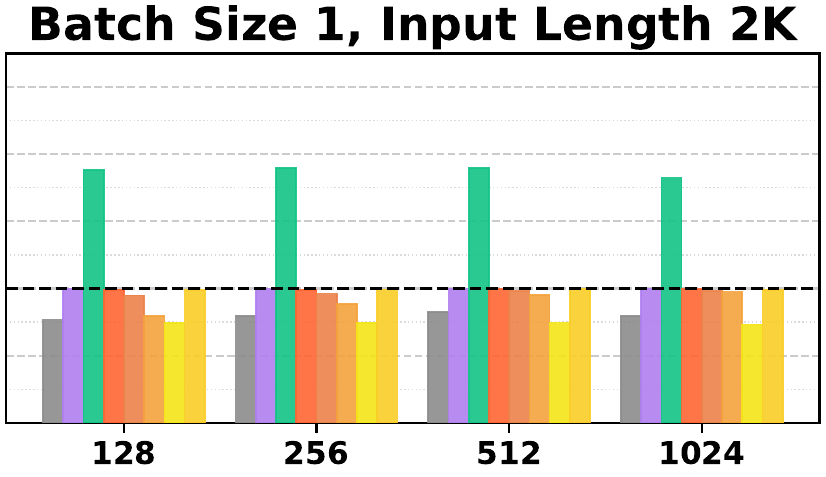}\hfill
\includegraphics[width=0.23449856\textwidth]{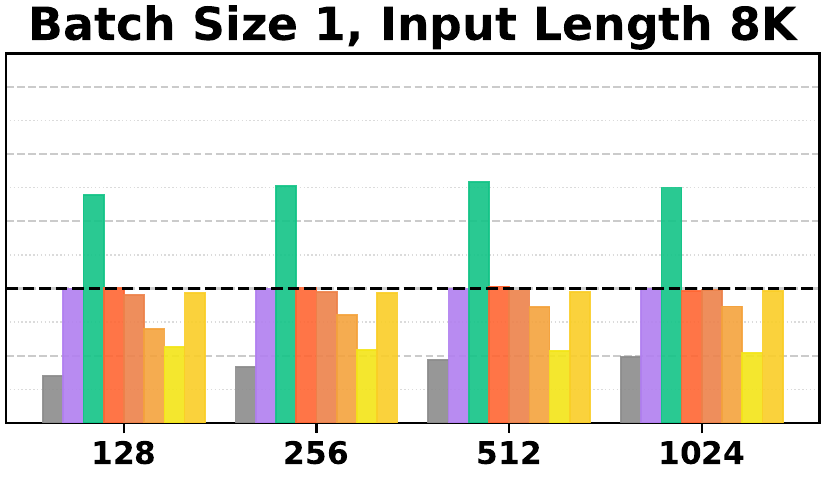}\hfill
\includegraphics[width=0.23449856\textwidth]{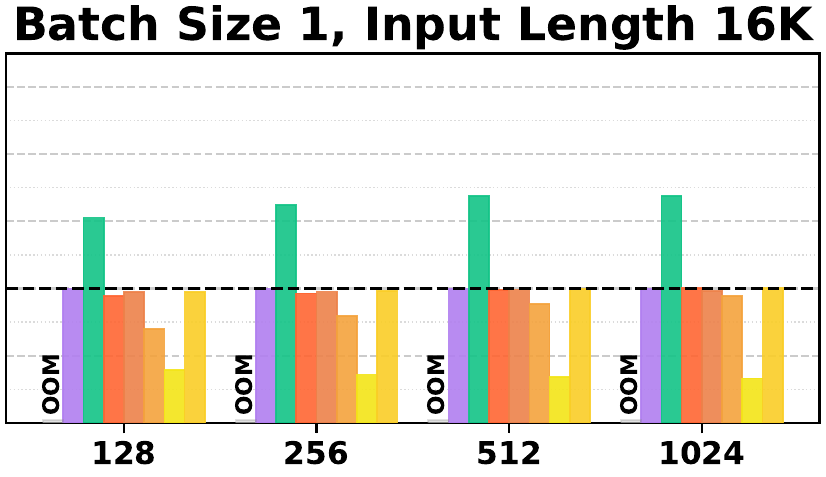}\\[-3pt]
\includegraphics[width=0.27650431\textwidth]{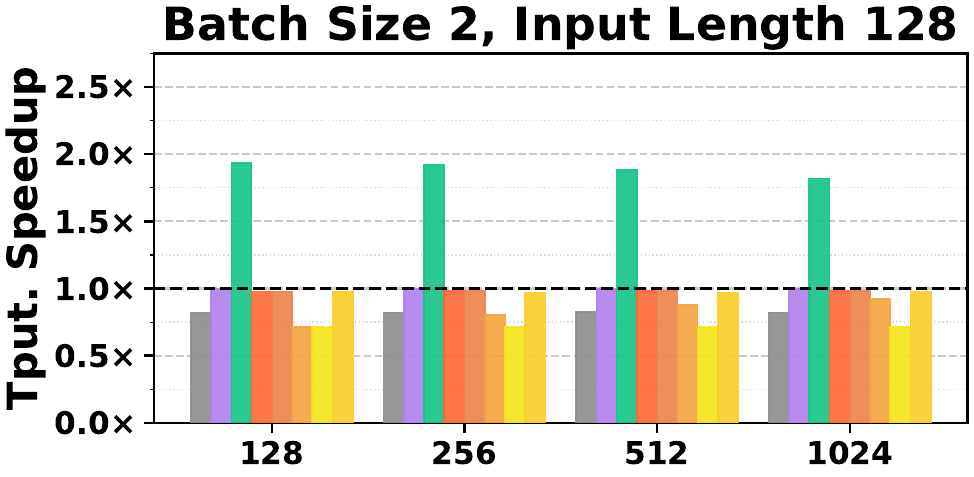}\hfill
\includegraphics[width=0.23449856\textwidth]{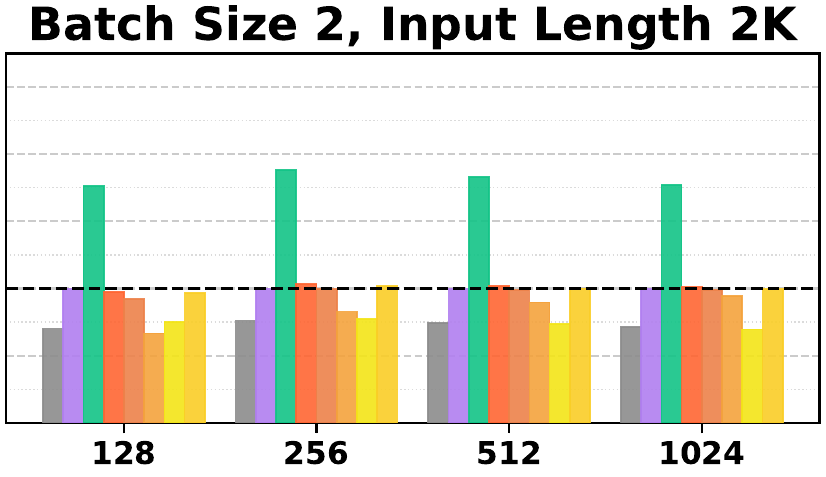}\hfill
\includegraphics[width=0.23449856\textwidth]{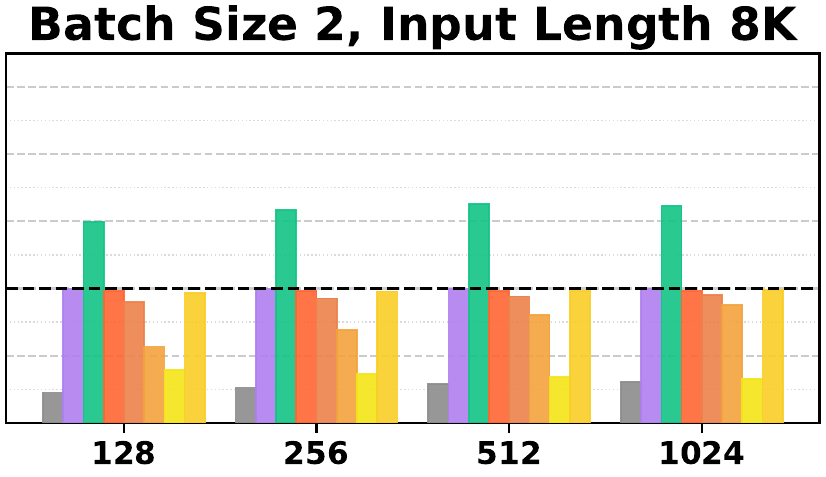}\hfill
\includegraphics[width=0.23449856\textwidth]{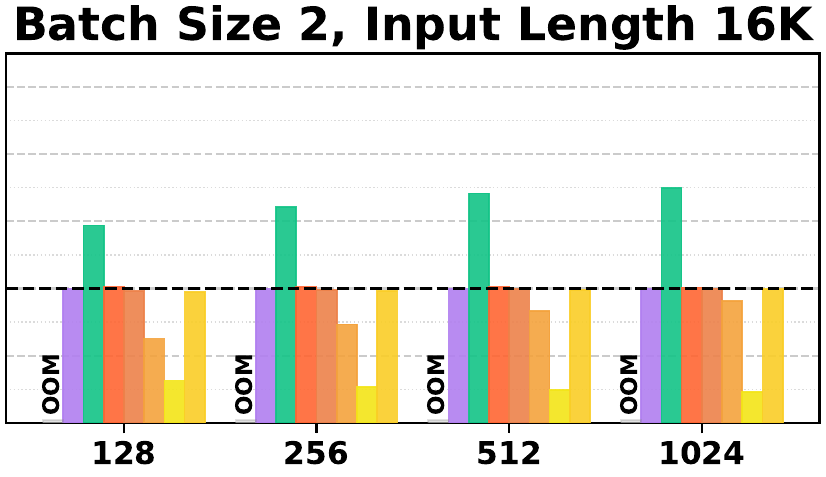}\\[-3pt]
\includegraphics[width=0.27650431\textwidth]{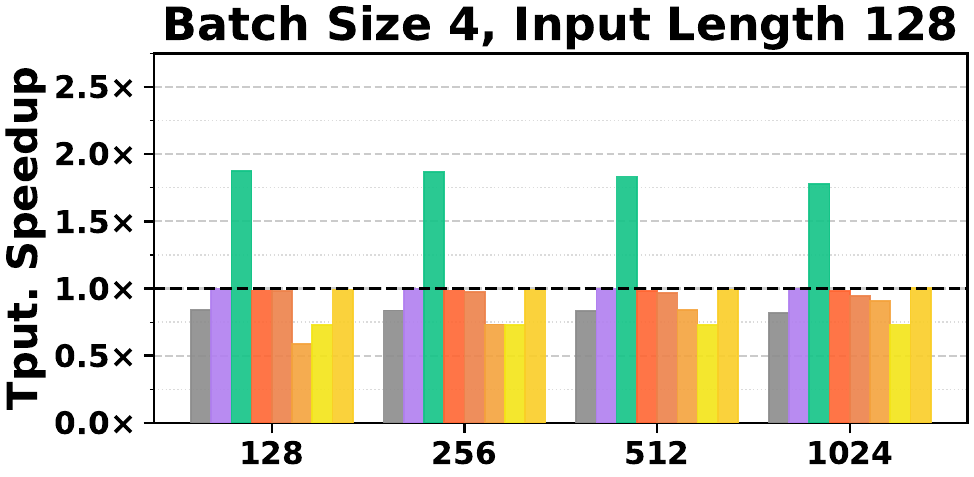}\hfill
\includegraphics[width=0.23449856\textwidth]{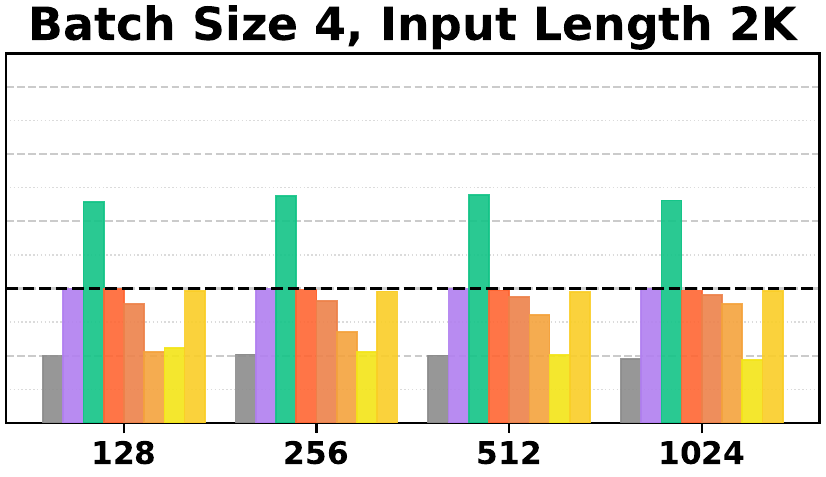}\hfill
\includegraphics[width=0.23449856\textwidth]{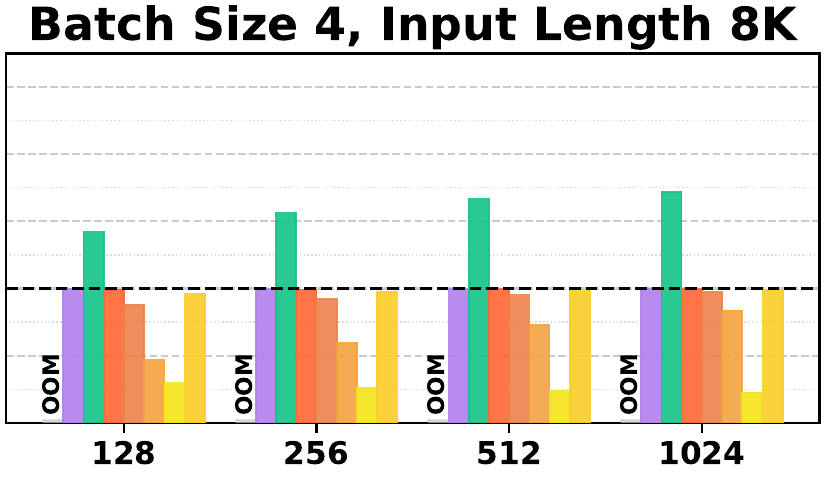}\hfill
\includegraphics[width=0.23449856\textwidth]{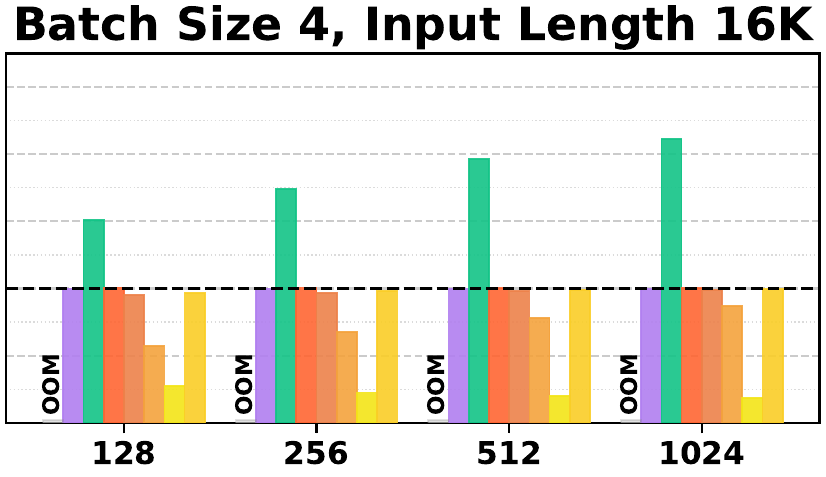}\\[-3pt]
\includegraphics[width=0.27650431\textwidth]{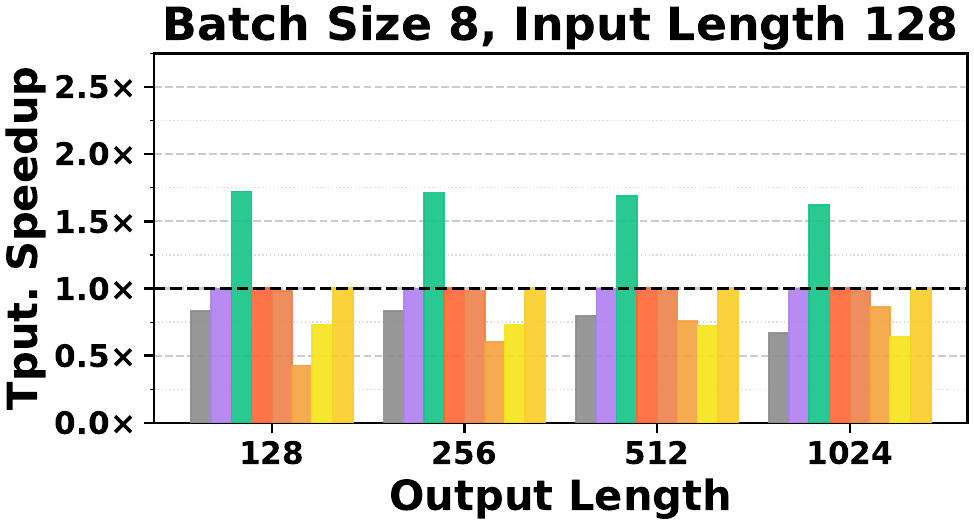}\hfill
\includegraphics[width=0.23449856\textwidth]{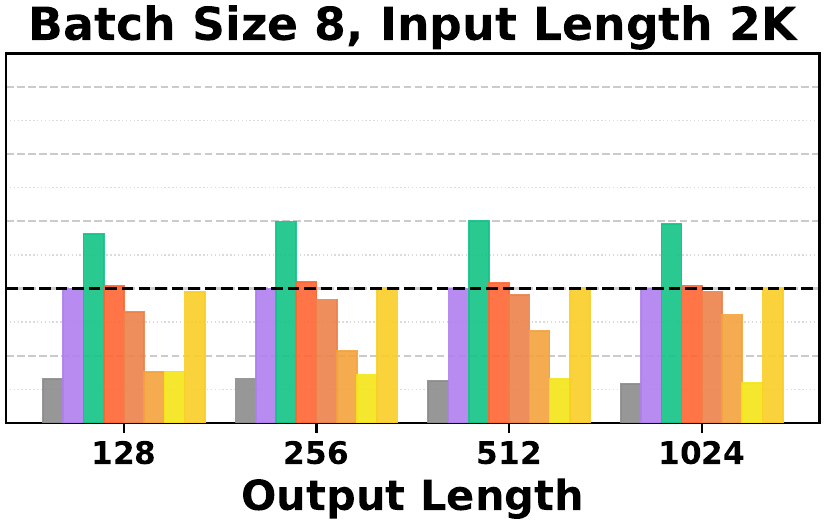}\hfill
\includegraphics[width=0.23449856\textwidth]{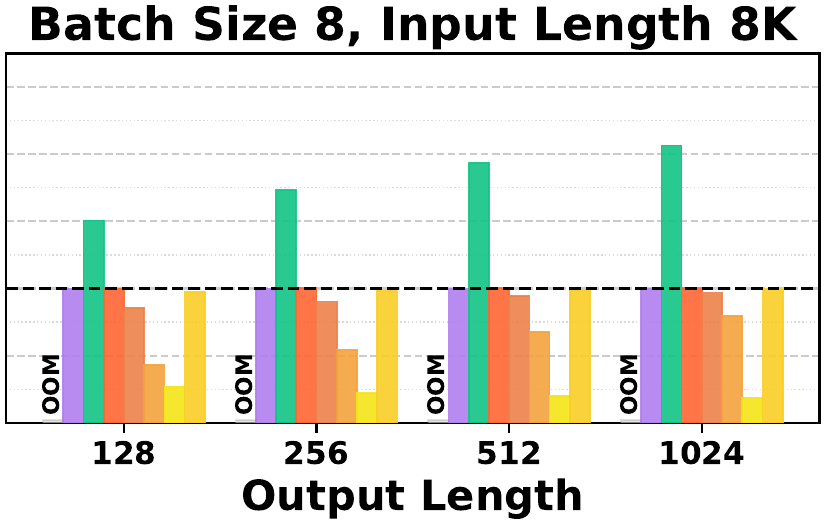}\hfill
\includegraphics[width=0.23449856\textwidth]{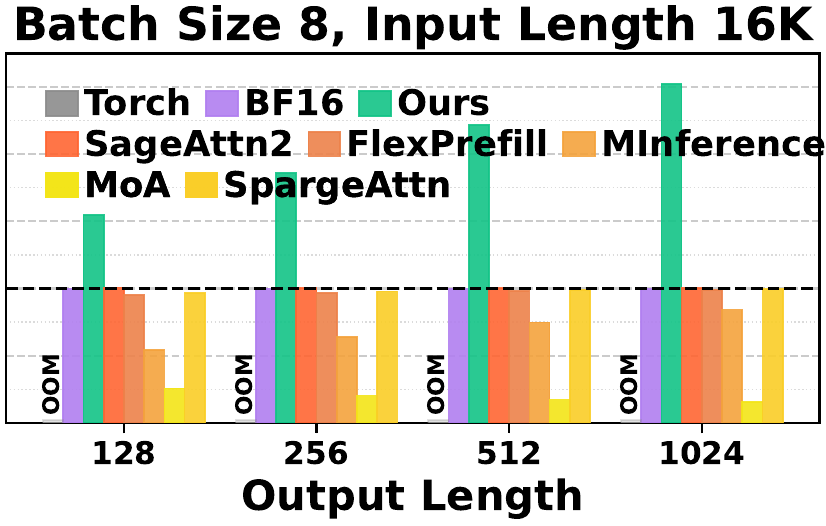}
\vspace{-10pt}
\caption{\textbf{Generation-throughput ratio on A100-SXM4-40GB.} Llama-3.2-3B-Instruct. Rows use batch sizes 1, 2, 4, and 8; columns use 128, 2,048, 8,192, 16,384 input tokens. Each panel varies output length over 128, 256, 512, and 1024 tokens. BF16 is normalized to $1\times$; higher is better. Method colors and order are fixed across panels.}
\label{fig:efficiency-grid-main}
\vspace{-5pt}
\end{figure*}

\section{Experiments}
\label{sec:experiments}

\subsection{Experimental Setup}
\label{sec:setup}

We evaluate inference efficiency and model quality for general and long-context tasks, with allocation controls examining the value of head-specific retention and precision.
The main evaluations focus on Llama-3.2-3B-Instruct, with computation-level ablations also covering Qwen2-7B-Instruct and Ministral-3-8B-Instruct-2512-BF16. Appendix~\ref{app:extended-quality} reports additional quality results for the latter two models.

\noindent\textbf{Datasets and Metrics.}
We evaluate language modeling using WikiText-2 word perplexity and language understanding using accuracy on HellaSwag, MMLU, ARC-Challenge, LAMBADA, PIQA, and WinoGrande.
Long-context question answering is assessed on nine LV-Eval~\citep{yuan2024lv} datasets at the 16K, 32K, and 64K length levels, using the benchmark's task-specific F1 and ROUGE-L metrics.

\begin{table*}[t]
\centering
\caption{General-task results for Llama-3.2-3B-Instruct. Accuracy is reported in percent (higher is better); WikiText-2 uses word perplexity (lower is better). BF16 is the upper reference; \texttt{all\_int4} is the lower reference. Colors compare with BF16: green denotes equal or better results, blue worse results, and gray BF16. Bold marks the best result among Ours and Others; displayed ties share the mark.}
\label{tab:general-llama}
\small
\setlength{\tabcolsep}{3pt}
\begin{adjustbox}{max width=\textwidth}
\begin{tabular*}{\textwidth}{@{\extracolsep{\fill}}llrrrrrrrr@{}}
\toprule
\multirow{2}{*}{\textbf{Benchmark}} & \multirow{2}{*}{\textbf{Metric}} & \multicolumn{1}{c}{\textbf{Upper Ref.}} & \multicolumn{1}{c}{\textbf{Lower Ref.}} & \multicolumn{1}{c}{\textbf{Ours}} & \multicolumn{5}{c}{\textbf{Others}}\\
\cmidrule(lr){3-3}\cmidrule(lr){4-4}\cmidrule(lr){5-5}\cmidrule(lr){6-10}
 & & \textbf{BF16} & \textbf{\texttt{all\_int4}} & \textbf{\modelname} & \textbf{MInference} & \textbf{FlexPrefill} & \textbf{SageAttn} & \textbf{MoA} & \textbf{SpargeAttn}\\
\midrule
WikiText-2 & PPL $\downarrow$ & \cellcolor{heatneutral}12.49 & \cellcolor{belowblue!25}12.59 & \cellcolor{belowblue!25}12.50 & \cellcolor{belowblue!25}12.53 & \cellcolor{belowblue!55}12.74 & \cellcolor{belowblue!25}12.52 & \cellcolor{belowblue!40}12.64 & \cellcolor{abovegreen!25}\textbf{12.43}\\
HellaSwag & Acc. $\uparrow$ & \cellcolor{heatneutral}71.52 & \cellcolor{abovegreen!25}71.53 & \cellcolor{abovegreen!70}\textbf{72.03} & \cellcolor{abovegreen!25}71.62 & \cellcolor{abovegreen!40}71.63 & \cellcolor{abovegreen!55}71.82 & \cellcolor{abovegreen!40}71.65 & \cellcolor{abovegreen!25}71.61\\
MMLU & Acc. $\uparrow$ & \cellcolor{heatneutral}60.66 & \cellcolor{belowblue!25}60.26 & \cellcolor{abovegreen!70}\textbf{61.32} & \cellcolor{belowblue!25}60.65 & \cellcolor{belowblue!25}60.64 & \cellcolor{abovegreen!40}60.86 & \cellcolor{abovegreen!70}61.28 & \cellcolor{abovegreen!40}60.79\\
ARC-Challenge & Acc. $\uparrow$ & \cellcolor{heatneutral}52.56 & \cellcolor{belowblue!40}51.71 & \cellcolor{abovegreen!55}52.97 & \cellcolor{abovegreen!40}52.73 & \cellcolor{belowblue!25}52.47 & \cellcolor{belowblue!40}51.79 & \cellcolor{abovegreen!70}\textbf{53.07} & \cellcolor{belowblue!25}52.05\\
LAMBADA & Acc. $\uparrow$ & \cellcolor{heatneutral}64.23 & \cellcolor{abovegreen!25}64.33 & \cellcolor{abovegreen!70}\textbf{64.83} & \cellcolor{abovegreen!40}64.39 & \cellcolor{abovegreen!55}64.49 & \cellcolor{abovegreen!25}64.25 & \cellcolor{abovegreen!70}64.82 & \cellcolor{abovegreen!40}64.39\\
PIQA & Acc. $\uparrow$ & \cellcolor{heatneutral}77.31 & \cellcolor{belowblue!25}77.15 & \cellcolor{abovegreen!55}\textbf{77.65} & \cellcolor{belowblue!25}76.71 & \cellcolor{belowblue!25}77.26 & \cellcolor{belowblue!25}76.88 & \cellcolor{belowblue!25}77.04 & \cellcolor{belowblue!25}76.71\\
WinoGrande & Acc. $\uparrow$ & \cellcolor{heatneutral}71.43 & \cellcolor{belowblue!25}71.11 & \cellcolor{abovegreen!55}\textbf{71.85} & \cellcolor{abovegreen!25}71.51 & \cellcolor{abovegreen!40}71.59 & \cellcolor{abovegreen!55}71.82 & \cellcolor{belowblue!25}70.80 & \cellcolor{belowblue!40}70.24\\
\bottomrule
\end{tabular*}
\end{adjustbox}
\vspace{-10pt}
\end{table*}

\begin{table*}[t]
\centering
\caption{LV-Eval QA results for Llama-3.2-3B-Instruct. Dataset identifiers follow Section~\ref{sec:model-performance}. Scores use official task-specific metrics on a 0--100 scale; higher is better. Length labels follow the official dataset levels. BF16 is the upper reference; \texttt{all\_int4} is the lower reference. Colors compare with BF16: green denotes equal or better results, blue worse results, and gray BF16. Bold marks the best result among Ours, Others, and Ablations; displayed ties share the mark.}
\label{tab:quality}
\small
\setlength{\tabcolsep}{3pt}
\begin{adjustbox}{max width=\textwidth}
\begin{tabular}{@{}ccrrrrrrrrrrr@{}}
\toprule
\multirow{2}{*}{\textbf{Dataset}} & \multirow{2}{*}{\textbf{Level}} & \multicolumn{1}{c}{\textbf{Upper Ref.}} & \multicolumn{1}{c}{\textbf{Lower Ref.}} & \multicolumn{1}{c}{\textbf{Ours}} & \multicolumn{5}{c}{\textbf{Others}} & \multicolumn{3}{c}{\textbf{Ablations}}\\
\cmidrule(lr){3-3}\cmidrule(lr){4-4}\cmidrule(lr){5-5}\cmidrule(lr){6-10}\cmidrule(lr){11-13}
 & & \textbf{BF16} & \textbf{\texttt{all\_int4}} & \textbf{\modelname} & \textbf{MInference} & \textbf{FlexPrefill} & \textbf{SageAttn} & \textbf{MoA} & \textbf{SpargeAttn} & \textbf{Layer Same} & \textbf{All Same} & \textbf{Random}\\
\midrule
\multirow{3}{*}{\ds{1}} & 16K & \cellcolor{heatneutral}21.33 & \cellcolor{belowblue!70}18.10 & \cellcolor{abovegreen!70}21.94 & \cellcolor{abovegreen!70}26.93 & \cellcolor{abovegreen!70}27.11 & \cellcolor{abovegreen!70}\textbf{29.79} & \cellcolor{belowblue!70}13.54 & \cellcolor{belowblue!25}20.85 & \cellcolor{belowblue!25}20.66 & \cellcolor{belowblue!40}20.05 & \cellcolor{belowblue!40}19.65\\
 & 32K & \cellcolor{heatneutral}18.49 & \cellcolor{belowblue!70}15.34 & \cellcolor{belowblue!25}\textbf{18.07} & \cellcolor{belowblue!70}11.55 & \cellcolor{belowblue!70}11.74 & \cellcolor{belowblue!70}13.50 & \cellcolor{belowblue!70}10.02 & \cellcolor{belowblue!55}16.40 & \cellcolor{belowblue!55}16.50 & \cellcolor{belowblue!40}17.47 & \cellcolor{belowblue!70}13.38\\
 & 64K & \cellcolor{heatneutral}13.95 & \cellcolor{belowblue!55}11.32 & \cellcolor{abovegreen!70}\textbf{14.61} & \cellcolor{belowblue!70}5.31 & \cellcolor{belowblue!70}5.39 & \cellcolor{belowblue!70}5.77 & \cellcolor{belowblue!70}6.21 & \cellcolor{belowblue!25}13.72 & \cellcolor{belowblue!55}11.63 & \cellcolor{belowblue!70}6.47 & \cellcolor{belowblue!70}4.30\\
\midrule
\multirow{3}{*}{\ds{2}} & 16K & \cellcolor{heatneutral}16.22 & \cellcolor{belowblue!40}15.48 & \cellcolor{abovegreen!70}\textbf{16.84} & \cellcolor{belowblue!40}14.61 & \cellcolor{belowblue!40}14.92 & \cellcolor{abovegreen!55}16.67 & \cellcolor{belowblue!40}14.57 & \cellcolor{belowblue!40}15.33 & \cellcolor{belowblue!25}15.84 & \cellcolor{belowblue!25}15.56 & \cellcolor{belowblue!25}16.20\\
 & 32K & \cellcolor{heatneutral}17.59 & \cellcolor{belowblue!55}15.16 & \cellcolor{belowblue!25}\textbf{17.01} & \cellcolor{belowblue!70}12.01 & \cellcolor{belowblue!70}11.64 & \cellcolor{belowblue!70}14.59 & \cellcolor{belowblue!55}14.96 & \cellcolor{belowblue!40}16.87 & \cellcolor{belowblue!55}15.10 & \cellcolor{belowblue!40}16.31 & \cellcolor{belowblue!55}15.14\\
 & 64K & \cellcolor{heatneutral}17.00 & \cellcolor{belowblue!55}15.25 & \cellcolor{abovegreen!70}\textbf{17.92} & \cellcolor{belowblue!70}7.96 & \cellcolor{belowblue!70}6.81 & \cellcolor{belowblue!70}11.22 & \cellcolor{belowblue!70}13.36 & \cellcolor{belowblue!25}16.39 & \cellcolor{belowblue!40}15.35 & \cellcolor{belowblue!70}14.20 & \cellcolor{belowblue!55}14.37\\
\midrule
\multirow{3}{*}{\ds{3}} & 16K & \cellcolor{heatneutral}23.76 & \cellcolor{belowblue!70}20.26 & \cellcolor{abovegreen!55}\textbf{24.22} & \cellcolor{belowblue!70}13.97 & \cellcolor{belowblue!70}13.16 & \cellcolor{belowblue!70}15.27 & \cellcolor{belowblue!70}15.91 & \cellcolor{belowblue!70}20.94 & \cellcolor{belowblue!70}18.68 & \cellcolor{belowblue!55}21.98 & \cellcolor{belowblue!40}22.66\\
 & 32K & \cellcolor{heatneutral}29.57 & \cellcolor{belowblue!40}28.78 & \cellcolor{belowblue!40}\textbf{28.81} & \cellcolor{belowblue!70}7.81 & \cellcolor{belowblue!70}7.20 & \cellcolor{belowblue!70}9.11 & \cellcolor{belowblue!70}19.04 & \cellcolor{belowblue!40}28.43 & \cellcolor{belowblue!70}21.63 & \cellcolor{belowblue!70}23.25 & \cellcolor{belowblue!70}24.61\\
 & 64K & \cellcolor{heatneutral}22.66 & \cellcolor{abovegreen!25}22.68 & \cellcolor{abovegreen!70}\textbf{25.53} & \cellcolor{belowblue!70}4.26 & \cellcolor{belowblue!70}3.86 & \cellcolor{belowblue!70}5.50 & \cellcolor{belowblue!70}19.20 & \cellcolor{abovegreen!25}22.75 & \cellcolor{abovegreen!55}22.89 & \cellcolor{belowblue!70}13.54 & \cellcolor{belowblue!70}17.63\\
\midrule
\multirow{3}{*}{\ds{4}} & 16K & \cellcolor{heatneutral}11.05 & \cellcolor{belowblue!40}10.12 & \cellcolor{abovegreen!55}\textbf{11.53} & \cellcolor{belowblue!40}9.82 & \cellcolor{belowblue!55}9.28 & \cellcolor{belowblue!25}10.69 & \cellcolor{belowblue!70}7.88 & \cellcolor{belowblue!25}10.75 & \cellcolor{belowblue!25}10.53 & \cellcolor{belowblue!40}9.50 & \cellcolor{abovegreen!25}11.09\\
 & 32K & \cellcolor{heatneutral}12.03 & \cellcolor{belowblue!55}10.01 & \cellcolor{abovegreen!70}\textbf{12.67} & \cellcolor{belowblue!70}4.88 & \cellcolor{belowblue!70}4.48 & \cellcolor{belowblue!70}5.69 & \cellcolor{belowblue!70}6.72 & \cellcolor{belowblue!55}9.72 & \cellcolor{belowblue!25}12.02 & \cellcolor{belowblue!25}11.35 & \cellcolor{belowblue!70}7.47\\
 & 64K & \cellcolor{heatneutral}9.57 & \cellcolor{belowblue!55}7.33 & \cellcolor{abovegreen!70}\textbf{10.85} & \cellcolor{belowblue!70}1.12 & \cellcolor{belowblue!70}1.07 & \cellcolor{belowblue!70}1.44 & \cellcolor{belowblue!70}4.15 & \cellcolor{belowblue!25}9.42 & \cellcolor{belowblue!55}7.41 & \cellcolor{belowblue!70}4.47 & \cellcolor{belowblue!70}1.93\\
\midrule
\multirow{3}{*}{\ds{5}} & 16K & \cellcolor{heatneutral}25.27 & \cellcolor{belowblue!40}23.86 & \cellcolor{abovegreen!55}\textbf{25.64} & \cellcolor{belowblue!70}15.59 & \cellcolor{belowblue!70}14.05 & \cellcolor{belowblue!70}19.94 & \cellcolor{belowblue!70}19.28 & \cellcolor{abovegreen!55}25.53 & \cellcolor{belowblue!40}23.85 & \cellcolor{belowblue!25}24.75 & \cellcolor{belowblue!40}23.87\\
 & 32K & \cellcolor{heatneutral}25.70 & \cellcolor{belowblue!55}23.41 & \cellcolor{abovegreen!25}\textbf{25.75} & \cellcolor{belowblue!70}12.23 & \cellcolor{belowblue!70}10.95 & \cellcolor{belowblue!70}15.92 & \cellcolor{belowblue!70}18.15 & \cellcolor{belowblue!25}25.39 & \cellcolor{belowblue!55}23.00 & \cellcolor{belowblue!70}21.29 & \cellcolor{belowblue!55}23.05\\
 & 64K & \cellcolor{heatneutral}24.83 & \cellcolor{belowblue!25}24.17 & \cellcolor{abovegreen!70}\textbf{25.46} & \cellcolor{belowblue!70}7.85 & \cellcolor{belowblue!70}6.67 & \cellcolor{belowblue!70}11.24 & \cellcolor{belowblue!70}18.43 & \cellcolor{belowblue!55}22.15 & \cellcolor{belowblue!70}20.57 & \cellcolor{belowblue!70}15.59 & \cellcolor{belowblue!70}14.62\\
\midrule
\multirow{3}{*}{\ds{6}} & 16K & \cellcolor{heatneutral}19.12 & \cellcolor{belowblue!25}18.90 & \cellcolor{abovegreen!70}\textbf{20.16} & \cellcolor{belowblue!70}12.32 & \cellcolor{belowblue!70}11.39 & \cellcolor{belowblue!70}13.93 & \cellcolor{belowblue!70}15.44 & \cellcolor{abovegreen!70}19.78 & \cellcolor{belowblue!55}17.21 & \cellcolor{belowblue!40}17.64 & \cellcolor{belowblue!55}17.17\\
 & 32K & \cellcolor{heatneutral}20.11 & \cellcolor{belowblue!40}18.81 & \cellcolor{belowblue!40}\textbf{19.17} & \cellcolor{belowblue!70}9.26 & \cellcolor{belowblue!70}7.40 & \cellcolor{belowblue!70}10.54 & \cellcolor{belowblue!70}13.68 & \cellcolor{belowblue!40}18.85 & \cellcolor{belowblue!70}16.12 & \cellcolor{belowblue!70}13.58 & \cellcolor{belowblue!70}13.11\\
 & 64K & \cellcolor{heatneutral}19.09 & \cellcolor{belowblue!25}18.89 & \cellcolor{abovegreen!70}\textbf{19.65} & \cellcolor{belowblue!70}7.24 & \cellcolor{belowblue!70}5.08 & \cellcolor{belowblue!70}9.45 & \cellcolor{belowblue!70}9.74 & \cellcolor{belowblue!40}17.58 & \cellcolor{belowblue!55}17.38 & \cellcolor{belowblue!70}7.94 & \cellcolor{belowblue!70}10.00\\
\midrule
\multirow{3}{*}{\ds{7}} & 16K & \cellcolor{heatneutral}54.01 & \cellcolor{belowblue!55}51.31 & \cellcolor{abovegreen!55}\textbf{54.42} & \cellcolor{belowblue!70}38.16 & \cellcolor{belowblue!70}35.24 & \cellcolor{belowblue!70}42.12 & \cellcolor{belowblue!70}37.20 & \cellcolor{belowblue!70}51.22 & \cellcolor{belowblue!70}49.34 & \cellcolor{belowblue!55}51.59 & \cellcolor{belowblue!70}50.49\\
 & 32K & \cellcolor{heatneutral}49.88 & \cellcolor{belowblue!70}46.84 & \cellcolor{abovegreen!70}\textbf{50.74} & \cellcolor{belowblue!70}20.09 & \cellcolor{belowblue!70}21.40 & \cellcolor{belowblue!70}22.50 & \cellcolor{belowblue!70}29.38 & \cellcolor{belowblue!55}47.21 & \cellcolor{belowblue!70}44.03 & \cellcolor{belowblue!70}35.39 & \cellcolor{belowblue!70}41.88\\
 & 64K & \cellcolor{heatneutral}48.40 & \cellcolor{belowblue!70}44.82 & \cellcolor{belowblue!25}\textbf{47.94} & \cellcolor{belowblue!70}10.24 & \cellcolor{belowblue!70}10.69 & \cellcolor{belowblue!70}13.21 & \cellcolor{belowblue!70}19.63 & \cellcolor{belowblue!70}45.45 & \cellcolor{belowblue!70}45.51 & \cellcolor{belowblue!70}17.07 & \cellcolor{belowblue!70}20.17\\
\midrule
\multirow{3}{*}{\ds{8}} & 16K & \cellcolor{heatneutral}34.19 & \cellcolor{belowblue!40}33.12 & \cellcolor{belowblue!40}33.26 & \cellcolor{belowblue!70}23.28 & \cellcolor{belowblue!70}23.43 & \cellcolor{belowblue!70}25.18 & \cellcolor{belowblue!70}23.59 & \cellcolor{abovegreen!70}\textbf{36.27} & \cellcolor{belowblue!55}32.08 & \cellcolor{belowblue!40}33.39 & \cellcolor{belowblue!40}33.07\\
 & 32K & \cellcolor{heatneutral}32.95 & \cellcolor{belowblue!55}30.55 & \cellcolor{belowblue!40}31.50 & \cellcolor{belowblue!70}12.70 & \cellcolor{belowblue!70}10.28 & \cellcolor{belowblue!70}13.94 & \cellcolor{belowblue!70}20.76 & \cellcolor{abovegreen!25}\textbf{32.99} & \cellcolor{belowblue!55}30.68 & \cellcolor{belowblue!70}23.18 & \cellcolor{belowblue!70}27.54\\
 & 64K & \cellcolor{heatneutral}36.77 & \cellcolor{belowblue!70}33.60 & \cellcolor{belowblue!40}35.19 & \cellcolor{belowblue!70}9.42 & \cellcolor{belowblue!70}7.68 & \cellcolor{belowblue!70}13.37 & \cellcolor{belowblue!70}23.98 & \cellcolor{belowblue!70}33.44 & \cellcolor{belowblue!40}\textbf{35.36} & \cellcolor{belowblue!70}22.53 & \cellcolor{belowblue!70}21.71\\
\midrule
\multirow{3}{*}{\ds{9}} & 16K & \cellcolor{heatneutral}18.42 & \cellcolor{abovegreen!70}19.79 & \cellcolor{abovegreen!70}19.54 & \cellcolor{belowblue!55}15.99 & \cellcolor{belowblue!70}12.64 & \cellcolor{belowblue!25}18.19 & \cellcolor{belowblue!55}16.02 & \cellcolor{abovegreen!70}19.56 & \cellcolor{abovegreen!70}\textbf{20.56} & \cellcolor{abovegreen!70}19.37 & \cellcolor{abovegreen!70}19.26\\
 & 32K & \cellcolor{heatneutral}12.81 & \cellcolor{belowblue!40}11.86 & \cellcolor{abovegreen!70}14.53 & \cellcolor{belowblue!40}11.33 & \cellcolor{belowblue!55}10.25 & \cellcolor{abovegreen!70}\textbf{14.57} & \cellcolor{belowblue!70}7.43 & \cellcolor{abovegreen!70}13.61 & \cellcolor{belowblue!55}10.87 & \cellcolor{abovegreen!70}14.31 & \cellcolor{belowblue!25}12.67\\
 & 64K & \cellcolor{heatneutral}11.59 & \cellcolor{belowblue!40}10.64 & \cellcolor{abovegreen!70}\textbf{12.23} & \cellcolor{belowblue!70}3.10 & \cellcolor{belowblue!70}4.02 & \cellcolor{belowblue!70}5.79 & \cellcolor{belowblue!70}6.66 & \cellcolor{belowblue!55}9.34 & \cellcolor{belowblue!25}11.29 & \cellcolor{belowblue!70}4.97 & \cellcolor{belowblue!70}8.64\\
\bottomrule
\end{tabular}
\end{adjustbox}
\end{table*}

\noindent\textbf{Configuration and Comparisons.}
\modelname{} uses $64\times64$ attention tiles.
For the full method, the target mean normalized computation cost is 0.80.
BF16 provides the original-model reference with scaled dot-product attention (SDPA), while Torch attention uses the math backend.
The low-bit reference, \texttt{all\_int4}, retains all causal tiles and uses INT4 QK outside the protected first and diagonal tiles.
We compare against MInference, FlexPrefill, SageAttention2, MoA, and SpargeAttn~\citep{jiang2024minference,lai2025flexprefill,zhang2024sageattention2,fu2024moa,zhang2025spargeattention}.
Layer-shared, model-shared, and randomized boundary assignments test allocation granularity and sensitivity-guided placement. Appendix~\ref{app:quality-protocols} provides model configurations, task settings, and allocation-control definitions.

\noindent\textbf{Hardware and Measurement.}
Primary efficiency measurements use an NVIDIA A100-SXM4 40\,GB GPU, with additional system evaluations on H100 NVL and RTX 4090 GPUs.
We report time per output token (TPOT), complete-request generation throughput, peak GPU memory, and energy per 1,000 generated tokens.
The workload sweeps vary prompt length, output length, and batch size, with complete-request measurements covering both prefill and decode. Timing, energy, and memory-measurement protocols are specified in Appendix~\ref{app:measurement}.

\subsection{Inference Efficiency}
\label{sec:efficiency}

We first examine whether the selected attention representation improves complete generation requests.
On Llama with a 64K prompt, \modelname{} reaches 22.80 tokens/s, compared with 14.06 for BF16 and 17.39 for FlexPrefill, yielding $1.62\times$ and $1.31\times$ throughput, respectively (Table~\ref{tab:efficiency}).
The improvement accompanies a reduction in TPOT from 26.64 to 10.06\,ms relative to BF16.
Peak allocation decreases from 15.65 to 11.76\,GiB, and energy per generated token falls by 36.5\%.
These measurements establish that the attention implementation's benefits extend to inference with the rest of the model included.

The comparison with FlexPrefill highlights the importance of carrying attention savings into generation.
Its TPOT is 26.79\,ms, compared with 10.06\,ms for \modelname{}, alongside the complete-request throughput improvement.
Figure~\ref{fig:efficiency-grid-main} evaluates this combination across different batch sizes and input/output lengths.
Prompt processing occurs once per request, while each generated token reuses the compressed prompt cache.
The workload sweep therefore evaluates how the prefill and decode paths contribute as the balance between prompt processing and generation changes. Appendix~\ref{app:workload-scaling} extends the throughput and TPOT comparisons across models, request shapes, and GPU platforms.

\findingbox{%
Joint retention and precision improve complete-request throughput while reducing token-generation latency, peak memory, and energy.
}

\subsection{General-Purpose Performance}
\label{sec:general-quality}

Having established inference gains, we examine whether the same boundary table preserves general prediction quality across tasks. The table selected through summary and retrieval calibration is reused across the benchmarks, testing whether its retention and precision choices support broader model capabilities. On Llama, \modelname{} has word perplexity 12.50, compared with 12.49 for BF16 (Table~\ref{tab:general-llama}). All six accuracy scores exceed the BF16 reference by less than 0.7 percentage points, with MMLU increasing from 60.66\% to 61.32\%. Near-reference quality therefore extends across both language modeling and the evaluated understanding tasks under one allocation.

The full-context \texttt{all\_int4} reference provides a complementary comparison by retaining all causal tiles and applying 4-bit QK outside the protected regions. Its perplexity is 12.59 and MMLU accuracy is 60.26\%, while \modelname{} achieves lower perplexity and higher accuracy on all six tasks. These results support jointly selecting retained context and score precision to preserve predictions alongside the measured inference gains. The following long-context QA evaluation examines whether this quality extends from prefill predictions to answer generation with repeated access to the compressed prompt cache.

\findingbox{%
One model-level retention-and-precision policy preserves general-task quality across benchmarks while supporting efficient inference.
}

\subsection{Long-Context Question Answering}
\label{sec:model-performance}

Long-context QA evaluates whether the context retained during prefill remains useful throughout answer generation. Table~\ref{tab:quality} reports nine tasks at the 16K, 32K, and 64K dataset levels using the same calibrated boundary table, alongside controls for boundary sharing and placement. Dataset identifiers are \ds{1} \texttt{cmrc\_mixup}, \ds{2} \texttt{dureader\_mixup}, \ds{3} \texttt{hotpotwikiqa\_mixup}, \ds{4} \texttt{lic\_mixup}, \ds{5} \texttt{loogle\_CR\_mixup}, \ds{6} \texttt{loogle\_MIR\_mixup}, \ds{7} \texttt{loogle\_SD\_mixup}, \ds{8} \texttt{multifieldqa\_en\_mixup}, and \ds{9} \texttt{multifieldqa\_zh\_mixup}.

LooGLE-SD shows how this policy preserves answer quality as context grows. \modelname{} scores 54.42, 50.74, and 47.94 at 16K, 32K, and 64K, respectively, following BF16's 54.01, 49.88, and 48.40 with differences below one point at every level. At 64K, its score also exceeds 44.82 for \texttt{all\_int4} and 45.45 for SpargeAttn. The comparison with \texttt{all\_int4}, which retains every causal tile, supports selecting score precision together with retained context. Selective mixed-precision attention preserves stronger answer quality here than retaining the full context with uniformly low QK precision outside the protected regions.

Tasks requiring multiple pieces of evidence provide a complementary check. At 64K, \modelname{} scores 19.65 on LooGLE-MIR, compared with 19.09 for BF16 and 17.58 for SpargeAttn. On LooGLE-CR, the corresponding scores are 25.46, 24.83, and 22.15. HotpotWikiQA also reaches 25.53, compared with 22.66 for BF16 and 22.75 for SpargeAttn. These comparisons extend quality preservation across different QA tasks under one model-level allocation. They assess the complete prefill-and-decode path: the retained context must support generated answers after being packed into the compressed prompt cache.

The allocation controls examine how this behavior depends on the freedom to assign different boundaries. Layer Same shares one boundary pair across the query heads of each layer, while All Same uses a common 32K 8-bit window throughout the model, with the first block protected. On LooGLE-SD at 64K, their scores are 45.51 and 17.07, compared with 47.94 for \modelname{}. On LooGLE-CR, the corresponding scores are 20.57, 15.59, and 25.46. Sharing boundaries couples the retained ranges and precision choices of heads with different sensitivities. Head-specific boundaries let the allocator preserve wider ranges or higher precision where they most reduce projected-output error, supporting stronger QA performance in these comparisons.

Random then tests the value of matching the selected configurations to head sensitivity. It permutes the deployed boundary pairs across layers and heads, preserving both their collection and the aggregate modeled computation cost. On LooGLE-SD, its gap from \modelname{} widens from 3.93 points at 16K to 27.77 points at 64K, where Random scores 20.17 against 47.94. LooGLE-MIR shows a corresponding separation at 64K, with 10.00 for Random and 19.65 for \modelname{}. The growing separation links long-context quality to which heads receive the wider ranges and higher precision. Sensitivity-guided placement therefore makes the selected computation more effective for QA within the same modeled budget.

\findingbox{%
The selected ranges preserve useful context throughout answer generation, while sensitivity-guided placement achieves stronger long-context QA than randomized assignment at the same modeled computation cost.
}

\subsection{Ablation and Analysis}
\label{sec:ablation}

The preceding QA controls examine the placement and sharing of boundary configurations. We now examine the computation choices available to the allocator: whether combining skipping with two retained precision levels reduces the prefill time needed to meet the same quality-selection requirements.

\begingroup
\clubpenalty=10000
\widowpenalty=10000
\textls[-10]{\textbf{Computation-level controls.}
Without 4-bit imposes $w_8=w_4$, leaving each non-protected region a choice between 8-bit QK and skipping. Without Skip imposes $w_4=\infty$, retaining the full causal context while assigning its 4-bit and 8-bit regions. Each control is independently reallocated and selected under the same quality-selection thresholds, allowing its remaining choices to adapt to the validation requirements. Table~\ref{tab:three_level_ablation} compares the resulting prefill attention latency on A100 across three models and contexts from 32K to 128K.}\looseness=-1\par
\endgroup

\par\addvspace{2pt}
\noindent\begin{minipage}{\columnwidth}
\setlength{\abovecaptionskip}{0pt}
\centering
\captionof{table}{Effect of three-level computation on A100 prefill attention latency. Latency includes Q/K preparation, V packing, and attention across all layers. Ablations are independently reallocated under the same quality-selection thresholds. Lower is better.}
\label{tab:three_level_ablation}
\vspace{2pt}
\fontsize{8}{9.5}\selectfont
\setlength{\tabcolsep}{2pt}
\renewcommand{\arraystretch}{1}
\begin{tabular*}{\columnwidth}{@{\extracolsep{\fill}}llccc@{}}
\toprule
\textbf{Model} & \textbf{Context} &
\shortstack{\textbf{\modelname{}}\\\textbf{(s)}} &
\shortstack{\textbf{Without}\\\textbf{4-bit (s)}} &
\shortstack{\textbf{Without}\\\textbf{Skip (s)}} \\
\midrule
\multirow{3}{*}{Llama}
& 32K  & \textbf{0.654} & 0.660  & 0.682 \\
& 64K  & \textbf{2.242} & 2.259  & 2.647 \\
& 128K & \textbf{7.347} & 7.568  & 10.324 \\
\midrule
\multirow{3}{*}{Qwen2}
& 32K  & \textbf{0.781}  & 0.810  & 0.845 \\
& 64K  & \textbf{2.752}  & 2.924  & 3.232 \\
& 128K & \textbf{10.134} & 11.151 & 13.010 \\
\midrule
\multirow{3}{*}{Ministral}
& 32K  & \textbf{1.066}  & 1.106  & 1.086 \\
& 64K  & \textbf{3.856}  & 4.078  & 4.182 \\
& 128K & \textbf{14.266} & 15.449 & 16.402 \\
\bottomrule
\end{tabular*}
\end{minipage}
\par\nopagebreak[2]

\begingroup
\clubpenalty=10000
\textbf{Lowering the cost of retained context.}
\modelname{} achieves the lowest latency in all nine settings. At 128K, the full design reduces Qwen2 latency from 11.151\,s without 4-bit to 10.134\,s, and Ministral latency from 15.449 to 14.266\,s, reductions of 9.1\% and 7.7\%, respectively. Without the intermediate level, retaining an interaction also commits it to 8-bit QK computation. The 4-bit region separates these choices, allowing lower-cost retention alongside an 8-bit region for more sensitive interactions. Since the measurements include Q/K preparation and V packing across all layers, these reductions reflect the benefit over complete attention calls after preparing the mixed-precision operands.\par
\endgroup

\textbf{Removing work as context grows.}
Skipping reduces latency by eliminating QK, normalization, and PV for omitted tiles. For Llama, the reduction relative to Without Skip increases from 4.1\% at 32K to 28.8\% at 128K, where latency falls from 10.324 to 7.347\,s. Qwen2 shows the same length-dependent pattern, with reductions increasing from 7.6\% to 22.1\%; the relative benefit also grows across the three lengths for Ministral. Full retention adds interactions throughout the causal map as context expands. Allowing selected distance regions to be skipped removes their complete tile cost, complementing the lower QK cost of interactions that remain.

\textbf{Adapting the retention--precision trade-off.}
The faster restricted configuration also changes with the workload. For Ministral at 32K, Without Skip takes 1.086\,s and Without 4-bit takes 1.106\,s. At 128K, their ordering reverses, with 16.402 and 15.449\,s, respectively, while the full design remains fastest at both lengths. Thus, the relative efficiency of retaining all context at mixed precision and retaining selected context at 8 bits varies even within one model. Independently adjustable boundaries let the allocator combine these alternatives. The measured advantage of the full design supports jointly choosing how far attention extends and which precision is used within that range.

\findingbox{%
\textls[20]{Three computation levels reduce both the cost of retained interactions and the number of executed tiles, lowering prefill latency under common quality-selection requirements.}
}

\section{Conclusion}
\label{sec:conclusion}

\modelname{} demonstrates that useful attention interactions can be preserved at lower cost by jointly allocating retention and QK precision according to regional sensitivity.
Compact distance boundaries carry these choices through fused prefill and compressed-cache decoding, yielding faster generation while preserving general-task and long-context quality.
Allocation controls further show that computation placement affects long-context quality at a fixed modeled cost.
These findings support a broader principle for efficient attention: evaluate which interactions to retain together with the precision they require, and make that joint policy govern both GPU computation and repeated context access.

\bibliography{reference}
\bibliographystyle{triattention}

\clearpage
\chead{\small\bfseries\modelname{}: Supplementary Material}
\appendix
\renewcommand{\dbltopfraction}{0.95}
\renewcommand{\dblfloatpagefraction}{0.80}
\renewcommand{\textfraction}{0.05}
\setcounter{dbltopnumber}{3}
\makeatletter
\setlength{\@dblfpsep}{1pt plus 1fil}
\makeatother

\section{Calibration and Implementation Details}
\label{app:implementation}

The boundary table connects offline sensitivity measurements to the operands used during inference. This appendix details how calibration selects that table, how low-bit products recover attention scores and outputs, and how execution addresses and merges the resulting representations.

\subsection{Calibration and Boundary-Table Selection}
\label{app:calibration}

\textbf{Calibration inputs and scored positions.}
We combine document summarization and archive-record retrieval to evaluate both distributed context use and access to specific evidence. The 16 calibration prompts span 16K, 32K, 64K, and 128K contexts. Summary prompts use Multi-News~\citep{fabbri2019multi} documents and reference summaries capped at 256 response tokens. Retrieval prompts place a unique numeric record at document depths 0.2, 0.5, or 0.8 and request its value at the end. Each prompt uses the model's chat template and reference response, with a 129,024-token budget for the 128K construction to accommodate both. Calibration and selection use separate documents, disjoint from LV-Eval and LongEval.

We score the final 1,024 positions, covering the prompt ending and teacher-forced response, so the evaluated queries attend over the long input. A dense BF16 pass supplies common Q/K/V for all candidates and propagates the reference output to the next layer. Candidate errors therefore reflect their retention and precision choices under the same layer input. Equation~\ref{eq:ta-error} normalizes each prompt's projected-output error before averaging, giving every prompt equal weight.

\textbf{Candidate ranges and computation costs.}
Geometrically spaced boundaries cover local through full-context configurations with a compact candidate set. Measured in 64-token blocks, the grid is \[
\begin{aligned}
\mathcal W_8 &= \{1,2,4,\ldots,512,\infty\},\\
\mathcal W_4 &= \{1,2,4,\ldots,2048,\infty\},\\
\mathcal C &= \{(w_8,w_4):w_8\in\mathcal W_8,\;\\
           &\qquad w_4\in\mathcal W_4,\;w_8\leq w_4\},
\end{aligned}
\] These 86 pairs jointly vary retained context and QK precision. Setting $w_4=\infty$ retains the full causal context, while $w_8=w_4$ assigns 8-bit QK to retained non-protected tiles.

The allocator evaluates non-protected tile costs at 65,536 tokens using the measured complete-tile ratio $\gamma=0.905$. For $N_H$ layer--query-head pairs, a target mean cost $\tau$ gives the total budget $B=N_H\tau$ in Equation~\ref{eq:ta-allocation}. We evaluate $\lambda=0$ and 400 logarithmically spaced positive values from $10^{-9}$ to $10^1$, retaining the feasible sampled table with the lowest summed calibration error for each target budget. The deployed mean-cost targets are 0.80 for Llama-3.2-3B-Instruct and 0.85 for Qwen2-7B-Instruct and Ministral-3-8B-Instruct-2512-BF16.

\textbf{Complete-model selection.}
The assembled boundary tables are evaluated on separate summary and retrieval prompts. KL divergence measures changes in the BF16 next-token distribution, while response NLL measures the likelihood assigned to the reference answers. We average each metric over response tokens within a prompt, then over prompts in the 16K/32K, 64K, and 128K groups. Each group must satisfy mean KL at most 0.002 and mean NLL increase at most 0.01 nats. This applies the quality requirements separately across context lengths before selecting the lowest-cost qualifying table for inference.

\subsection{Operand Quantization and Shared Normalization}
\label{app:numerics}

\textbf{Queries and keys in common score coordinates.}
Key centering removes the prompt-wide key offset, and a shared orthogonal transform redistributes channel magnitudes before quantization~\citep{zhang2024sageattention2,shah2024flashattention,zhang2026sageattention3}. These operations prepare the same Q/K inputs for both precision paths. After rotary position embedding, let $\mu_K$ be the mean prompt key vector for one KV head and $U$ the fixed orthogonal signed Hadamard transform. The prepared operands are $\widetilde Q=QU$ and $\widetilde K=(K-\mathbf 1\mu_K^{\top})U$. Since $UU^{\top}=I$, \begin{equation}
\widetilde Q\widetilde K^{\top}
=QK^{\top}-(Q\mu_K)\mathbf 1^{\top}.
\label{eq:app-key-coordinates}
\end{equation} Centering subtracts the same quantity from every legal score in a query row and therefore preserves its softmax before quantization. The orthogonal transform redistributes channel magnitudes without changing the centered dot products.

For bit-width $r\in\{4,8\}$, symmetric quantization uses $a_r=2^{r-1}-1$ and absmax scale $s_X^{(r)}=\max|X|/a_r$ per quantization group~\citep{zhang2025sageattention}. Codes round $X/s_X^{(r)}$ to the nearest integer, with zero-valued groups producing zero codes. Queries use one scale per token and query head; keys use one scale per KV head and group of 16 consecutive tokens across their channels. The 4-bit and 8-bit codes represent the same transformed operands with separate scales. For query-token index $t$ and key-token index $k$, the restored score is \begin{equation}
\widetilde s_{tk}^{(r)}
=\frac{s_{Q,t}^{(r)}s_{K,g(k)}^{(r)}}{\sqrt d}
\left\langle\widehat Q_t^{(r)},\widehat K_k^{(r)}\right\rangle,
\label{eq:app-score-restoration}
\end{equation} where $g(k)$ is key $k$'s 16-token group. Restored scores at both precisions enter the same probability computation.

\textbf{Values in shared reconstruction coordinates.}
Sharing channel statistics across value formats lets the kernel accumulate their contributions together and reconstruct the output once. We center and scale each prompt value channel to prepare the FP16, INT8, and INT4 representations~\citep{zhang2024sageattention2}. For each KV head and channel $c$, preparation computes the prompt mean $\mu_{V,c}$ and \[
\sigma_{V,c}=\frac{\max_k|V_{k,c}-\mu_{V,c}|}{127}.
\] A constant channel uses scale one. The normalized values are $V'=(V-\mu_V)\oslash\sigma_V$. Protected prefill tiles store $V'$ in FP16, while other retained prefill tiles use the rounded INT8 codes $\widehat V_8$. For distant decode tiles, $\widehat V_4=\operatorname{round}(7V'/127)$ is stored in signed 4-bit form and reconstructed in the same normalized coordinates as $(127/7)\widehat V_4$.

\textbf{Probability coding and accumulation.}
Tile-local scaling gives each tile its own probability-coding range, while a separate factor preserves its weight relative to other tiles. For one query row, we encode the restored scores in base-two units and merge their contributions into the shared attention state. For one query row and the current key tile, let $s_k$ be the restored score multiplied by $\log_2 e$, and let $b=\max_{k\in\mathcal K}s_k$, where $\mathcal K$ contains this row's legal keys in the tile.

Using the row's query scale and the four key-group scales at QK precision $r$, the largest score multiplier is
\begin{equation}
\eta_{\max}
=\frac{\log_2 e}{\sqrt d}\,s_Q^{(r)}
\max_{u\in\{0,1,2,3\}}s_{K,u}^{(r)}.
\end{equation}
We use the probability-coding base
\begin{equation}
\begin{aligned}
\varepsilon_{\mathrm{tile}}
&=0.75\,\eta_{\max}+2.4\times10^{-7}|b|,\\
b'&=b+\varepsilon_{\mathrm{tile}}.
\end{aligned}
\label{eq:app-probability-margin}
\end{equation}
The shared base provides rounding headroom while keeping all four key groups on the same probability scale.

The probability codes are
\[
\widehat p_k=
\left\lfloor
\operatorname{round}_{1/256}
\left(a_P\,2^{s_k-b'}+\rho_k\right)
\right\rfloor,
\]
where $\operatorname{round}_{1/256}$ denotes rounding to the nearest multiple of $1/256$ in the FP32 packing step. Prefill uses $\rho_k=1/2$. Decode uses $\rho_k=(D_k\bmod256)/256$, where $D_k$ is the integer QK dot product for key $k$. The maximum code is $a_P=255$ for 8-bit PV and $a_P=15$ for 4-bit PV. Masked keys receive zero probability.

Let $z_k$ be the stored value expressed in $V'$ coordinates. The factors $\alpha$ and $\beta$ align the accumulated state and the current tile to the updated maximum~\citep{dao2022flashattention}: \begin{equation}
\begin{aligned}
m'&=\max(m,b),\\
\alpha&=2^{m-m'},\qquad \beta=2^{b'-m'}/a_P,\\
l'&=\alpha l+\beta\sum_k\widehat p_k,\\
o'&=\alpha o+\beta\sum_k\widehat p_k z_k.
\end{aligned}
\label{eq:app-integer-accumulation}
\end{equation} The same quantized weights update the denominator and weighted value sum. Protected tiles contribute through FP16 PV into the same FP32 attention state. After the retained tiles have been processed, value restoration gives \begin{equation}
O=\mu_V+\sigma_V\odot(o/l).
\label{eq:app-value-restoration}
\end{equation}

\subsection{Compact Traversal and Compressed-Cache Execution}
\label{app:execution}

\textbf{From boundaries to local work.}
The contiguous distance ranges let each query block construct its traversal from the two stored boundaries. Processing the protected, 8-bit QK, and 4-bit QK groups separately reuses the corresponding query representation while successive K/V tiles are staged. On RTX 4090, direct index calculation and reduced value staging lower the per-thread-block shared-memory footprint. Across thread blocks, KV-group-major scheduling places heads sharing K/V together and visits query blocks in descending order to group accesses to common operands.

\textbf{Mapping logical retention to physical storage.}
GQA query heads share physical K/V, so cache construction combines their requested prompt tiles into $\mathcal R_g=\bigcup_{h\in\mathcal H_g}\mathcal R_h$, where $\mathcal R_h$ is head $h$'s retained set at the final prompt query block. Each tile is stored once at the highest requested key precision. A head mask preserves the participating query heads, and per-KV-head offsets locate the packed tiles in contiguous buffers. This shares stored operands while retaining each head's selected context.

Values use INT8 for the first and most recent 16 prompt blocks and packed INT4 elsewhere in the retained set. At head dimension 128, a 64-token INT4 K or V tile uses 4\,KiB, half the 8\,KiB required by INT8, before scales and metadata. The packed prompt cache is reused throughout generation, while new K/V enter the BF16 tail.

\textbf{Combining prompt and generated context.}
The prompt numerator is accumulated in normalized $V'$ coordinates, whereas the BF16 tail numerator uses original values. They must be reconciled before the final normalization. Let $(m_{\mathrm p},l_{\mathrm p},o_{\mathrm p})$ be the prompt state obtained by aligning and merging the thread-block partials, and let $(m_{\mathrm t},l_{\mathrm t},o_{\mathrm t})$ be the tail state. Tail scores use the prompt's key-centering offset. With $m=\max(m_{\mathrm p},m_{\mathrm t})$, the final output is \begin{equation}
O=
\frac{
\begin{gathered}
2^{m_{\mathrm p}-m}\bigl(\sigma_V\odot o_{\mathrm p}+\mu_V l_{\mathrm p}\bigr)\\
{}+2^{m_{\mathrm t}-m}o_{\mathrm t}
\end{gathered}
}{
2^{m_{\mathrm p}-m}l_{\mathrm p}+2^{m_{\mathrm t}-m}l_{\mathrm t}
}.
\label{eq:app-prompt-tail-merge}
\end{equation} The term $\sigma_V\odot o_{\mathrm p}+\mu_V l_{\mathrm p}$ restores the unnormalized prompt numerator; the exponential factors align prompt and tail maxima before combining numerators and denominators.

\section{Experimental Protocols}
\label{app:protocols}

\subsection{Models, Tasks, and Comparison Settings}
\label{app:quality-protocols}

The evaluated checkpoints are Llama-3.2-3B-Instruct, Qwen2-7B-Instruct, and Ministral-3-8B-Instruct-2512-BF16. Their respective numbers of query heads, KV heads, and decoder layers are $(24,8,28)$, $(28,4,28)$, and $(32,8,34)$, with head dimension 128. The original projection weights are retained, and the attention implementation consumes their Q/K/V outputs.

\textbf{General quality.}
General evaluation uses the Language Model Evaluation Harness~\citep{eval-harness} with the recorded task configurations and dataset snapshots. MMLU uses five-shot prompts and subject-macro accuracy. These likelihood-based evaluations run the prefill attention path with the model's native positional configuration.

\textbf{Long-context QA.}
LV-Eval~\citep{yuan2024lv} uses the official dataset prompts and task-specific scorers at the 16K, 32K, and 64K dataset levels, with batch size one, greedy generation, and each task's configured response limit. Each prompt is processed in a single prefill pass before answer generation. Scores are reported separately for each dataset--length pair on a 0--100 scale, using keyword-aware F1 or DuReader ROUGE-L as specified by the benchmark. The length levels identify the supplied datasets; actual tokenized lengths follow the model's tokenizer.

\textbf{References and allocation controls.}
BF16 uses default PyTorch SDPA dispatch, while Torch attention restricts SDPA to its math backend. The \texttt{all\_int4} reference retains all causal tiles and changes non-protected QK computation to INT4; the protected first and diagonal tiles retain INT8 QK and FP16 PV. MoA loads FP16 model weights, while the other quality comparisons use BF16 weights. The H100 quality environment uses Python 3.12, PyTorch 2.10 with CUDA 12.6, and Transformers 5.18.

Layer Same selects one boundary pair per layer by aggregating its heads' candidate errors and sharing the resulting pair across those heads. All Same assigns $(w_8,w_4)=(512,512)$ throughout the model, corresponding to a 32K local 8-bit range with the first block protected. Random uses seed zero to permute the deployed boundary pairs across layers and heads, preserving the configuration collection and its aggregate modeled prefill cost.

\subsection{Measurement and Ablation Protocols}
\label{app:measurement}

\textbf{Complete-request measurements.}
System evaluation uses an NVIDIA A100-SXM4 40\,GB GPU, H100 NVL, and GeForce RTX 4090. Within each model, comparisons reuse the same stored input token IDs across methods and GPUs. The core grid varies prompt length over $\{128,2048,8192,16384,32768,65536,131072\}$, batch size over $\{1,2,4,8\}$, and output length over $\{128,256,512,1024\}$. The figures display the indicated prompt groups, with out-of-memory outcomes marked separately. Qwen2 system runs apply fourfold YaRN scaling~\citep{peng2024yarn} across methods. For prompts of at least 64K tokens, feed-forward computation is processed in 16K-position chunks across methods.

Two warm-up runs precede five timed repetitions. Each repetition generates 1,024 tokens greedily, with synchronized checkpoints at the reported output lengths. For a checkpoint containing $L_{\mathrm{out}}$ generated tokens per request, TPOT is the elapsed time from the first token to that checkpoint divided by $L_{\mathrm{out}}-1$. Complete-request throughput divides the batch's generated-token count by elapsed time from request start, including prefill. Latency, throughput, and energy use medians across repetitions, while peak GPU allocation uses the maximum observed allocation. NVIDIA Management Library (NVML) measures GPU energy during generation, reported in joules per 1,000 generated tokens.

Before timing, CUDA graphs capture decode and prefill with prompts up to 16K tokens. Timed requests include operand preparation, attention, and prompt-cache construction. Baselines use released generation interfaces.

\textbf{Attention-level timing and storage.}
The efficiency ablations use seeded BF16 Q/K/V tensors with each model's head geometry and per-layer boundary table. CUDA events time complete attention calls, and per-layer median times are summed over the indicated layers. Prefill timing includes Q/K preparation, V packing, and attention computation; decode timing measures one attention step after prompt-cache construction. The PV comparison uses the first, middle, and last decoder layers. Prompt-cache storage counts retained K/V and associated metadata after prefill, excluding model weights, generated-token storage, and temporary workspace.

\section{Design Validation}
\label{app:design-validation}
\raggedbottom
\clubpenalties 2 10000 0
\widowpenalties 2 10000 0
\makeatletter
\let\designsavedparagraph\paragraph
\renewcommand{\paragraph}{\@startsection{paragraph}{4}{\z@}{5pt}{-1em}{\normalfont\normalsize\bfseries}}
\makeatother

Head-specific allocation determines where attention computation is retained. Its execution cost also depends on the available precision levels, operand-staging space, and the representation accessed at every decode step.

\subsection{Value-Aggregation Precision}
\label{app:three-level-pv}
\vspace{-3pt}

\textbf{Value-aggregation precision.}
Retained tiles still require PV computation, making its precision another source of execution cost. With retained tiles and 8-bit QK fixed, Table~\ref{tab:pv_precision_ablation} changes non-protected PV from 16 to 8 bits over the first, middle, and last layers. The 8-bit path is faster in all settings, reducing Ministral's 128K latency from 882.1 to 762.2\,ms. Lower-bit value aggregation therefore complements the savings from retention and QK precision.

\par\addvspace{2pt}
\noindent\begin{minipage}{\columnwidth}
\setlength{\abovecaptionskip}{0pt}
\centering
\captionof{table}{Effect of value-aggregation precision on A100 prefill latency. The comparison covers the first, middle, and last attention layers with identical retained tiles and 8-bit QK computation. Non-protected tiles use either 8-bit or 16-bit PV. Lower is better.}
\label{tab:pv_precision_ablation}
\vspace{2pt}
\fontsize{8}{9.5}\selectfont
\setlength{\tabcolsep}{2pt}
\renewcommand{\arraystretch}{1}
\begin{tabular*}{\columnwidth}{@{\extracolsep{\fill}}llcc@{}}
\toprule
\textbf{Model} & \textbf{Context} &
\shortstack{\textbf{8-bit PV}\\\textbf{(ms)}} &
\shortstack{\textbf{16-bit PV}\\\textbf{(ms)}} \\
\midrule
\multirow{3}{*}{Llama}
& 32K  & \textbf{72.9}  & 79.2 \\
& 64K  & \textbf{251.7} & 275.3 \\
& 128K & \textbf{865.0} & 992.7 \\
\midrule
\multirow{3}{*}{Qwen2}
& 32K  & \textbf{77.1}  & 83.6 \\
& 64K  & \textbf{269.4} & 295.1 \\
& 128K & \textbf{985.2} & 1,089.4 \\
\midrule
\multirow{3}{*}{Ministral}
& 32K  & \textbf{64.5}  & 69.7 \\
& 64K  & \textbf{216.0} & 236.1 \\
& 128K & \textbf{762.2} & 882.1 \\
\bottomrule
\end{tabular*}
\end{minipage}
\par

\findingbox{%
Low-bit PV reduces value-aggregation cost with retained tiles and QK precision fixed.
}

\subsection{Compact On-Chip Execution and Scheduling}
\label{app:on-chip-execution}
\vspace{-3pt}

\par\addvspace{2pt}
\noindent\begin{minipage}{\columnwidth}
\setlength{\abovecaptionskip}{0pt}
\centering
\captionof{table}{Shared-memory footprint and prefill execution on RTX 4090. Direct boundary-based indexing and reduced value staging decrease per-CTA shared-memory usage, allowing more concurrent CTAs per SM. Both layouts preserve equivalent attention computation. Lower latency is better. CTA denotes a GPU thread block, and SM denotes a streaming multiprocessor.}
\label{tab:shared_memory_ablation}
\vspace{2pt}
\fontsize{8}{9.5}\selectfont
\setlength{\tabcolsep}{2pt}
\renewcommand{\arraystretch}{1}
\begin{minipage}[t]{\linewidth}
\centering
\textbf{(a) On-chip resource usage}\\[5pt]
\begin{tabular*}{\columnwidth}{@{\extracolsep{\fill}}lcc@{}}
\toprule
\textbf{Metric} &
\textbf{Original} &
\textbf{Compact} \\
\midrule
Shared memory/CTA (KiB) & 37 & \textbf{30} \\
Resident CTAs/SM & 2 & \textbf{3} \\
Resident warps/SM & 8 & \textbf{12} \\
\bottomrule
\end{tabular*}
\end{minipage}
\par\vspace{6pt}
\begin{minipage}[t]{\linewidth}
\centering
\textbf{(b) Prefill attention latency}\\[5pt]
\begin{tabular*}{\columnwidth}{@{\extracolsep{\fill}}llcc@{}}
\toprule
\textbf{Model} &
\textbf{Context} &
\textbf{Original (s)} &
\textbf{Compact (s)} \\
\midrule
\multirow{2}{*}{Llama}
& 32K  & 0.505 & \textbf{0.466} \\
& 128K & 5.609 & \textbf{5.149} \\
\midrule
\multirow{2}{*}{Qwen2}
& 32K  & 0.588 & \textbf{0.546} \\
& 128K & 7.565 & \textbf{7.036} \\
\midrule
\multirow{2}{*}{Ministral}
& 32K  & 0.810  & \textbf{0.743} \\
& 128K & 10.725 & \textbf{9.813} \\
\bottomrule
\end{tabular*}
\end{minipage}
\end{minipage}
\par

\textbf{Shared-memory organization.}
Traversal metadata and staged K/V share the thread block's limited on-chip storage, so reducing their combined footprint allows more blocks to reside on an SM. On RTX 4090, direct boundary-based indexing together with reduced value staging lowers shared memory per thread block from 37 to 30\,KiB (Table~\ref{tab:shared_memory_ablation}). The resource reduction changes how many thread blocks fit on an SM: a 7\,KiB decrease per block raises residency from two blocks to three. At 128K, the compact layout reduces prefill latency by 8.2\%, 7.0\%, and 8.5\% for Llama, Qwen2, and Ministral, respectively; Ministral latency falls from 10.725 to 9.813\,s. This links traversal metadata and value staging through their shared resource budget: reducing their combined footprint enables a different execution occupancy for the same attention computation.

\textbf{Ordering work that shares K/V.}
Query heads sharing a KV head repeatedly access common operands, motivating KV-group-major scheduling. With the selected tiles fixed and outputs bitwise identical, this ordering reduces 128K prefill latency from 7.803 to 7.347\,s for Llama and from 15.092 to 14.266\,s for Ministral (Table~\ref{tab:kv_group_scheduling}).

\par\addvspace{2pt}
\noindent\begin{minipage}{\columnwidth}
\setlength{\abovecaptionskip}{0pt}
\centering
\captionof{table}{Effect of KV-group-aware CTA scheduling on A100 prefill attention latency. Both schedules execute the same selected tiles and arithmetic operations, producing bitwise-identical outputs. KV-group-major scheduling places query heads sharing a KV head together to improve K/V data reuse. Lower is better.}
\label{tab:kv_group_scheduling}
\vspace{2pt}
\fontsize{8}{9.5}\selectfont
\setlength{\tabcolsep}{2pt}
\renewcommand{\arraystretch}{1}
\begin{tabular*}{\columnwidth}{@{\extracolsep{\fill}}llcc@{}}
\toprule
\textbf{Model} & \textbf{Context} &
\shortstack{\textbf{KV-Group-Major}\\\textbf{(s)}} &
\shortstack{\textbf{Heaviest-First}\\\textbf{(s)}} \\
\midrule
\multirow{3}{*}{Llama}
& 32K  & \textbf{0.654} & 0.681 \\
& 64K  & \textbf{2.242} & 2.359 \\
& 128K & \textbf{7.347} & 7.803 \\
\midrule
\multirow{3}{*}{Qwen2}
& 32K  & \textbf{0.781}  & 0.782 \\
& 64K  & \textbf{2.752}  & 2.790 \\
& 128K & \textbf{10.134} & 10.339 \\
\midrule
\multirow{3}{*}{Ministral}
& 32K  & \textbf{1.066}  & 1.102 \\
& 64K  & \textbf{3.856}  & 4.053 \\
& 128K & \textbf{14.266} & 15.092 \\
\bottomrule
\end{tabular*}
\end{minipage}
\par

\findingbox{%
Reducing the combined footprint of traversal metadata and operand staging increases thread-block residency and accelerates the same computation.
}

\subsection{Prompt-Cache Storage and Decode}
\label{app:cache-validation}
\vspace{-3pt}

\par\addvspace{2pt}
\noindent\begin{minipage}{\columnwidth}
\setlength{\abovecaptionskip}{0pt}
\centering
\captionof{table}{Prompt KV-cache storage after prefill. The full-retention compressed cache stores all prompt tiles with 8-bit keys and mixed-bit values. \modelname{} additionally applies its selected retention and key-precision policy. Sizes exclude model weights and temporary inference workspace. Lower is better.}
\label{tab:kv_cache_storage}
\vspace{2pt}
\fontsize{8}{9.5}\selectfont
\setlength{\tabcolsep}{2pt}
\renewcommand{\arraystretch}{1}
\begin{tabular*}{\columnwidth}{@{\extracolsep{\fill}}llccc@{}}
\toprule
\textbf{Model} & \textbf{Context} &
\shortstack{\textbf{BF16}\\\textbf{(GiB)}} &
\shortstack{\textbf{Full-Retention}\\\textbf{Compressed (GiB)}} &
\shortstack{\textbf{\modelname{}}\\\textbf{(GiB)}} \\
\midrule
\multirow{3}{*}{Llama}
& 32K  & 3.50  & 1.33 & \textbf{1.21} \\
& 64K  & 7.00  & 2.64 & \textbf{2.10} \\
& 128K & 14.00 & 5.27 & \textbf{3.74} \\
\midrule
\multirow{3}{*}{Qwen2}
& 32K  & 1.75 & \textbf{0.66} & \textbf{0.66} \\
& 64K  & 3.50 & 1.32 & \textbf{1.28} \\
& 128K & 7.00 & 2.64 & \textbf{2.51} \\
\midrule
\multirow{3}{*}{Ministral}
& 32K  & 4.25  & 1.61 & \textbf{1.47} \\
& 64K  & 8.50  & 3.21 & \textbf{2.80} \\
& 128K & 17.00 & 6.40 & \textbf{5.45} \\
\bottomrule
\end{tabular*}
\end{minipage}
\par

\textbf{Persistent prompt storage.}
Cache size depends on both the number of retained tiles and their stored precision.

\begin{figure*}[!tp]
\centering
\begin{minipage}{\textwidth}
\normalsize
\clubpenalties 0
\widowpenalties 0
\setlength{\multicolsep}{0pt}
\begin{multicols}{2}
\raggedcolumns
At 128K, Llama's prompt cache occupies 14.00\,GiB in BF16, 5.27\,GiB with all tiles retained using 8-bit keys and mixed-bit values, and 3.74\,GiB with the selected policy (Table~\ref{tab:kv_cache_storage}). Low-bit representation supplies the first reduction; retention and key-precision selection save a further 29.0\% over the full-retention compressed cache.

\par\addvspace{2pt}
\noindent\begin{minipage}{\columnwidth}
\setlength{\abovecaptionskip}{0pt}
\centering
\captionof{table}{Effect of prompt-cache retention and value precision on decode attention latency on A100. Each entry measures one decode step summed across all layers. Full Retention keeps all prompt tiles with 8-bit keys. The 8-bit Values control retains the selected context and key precision and uses 8-bit values with corresponding 8-bit probability codes on every retained tile. Lower is better.}
\label{tab:decode_cache_ablation}
\vspace{2pt}
\fontsize{8}{9.5}\selectfont
\setlength{\tabcolsep}{2pt}
\renewcommand{\arraystretch}{1}
\begin{tabular*}{\columnwidth}{@{\extracolsep{\fill}}llccc@{}}
\toprule
\textbf{Model} & \textbf{Context} &
\shortstack{\textbf{\modelname{}}\\\textbf{(ms)}} &
\shortstack{\textbf{Full Retention}\\\textbf{(ms)}} &
\shortstack{\textbf{8-bit Values}\\\textbf{(ms)}} \\
\midrule
\multirow{3}{*}{Llama}
& 32K  & \textbf{1.62} & 1.68 & 1.95 \\
& 64K  & \textbf{2.39} & 2.69 & 2.93 \\
& 128K & \textbf{3.79} & 4.77 & 4.72 \\
\midrule
\multirow{3}{*}{Qwen2}
& 32K  & \textbf{0.93} & \textbf{0.93} & 1.32 \\
& 64K  & \textbf{1.81} & 1.82 & 2.15 \\
& 128K & \textbf{2.79} & 2.84 & 3.41 \\
\midrule
\multirow{3}{*}{Ministral}
& 32K  & \textbf{1.96} & 2.07 & 2.37 \\
& 64K  & \textbf{3.02} & 3.29 & 3.73 \\
& 128K & \textbf{5.19} & 5.82 & 6.52 \\
\bottomrule
\end{tabular*}
\end{minipage}
\par

\textbf{Repeated decode work.}
The compressed prompt cache is read at every generation step, so its retention and precision choices also affect recurring attention cost. Full Retention keeps all prompt tiles with 8-bit keys, while 8-bit Values preserves the selected context and key precision and uses 8-bit values with matching 8-bit probability codes. At 128K, Ministral takes 5.19\,ms with mixed-bit values and 6.52\,ms with 8-bit Values (Table~\ref{tab:decode_cache_ablation}).

The additional benefit of the selected cache policy grows with context length. For Llama, its storage reduction over the full-retention compressed cache increases from 9.0\% at 32K to 29.0\% at 128K. Over the same lengths, decode attention latency decreases from 1.68 to 1.62\,ms and from 4.77 to 3.79\,ms, corresponding to reductions of 3.6\% and 20.5\%. The paired measurements link the policy's smaller persistent prompt cache to lower per-step attention cost.

\findingbox{%
Applying the selected policy to compressed KV storage reduces the persistent prompt working set and the attention cost repeated during generation.
}

\clubpenalties 0
\widowpenalties 0
\let\paragraph\designsavedparagraph

\vspace{-6pt}
\section{Extended Evaluation}
\label{app:extended-evaluation}
\vspace{-4pt}

The selected boundary table is reused across tasks and request shapes. Additional models test whether reuse preserves prediction and evidence use, while the workload grids show how complete-request gains vary with generation length and GPU platform.

\vspace{-4pt}
\subsection{Workload and Hardware Scaling}
\label{app:workload-scaling}
\vspace{-3pt}

\textls[10]{The workload grids vary prompt processing, generation length, and batch-level parallelism. Each comparison uses the same model, GPU, and request shape. Torch attention provides the SDPA math-backend reference. Generation-throughput ratios divide method throughput by BF16 throughput, while TPOT speedups divide BF16 \mbox{TPOT by method TPOT.}}\looseness=-1\par
\end{multicols}
\end{minipage}
\par\vspace{2pt}
\begin{adjustbox}{max width=\textwidth}
\begin{minipage}{\textwidth}
\centering
\includegraphics[width=0.27650431\textwidth]{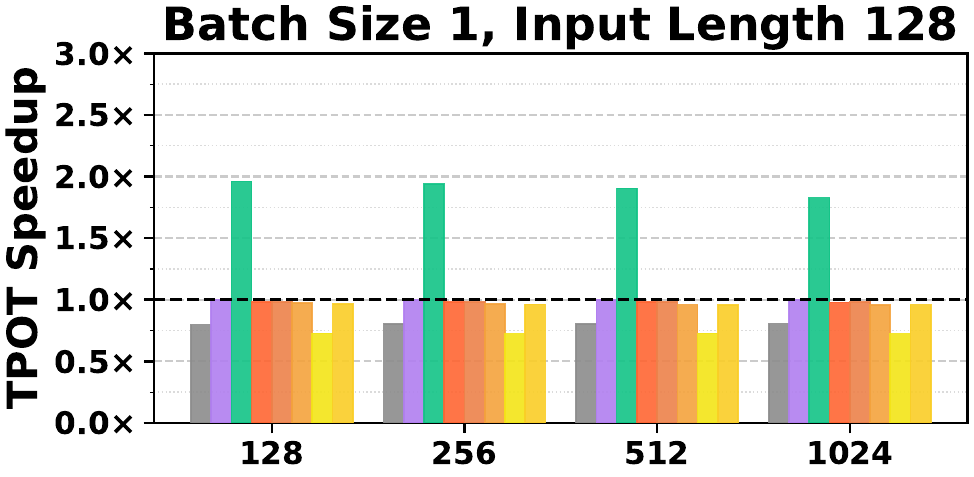}\hfill
\includegraphics[width=0.23449856\textwidth]{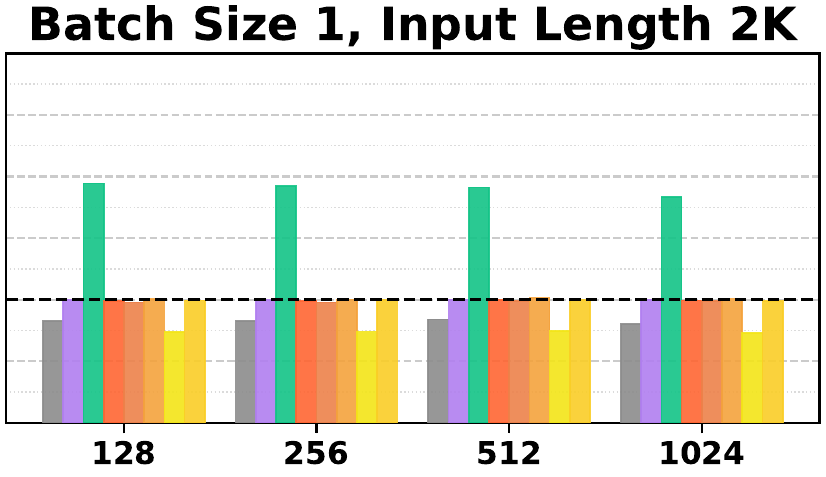}\hfill
\includegraphics[width=0.23449856\textwidth]{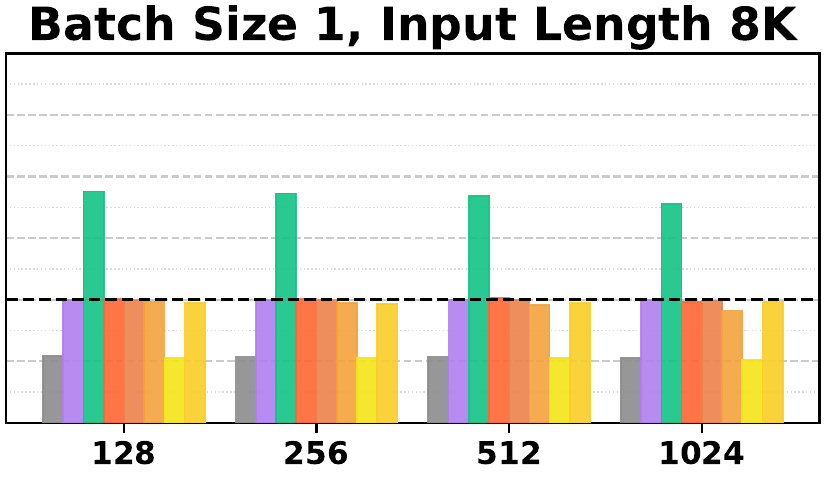}\hfill
\includegraphics[width=0.23449856\textwidth]{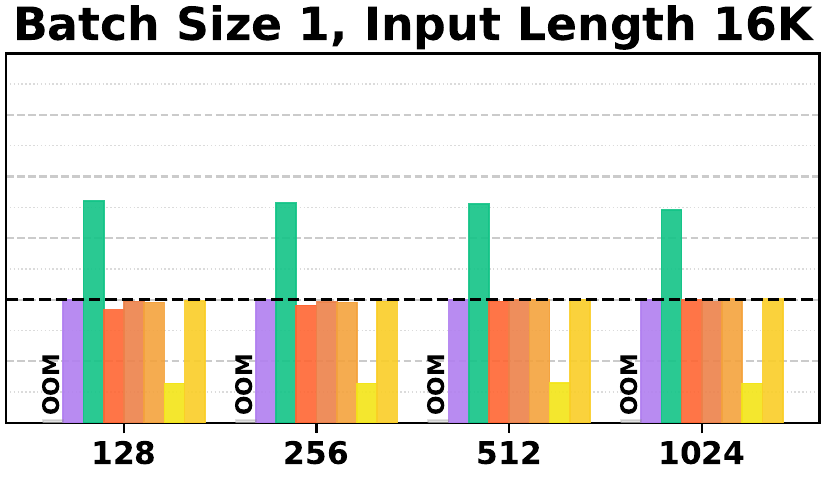}\\[-3pt]
\includegraphics[width=0.27650431\textwidth]{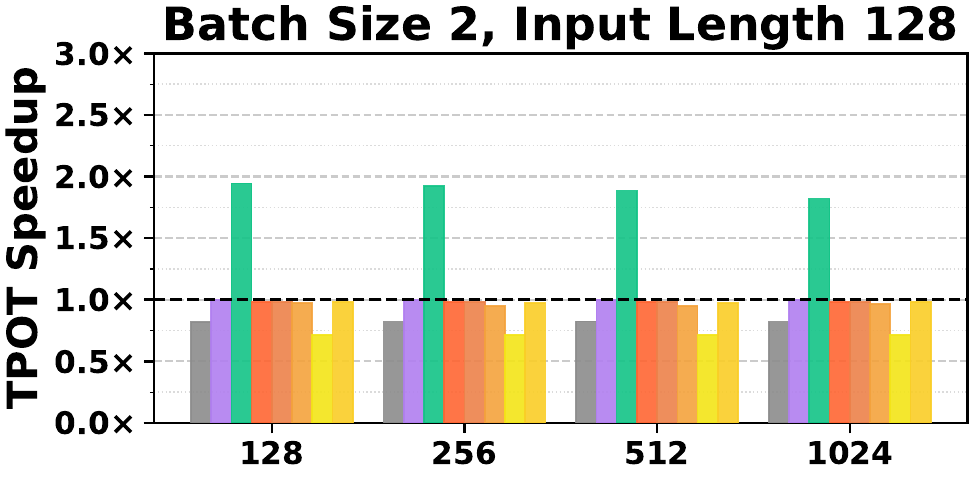}\hfill
\includegraphics[width=0.23449856\textwidth]{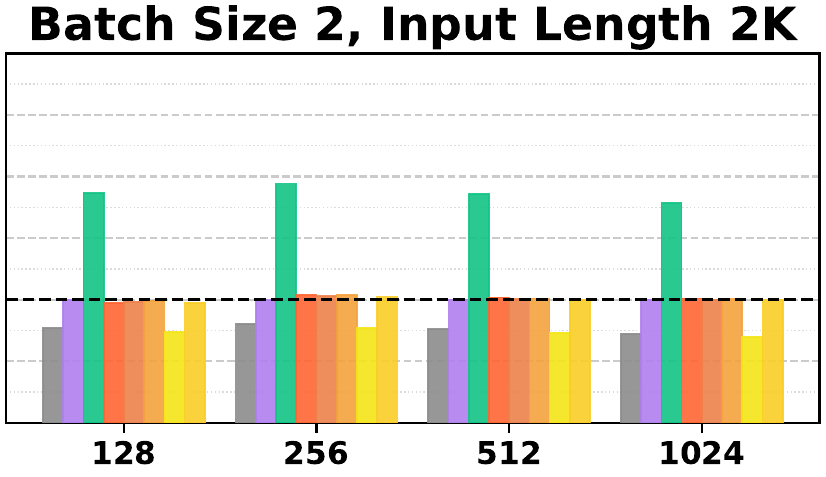}\hfill
\includegraphics[width=0.23449856\textwidth]{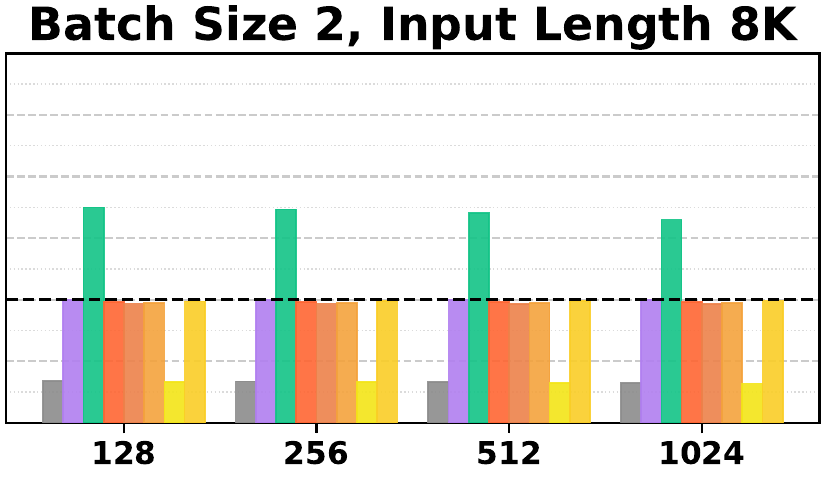}\hfill
\includegraphics[width=0.23449856\textwidth]{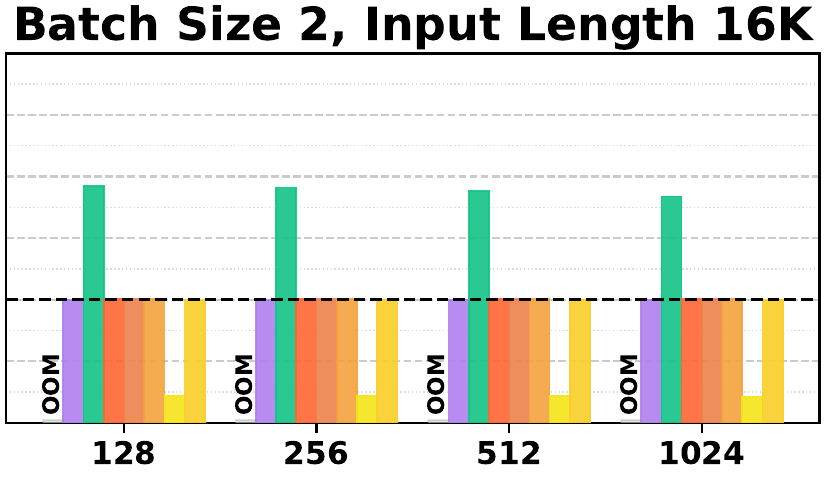}\\[-3pt]
\includegraphics[width=0.27650431\textwidth]{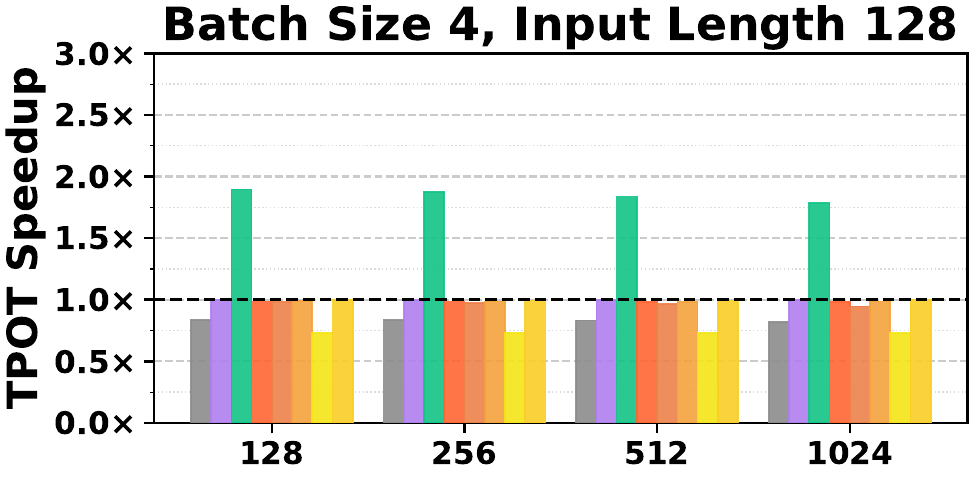}\hfill
\includegraphics[width=0.23449856\textwidth]{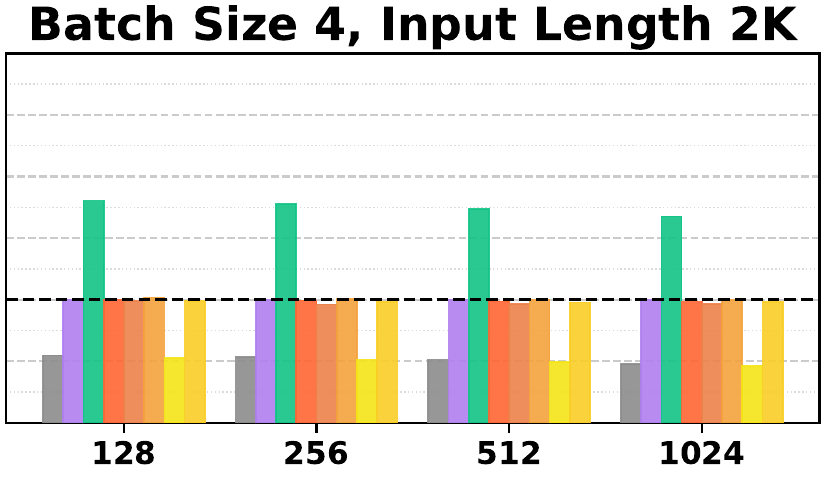}\hfill
\includegraphics[width=0.23449856\textwidth]{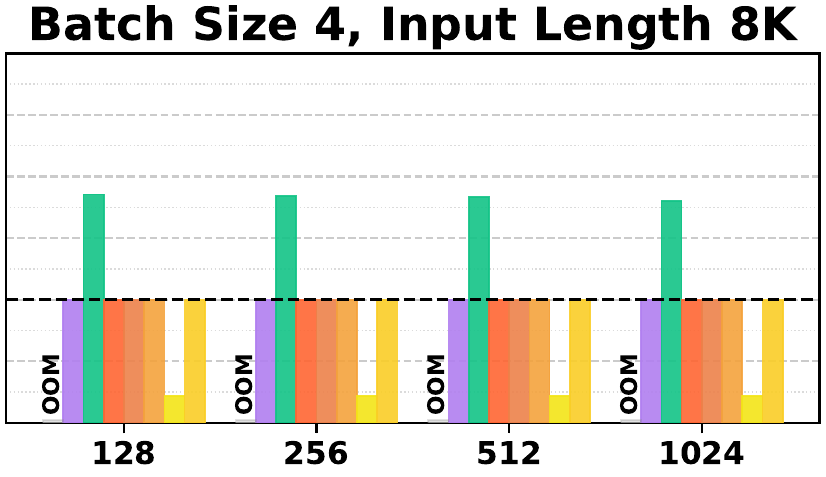}\hfill
\includegraphics[width=0.23449856\textwidth]{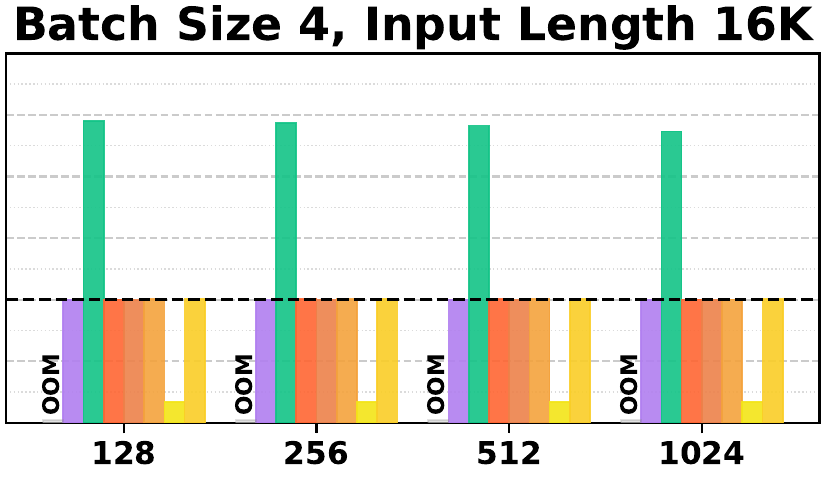}\\[-3pt]
\includegraphics[width=0.27650431\textwidth]{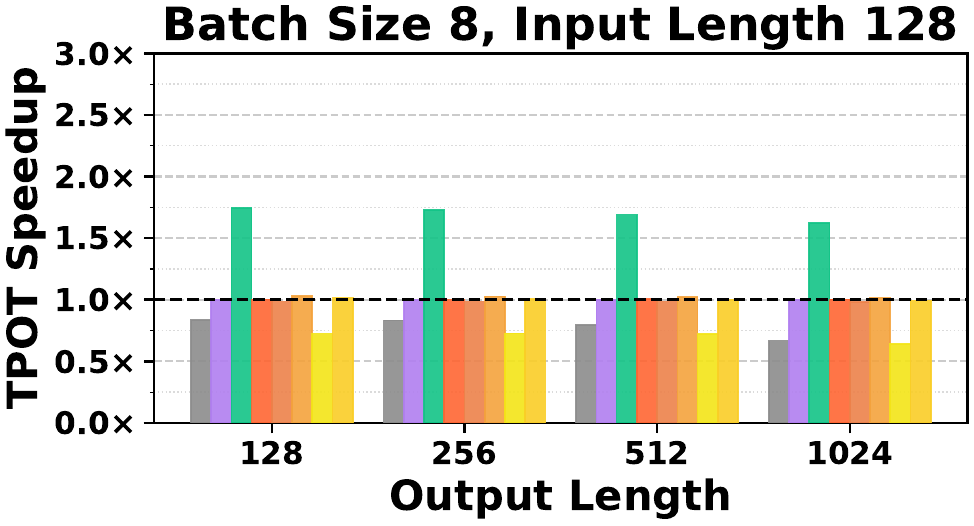}\hfill
\includegraphics[width=0.23449856\textwidth]{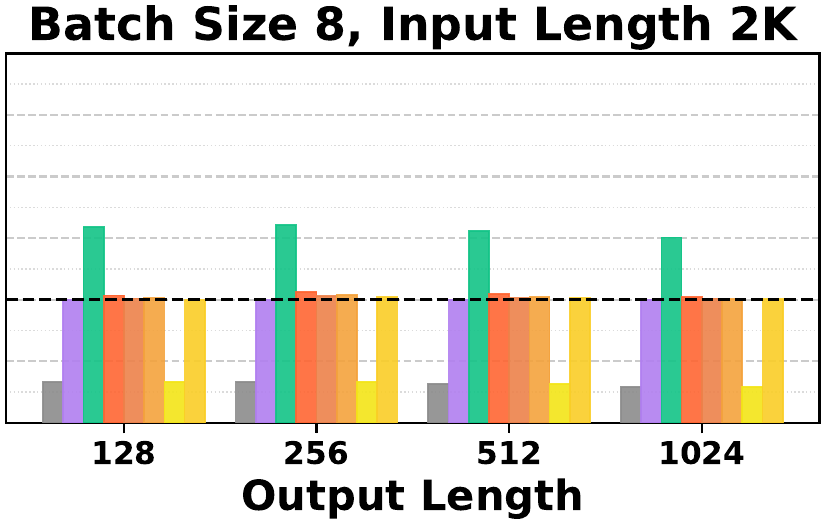}\hfill
\includegraphics[width=0.23449856\textwidth]{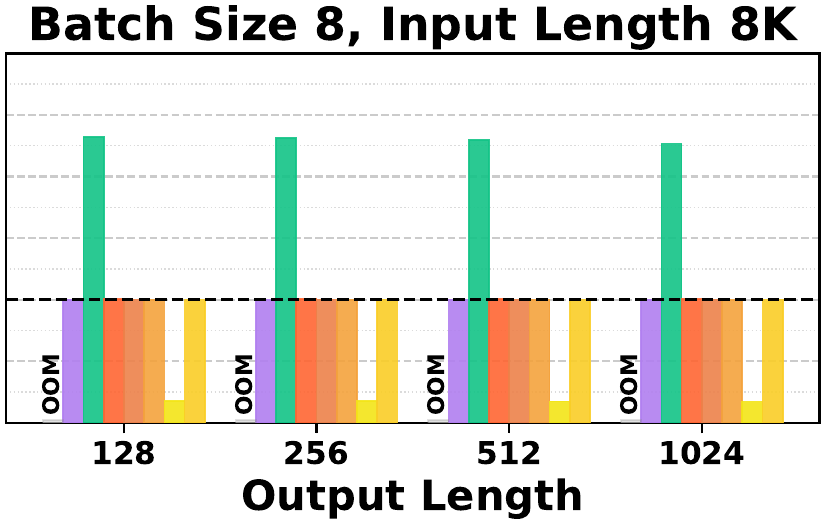}\hfill
\includegraphics[width=0.23449856\textwidth]{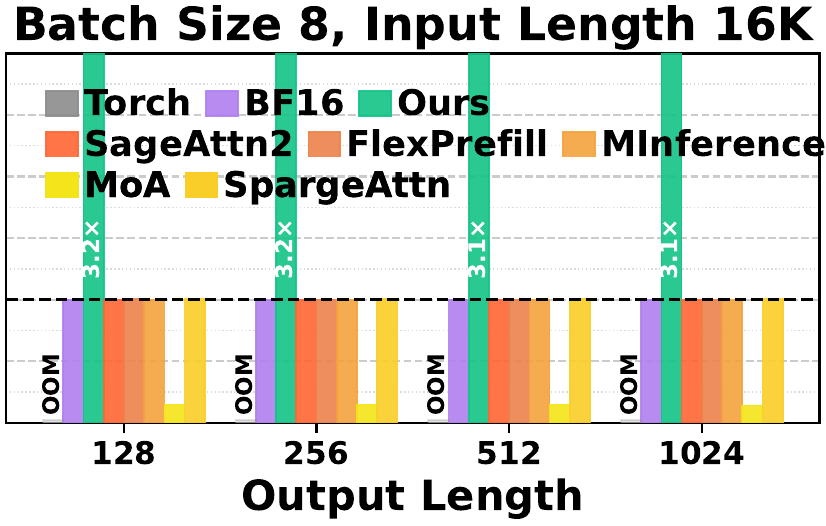}
\vspace{-8pt}
\end{minipage}
\end{adjustbox}
\caption{\textbf{TPOT speedup on A100-SXM4-40GB.} Llama-3.2-3B-Instruct. Rows use batch sizes 1, 2, 4, and 8; columns use 128, 2,048, 8,192, 16,384 input tokens. Each panel varies output length over 128, 256, 512, and 1024 tokens. BF16 is normalized to $1\times$; higher is better. Method colors and order are fixed across panels.}
\label{fig:system-a100-llama32_3b-tpot}
\end{figure*}
\clubpenalties 0
\widowpenalties 0
\let\paragraph\designsavedparagraph

\begin{figure*}[!tp]
\centering
\includegraphics[width=0.36347914\textwidth]{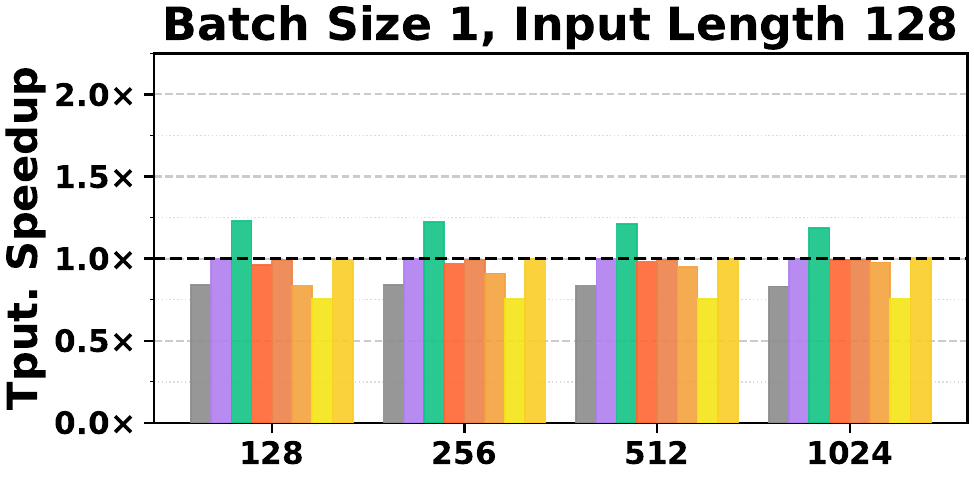}\hfill
\includegraphics[width=0.30826043\textwidth]{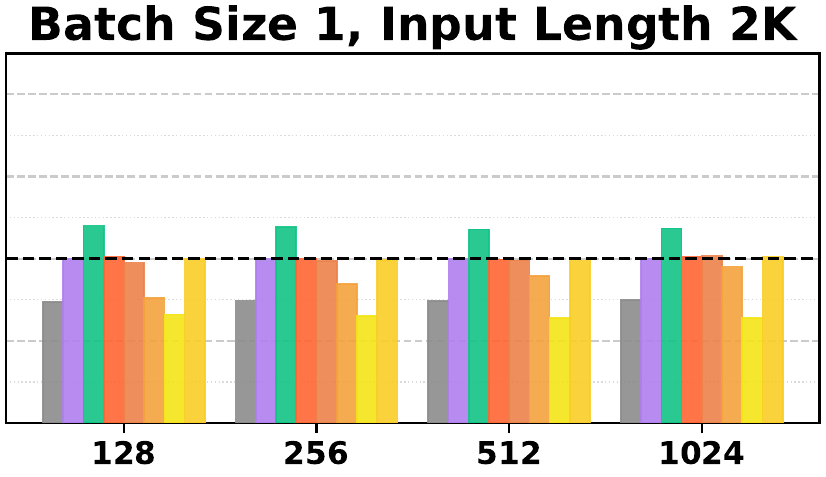}\hfill
\includegraphics[width=0.30826043\textwidth]{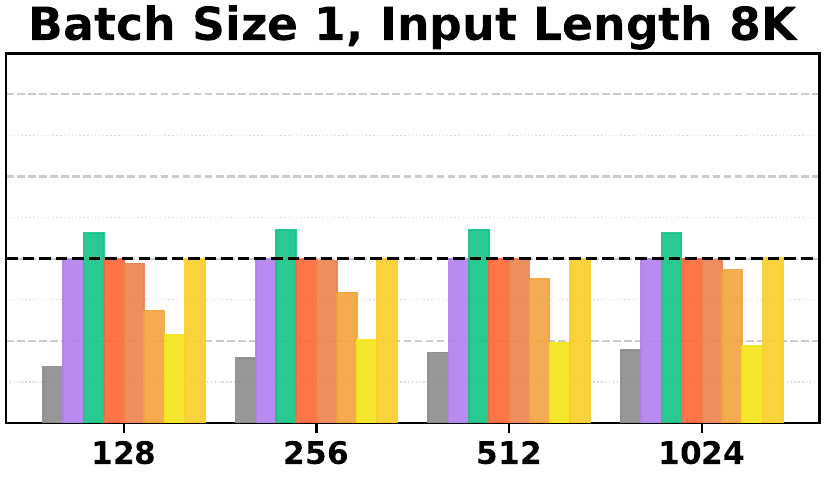}\\[-3pt]
\includegraphics[width=0.36347914\textwidth]{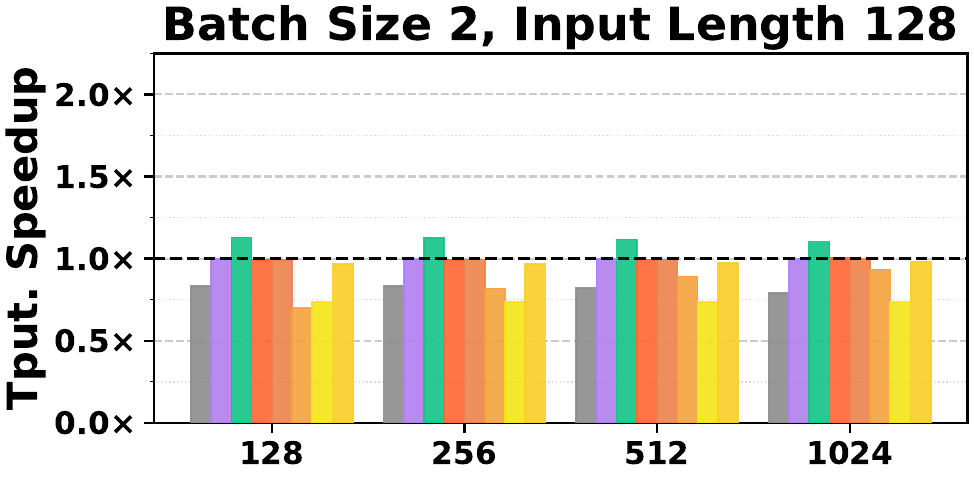}\hfill
\includegraphics[width=0.30826043\textwidth]{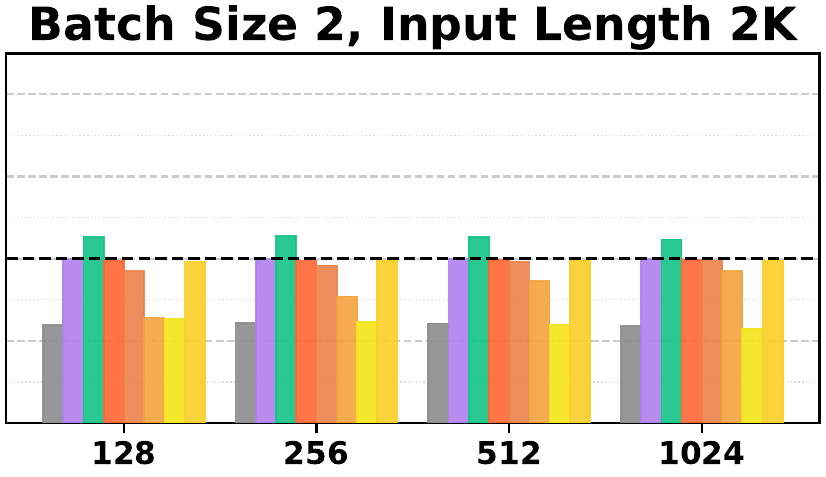}\hfill
\includegraphics[width=0.30826043\textwidth]{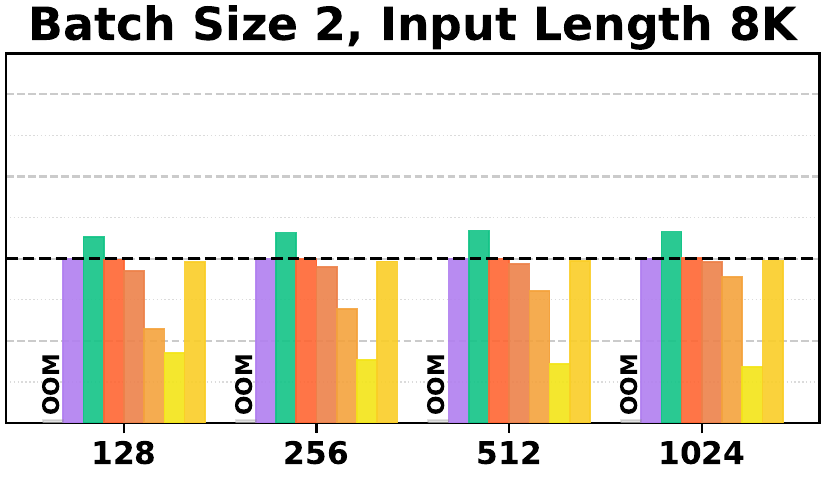}\\[-3pt]
\includegraphics[width=0.36347914\textwidth]{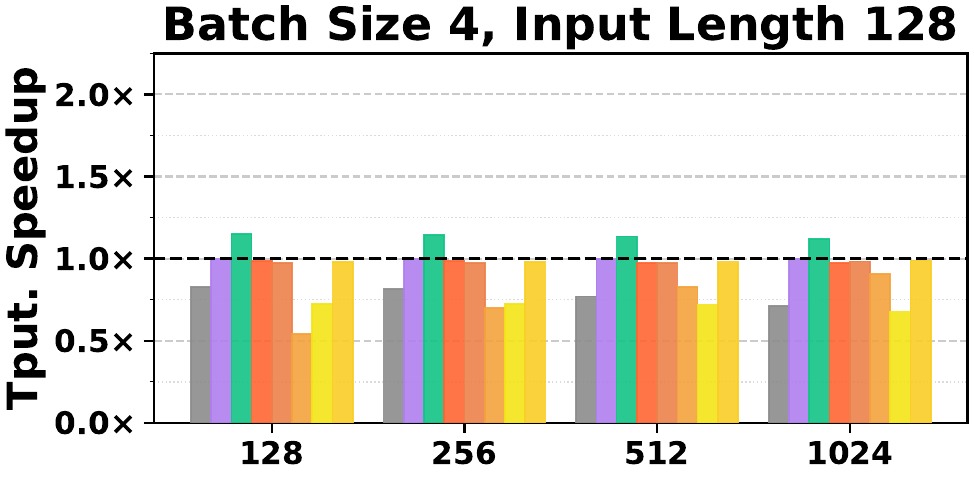}\hfill
\includegraphics[width=0.30826043\textwidth]{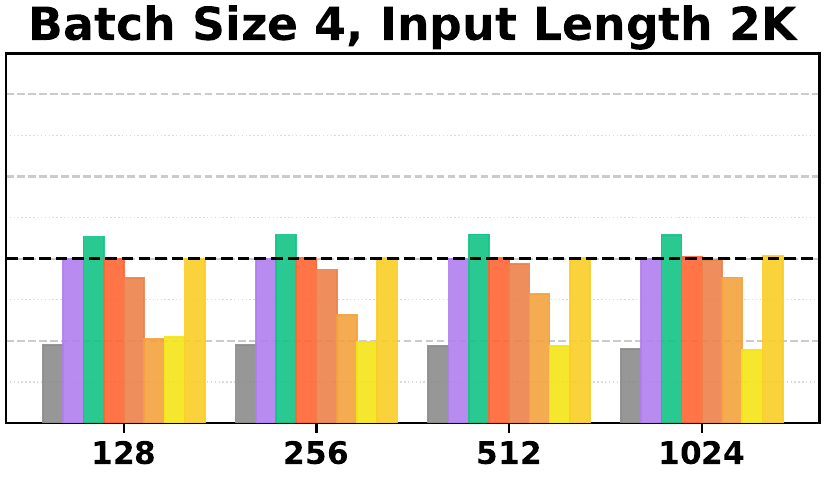}\hfill
\includegraphics[width=0.30826043\textwidth]{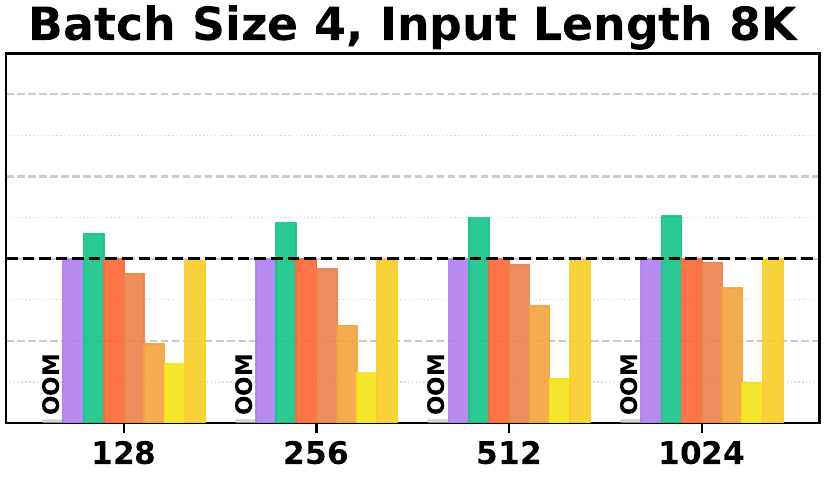}\\[-3pt]
\includegraphics[width=0.36347914\textwidth]{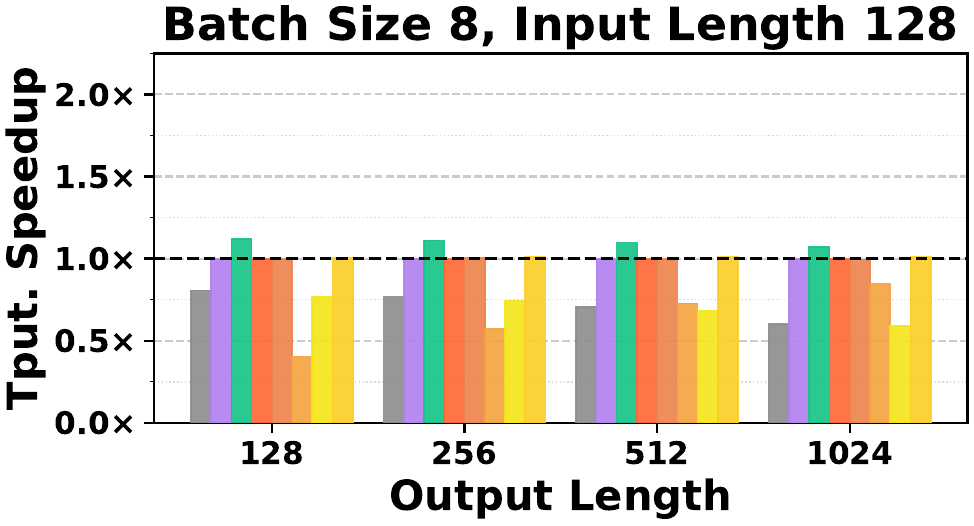}\hfill
\includegraphics[width=0.30826043\textwidth]{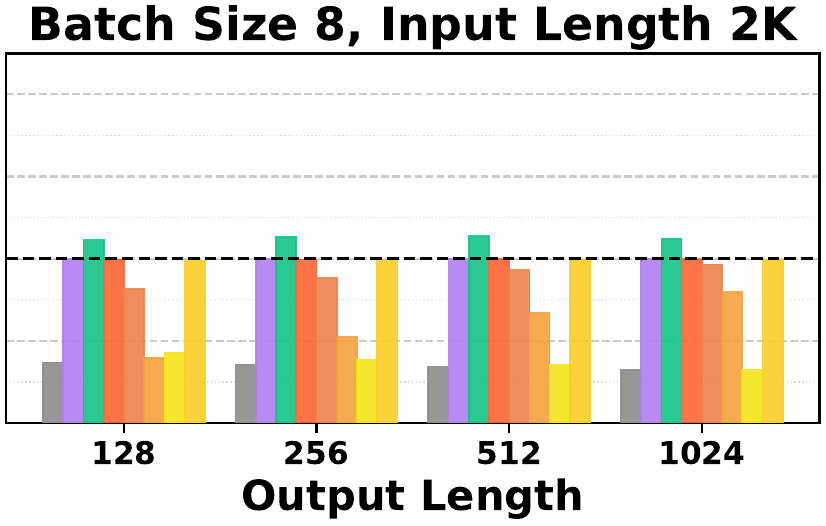}\hfill
\includegraphics[width=0.30826043\textwidth]{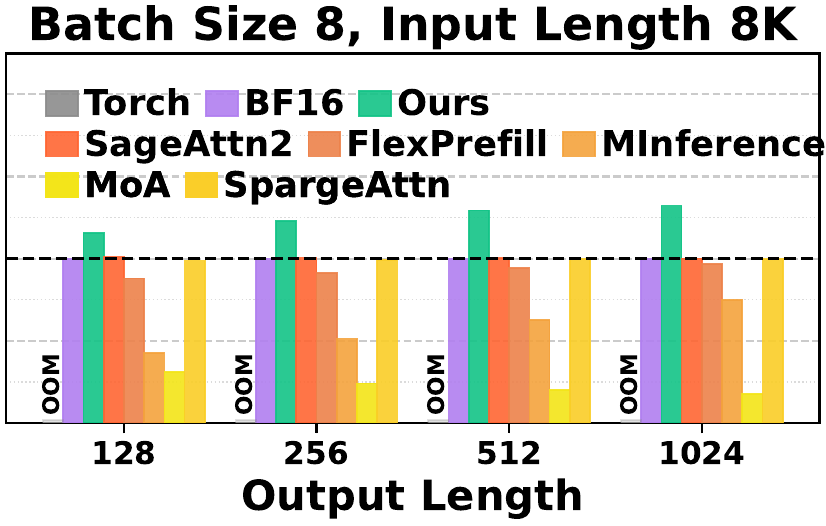}
\vspace{-8pt}
\caption{\textbf{Generation-throughput ratio on A100-SXM4-40GB.} Qwen2-7B-Instruct. Rows use batch sizes 1, 2, 4, and 8; columns use 128, 2,048, 8,192 input tokens. Each panel varies output length over 128, 256, 512, and 1024 tokens. BF16 is normalized to $1\times$; higher is better. Method colors and order are fixed across panels.}
\label{fig:system-a100-qwen2_7b-throughput}
\par\vspace{0pt}
\begin{minipage}{\textwidth}
\normalsize
\setlength{\multicolsep}{0pt}
\makeatletter
\renewcommand{\paragraph}{\@startsection{paragraph}{4}{\z@}{5pt}{-1em}{\normalfont\normalsize\bfseries}}
\makeatother
\begin{multicols}{2}
\raggedcolumns
\textbf{A100.}
Output length changes the balance between one-time prompt processing and repeated token generation. For Llama with a 16K prompt and batch size eight, increasing the output length from 128 to 1,024 tokens changes complete-request throughput from 137.07 to 394.97 tokens/s, compared with 88.55 to 156.58 tokens/s for BF16. The throughput advantage grows from $1.55\times$ to $2.52\times$ as the prompt cost is distributed across a longer continuation.\par
\end{multicols}
\end{minipage}
\end{figure*}

\begin{figure*}[!tp]
\centering
\includegraphics[width=0.36347914\textwidth]{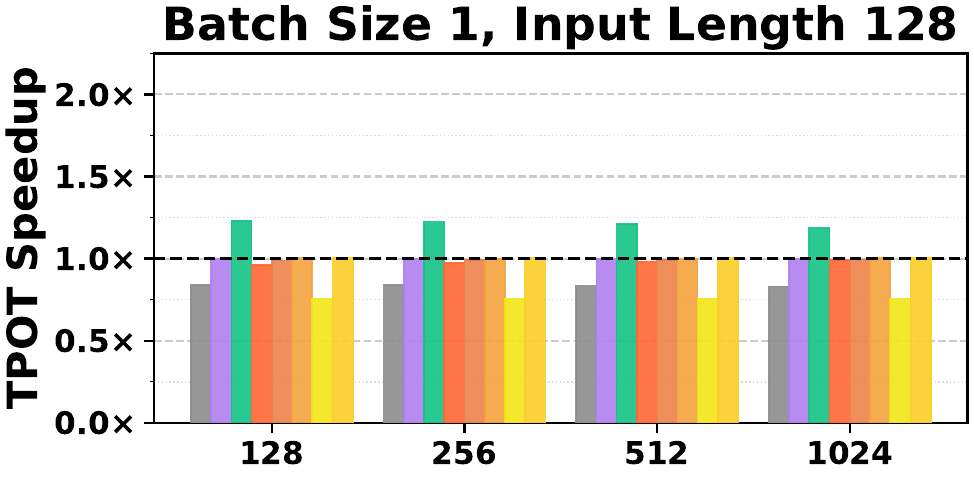}\hfill
\includegraphics[width=0.30826043\textwidth]{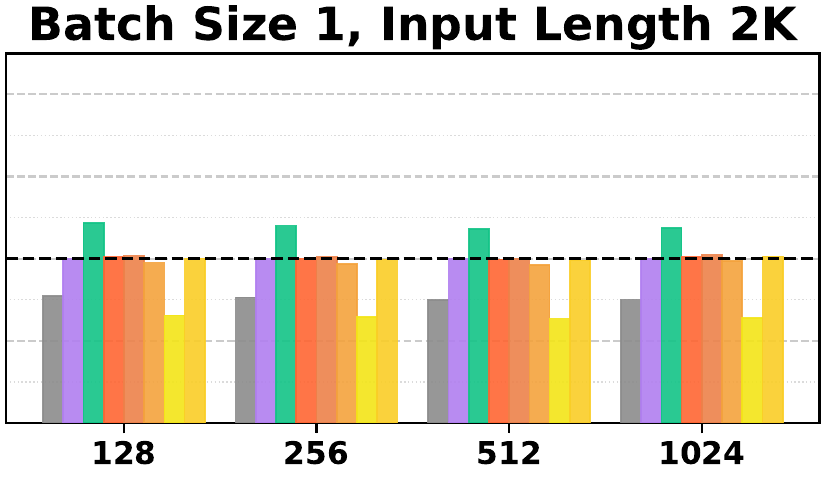}\hfill
\includegraphics[width=0.30826043\textwidth]{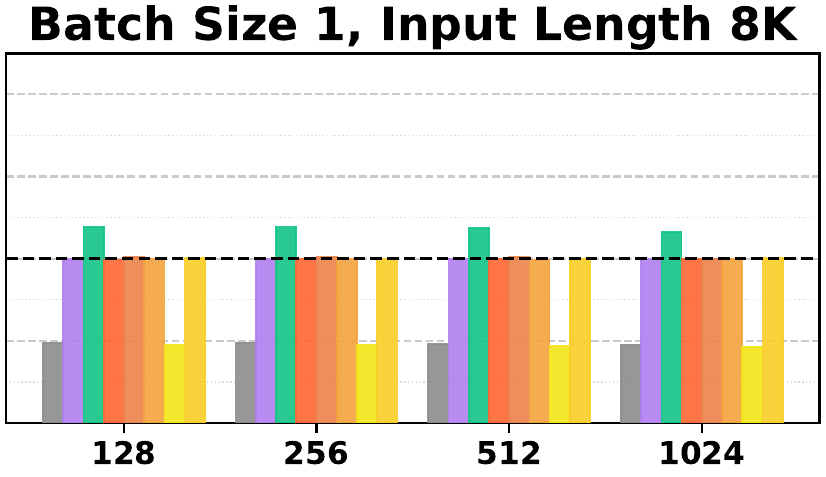}\\[-3pt]
\includegraphics[width=0.36347914\textwidth]{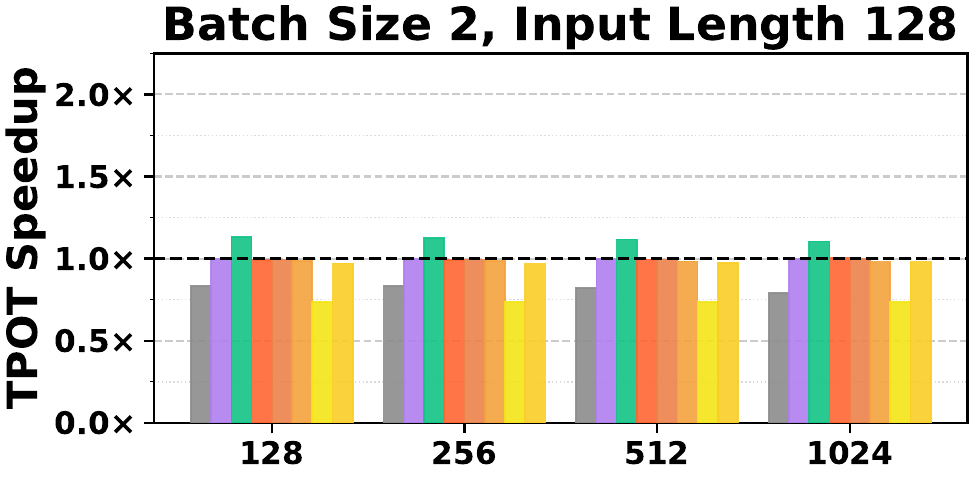}\hfill
\includegraphics[width=0.30826043\textwidth]{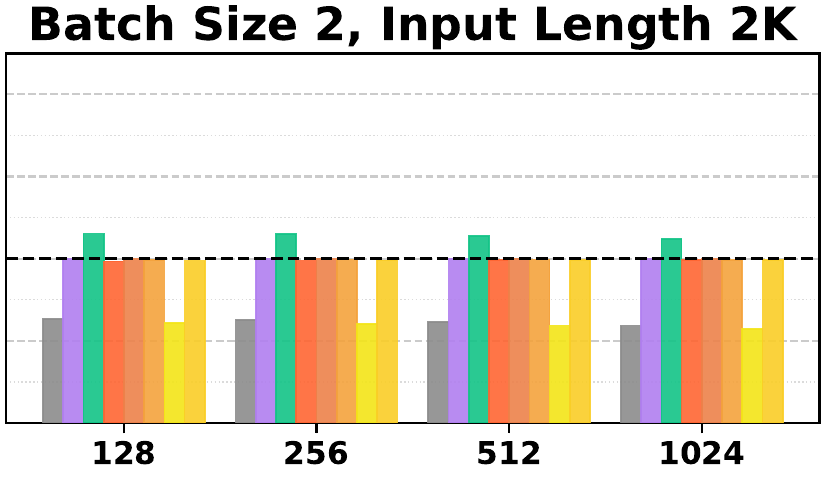}\hfill
\includegraphics[width=0.30826043\textwidth]{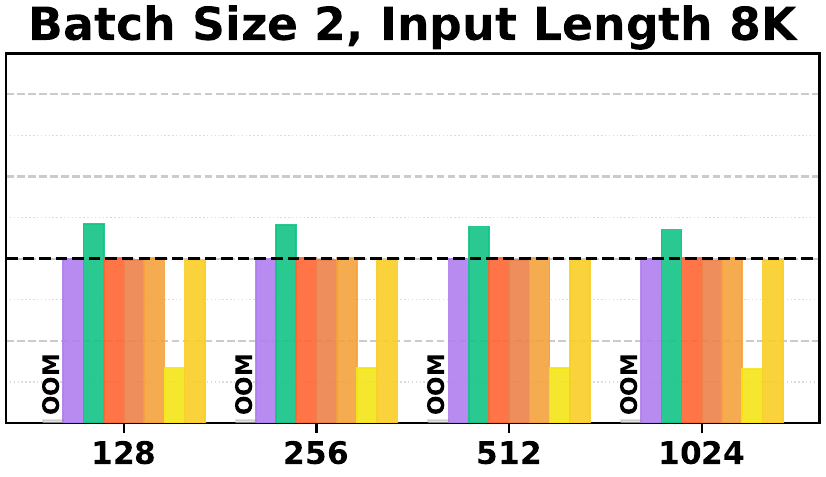}
\vspace{-8pt}
\caption{\textbf{TPOT speedup on A100-SXM4-40GB.} Qwen2-7B-Instruct. Rows use batch sizes 1 and 2; columns use 128, 2,048, 8,192 input tokens. Each panel varies output length over 128, 256, 512, and 1024 tokens. BF16 is normalized to $1\times$; higher is better. Method colors and order are fixed across panels.}
\label{fig:system-a100-qwen2_7b-tpot}
\end{figure*}

\begin{figure*}[!tp]
\centering
\includegraphics[width=0.36347914\textwidth]{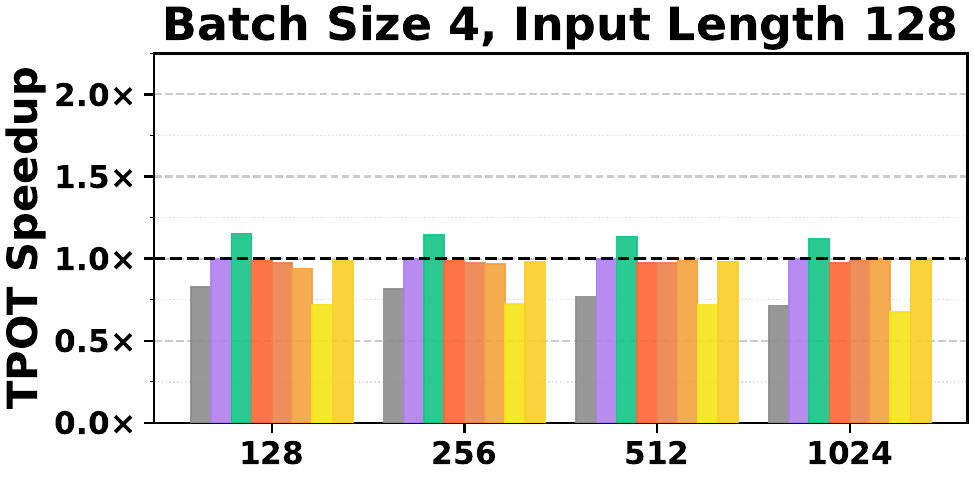}\hfill
\includegraphics[width=0.30826043\textwidth]{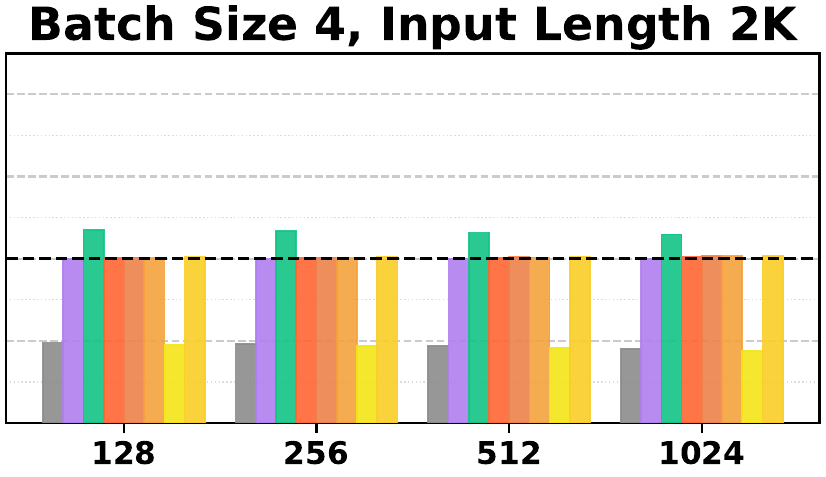}\hfill
\includegraphics[width=0.30826043\textwidth]{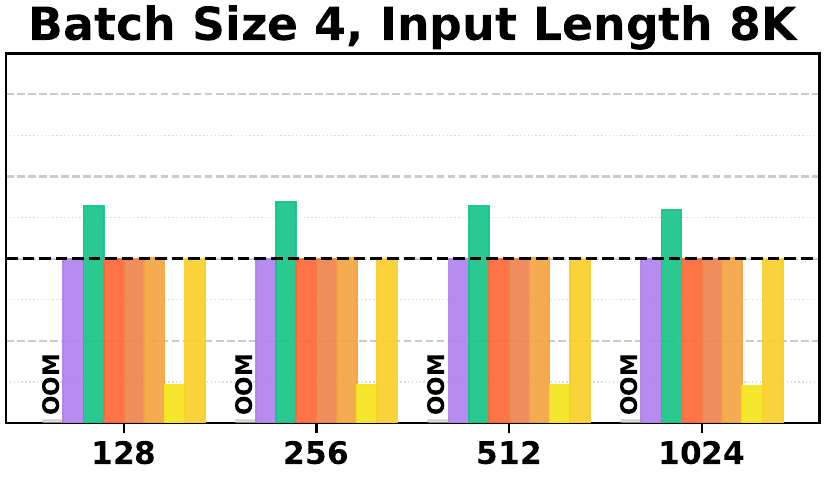}\\[-3pt]
\includegraphics[width=0.36347914\textwidth]{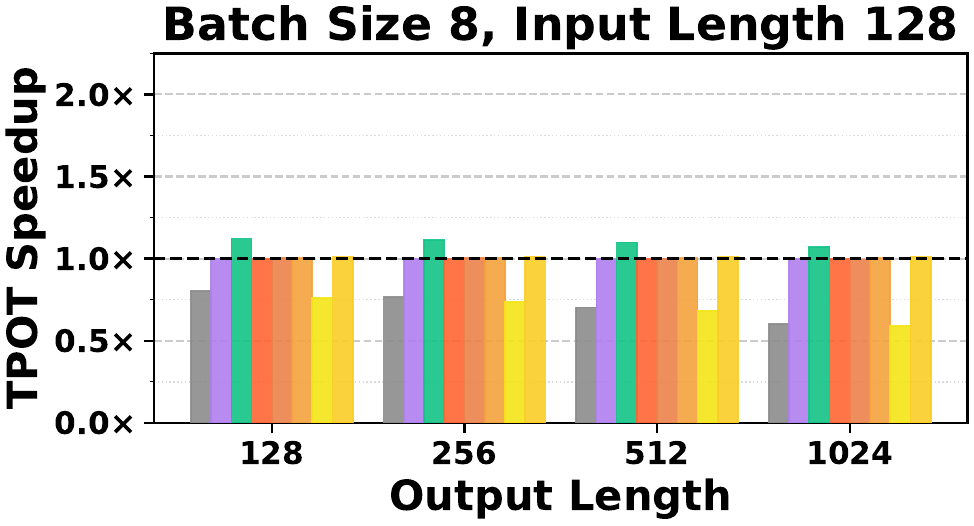}\hfill
\includegraphics[width=0.30826043\textwidth]{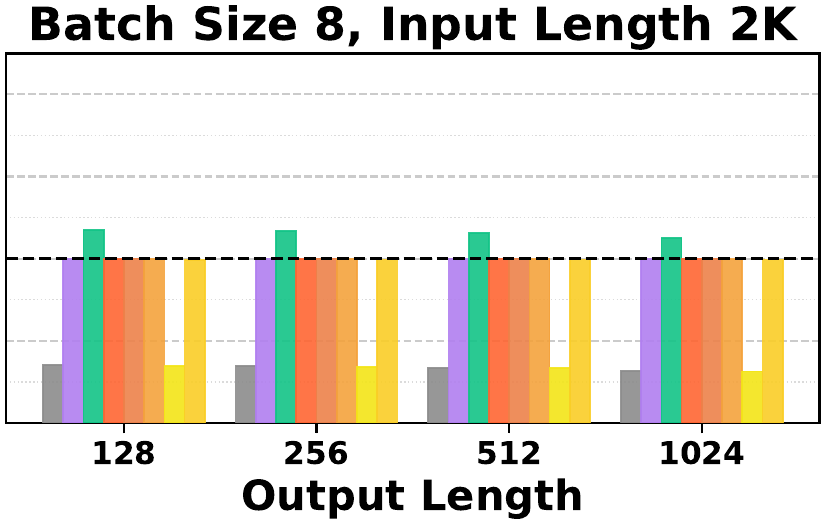}\hfill
\includegraphics[width=0.30826043\textwidth]{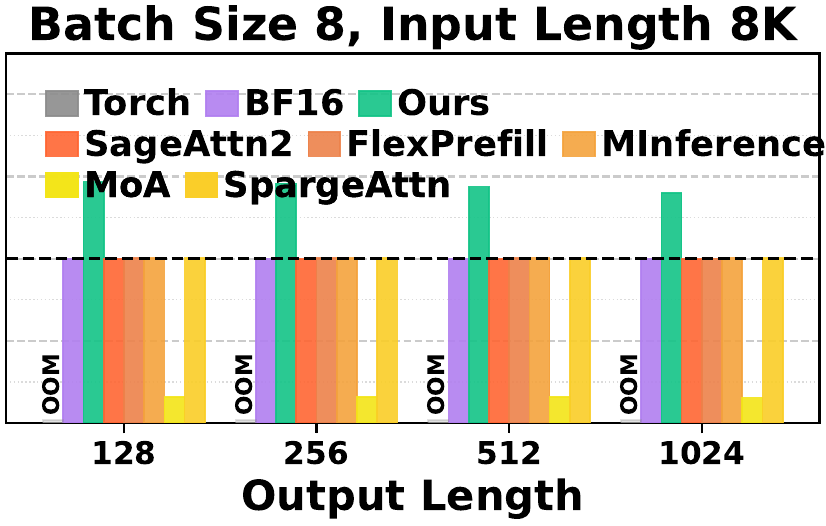}
\vspace{-8pt}
\addtocounter{figure}{-1}
\renewcommand{\theHfigure}{\arabic{figure}.continued}
\caption[]{\textbf{Continued.} Rows use batch sizes 4 and 8; panel settings follow the first part.}
\end{figure*}

\begin{figure*}[!tp]
\centering
\includegraphics[width=0.36347914\textwidth]{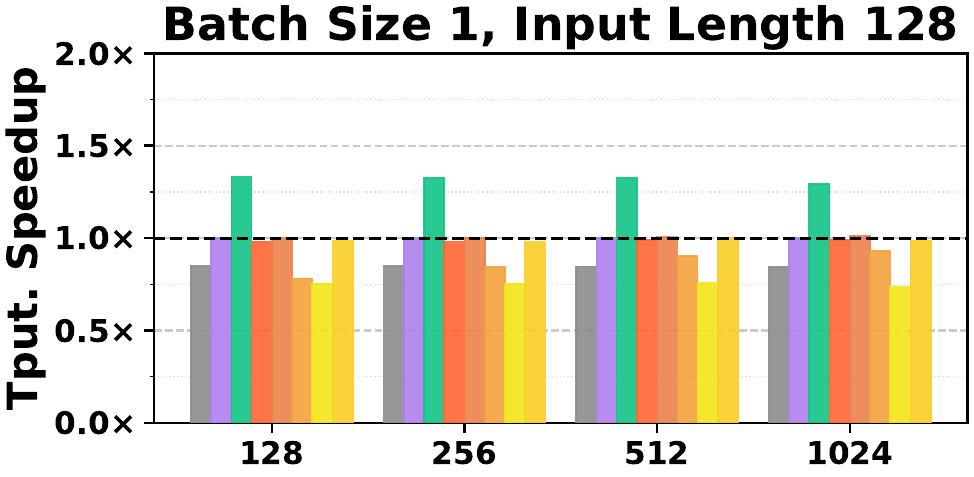}\hfill
\includegraphics[width=0.30826043\textwidth]{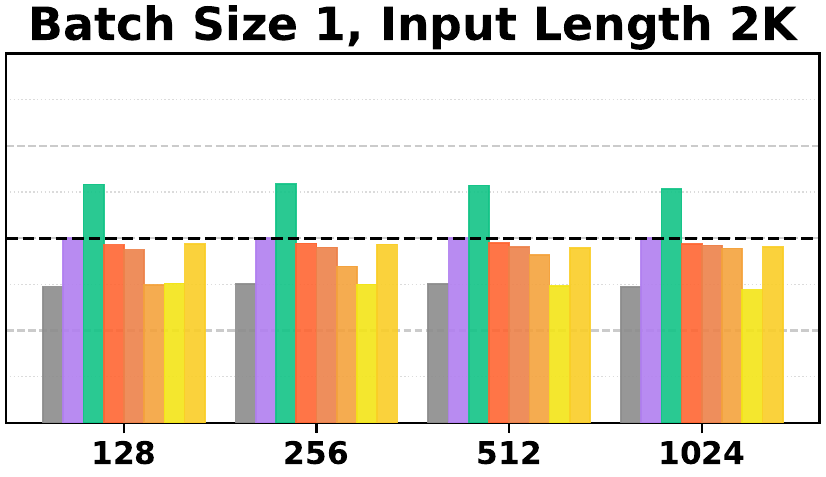}\hfill
\includegraphics[width=0.30826043\textwidth]{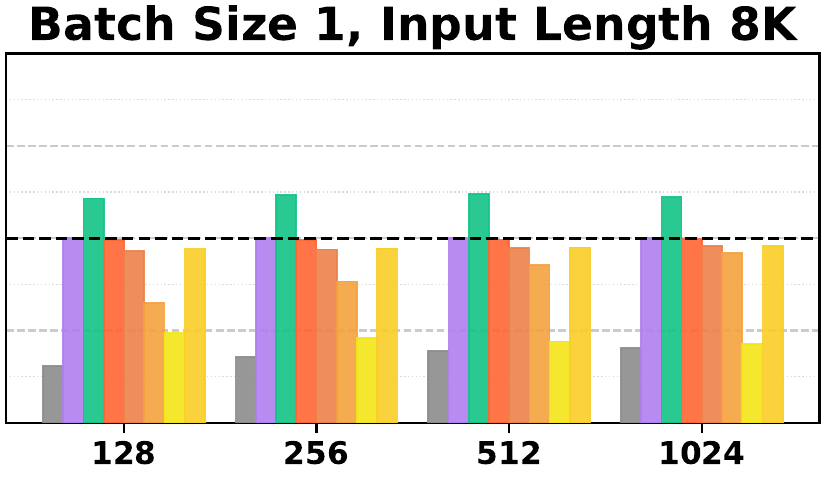}\\[-3pt]
\includegraphics[width=0.36347914\textwidth]{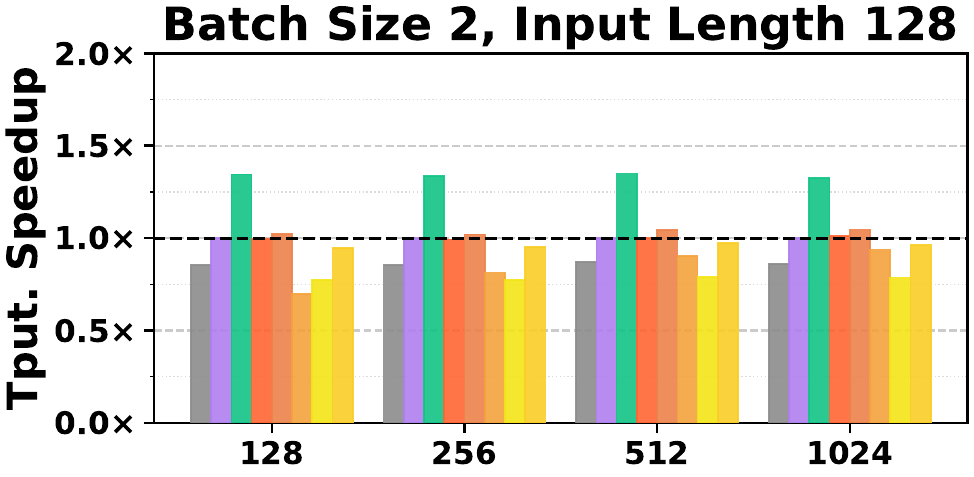}\hfill
\includegraphics[width=0.30826043\textwidth]{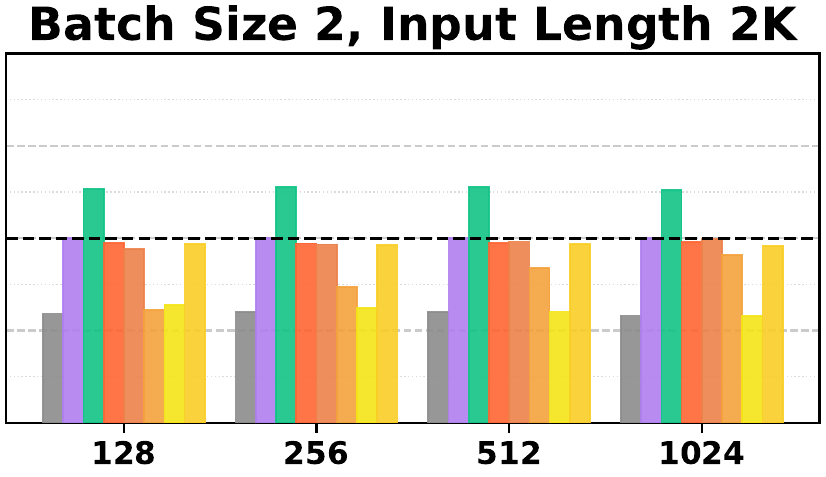}\hfill
\includegraphics[width=0.30826043\textwidth]{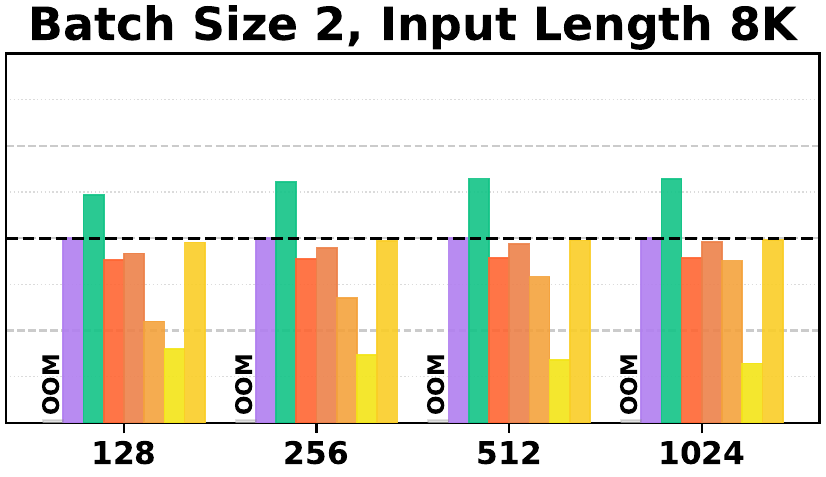}\\[-3pt]
\includegraphics[width=0.36347914\textwidth]{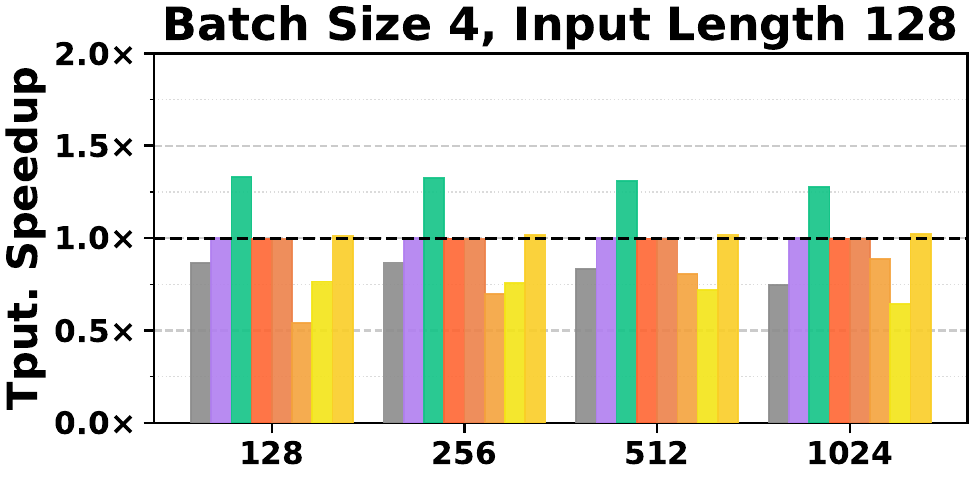}\hfill
\includegraphics[width=0.30826043\textwidth]{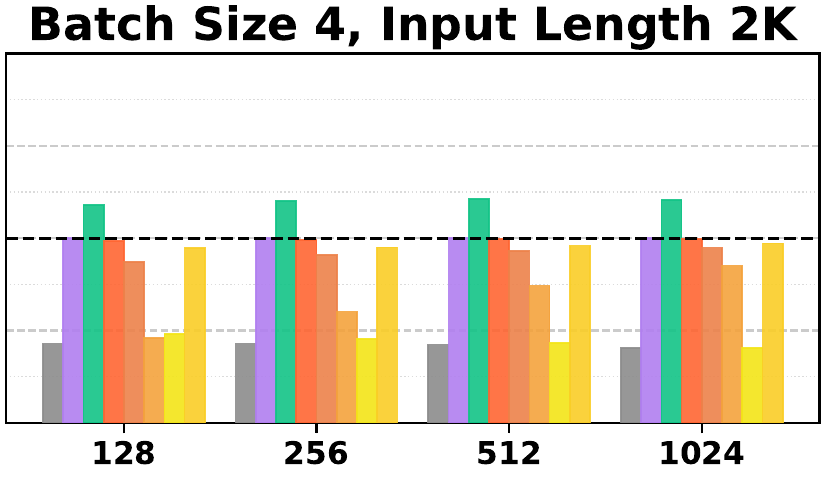}\hfill
\includegraphics[width=0.30826043\textwidth]{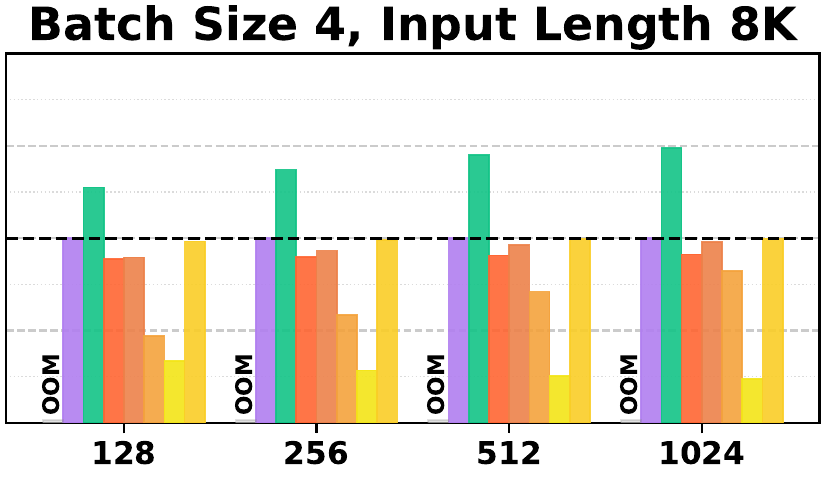}\\[-3pt]
\includegraphics[width=0.36347914\textwidth]{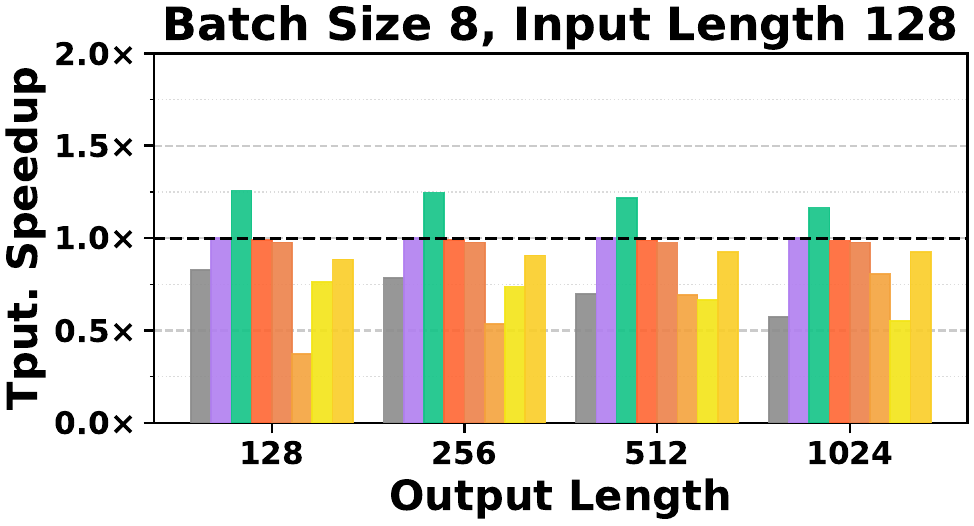}\hfill
\includegraphics[width=0.30826043\textwidth]{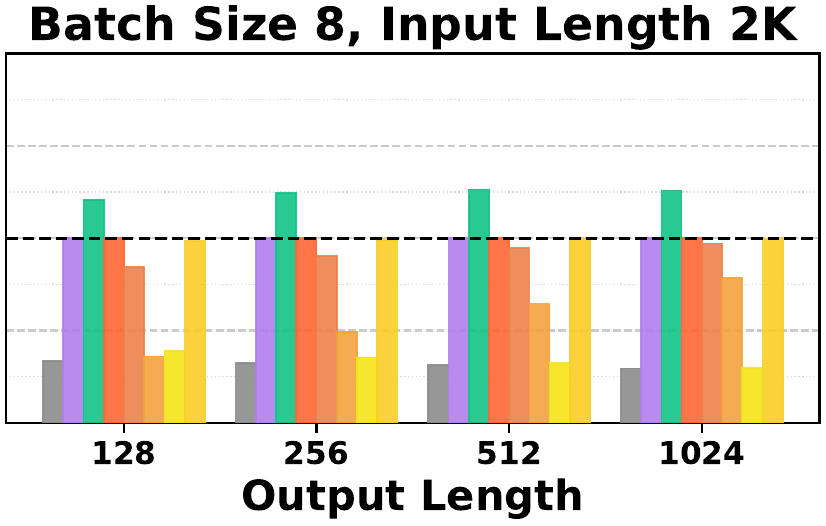}\hfill
\includegraphics[width=0.30826043\textwidth]{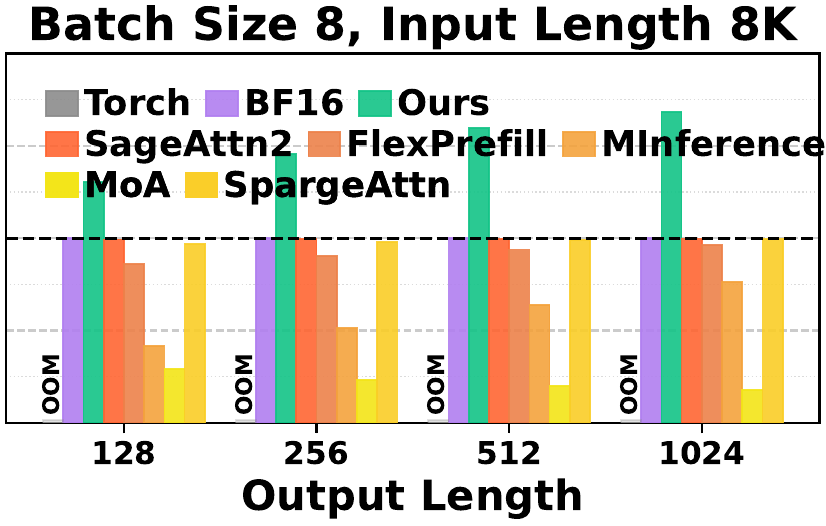}
\vspace{-8pt}
\caption{\textbf{Generation-throughput ratio on A100-SXM4-40GB.} Ministral-3-8B-Instruct-2512-BF16. Rows use batch sizes 1, 2, 4, and 8; columns use 128, 2,048, 8,192 input tokens. Each panel varies output length over 128, 256, 512, and 1024 tokens. BF16 is normalized to $1\times$; higher is better. Method colors and order are fixed across panels.}
\label{fig:system-a100-ministral3_8b-throughput}
\end{figure*}

\begin{figure*}[!tp]
\centering
\includegraphics[width=0.36347914\textwidth]{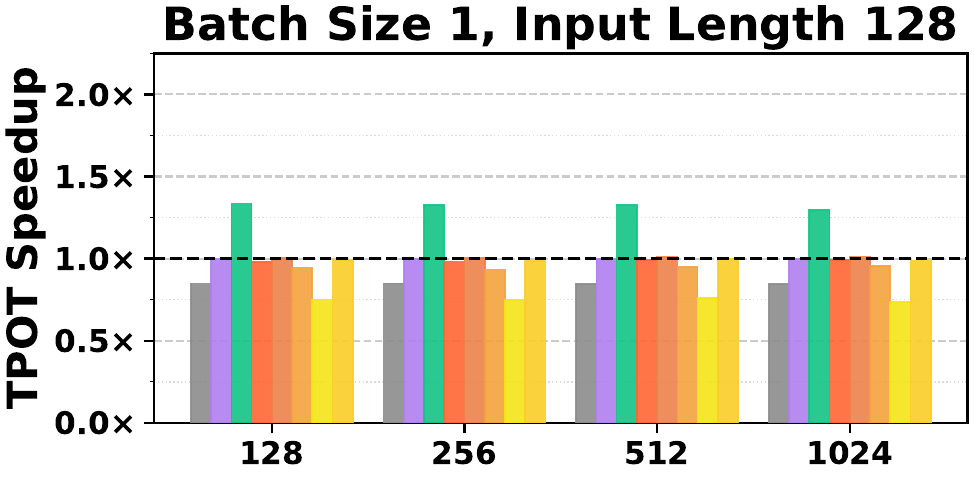}\hfill
\includegraphics[width=0.30826043\textwidth]{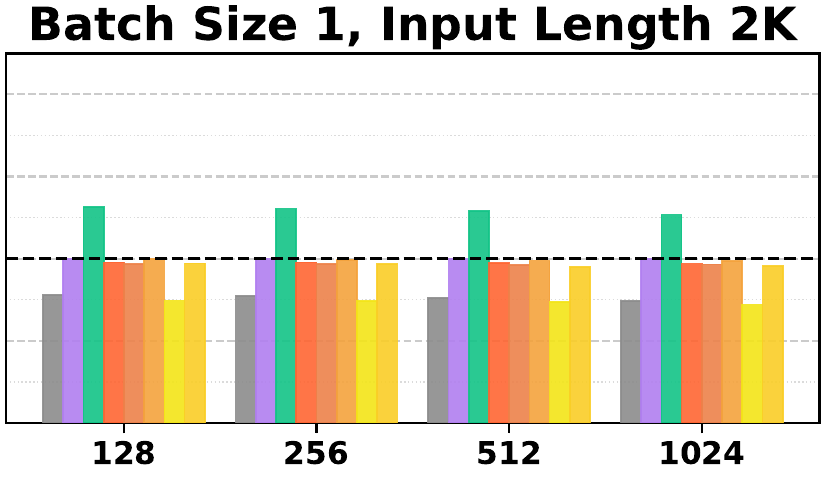}\hfill
\includegraphics[width=0.30826043\textwidth]{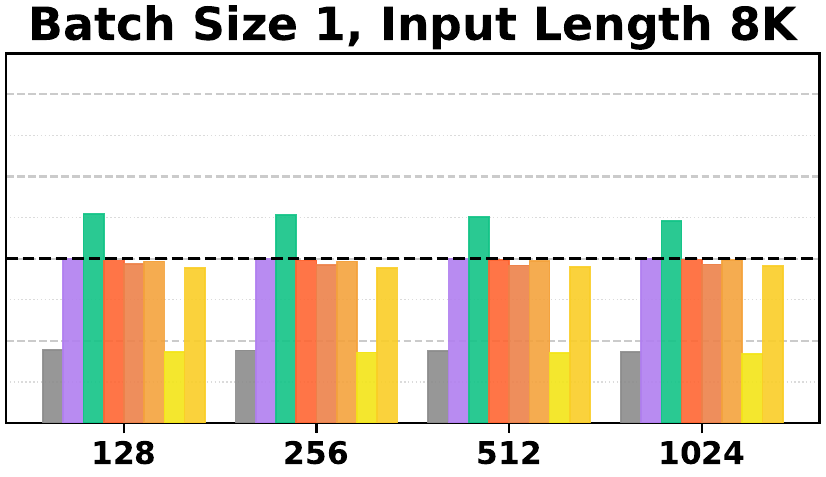}\\[-3pt]
\includegraphics[width=0.36347914\textwidth]{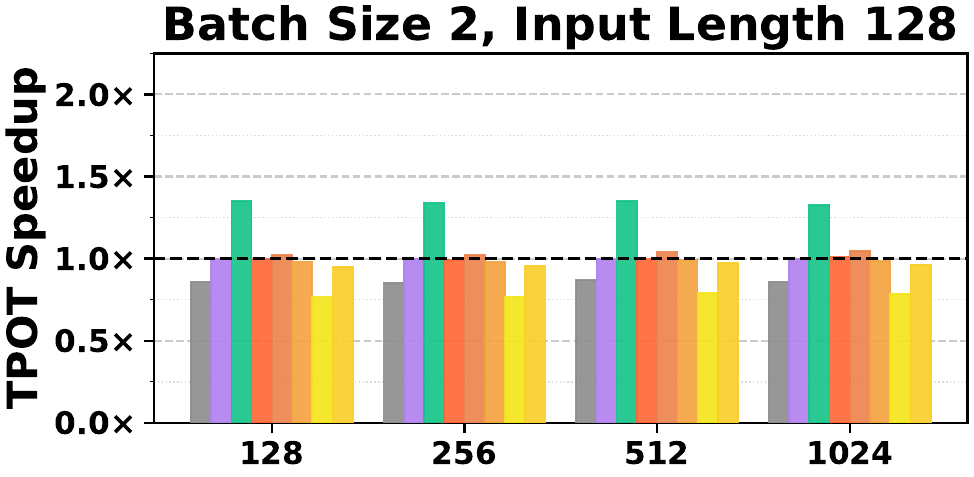}\hfill
\includegraphics[width=0.30826043\textwidth]{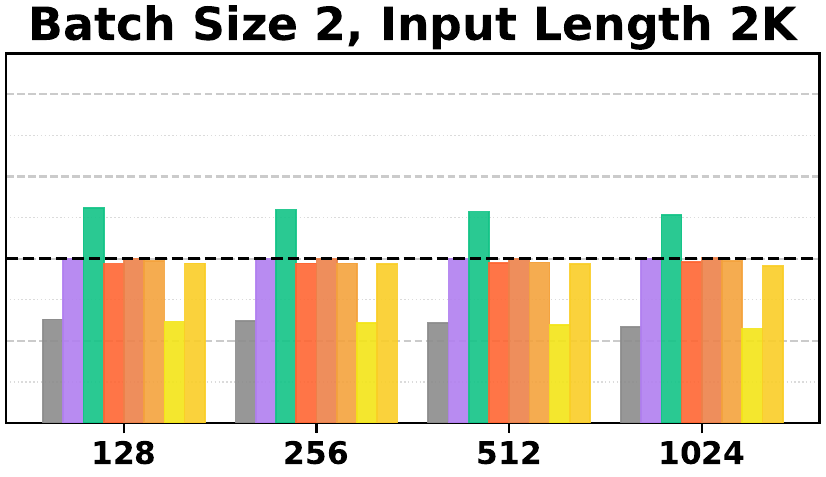}\hfill
\includegraphics[width=0.30826043\textwidth]{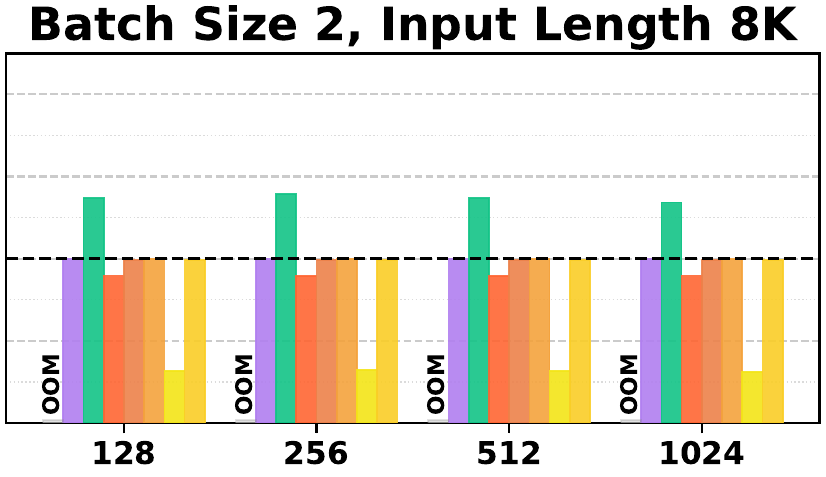}\\[-3pt]
\includegraphics[width=0.36347914\textwidth]{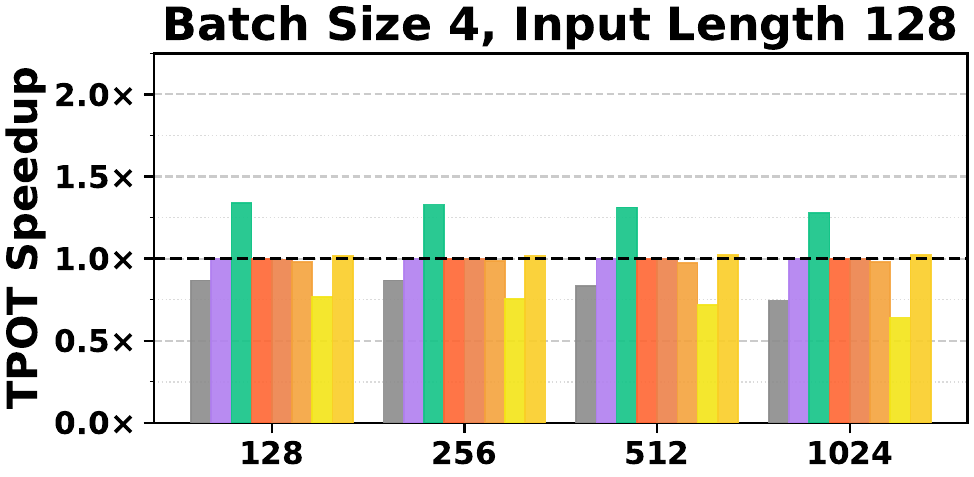}\hfill
\includegraphics[width=0.30826043\textwidth]{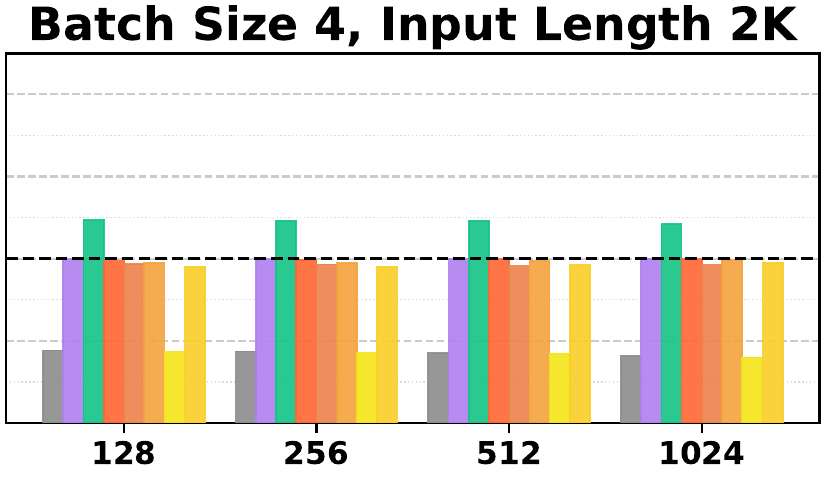}\hfill
\includegraphics[width=0.30826043\textwidth]{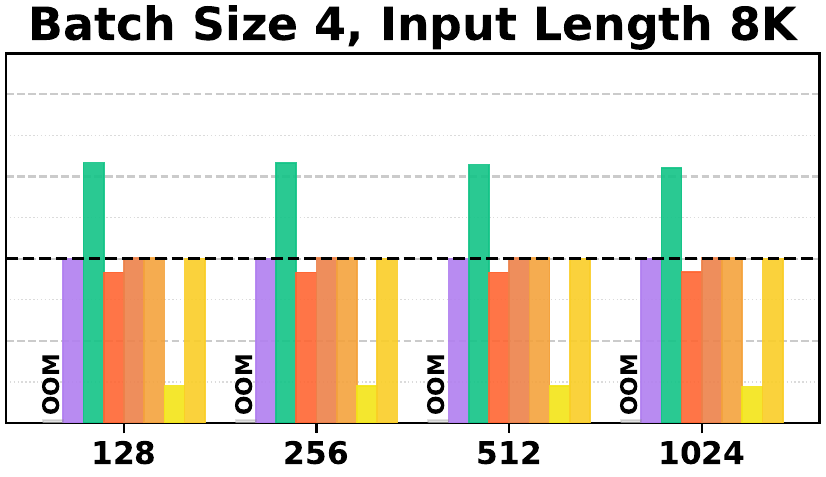}\\[-3pt]
\includegraphics[width=0.36347914\textwidth]{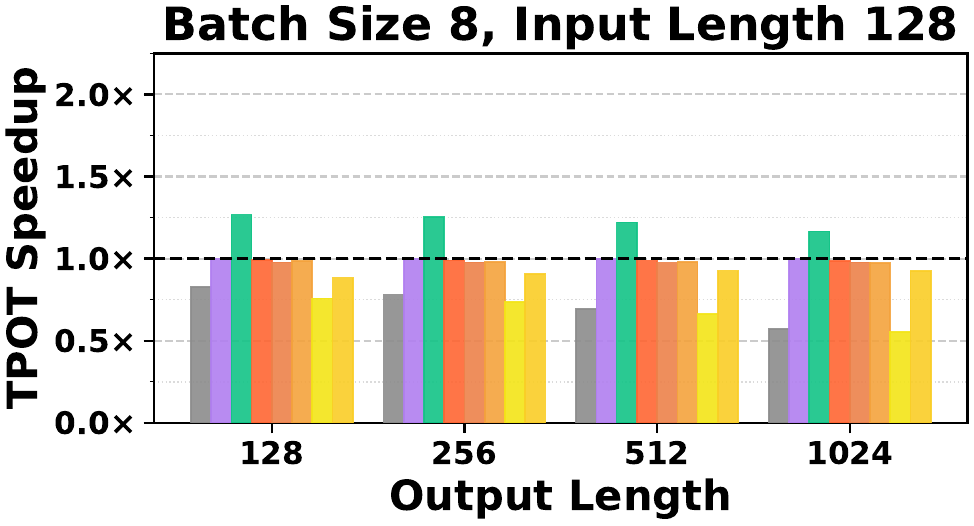}\hfill
\includegraphics[width=0.30826043\textwidth]{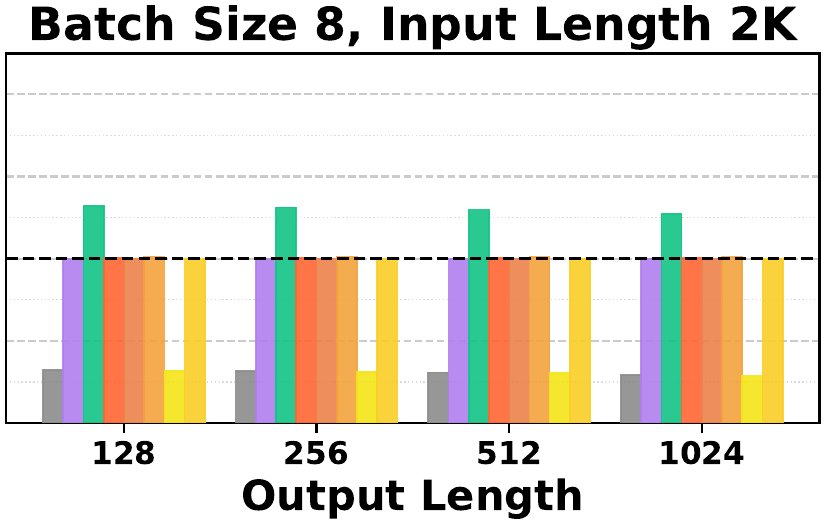}\hfill
\includegraphics[width=0.30826043\textwidth]{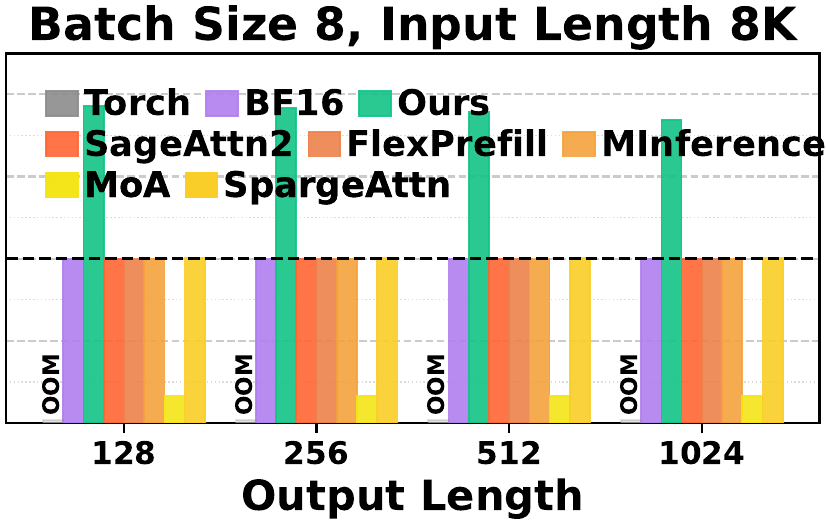}
\vspace{-8pt}
\caption{\textbf{TPOT speedup on A100-SXM4-40GB.} Ministral-3-8B-Instruct-2512-BF16. Rows use batch sizes 1, 2, 4, and 8; columns use 128, 2,048, 8,192 input tokens. Each panel varies output length over 128, 256, 512, and 1024 tokens. BF16 is normalized to $1\times$; higher is better. Method colors and order are fixed across panels.}
\label{fig:system-a100-ministral3_8b-tpot}
\par\vspace{4pt}
\begin{minipage}{\textwidth}
\normalsize
\setlength{\multicolsep}{0pt}
\makeatletter
\renewcommand{\paragraph}{\@startsection{paragraph}{4}{\z@}{5pt}{-1em}{\normalfont\normalsize\bfseries}}
\makeatother
\begin{multicols}{2}
\raggedcolumns
The long-context sweep reaches 42.95 tokens/s for Llama with a 128K prompt, 1,024 generated tokens, and batch size one, compared with 16.19 tokens/s for BF16, giving the $2.65\times$ speedup. Figure~\ref{fig:system-a100-llama32_3b-tpot} complements the main-text throughput grid, and Figures~\ref{fig:system-a100-qwen2_7b-throughput}--\ref{fig:system-a100-ministral3_8b-tpot} cover the other models.\par
\end{multicols}
\end{minipage}
\end{figure*}

\begin{figure*}[!tp]
\centering
\includegraphics[width=0.36347914\textwidth]{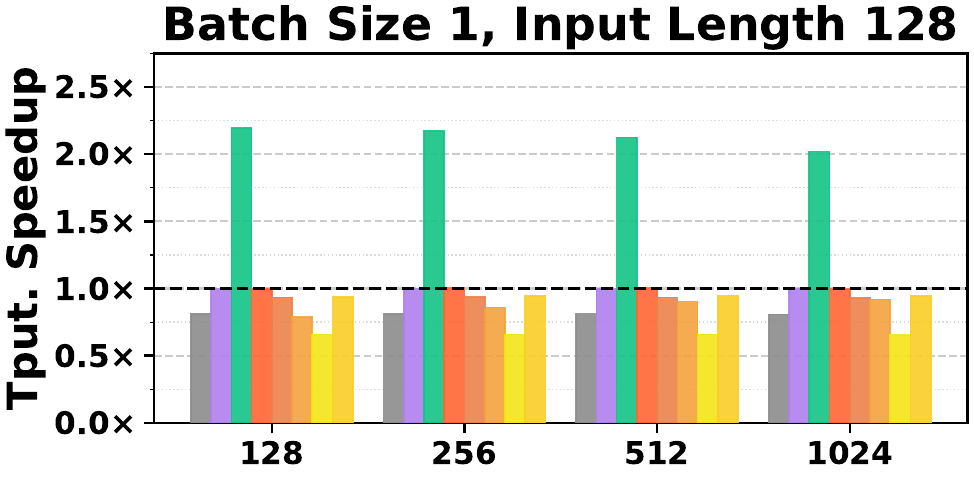}\hfill
\includegraphics[width=0.30826043\textwidth]{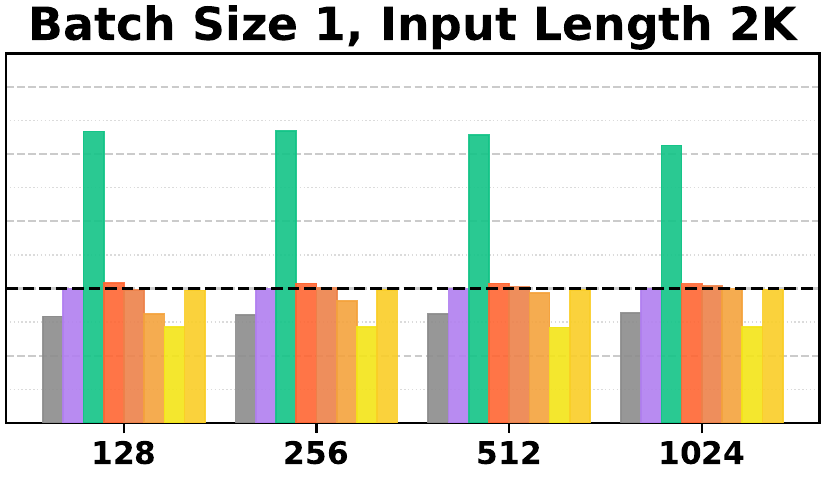}\hfill
\includegraphics[width=0.30826043\textwidth]{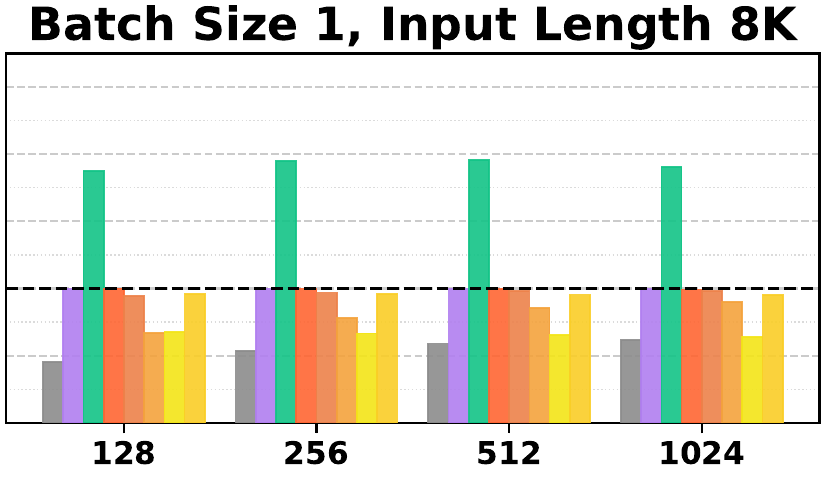}\\[-3pt]
\includegraphics[width=0.36347914\textwidth]{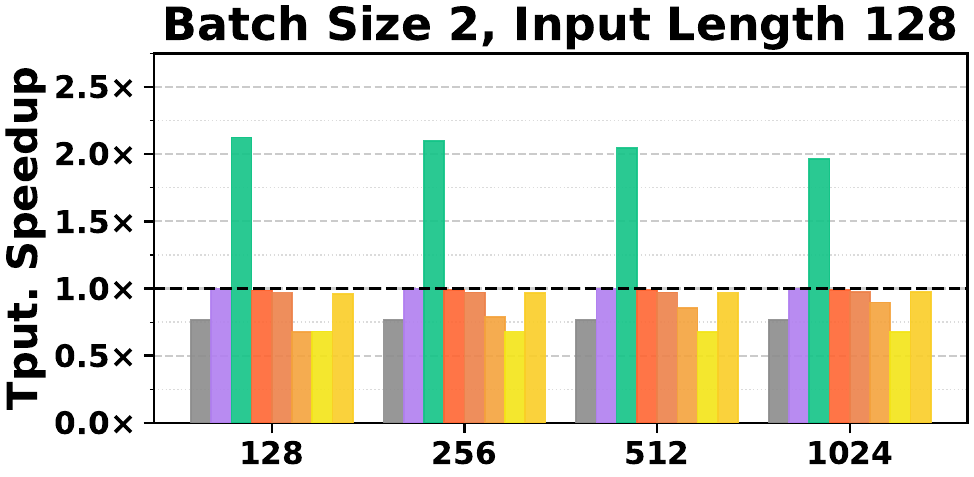}\hfill
\includegraphics[width=0.30826043\textwidth]{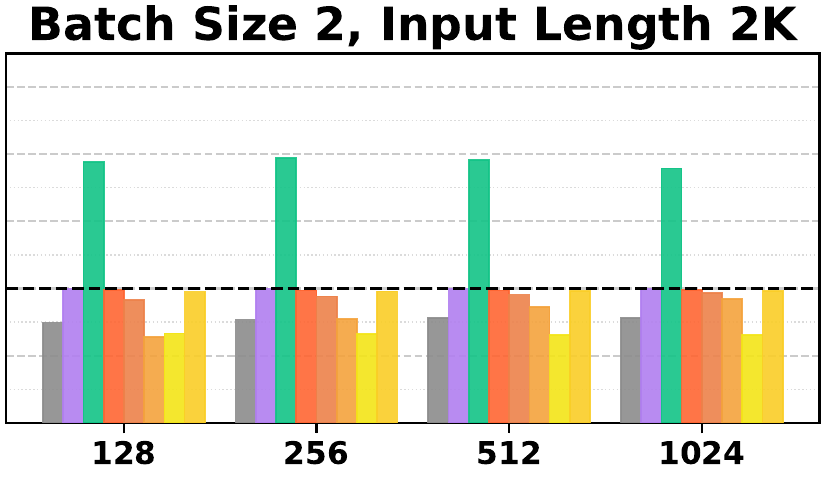}\hfill
\includegraphics[width=0.30826043\textwidth]{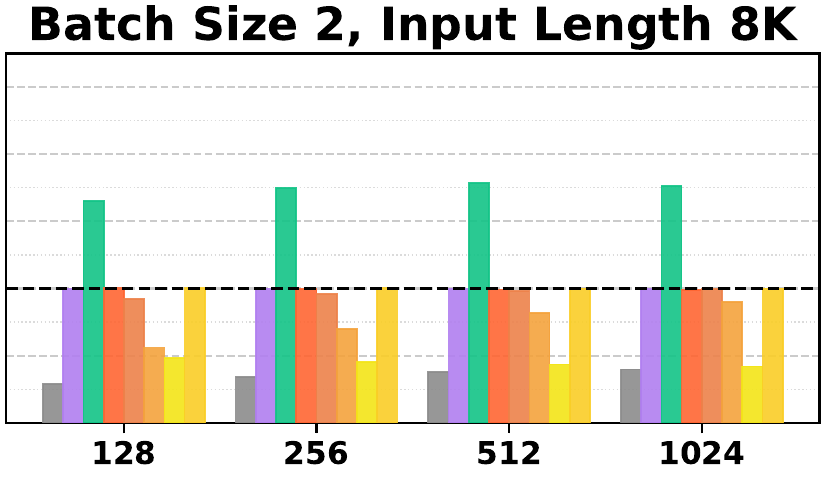}
\vspace{-8pt}
\caption{\textbf{Generation-throughput ratio on H100 NVL.} Llama-3.2-3B-Instruct (prompt group 1 of 2). Rows use batch sizes 1 and 2; columns use 128, 2,048, 8,192 input tokens. Each panel varies output length over 128, 256, 512, and 1024 tokens. BF16 is normalized to $1\times$; higher is better. Method colors and order are fixed across panels.}
\label{fig:system-h100-llama32_3b-throughput-1}
\end{figure*}

\begin{figure*}[!tp]
\centering
\includegraphics[width=0.36347914\textwidth]{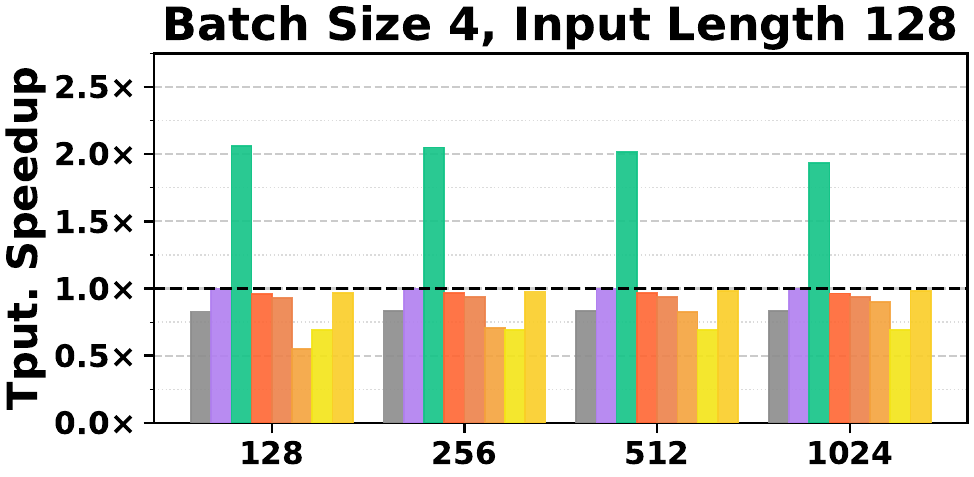}\hfill
\includegraphics[width=0.30826043\textwidth]{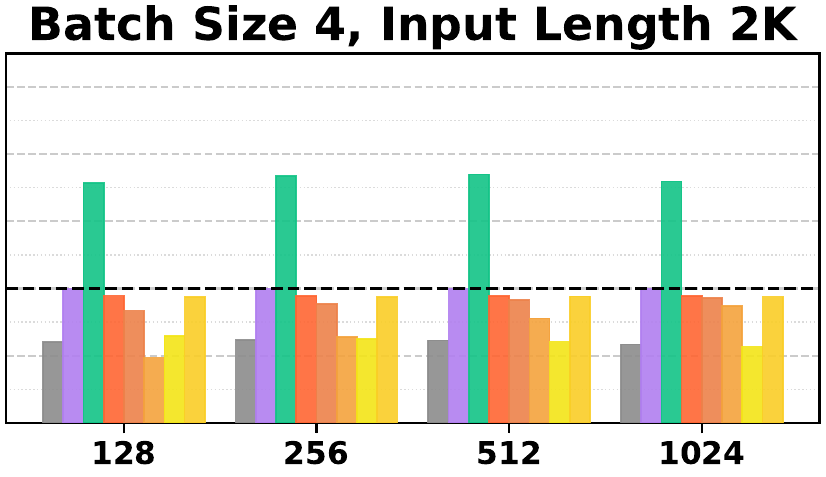}\hfill
\includegraphics[width=0.30826043\textwidth]{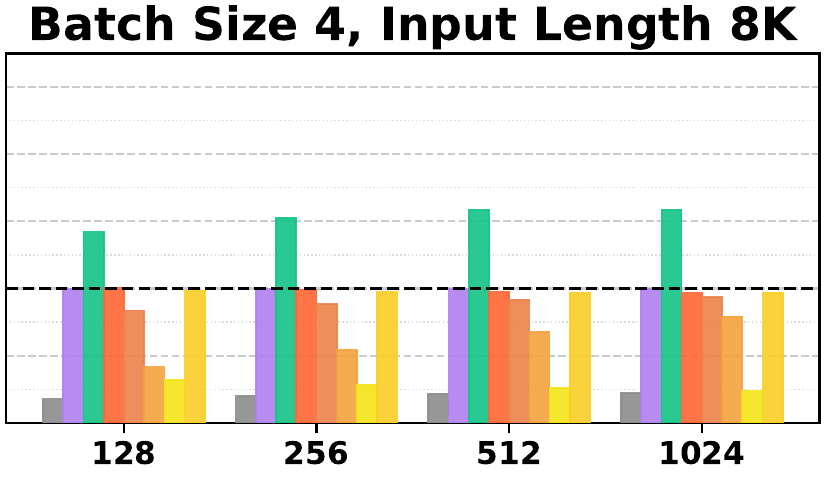}\\[-3pt]
\includegraphics[width=0.36347914\textwidth]{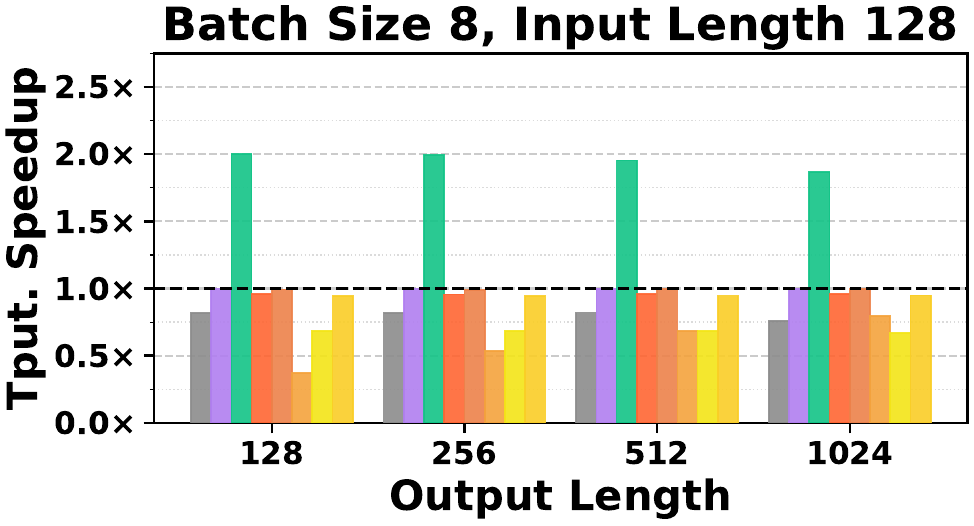}\hfill
\includegraphics[width=0.30826043\textwidth]{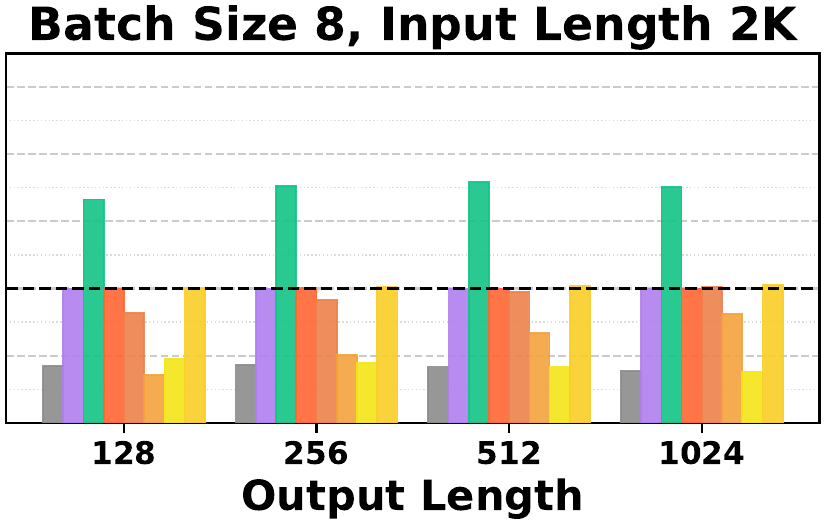}\hfill
\includegraphics[width=0.30826043\textwidth]{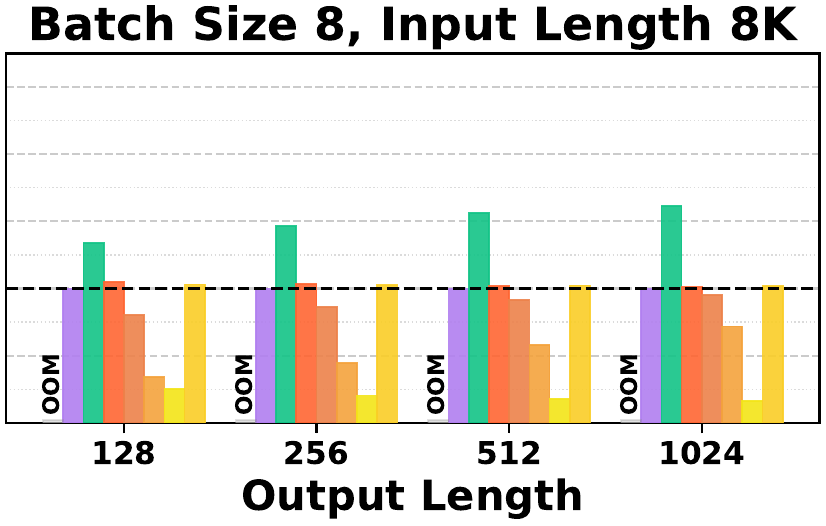}
\vspace{-8pt}
\addtocounter{figure}{-1}
\renewcommand{\theHfigure}{\arabic{figure}.continued}
\caption[]{\textbf{Continued.} Rows use batch sizes 4 and 8; panel settings follow the first part.}
\end{figure*}

\begin{figure*}[!tp]
\centering
\includegraphics[width=0.36347914\textwidth]{figures/efficiency/h100/llama32_3b/11_workload_generation_throughput/a_baselines/batch_1/input_128.pdf}\hfill
\includegraphics[width=0.30826043\textwidth]{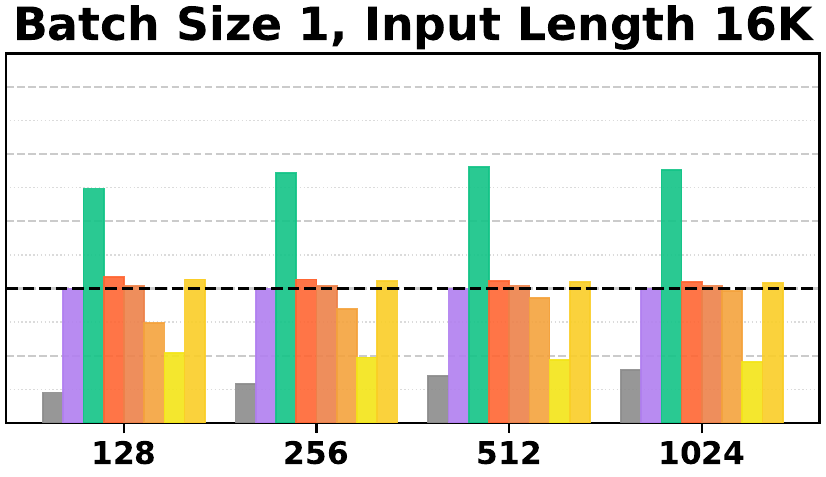}\hfill
\includegraphics[width=0.30826043\textwidth]{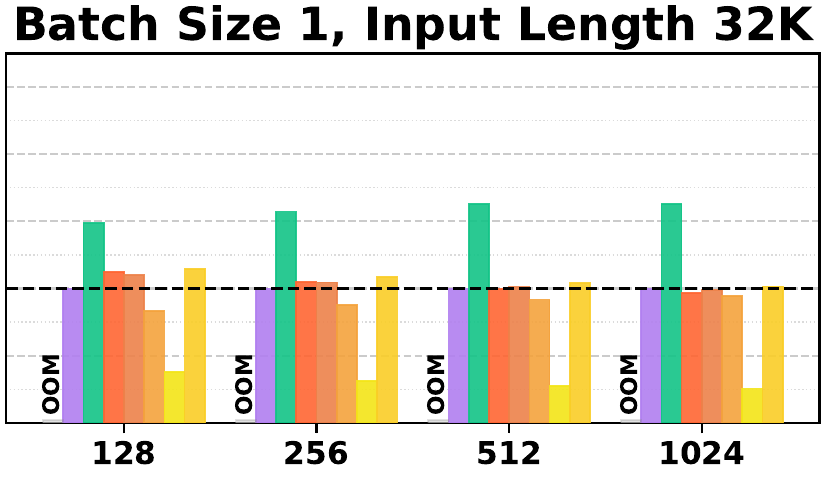}\\[-3pt]
\includegraphics[width=0.36347914\textwidth]{figures/efficiency/h100/llama32_3b/11_workload_generation_throughput/a_baselines/batch_2/input_128.pdf}\hfill
\includegraphics[width=0.30826043\textwidth]{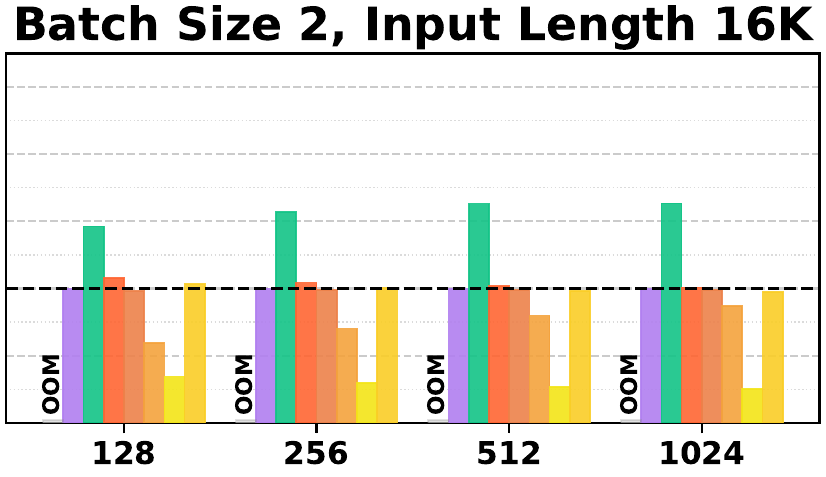}\hfill
\includegraphics[width=0.30826043\textwidth]{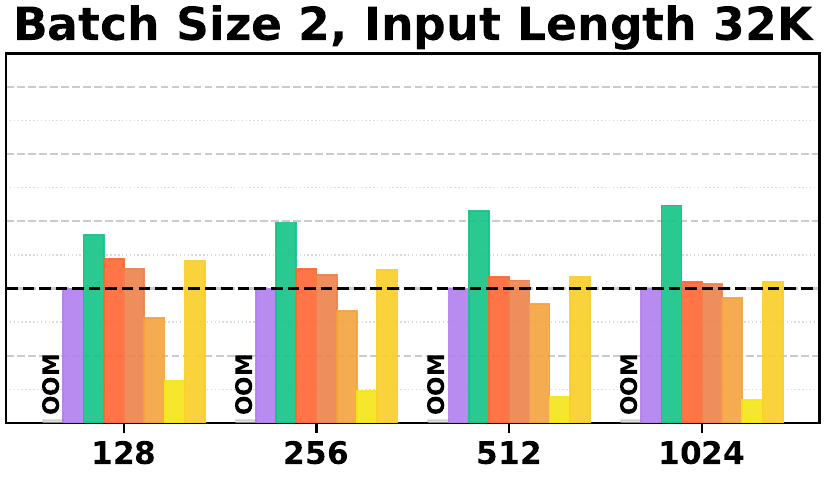}\\[-3pt]
\includegraphics[width=0.36347914\textwidth]{figures/efficiency/h100/llama32_3b/11_workload_generation_throughput/a_baselines/batch_4/input_128.pdf}\hfill
\includegraphics[width=0.30826043\textwidth]{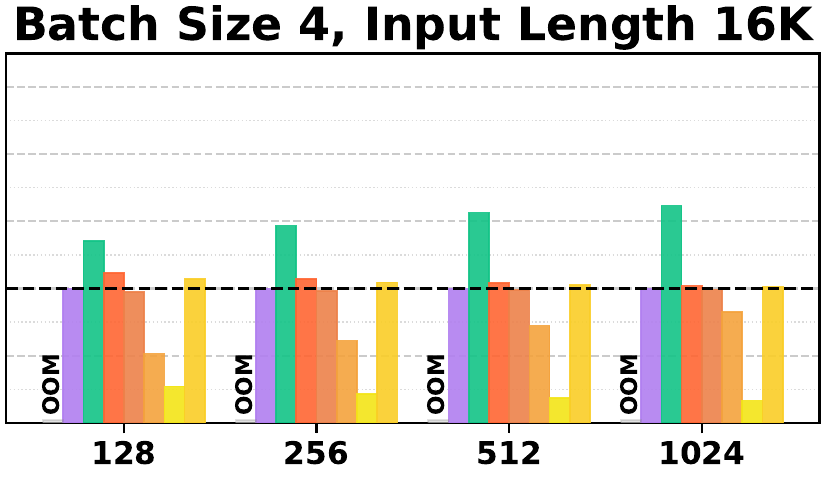}\hfill
\includegraphics[width=0.30826043\textwidth]{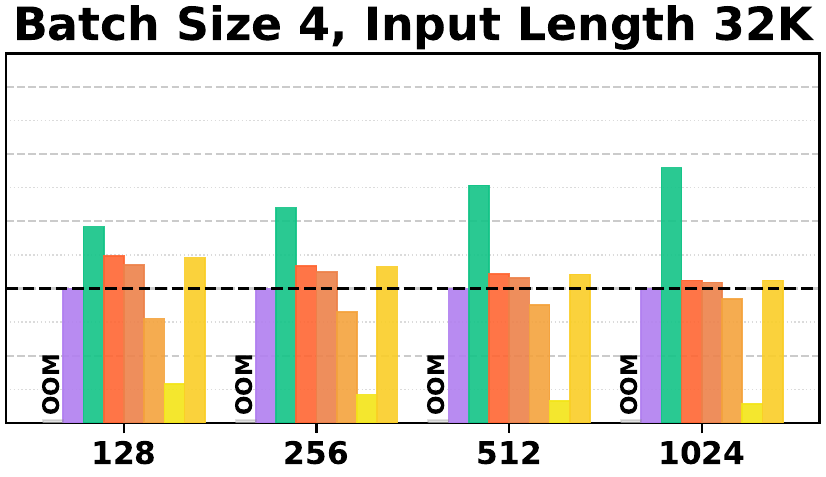}\\[-3pt]
\includegraphics[width=0.36347914\textwidth]{figures/efficiency/h100/llama32_3b/11_workload_generation_throughput/a_baselines/batch_8/input_128.pdf}\hfill
\includegraphics[width=0.30826043\textwidth]{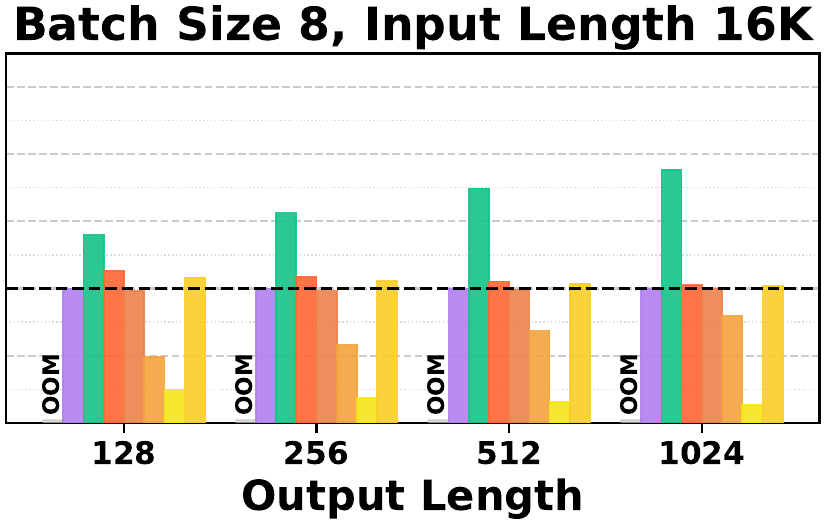}\hfill
\includegraphics[width=0.30826043\textwidth]{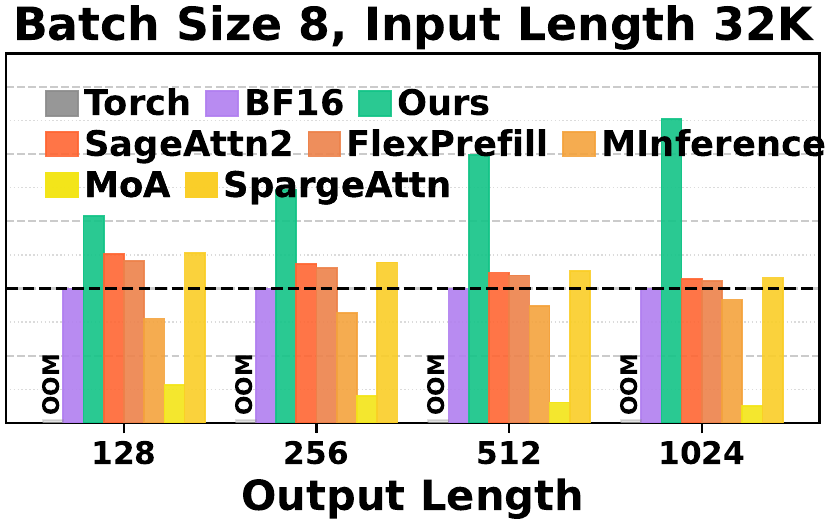}
\vspace{-8pt}
\caption{\textbf{Generation-throughput ratio on H100 NVL.} Llama-3.2-3B-Instruct (prompt group 2 of 2). Rows use batch sizes 1, 2, 4, and 8; columns use 128, 16,384, 32,768 input tokens. Each panel varies output length over 128, 256, 512, and 1024 tokens. BF16 is normalized to $1\times$; higher is better. The 128-token column repeats the first group as a shared reference. Panels share colors and order.}
\label{fig:system-h100-llama32_3b-throughput-2}
\end{figure*}

\begin{figure*}[!tp]
\centering
\includegraphics[width=0.36347914\textwidth]{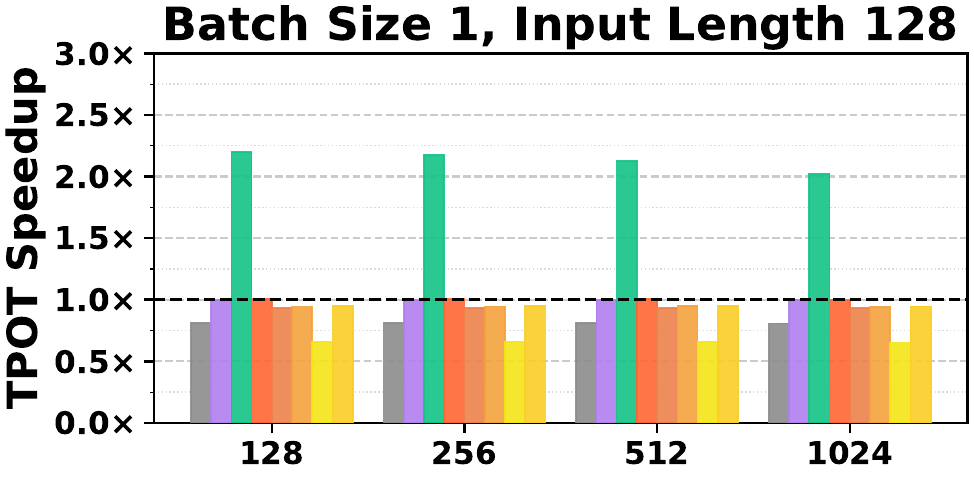}\hfill
\includegraphics[width=0.30826043\textwidth]{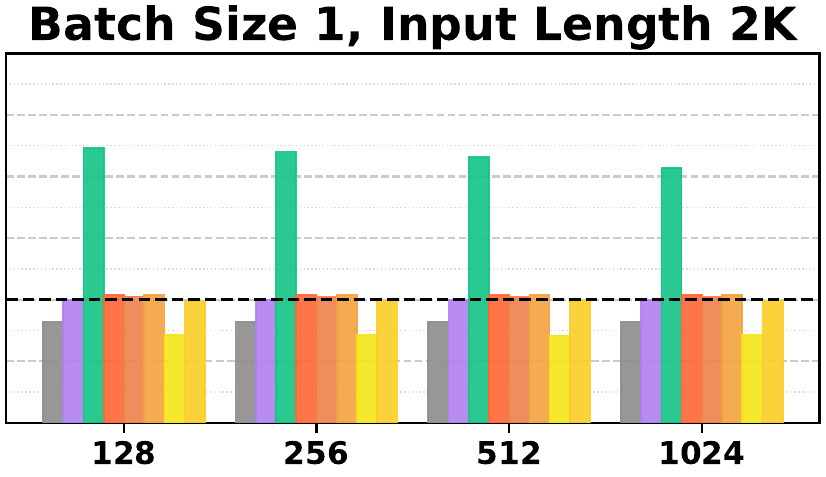}\hfill
\includegraphics[width=0.30826043\textwidth]{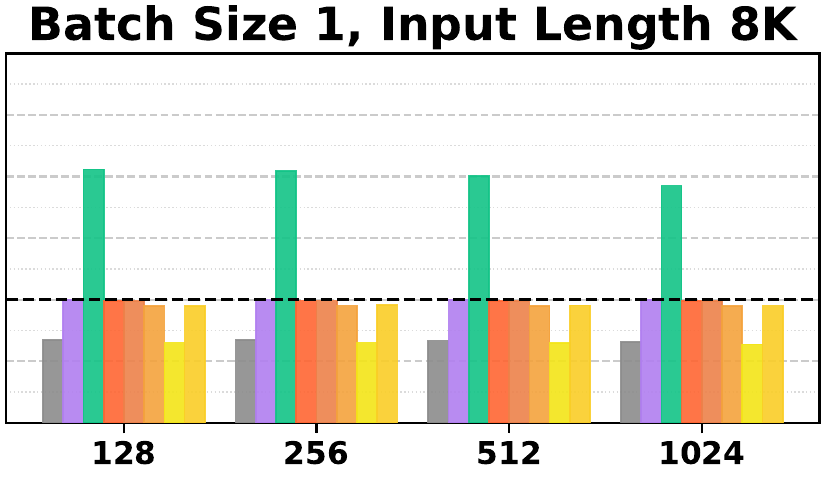}\\[-3pt]
\includegraphics[width=0.36347914\textwidth]{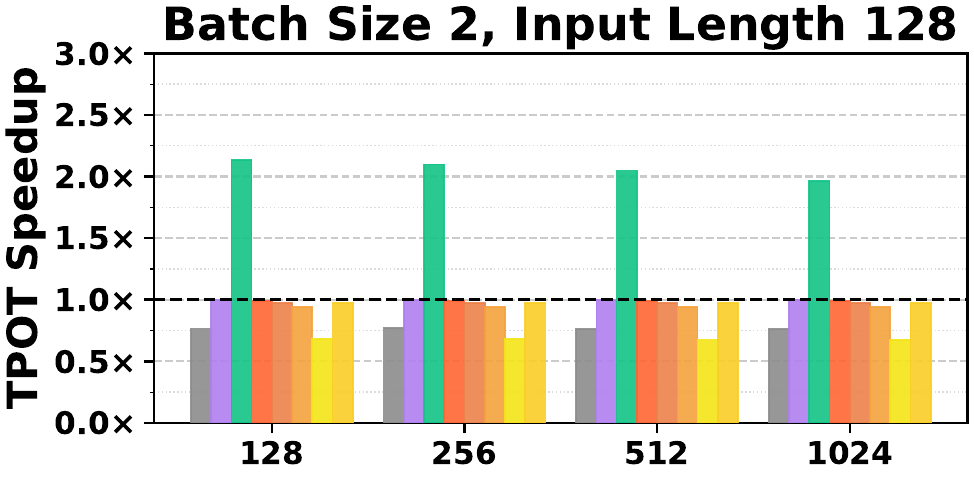}\hfill
\includegraphics[width=0.30826043\textwidth]{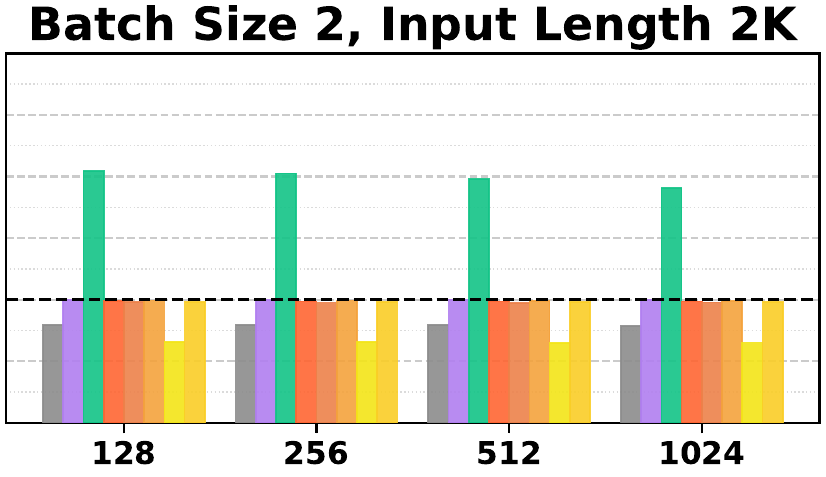}\hfill
\includegraphics[width=0.30826043\textwidth]{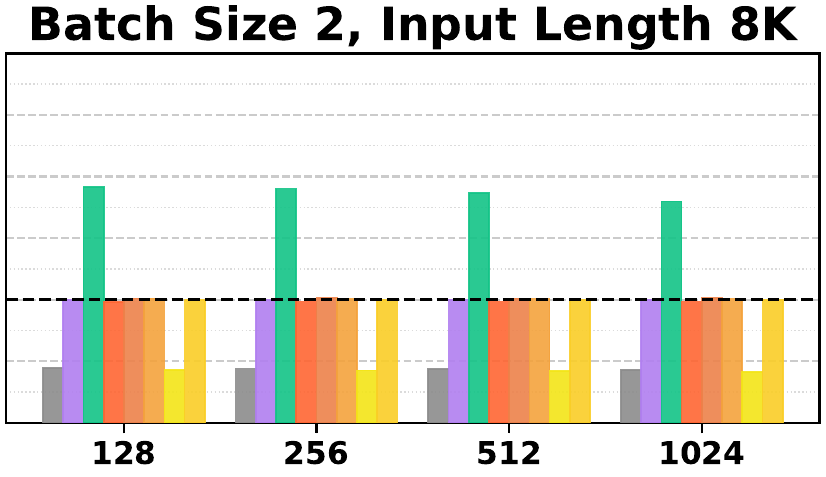}\\[-3pt]
\includegraphics[width=0.36347914\textwidth]{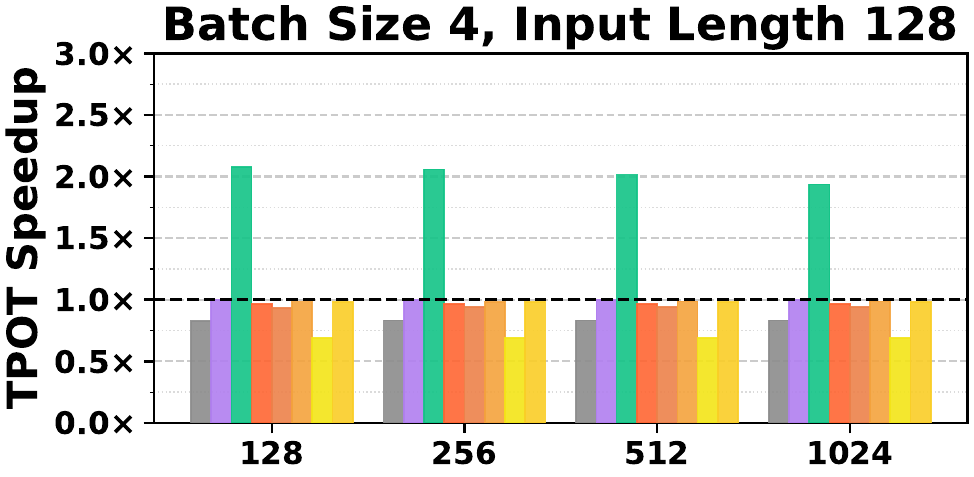}\hfill
\includegraphics[width=0.30826043\textwidth]{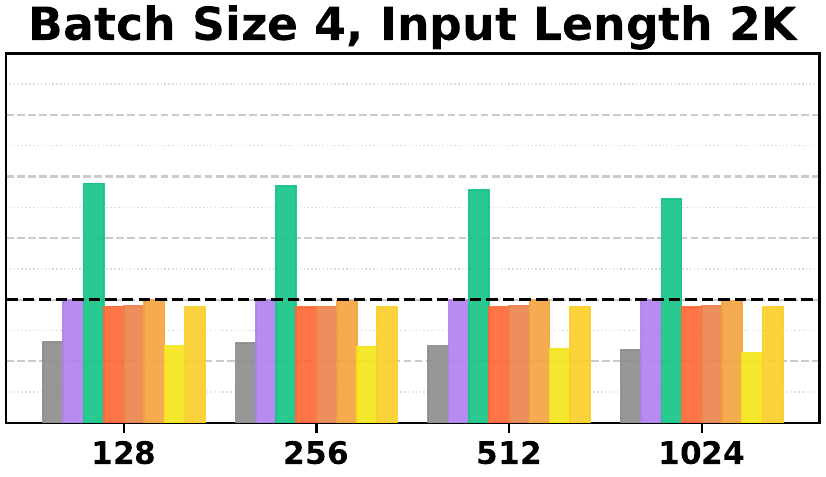}\hfill
\includegraphics[width=0.30826043\textwidth]{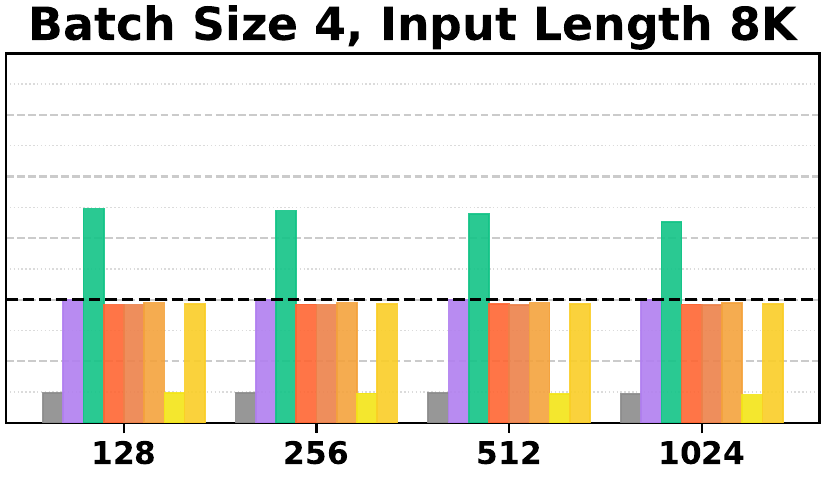}\\[-3pt]
\includegraphics[width=0.36347914\textwidth]{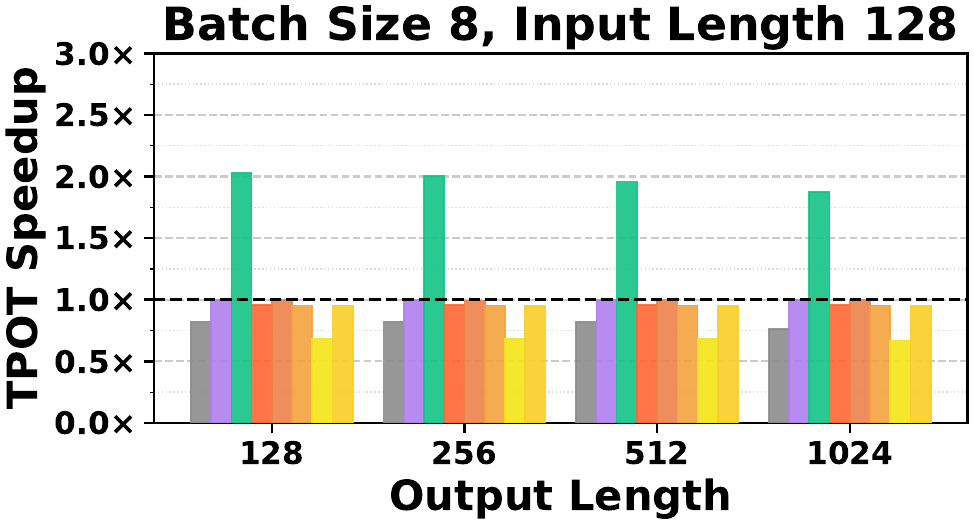}\hfill
\includegraphics[width=0.30826043\textwidth]{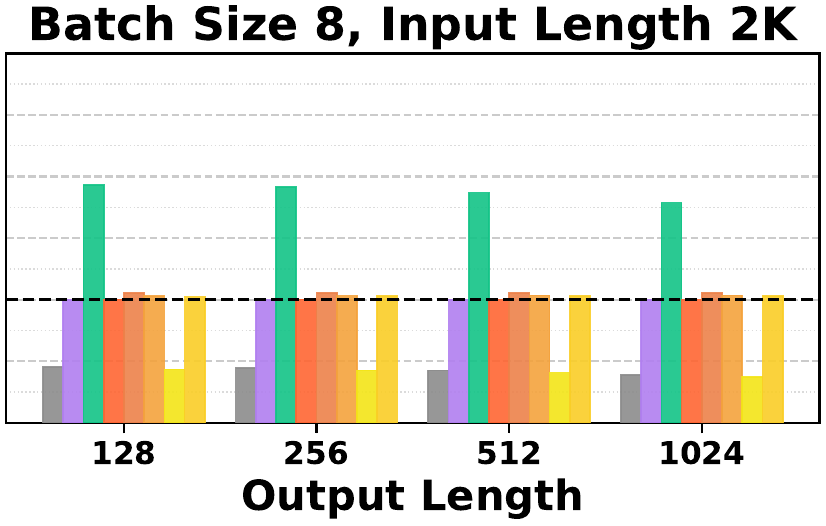}\hfill
\includegraphics[width=0.30826043\textwidth]{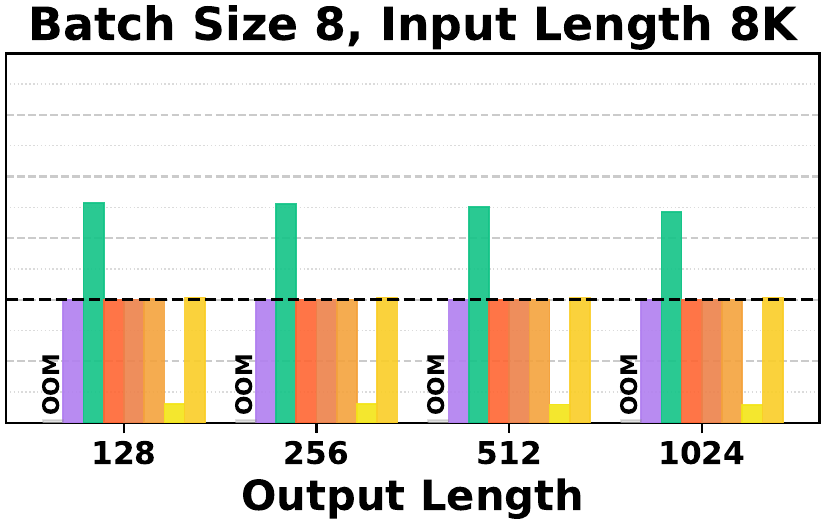}
\vspace{-8pt}
\caption{\textbf{TPOT speedup on H100 NVL.} Llama-3.2-3B-Instruct (prompt group 1 of 2). Rows use batch sizes 1, 2, 4, and 8; columns use 128, 2,048, 8,192 input tokens. Each panel varies output length over 128, 256, 512, and 1024 tokens. BF16 is normalized to $1\times$; higher is better. Method colors and order are fixed across panels.}
\label{fig:system-h100-llama32_3b-tpot-1}
\par\vspace{4pt}
\begin{minipage}{\textwidth}
\normalsize
\setlength{\multicolsep}{0pt}
\makeatletter
\renewcommand{\paragraph}{\@startsection{paragraph}{4}{\z@}{5pt}{-1em}{\normalfont\normalsize\bfseries}}
\makeatother
\begin{multicols}{2}
\raggedcolumns
\textbf{H100 NVL.}
On H100 NVL, Llama reaches 318.36 tokens/s with an 8K prompt, 128 generated tokens, and batch size four, compared with 40.42 tokens/s for Torch attention and 223.61 tokens/s for BF16 (Figures~\ref{fig:system-h100-llama32_3b-throughput-1}--\ref{fig:system-h100-qwen2_7b-tpot-1} and~\ref{fig:system-h100-qwen2_7b-throughput-2}--\ref{fig:system-h100-ministral3_8b-tpot}).\par
\columnbreak
These complete-request measurements give throughput gains of $7.88\times$ and $1.42\times$ over the respective references.\par
\end{multicols}
\end{minipage}
\end{figure*}

\begin{figure*}[!tp]
\centering
\includegraphics[width=0.36347914\textwidth]{figures/efficiency/h100/llama32_3b/10_workload_tpot/a_baselines/batch_1/input_128.pdf}\hfill
\includegraphics[width=0.30826043\textwidth]{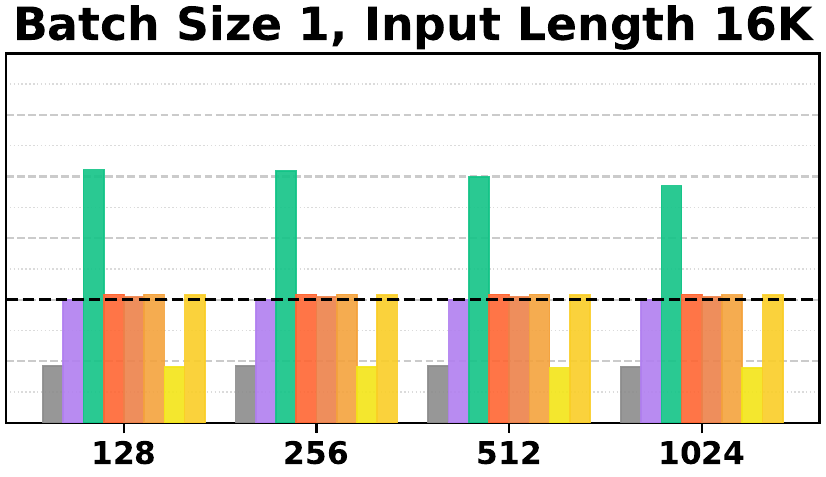}\hfill
\includegraphics[width=0.30826043\textwidth]{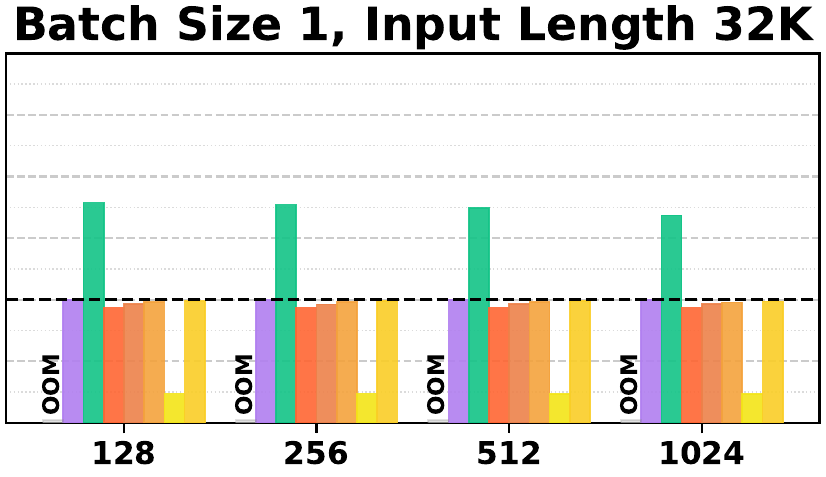}\\[-3pt]
\includegraphics[width=0.36347914\textwidth]{figures/efficiency/h100/llama32_3b/10_workload_tpot/a_baselines/batch_2/input_128.pdf}\hfill
\includegraphics[width=0.30826043\textwidth]{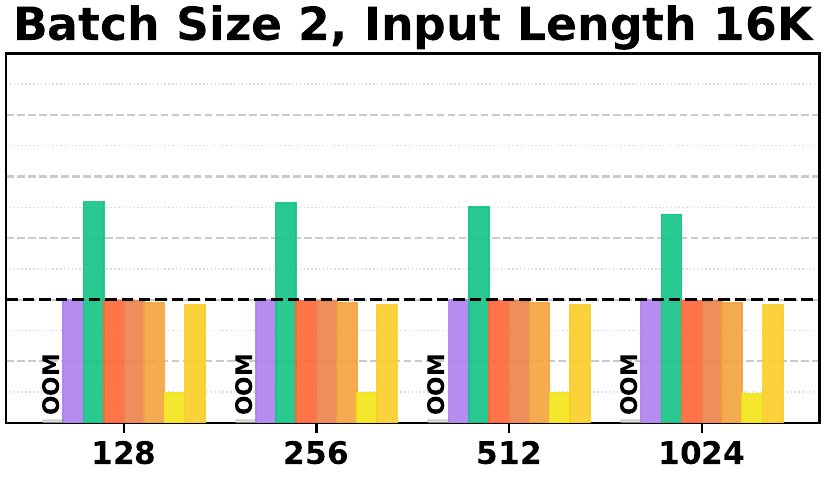}\hfill
\includegraphics[width=0.30826043\textwidth]{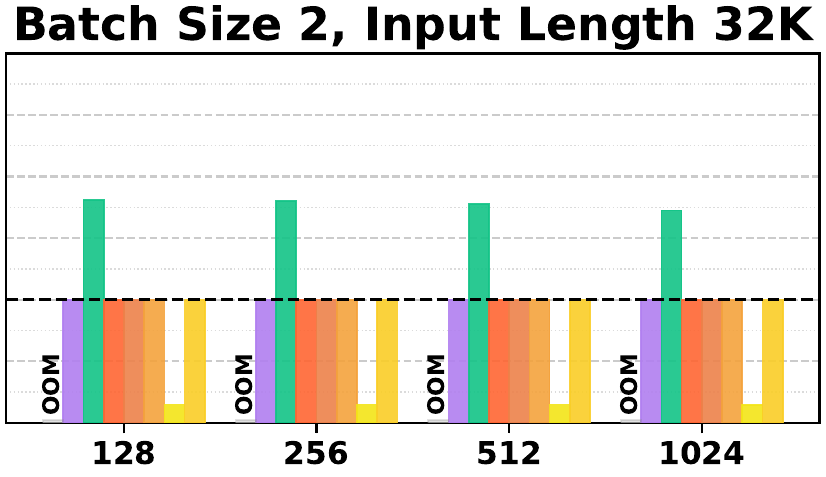}
\vspace{-8pt}
\caption{\textbf{TPOT speedup on H100 NVL.} Llama-3.2-3B-Instruct (prompt group 2 of 2). Rows use batch sizes 1 and 2; columns use 128, 16,384, 32,768 input tokens. Each panel varies output length over 128, 256, 512, and 1024 tokens. BF16 is normalized to $1\times$; higher is better. The 128-token column repeats the first group as a shared reference. Method colors and order are fixed across panels.}
\label{fig:system-h100-llama32_3b-tpot-2}
\end{figure*}

\begin{figure*}[!tp]
\centering
\includegraphics[width=0.36347914\textwidth]{figures/efficiency/h100/llama32_3b/10_workload_tpot/a_baselines/batch_4/input_128.pdf}\hfill
\includegraphics[width=0.30826043\textwidth]{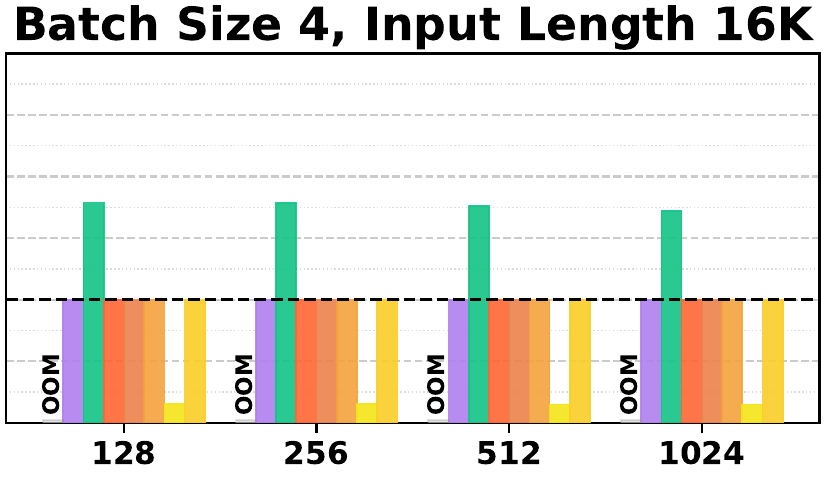}\hfill
\includegraphics[width=0.30826043\textwidth]{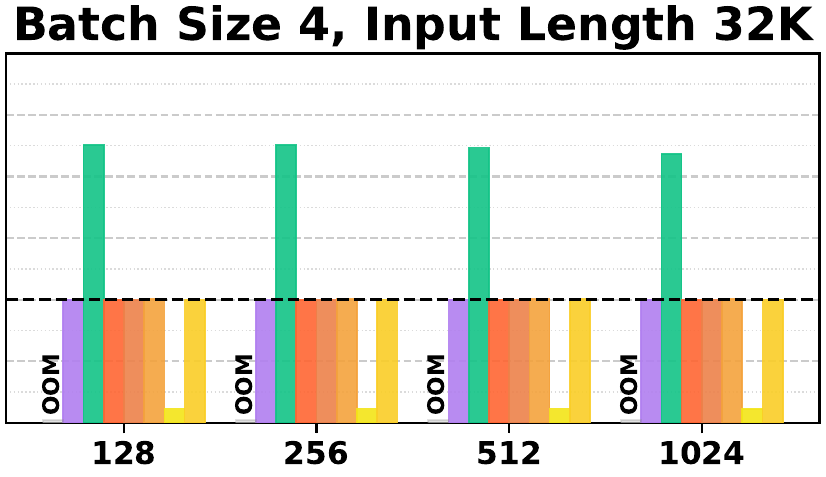}\\[-3pt]
\includegraphics[width=0.36347914\textwidth]{figures/efficiency/h100/llama32_3b/10_workload_tpot/a_baselines/batch_8/input_128.pdf}\hfill
\includegraphics[width=0.30826043\textwidth]{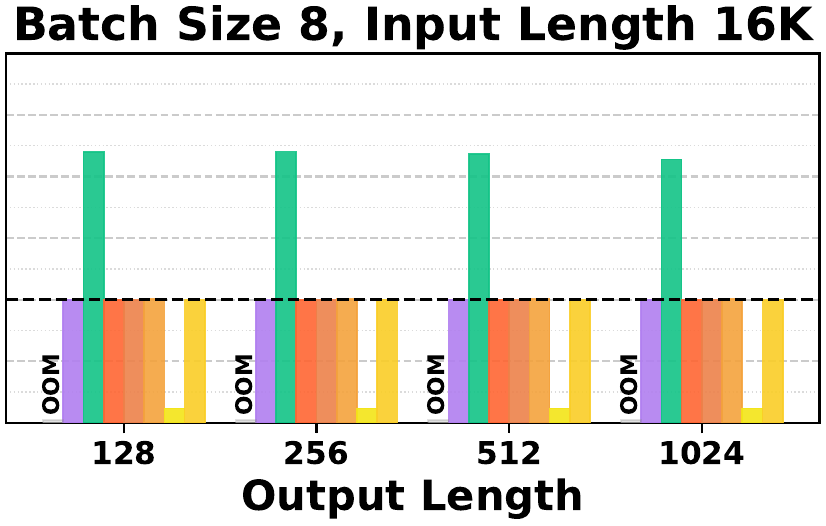}\hfill
\includegraphics[width=0.30826043\textwidth]{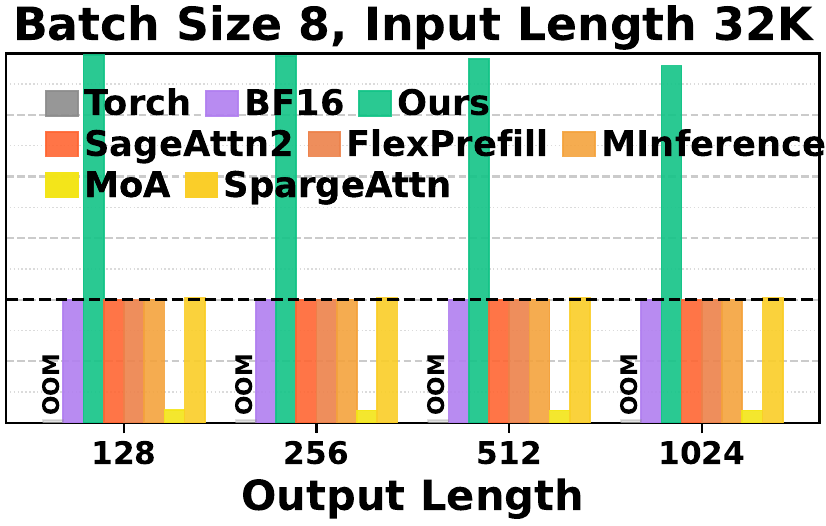}
\vspace{-8pt}
\addtocounter{figure}{-1}
\renewcommand{\theHfigure}{\arabic{figure}.continued}
\caption[]{\textbf{Continued.} Rows use batch sizes 4 and 8; panel settings follow the first part.}
\end{figure*}

\begin{figure*}[!tp]
\centering
\includegraphics[width=0.36347914\textwidth]{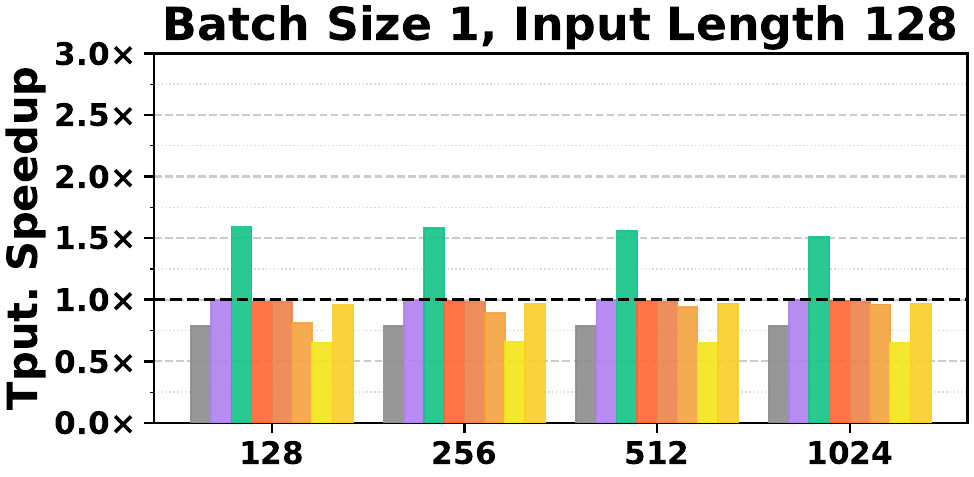}\hfill
\includegraphics[width=0.30826043\textwidth]{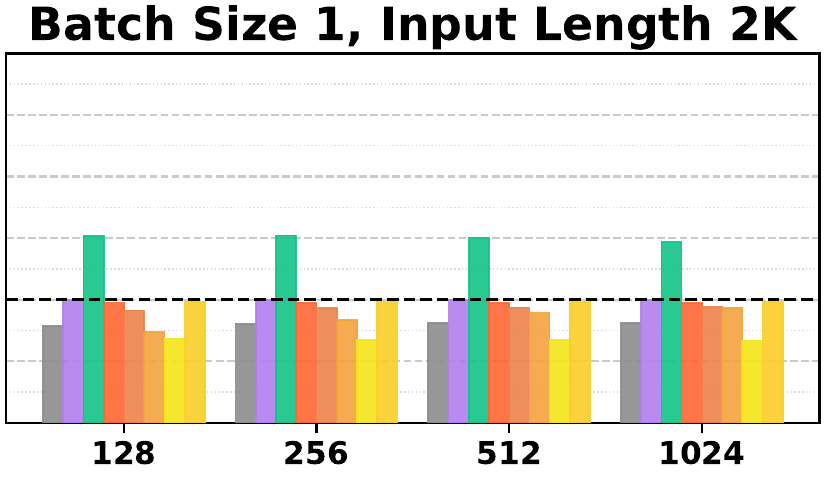}\hfill
\includegraphics[width=0.30826043\textwidth]{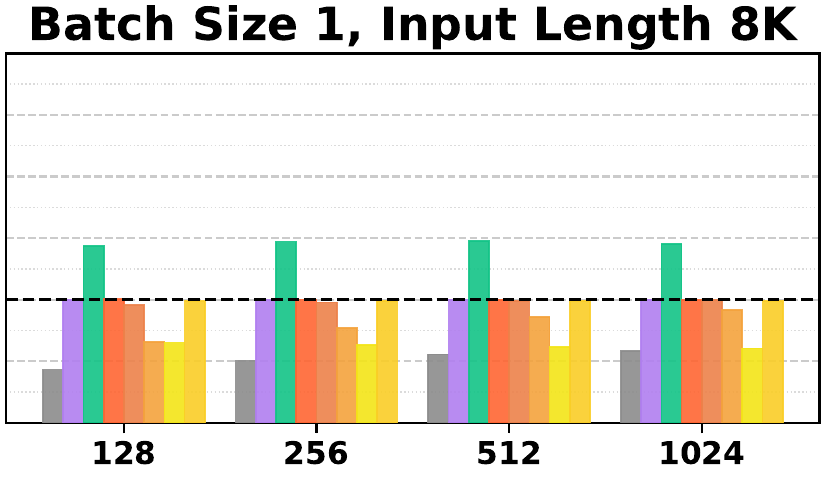}\\[-3pt]
\includegraphics[width=0.36347914\textwidth]{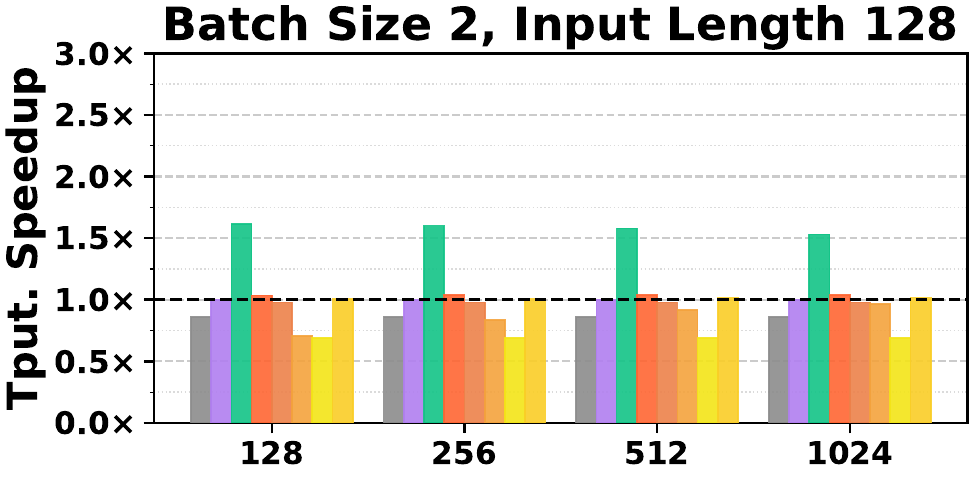}\hfill
\includegraphics[width=0.30826043\textwidth]{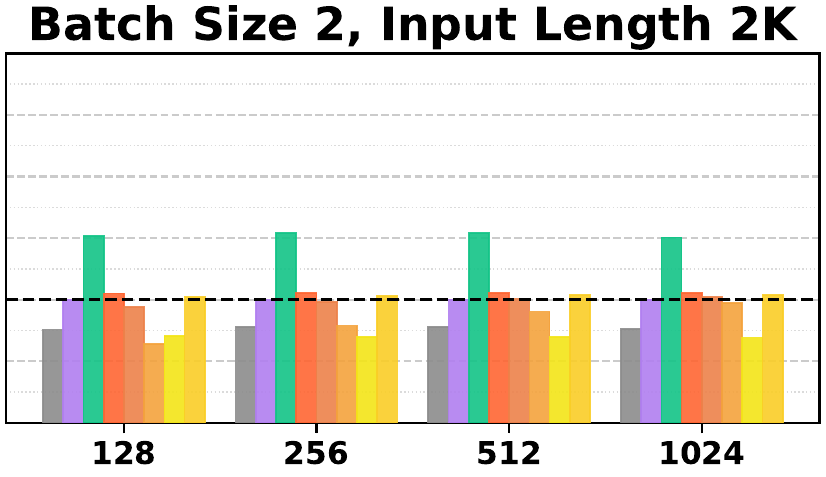}\hfill
\includegraphics[width=0.30826043\textwidth]{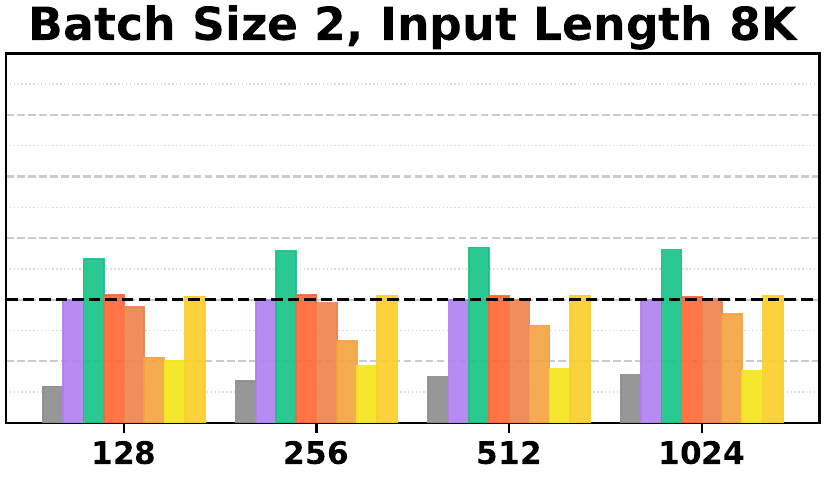}\\[-3pt]
\includegraphics[width=0.36347914\textwidth]{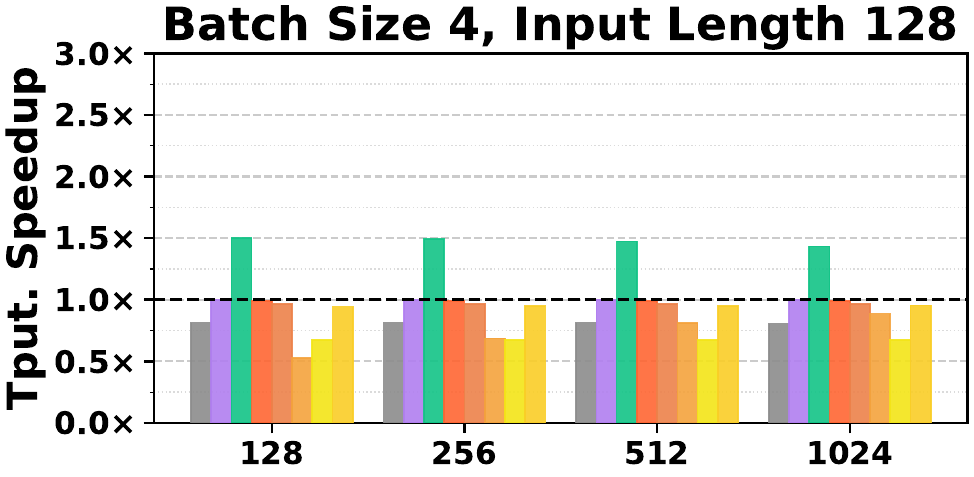}\hfill
\includegraphics[width=0.30826043\textwidth]{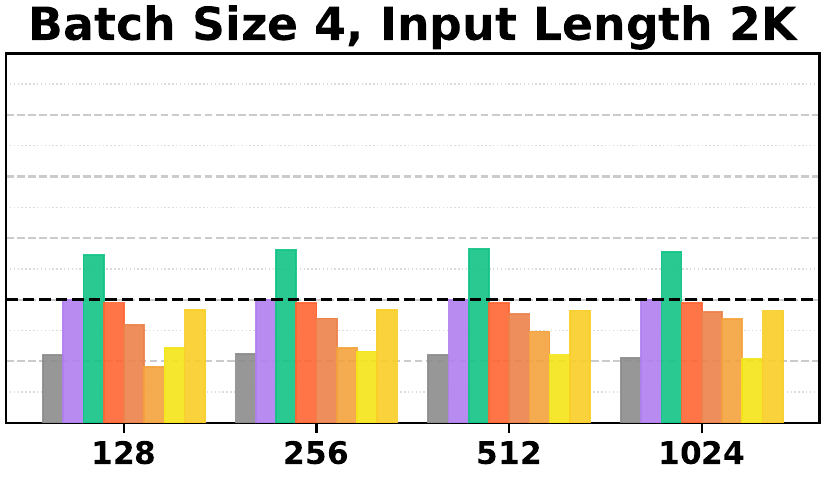}\hfill
\includegraphics[width=0.30826043\textwidth]{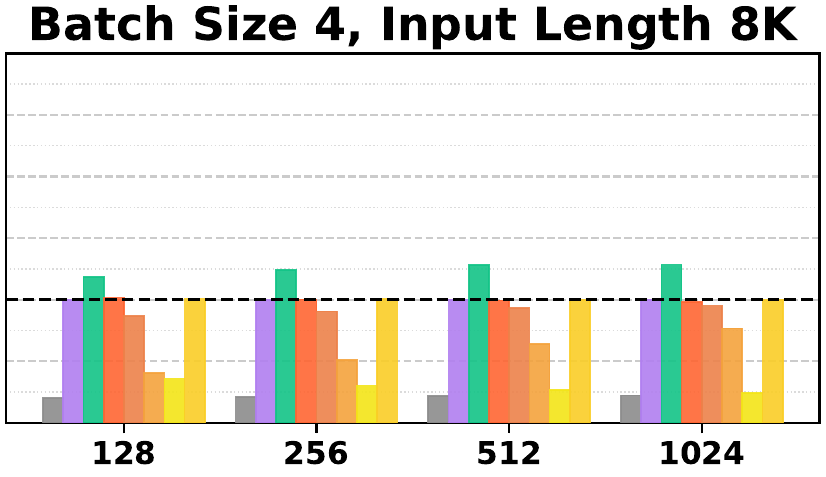}\\[-3pt]
\includegraphics[width=0.36347914\textwidth]{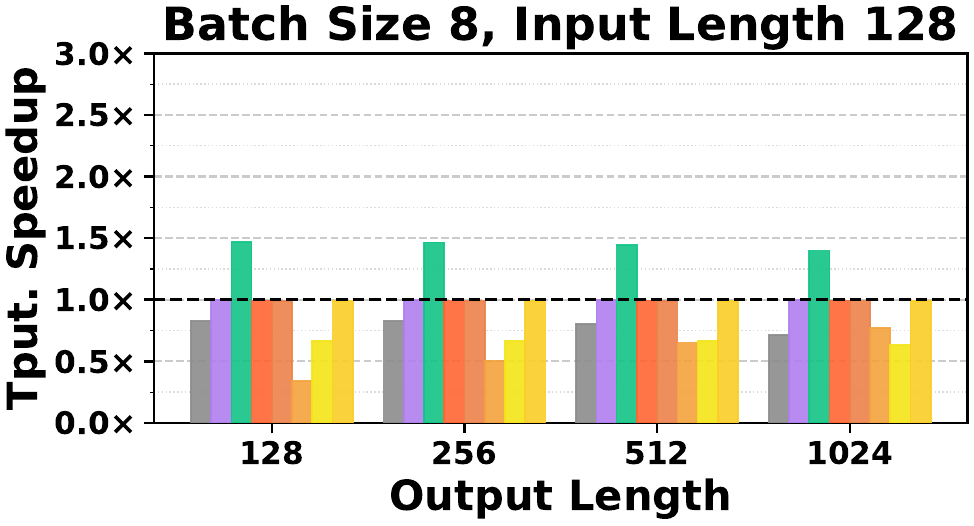}\hfill
\includegraphics[width=0.30826043\textwidth]{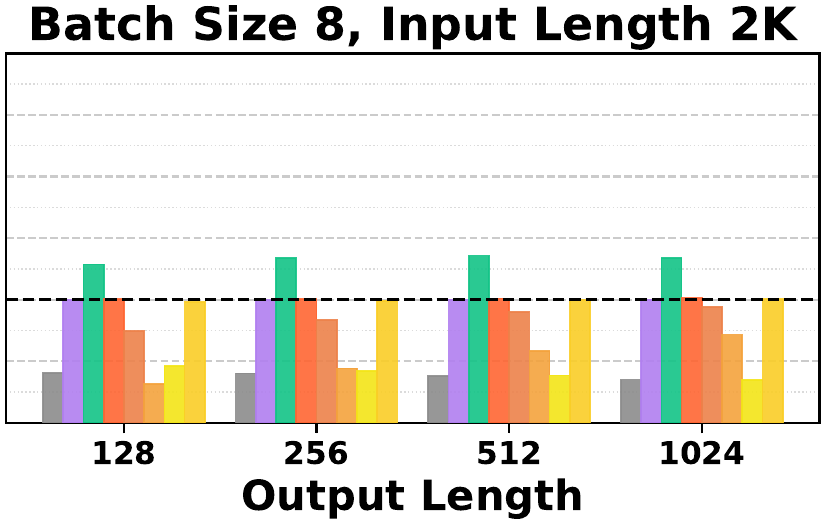}\hfill
\includegraphics[width=0.30826043\textwidth]{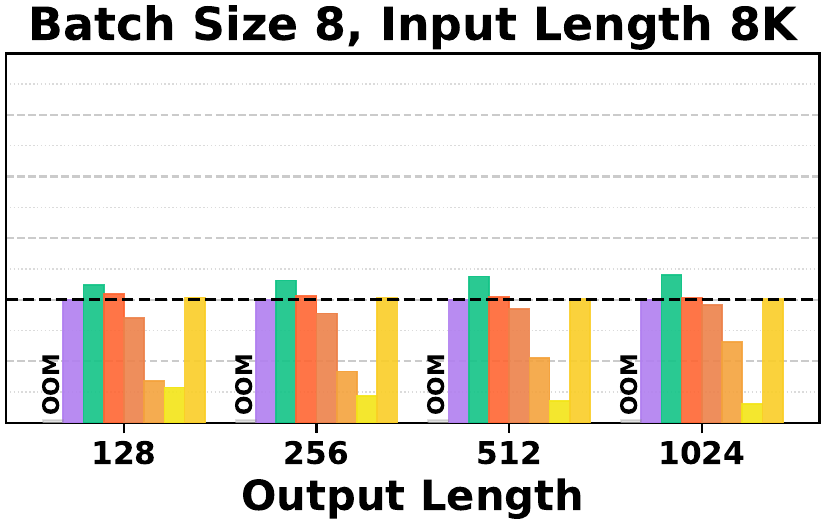}
\vspace{-8pt}
\caption{\textbf{Generation-throughput ratio on H100 NVL.} Qwen2-7B-Instruct (prompt group 1 of 2). Rows use batch sizes 1, 2, 4, and 8; columns use 128, 2,048, 8,192 input tokens. Each panel varies output length over 128, 256, 512, and 1024 tokens. BF16 is normalized to $1\times$; higher is better. Method colors and order are fixed across panels.}
\label{fig:system-h100-qwen2_7b-throughput-1}
\end{figure*}

\begin{figure*}[!tp]
\centering
\includegraphics[width=0.36347914\textwidth]{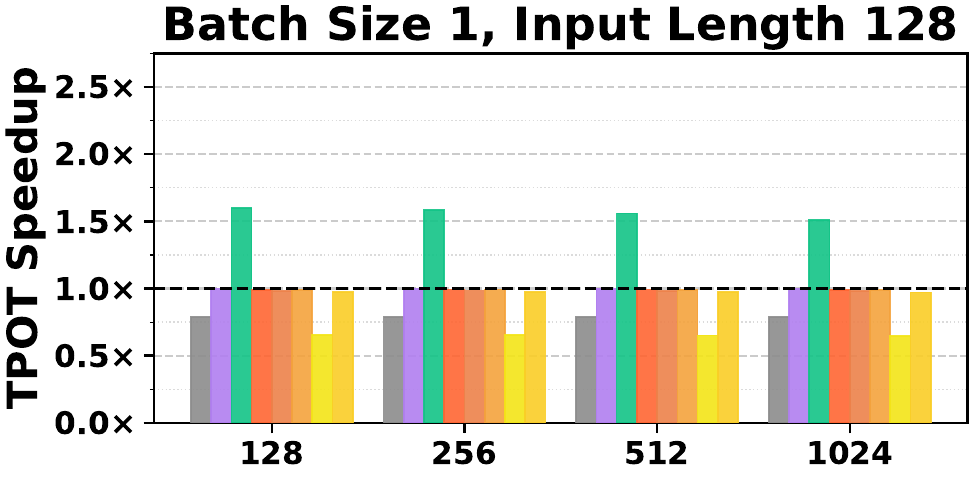}\hfill
\includegraphics[width=0.30826043\textwidth]{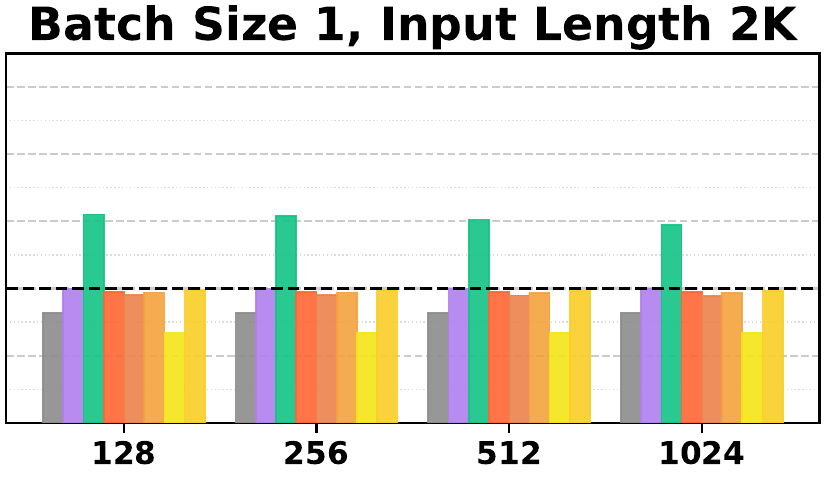}\hfill
\includegraphics[width=0.30826043\textwidth]{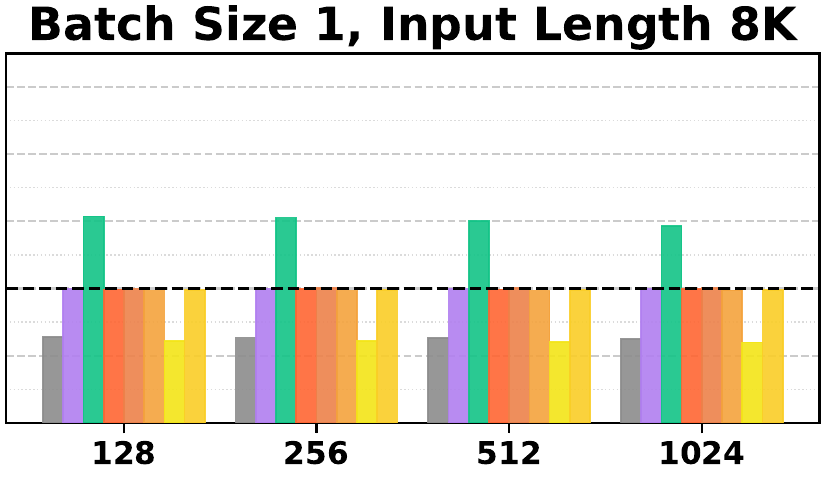}\\[-3pt]
\includegraphics[width=0.36347914\textwidth]{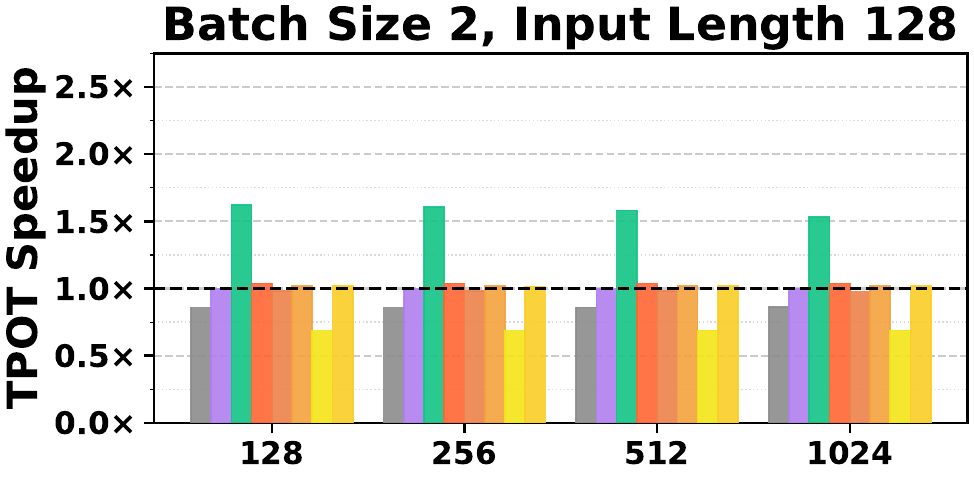}\hfill
\includegraphics[width=0.30826043\textwidth]{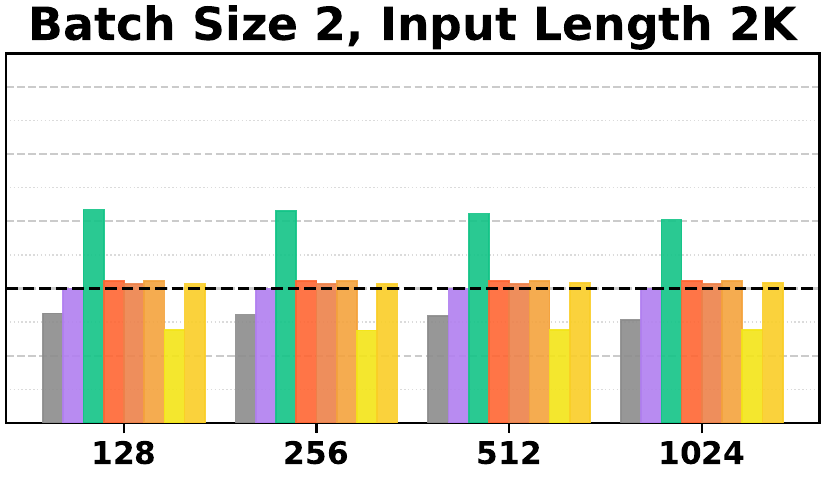}\hfill
\includegraphics[width=0.30826043\textwidth]{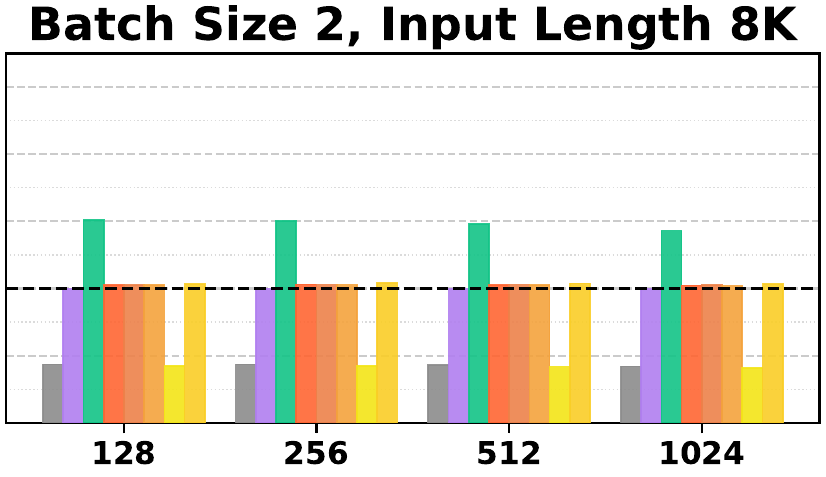}\\[-3pt]
\includegraphics[width=0.36347914\textwidth]{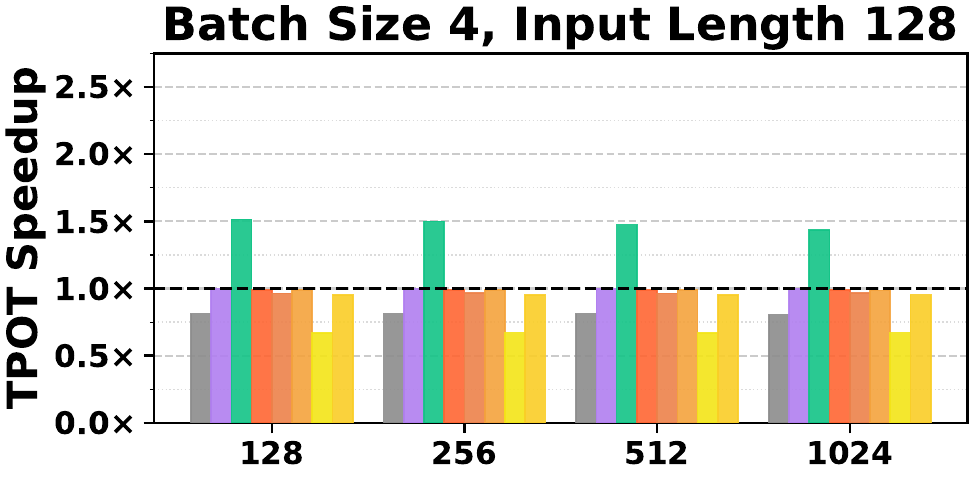}\hfill
\includegraphics[width=0.30826043\textwidth]{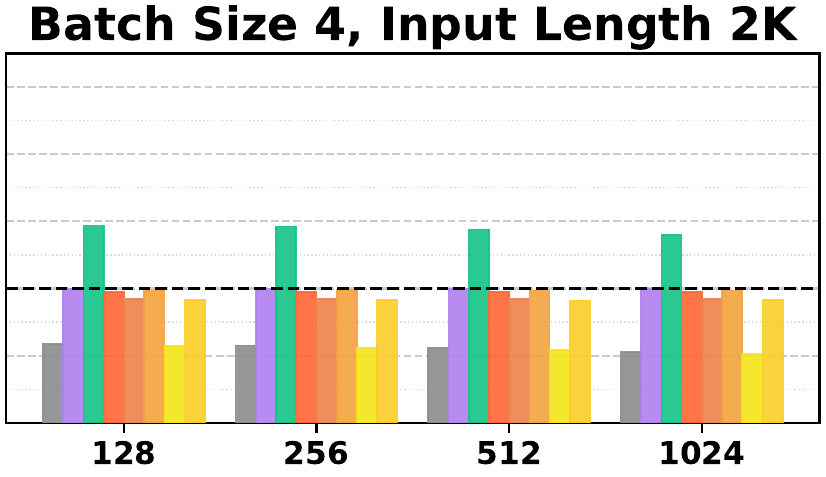}\hfill
\includegraphics[width=0.30826043\textwidth]{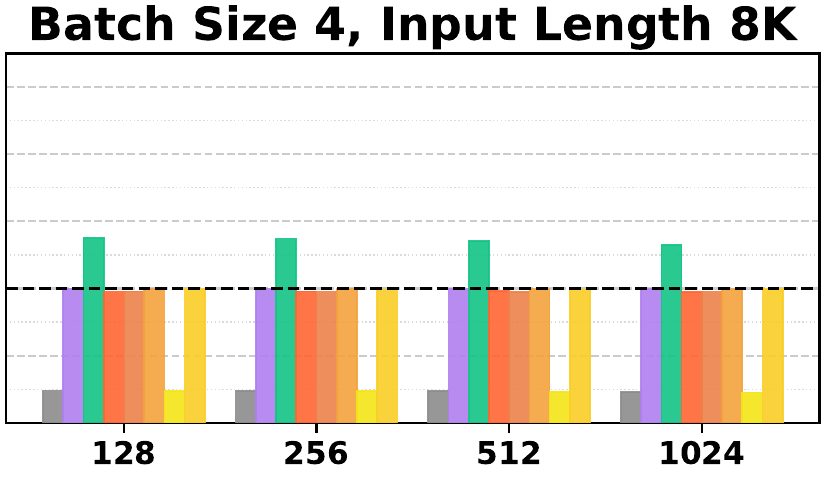}\\[-3pt]
\includegraphics[width=0.36347914\textwidth]{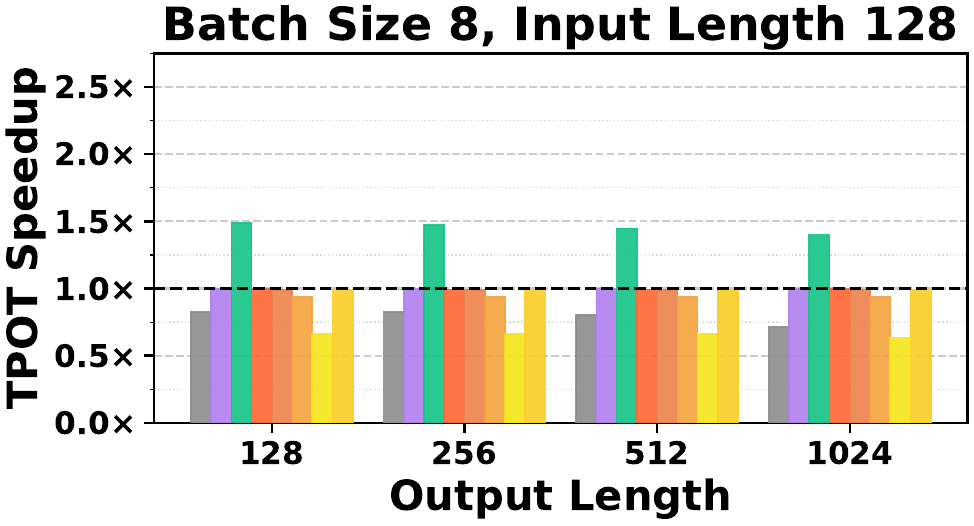}\hfill
\includegraphics[width=0.30826043\textwidth]{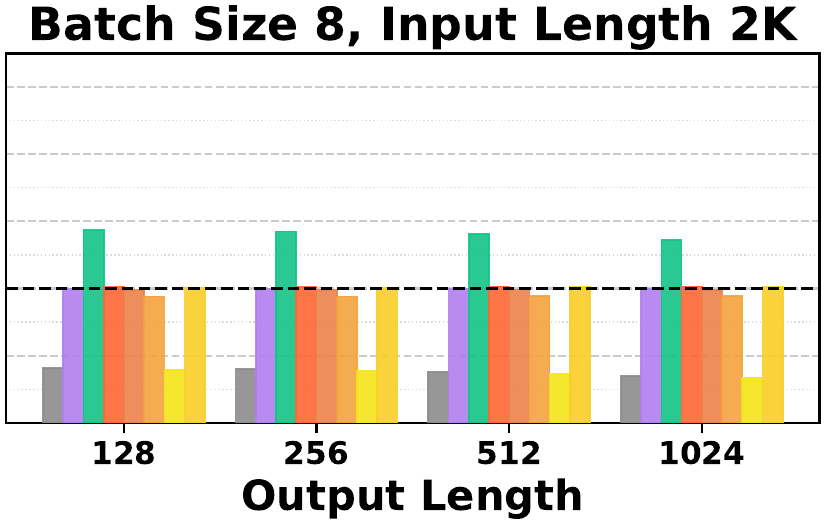}\hfill
\includegraphics[width=0.30826043\textwidth]{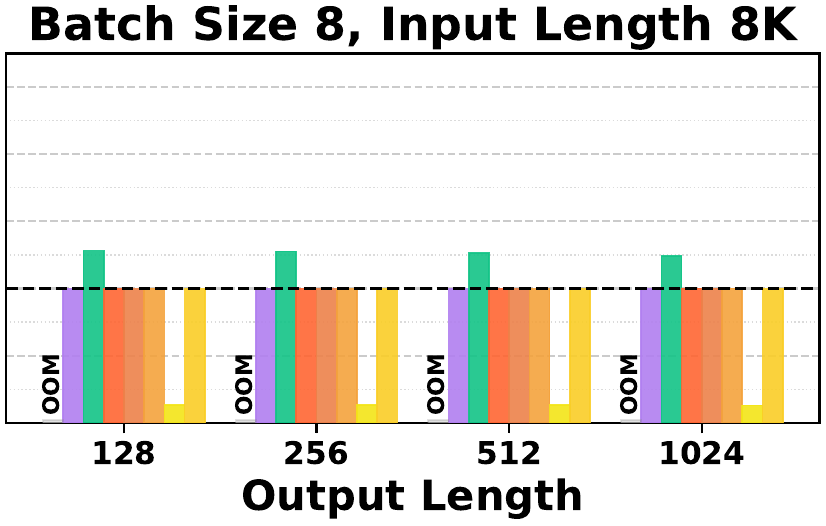}
\vspace{-8pt}
\caption{\textbf{TPOT speedup on H100 NVL.} Qwen2-7B-Instruct (prompt group 1 of 2). Rows use batch sizes 1, 2, 4, and 8; columns use 128, 2,048, 8,192 input tokens. Each panel varies output length over 128, 256, 512, and 1024 tokens. BF16 is normalized to $1\times$; higher is better. Method colors and order are fixed across panels.}
\label{fig:system-h100-qwen2_7b-tpot-1}
\end{figure*}

\begin{figure*}[!tp]
\centering
\includegraphics[width=0.36347914\textwidth]{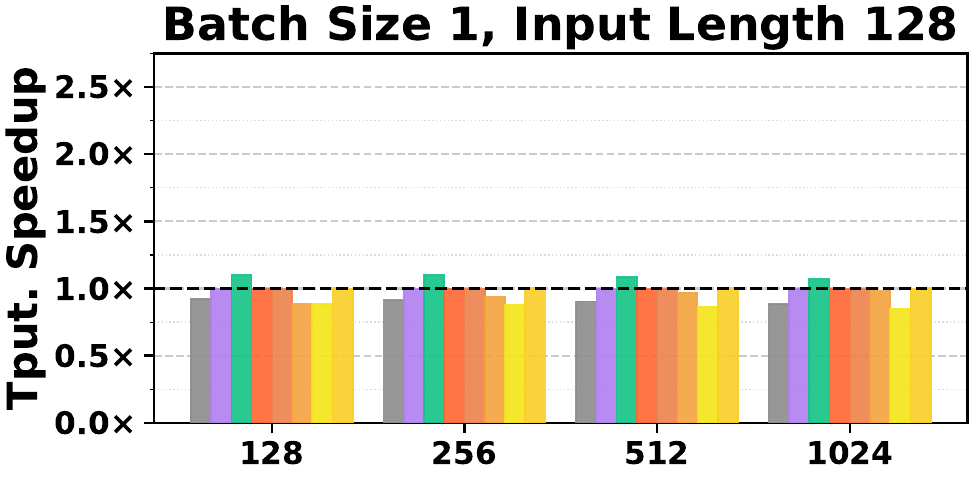}\hfill
\includegraphics[width=0.30826043\textwidth]{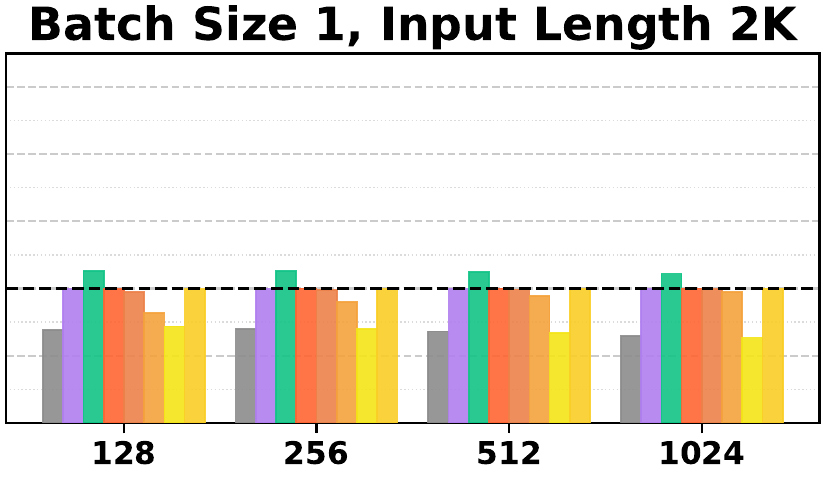}\hfill
\includegraphics[width=0.30826043\textwidth]{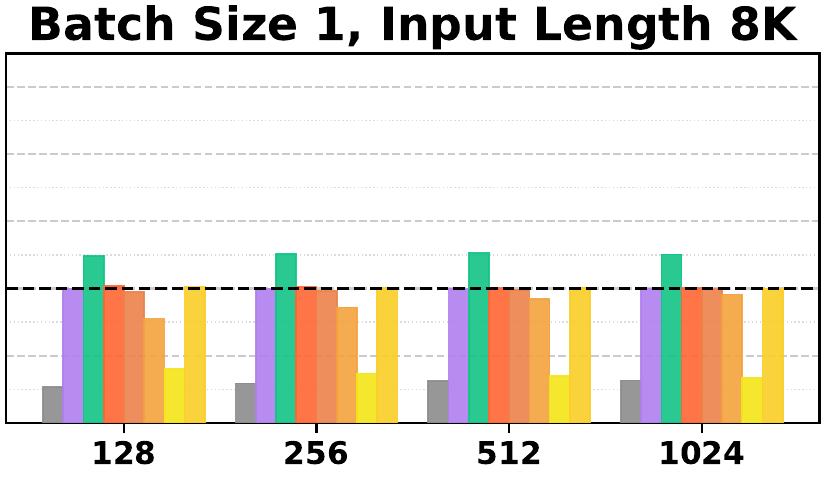}\\[-3pt]
\includegraphics[width=0.36347914\textwidth]{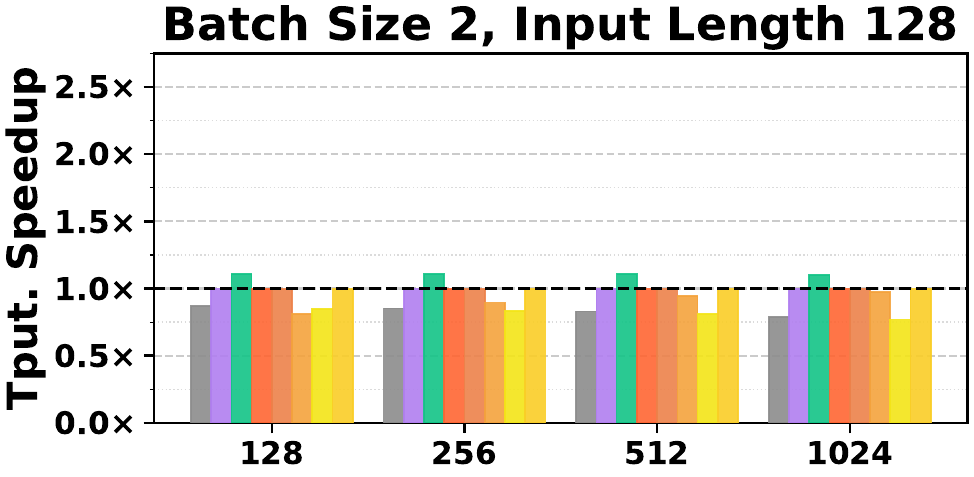}\hfill
\includegraphics[width=0.30826043\textwidth]{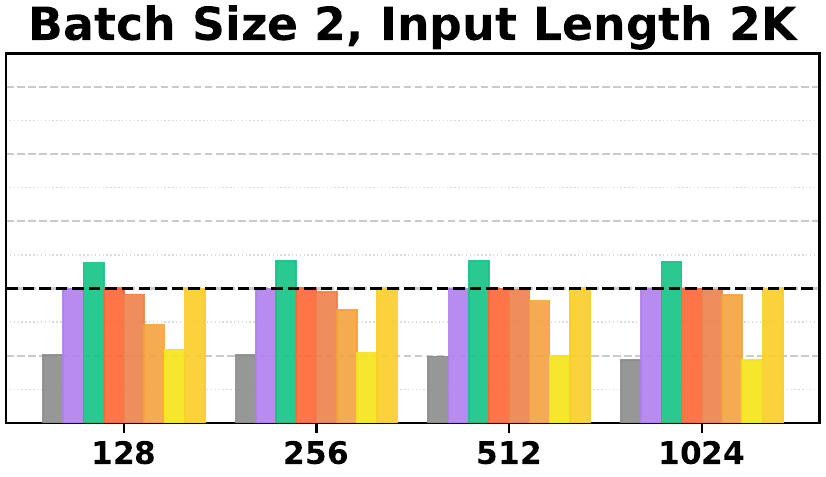}\hfill
\includegraphics[width=0.30826043\textwidth]{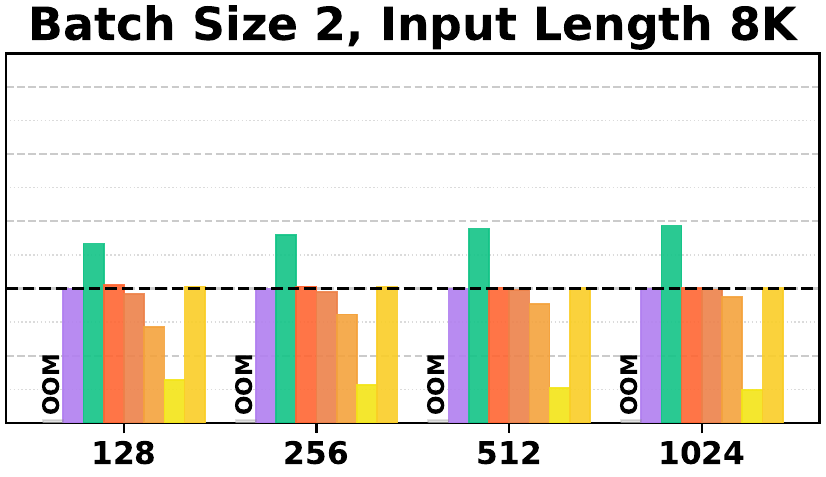}
\vspace{-8pt}
\caption{\textbf{Generation-throughput ratio on RTX 4090.} Llama-3.2-3B-Instruct. Rows use batch sizes 1 and 2; columns use 128, 2,048, 8,192 input tokens. Each panel varies output length over 128, 256, 512, and 1024 tokens. BF16 is normalized to $1\times$; higher is better. Method colors and order are fixed across panels.}
\label{fig:system-rtx4090-llama32_3b-throughput}
\par\vspace{4pt}
\begin{minipage}{\textwidth}
\normalsize
\setlength{\multicolsep}{0pt}
\begin{multicols}{2}
\raggedcolumns
\textbf{RTX 4090.}
The compact-layout implementation also improves complete-request throughput on RTX 4090 (Figures~\ref{fig:system-rtx4090-llama32_3b-throughput}--\ref{fig:system-rtx4090-llama32_3b-tpot} and~\ref{fig:system-rtx4090-qwen2_7b-throughput}--\ref{fig:system-rtx4090-ministral3_8b-tpot}). With an 8K prompt and batch size four, Llama reaches 167.69 tokens/s for 128 output tokens and 316.54 tokens/s for 1,024 output tokens, compared with 111.39 and 164.97 for BF16.\par
\end{multicols}
\end{minipage}
\end{figure*}

\begin{figure*}[!tp]
\centering
\includegraphics[width=0.36347914\textwidth]{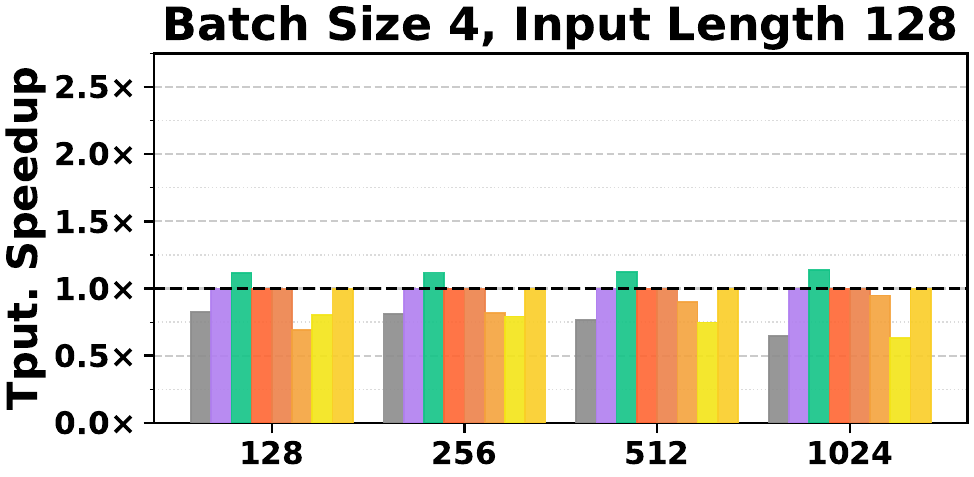}\hfill
\includegraphics[width=0.30826043\textwidth]{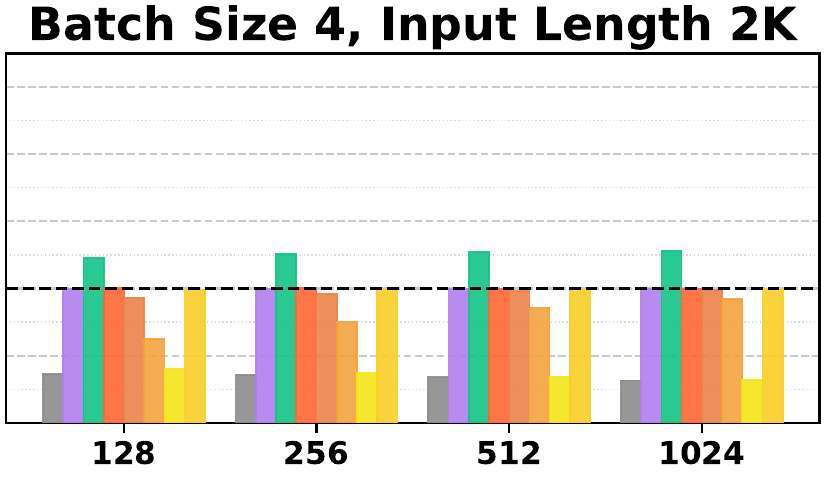}\hfill
\includegraphics[width=0.30826043\textwidth]{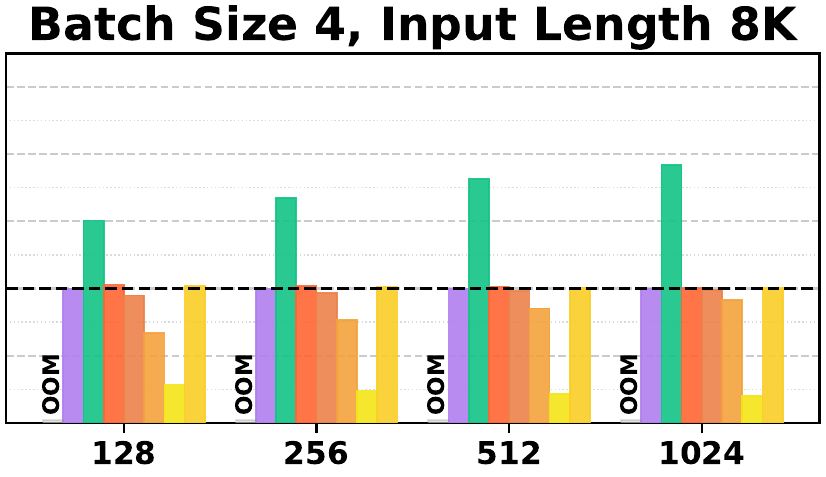}\\[-3pt]
\includegraphics[width=0.36347914\textwidth]{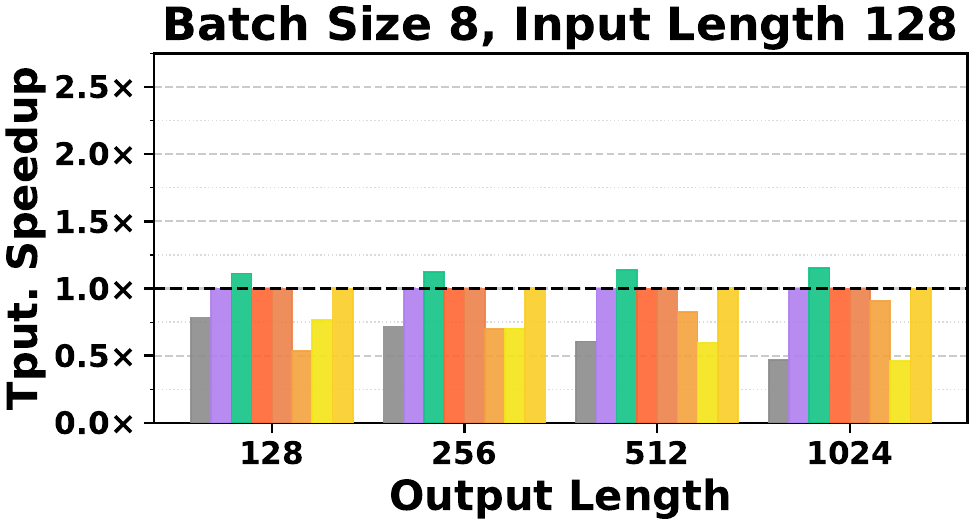}\hfill
\includegraphics[width=0.30826043\textwidth]{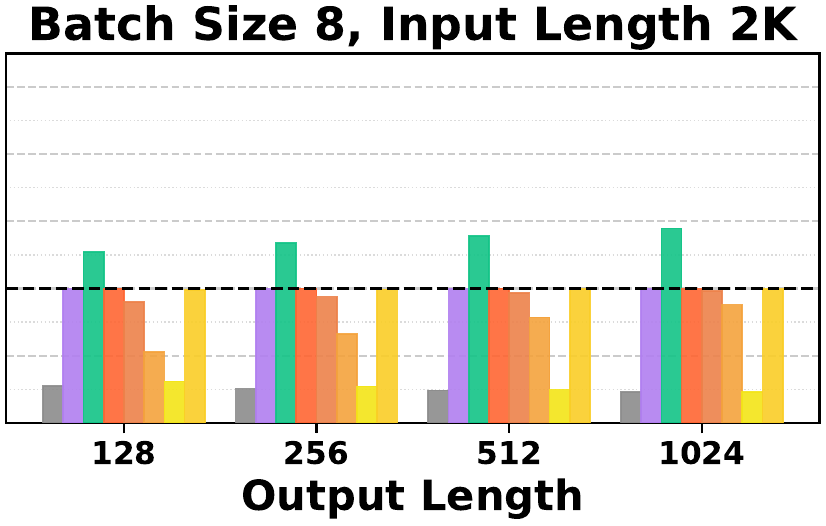}\hfill
\includegraphics[width=0.30826043\textwidth]{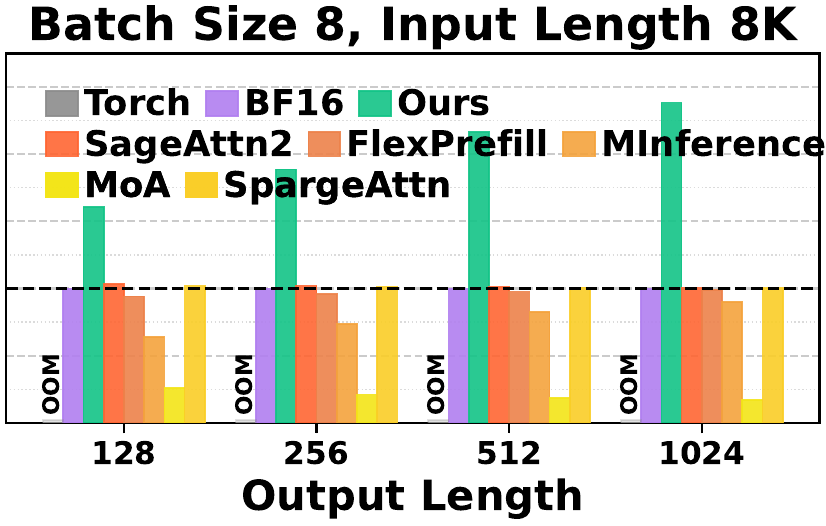}
\vspace{-8pt}
\addtocounter{figure}{-1}
\renewcommand{\theHfigure}{\arabic{figure}.continued}
\caption[]{\textbf{Continued.} Rows use batch sizes 4 and 8; panel settings follow the first part.}
\par\vspace{8pt}
\includegraphics[width=0.36347914\textwidth]{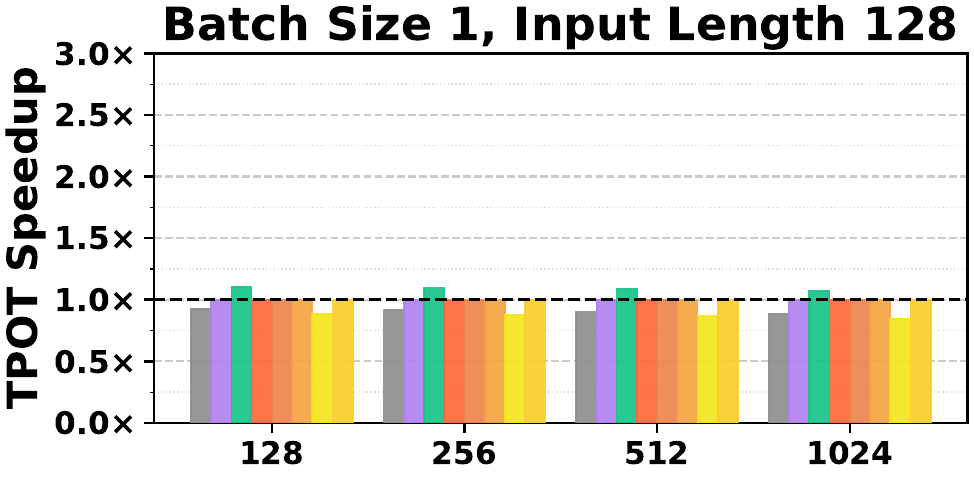}\hfill
\includegraphics[width=0.30826043\textwidth]{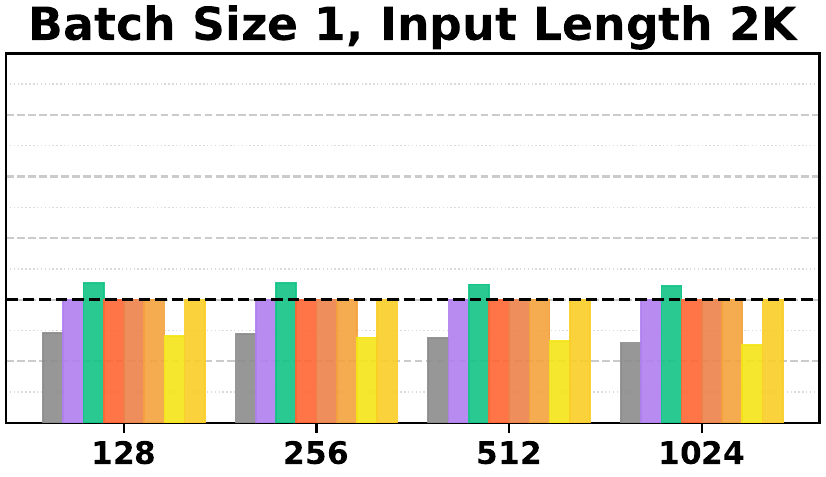}\hfill
\includegraphics[width=0.30826043\textwidth]{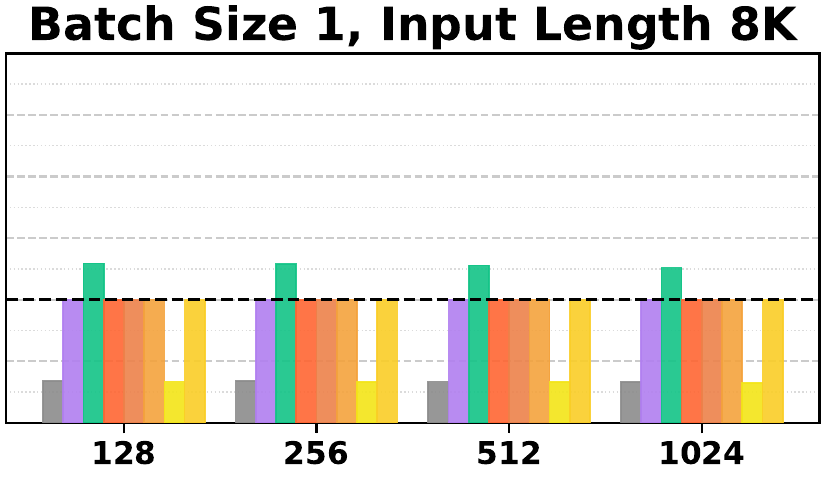}\\[-3pt]
\includegraphics[width=0.36347914\textwidth]{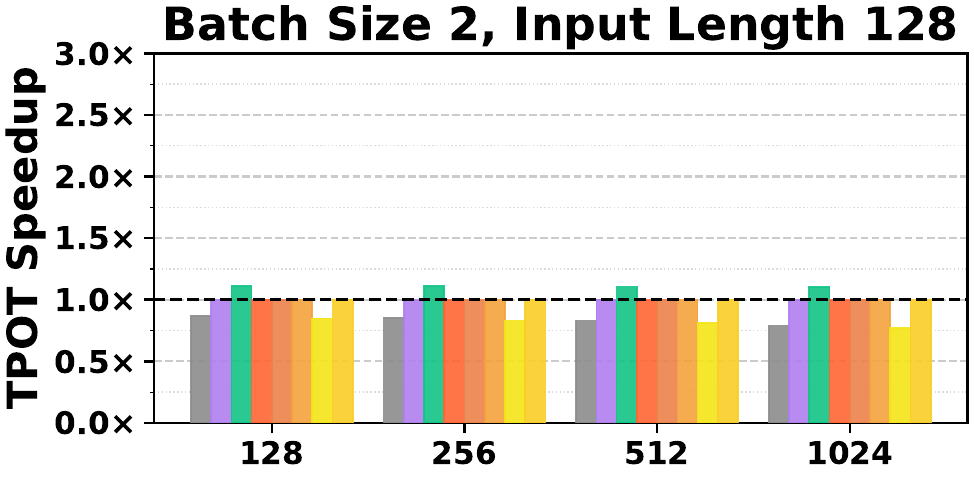}\hfill
\includegraphics[width=0.30826043\textwidth]{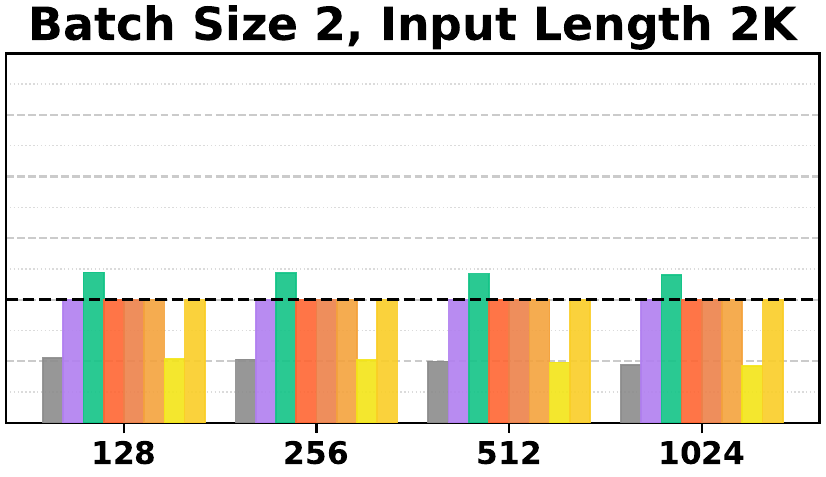}\hfill
\includegraphics[width=0.30826043\textwidth]{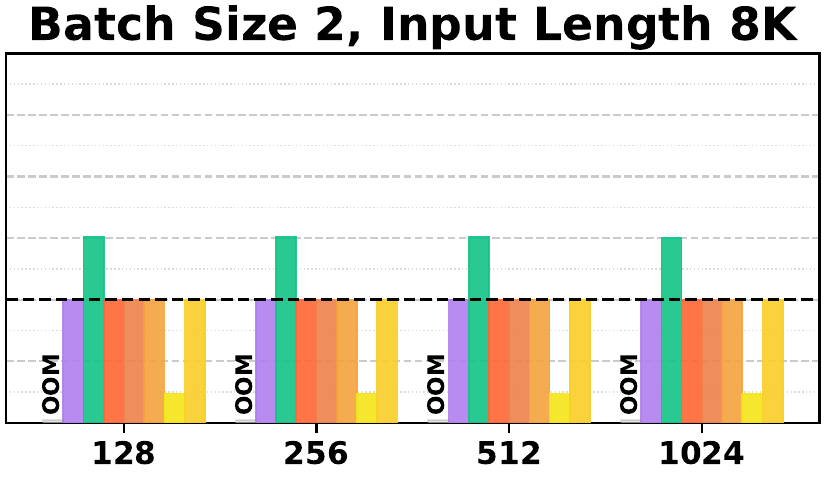}\\[-3pt]
\includegraphics[width=0.36347914\textwidth]{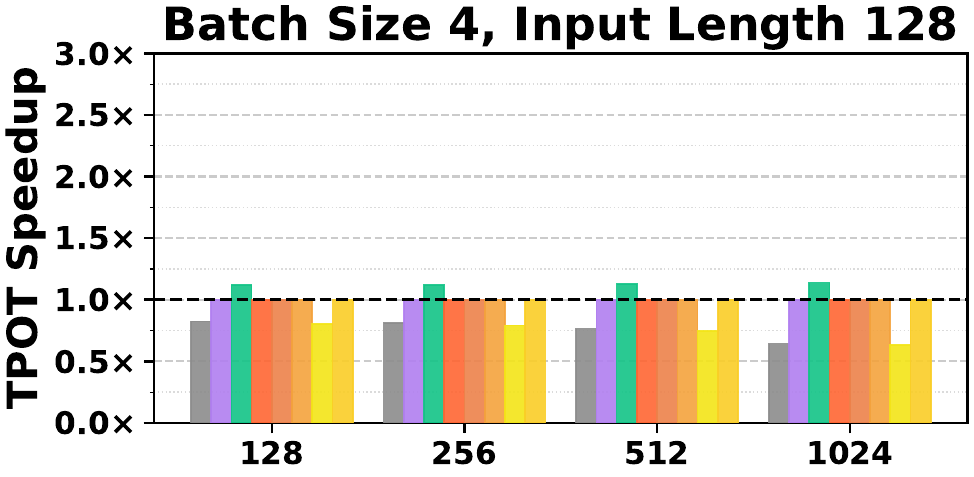}\hfill
\includegraphics[width=0.30826043\textwidth]{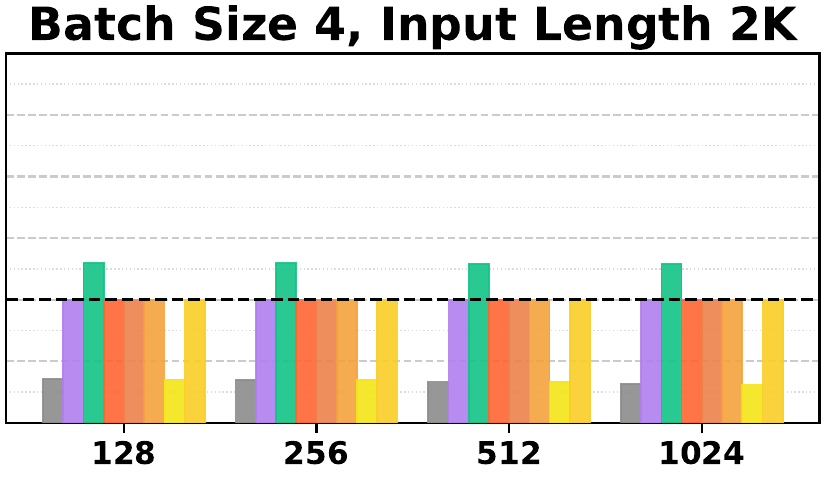}\hfill
\includegraphics[width=0.30826043\textwidth]{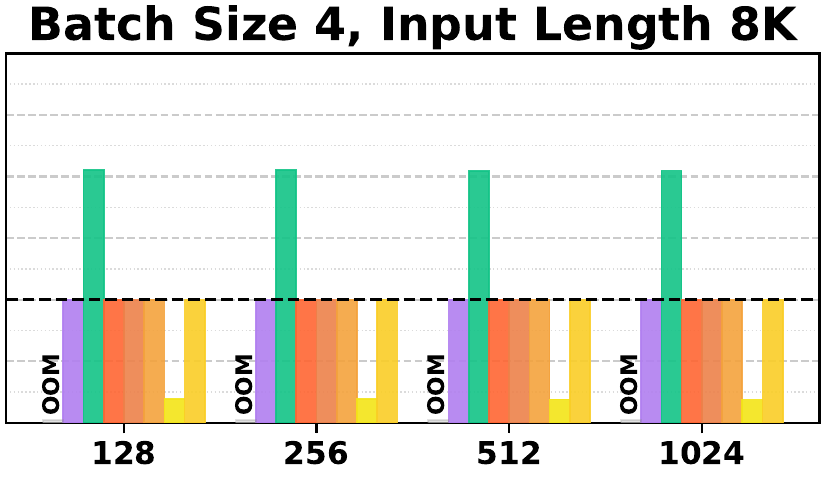}\\[-3pt]
\includegraphics[width=0.36347914\textwidth]{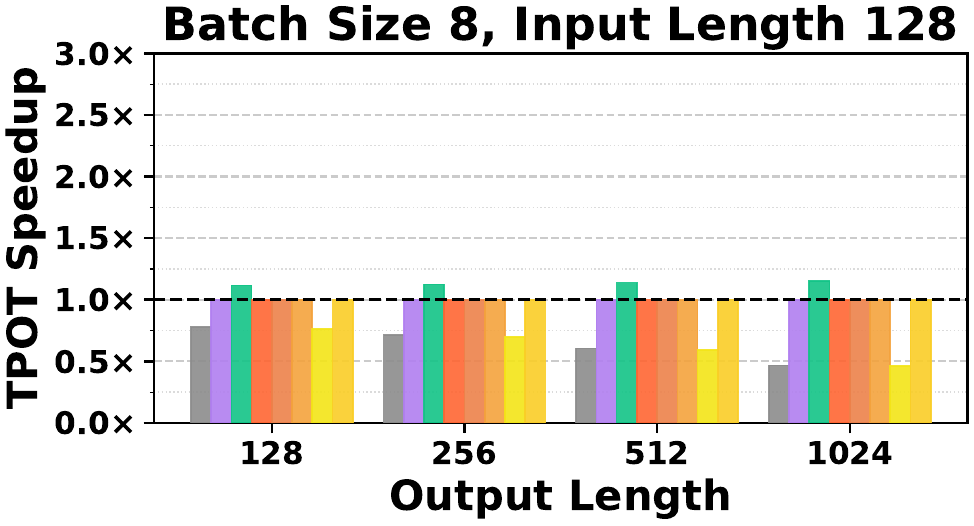}\hfill
\includegraphics[width=0.30826043\textwidth]{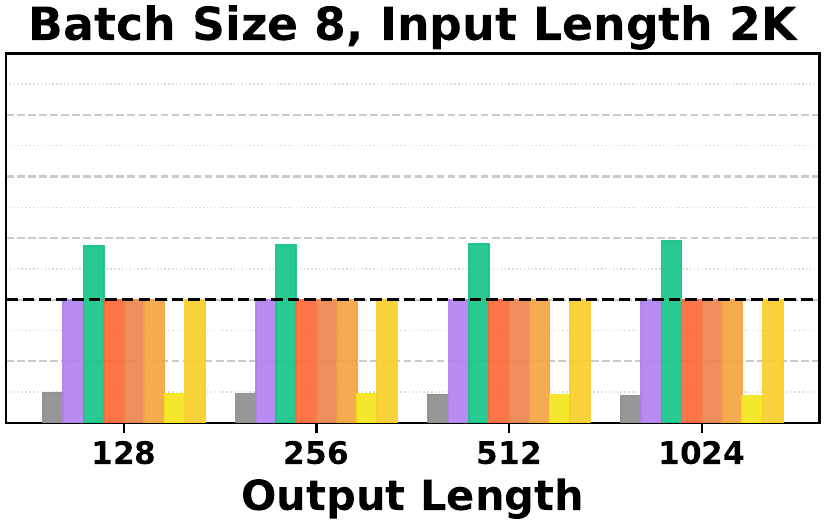}\hfill
\includegraphics[width=0.30826043\textwidth]{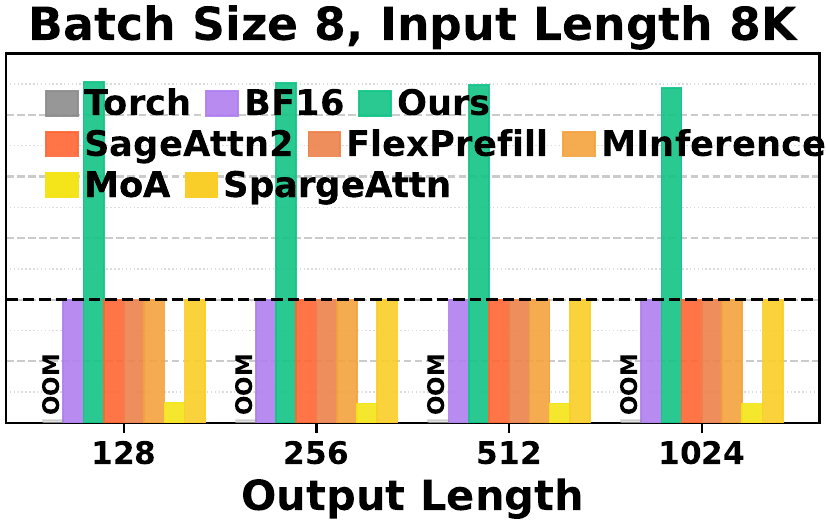}
\vspace{-8pt}
\caption{\textbf{TPOT speedup on RTX 4090.} Llama-3.2-3B-Instruct. Rows use batch sizes 1, 2, 4, and 8; columns use 128, 2,048, 8,192 input tokens. Each panel varies output length over 128, 256, 512, and 1024 tokens. BF16 is normalized to $1\times$; higher is better. Method colors and order are fixed across panels.}
\label{fig:system-rtx4090-llama32_3b-tpot}
\par\vspace{4pt}
\begin{minipage}{\textwidth}
\normalsize
\setlength{\multicolsep}{0pt}
\begin{multicols}{2}
\raggedcolumns
\textls[-20]{The corresponding ratios increase from $1.51\times$ to $1.92\times$, extending this trend to this GPU.}\looseness=-1\par
\columnbreak
\findingbox{%
In A100, H100 NVL, and RTX 4090 comparisons, throughput gains over BF16 grow with continuation length. Longer continuations increase decode efficiency's role in complete-request performance.
}
\end{multicols}
\end{minipage}
\end{figure*}

\begin{figure*}[!tp]
\centering
\begin{adjustbox}{max width=\textwidth,max totalheight=0.40\textheight}
\begin{minipage}{\textwidth}
\centering
\includegraphics[width=0.27650431\textwidth]{figures/efficiency/h100/qwen2_7b/11_workload_generation_throughput/a_baselines/batch_1/input_128.pdf}\hfill
\includegraphics[width=0.23449856\textwidth]{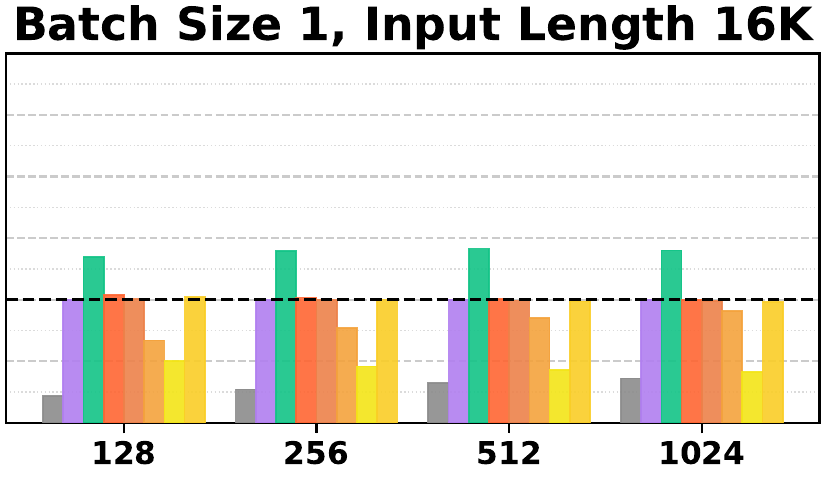}\hfill
\includegraphics[width=0.23449856\textwidth]{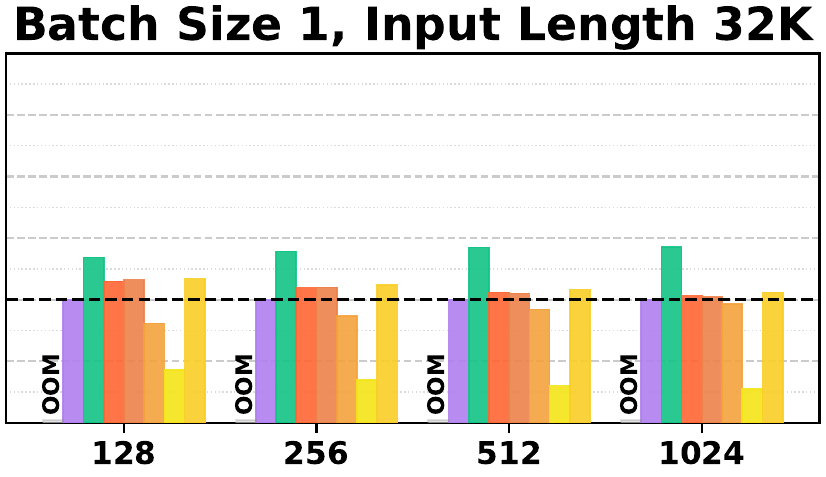}\hfill
\includegraphics[width=0.23449856\textwidth]{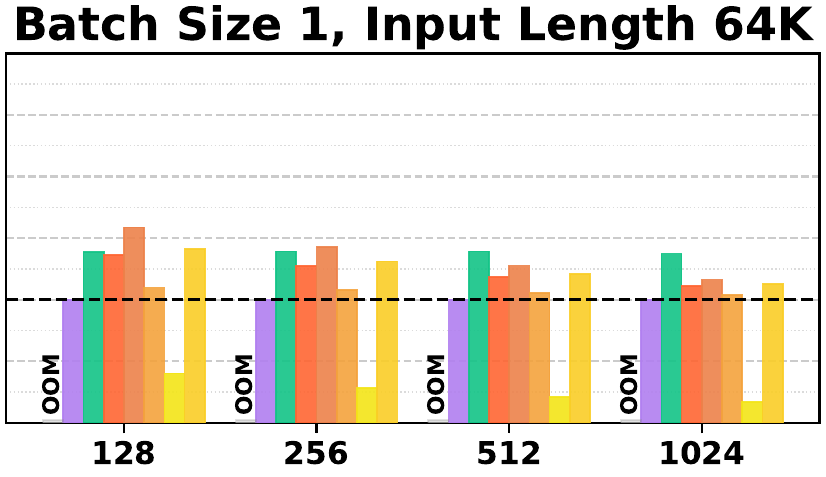}\\[-3pt]
\includegraphics[width=0.27650431\textwidth]{figures/efficiency/h100/qwen2_7b/11_workload_generation_throughput/a_baselines/batch_2/input_128.pdf}\hfill
\includegraphics[width=0.23449856\textwidth]{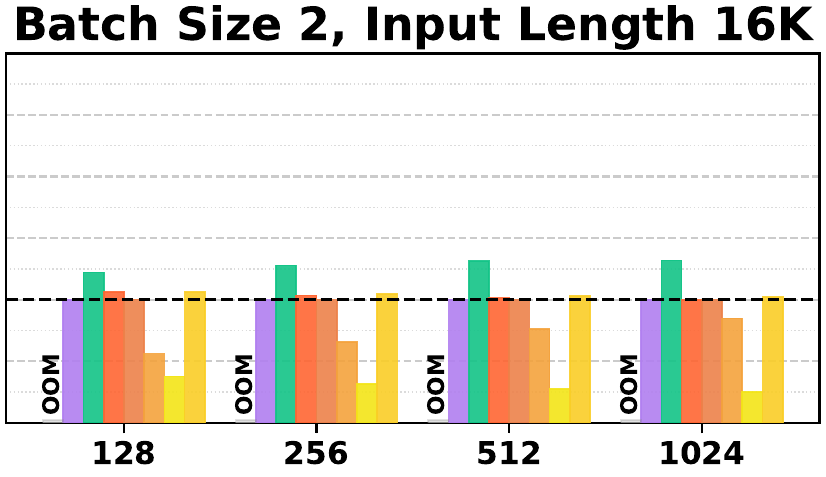}\hfill
\includegraphics[width=0.23449856\textwidth]{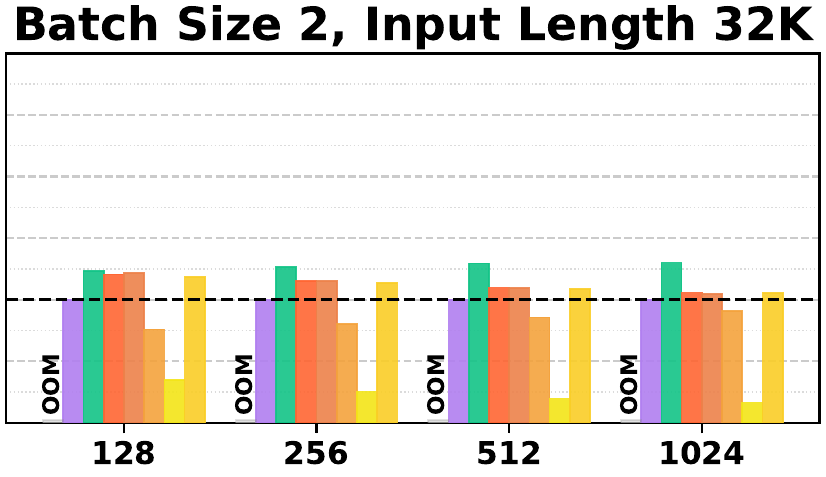}\hfill
\includegraphics[width=0.23449856\textwidth]{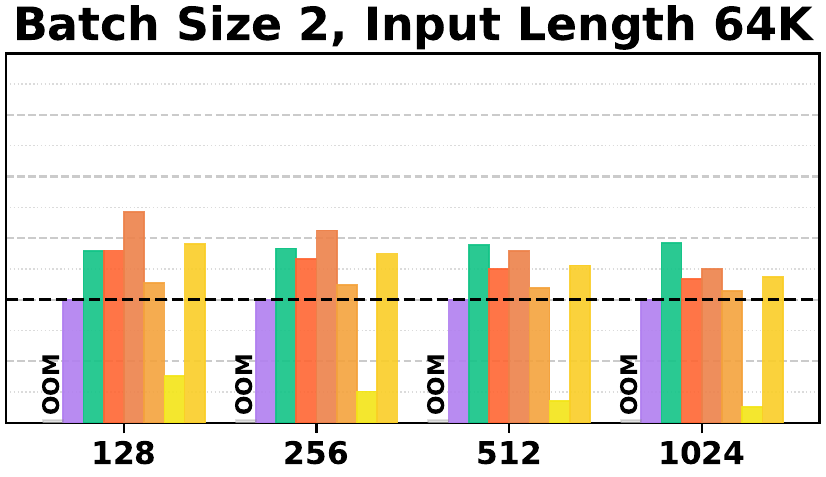}\\[-3pt]
\includegraphics[width=0.27650431\textwidth]{figures/efficiency/h100/qwen2_7b/11_workload_generation_throughput/a_baselines/batch_4/input_128.pdf}\hfill
\includegraphics[width=0.23449856\textwidth]{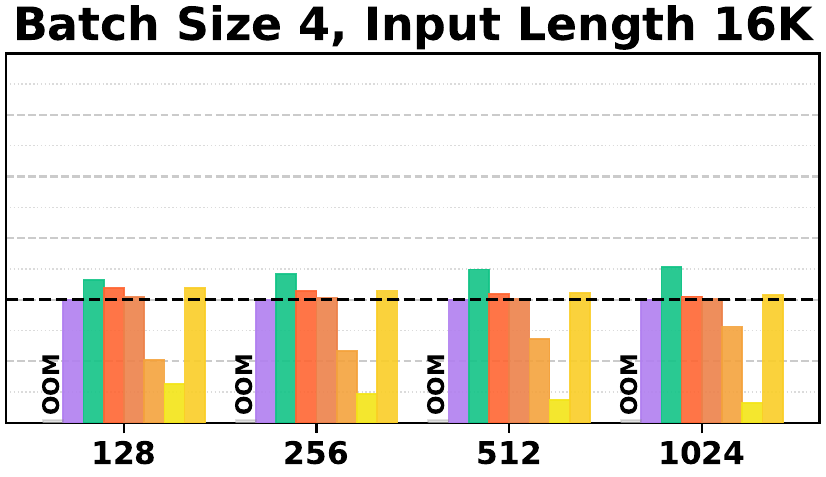}\hfill
\includegraphics[width=0.23449856\textwidth]{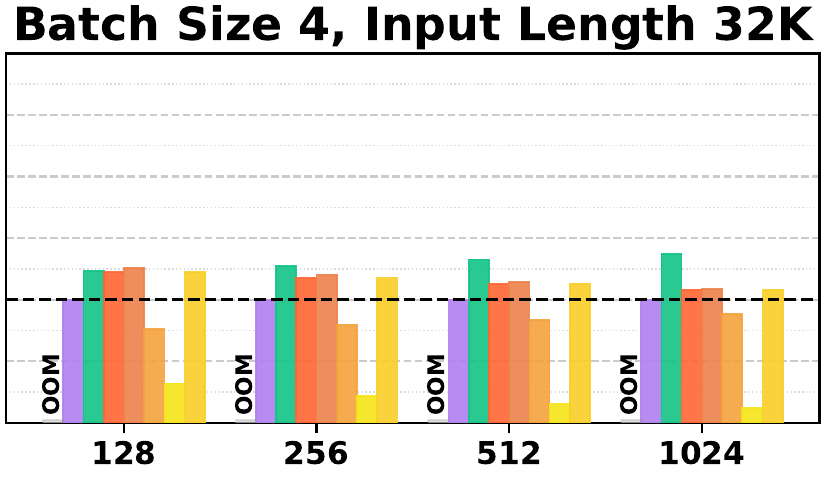}\hfill
\includegraphics[width=0.23449856\textwidth]{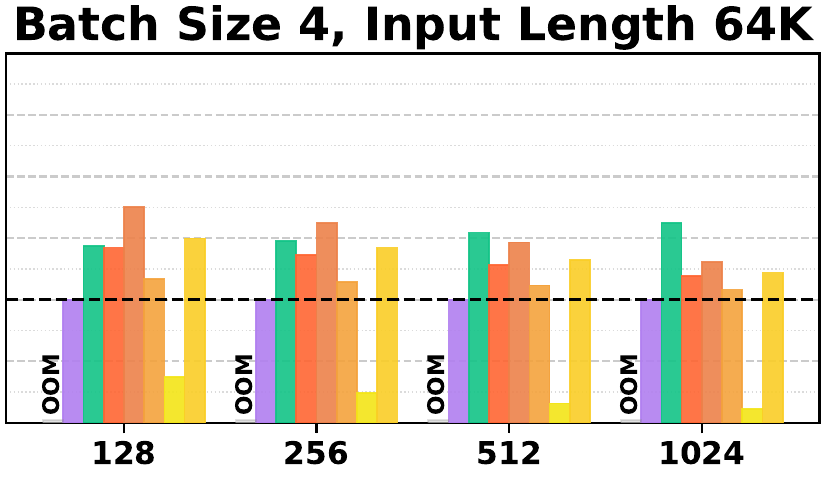}\\[-3pt]
\includegraphics[width=0.27650431\textwidth]{figures/efficiency/h100/qwen2_7b/11_workload_generation_throughput/a_baselines/batch_8/input_128.pdf}\hfill
\includegraphics[width=0.23449856\textwidth]{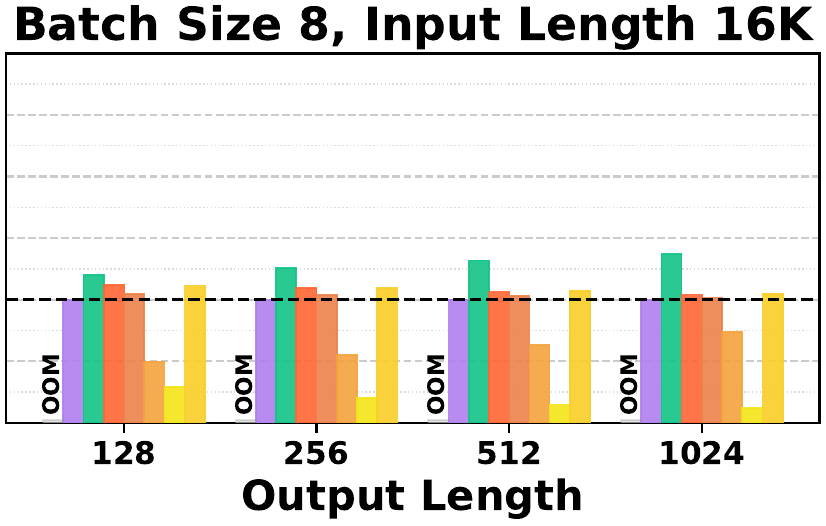}\hfill
\includegraphics[width=0.23449856\textwidth]{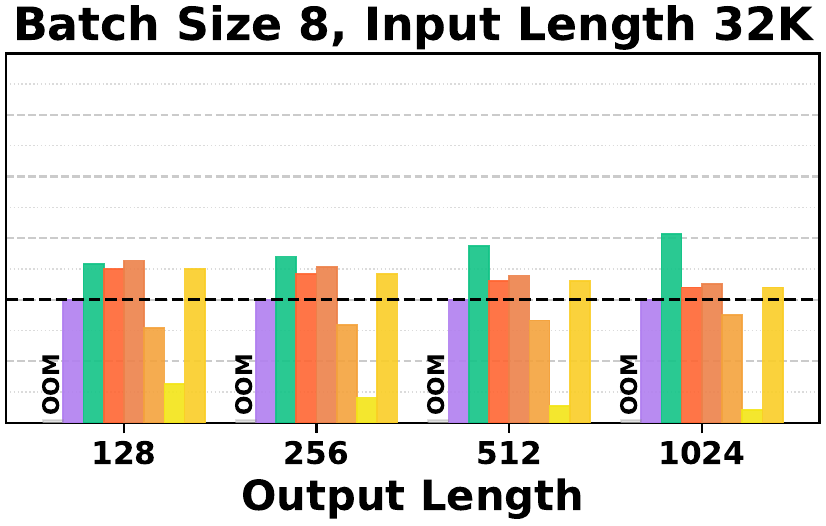}\hfill
\includegraphics[width=0.23449856\textwidth]{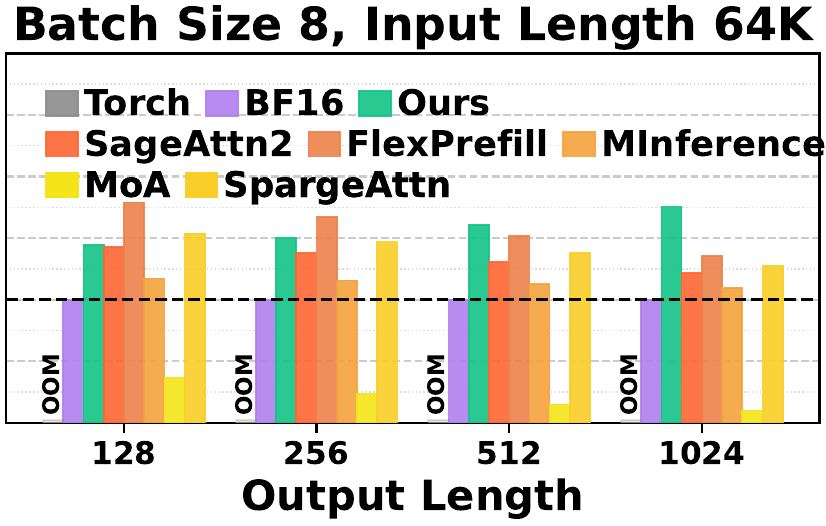}
\vspace{-8pt}
\end{minipage}
\end{adjustbox}
\caption{\textbf{Generation-throughput ratio on H100 NVL.} Qwen2-7B-Instruct (prompt group 2 of 2). Rows use batch sizes 1, 2, 4, and 8; columns use 128, 16,384, 32,768, 65,536 input tokens. Each panel varies output length over 128, 256, 512, and 1024 tokens. BF16 is normalized to $1\times$; higher is better. The 128-token column repeats the first group as a shared reference. Panels share colors and order.}
\label{fig:system-h100-qwen2_7b-throughput-2}
\par\vspace{8pt}
\begin{adjustbox}{max width=\textwidth,max totalheight=0.40\textheight}
\begin{minipage}{\textwidth}
\centering
\includegraphics[width=0.27650431\textwidth]{figures/efficiency/h100/qwen2_7b/10_workload_tpot/a_baselines/batch_1/input_128.pdf}\hfill
\includegraphics[width=0.23449856\textwidth]{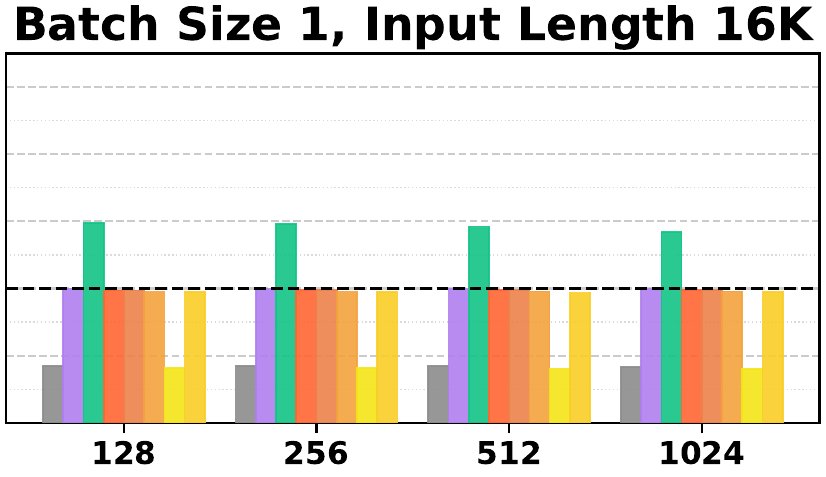}\hfill
\includegraphics[width=0.23449856\textwidth]{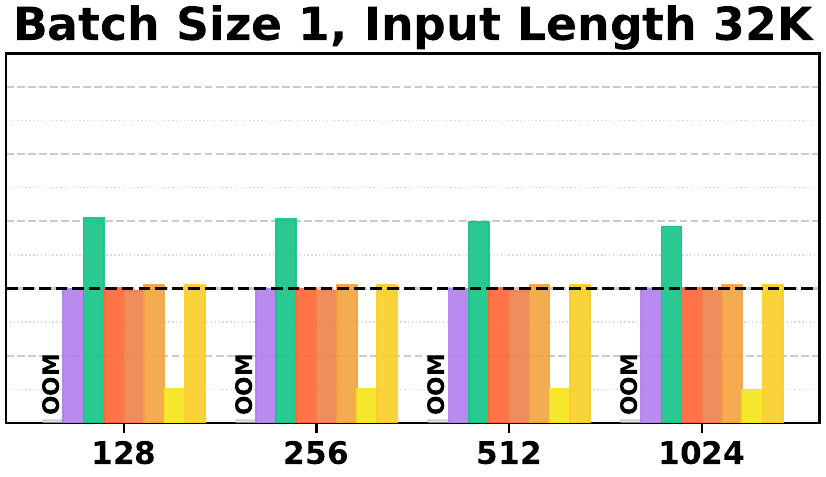}\hfill
\includegraphics[width=0.23449856\textwidth]{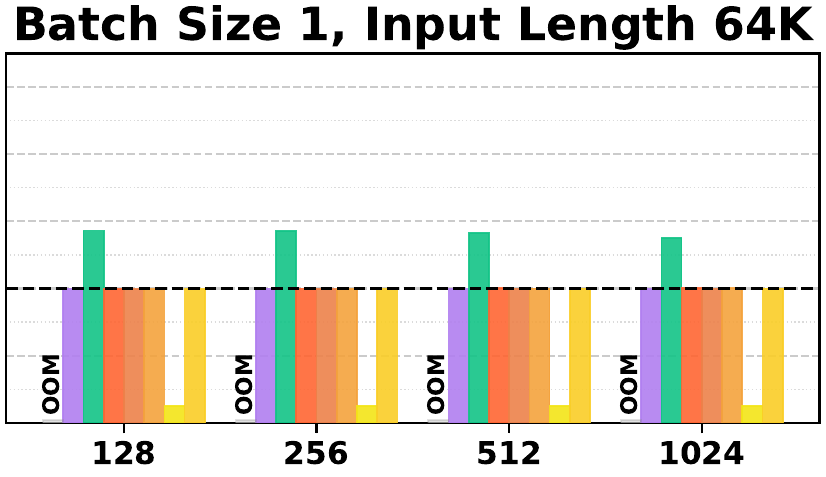}\\[-3pt]
\includegraphics[width=0.27650431\textwidth]{figures/efficiency/h100/qwen2_7b/10_workload_tpot/a_baselines/batch_2/input_128.pdf}\hfill
\includegraphics[width=0.23449856\textwidth]{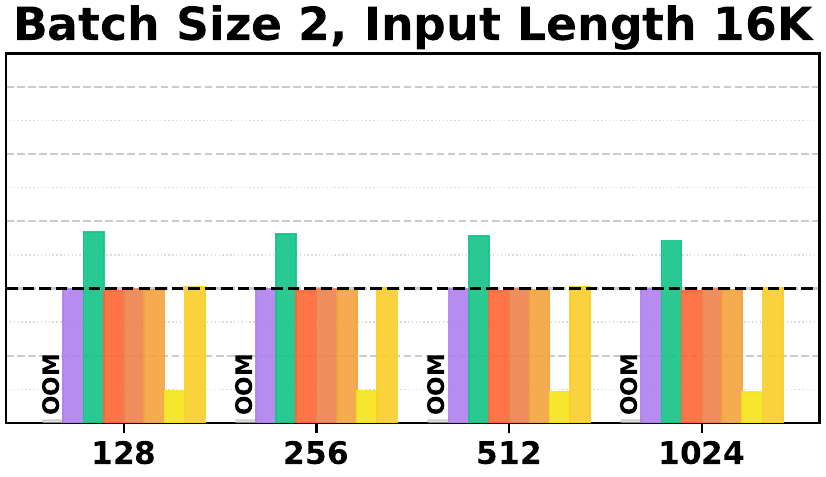}\hfill
\includegraphics[width=0.23449856\textwidth]{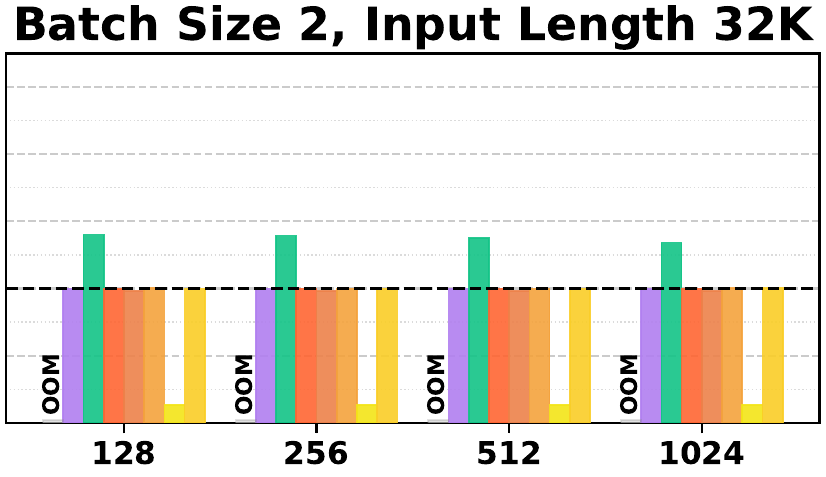}\hfill
\includegraphics[width=0.23449856\textwidth]{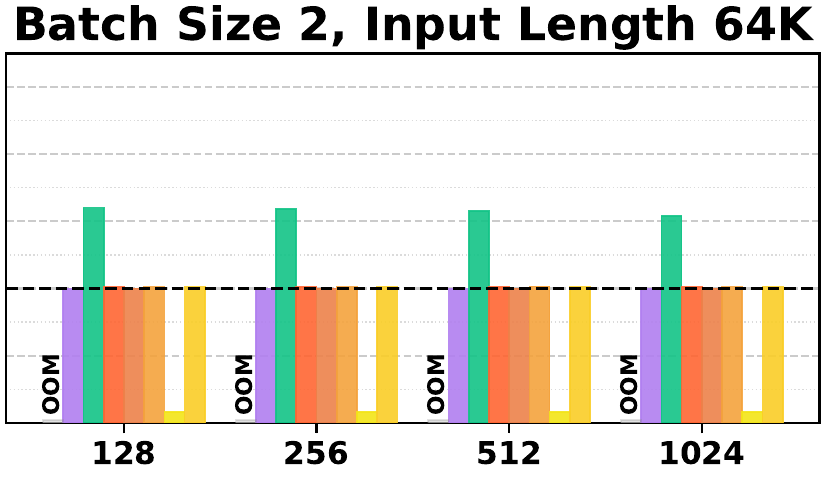}\\[-3pt]
\includegraphics[width=0.27650431\textwidth]{figures/efficiency/h100/qwen2_7b/10_workload_tpot/a_baselines/batch_4/input_128.pdf}\hfill
\includegraphics[width=0.23449856\textwidth]{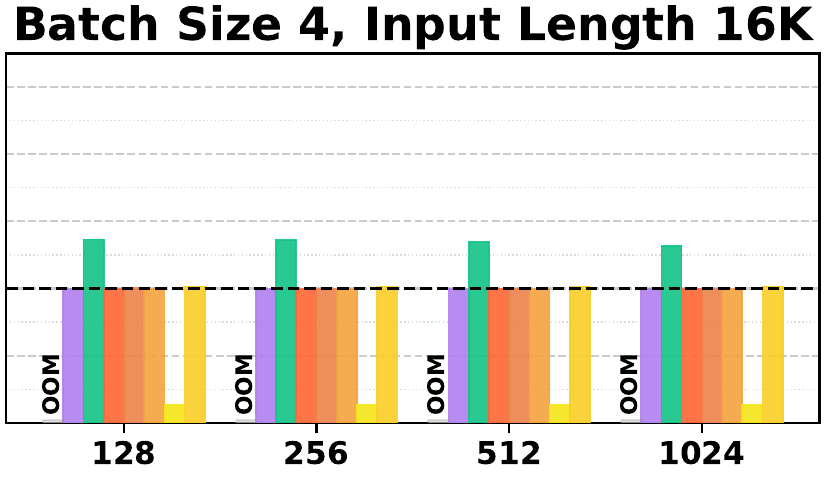}\hfill
\includegraphics[width=0.23449856\textwidth]{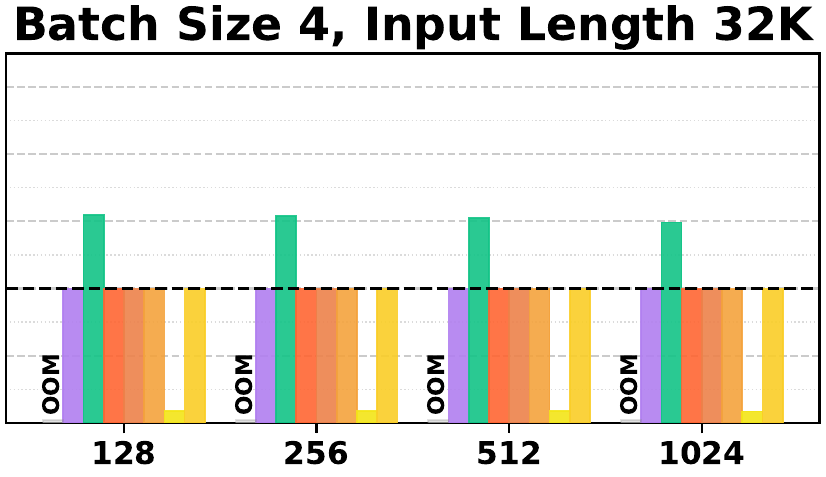}\hfill
\includegraphics[width=0.23449856\textwidth]{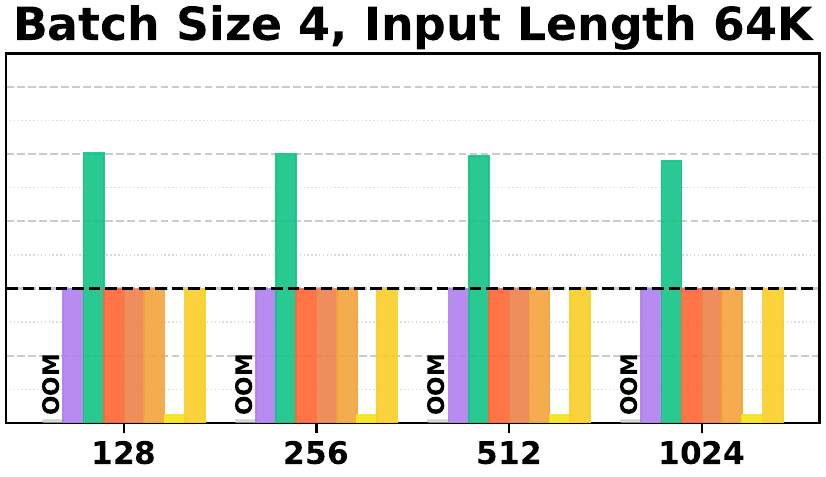}\\[-3pt]
\includegraphics[width=0.27650431\textwidth]{figures/efficiency/h100/qwen2_7b/10_workload_tpot/a_baselines/batch_8/input_128.pdf}\hfill
\includegraphics[width=0.23449856\textwidth]{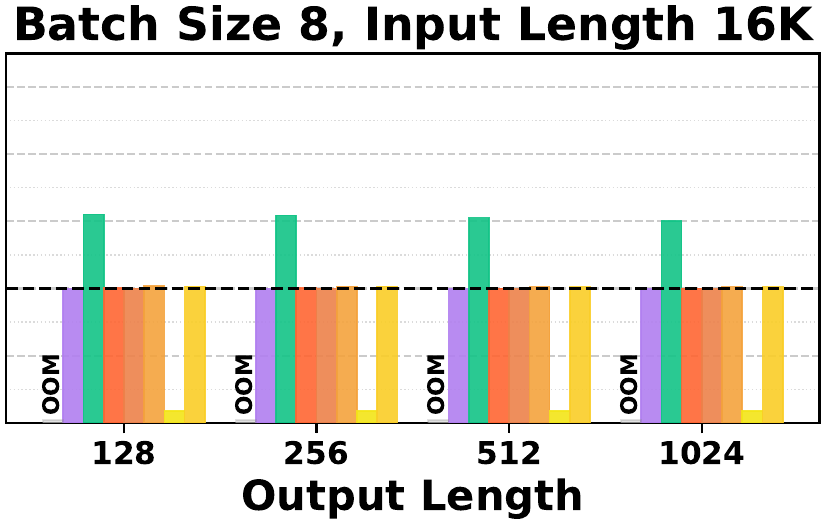}\hfill
\includegraphics[width=0.23449856\textwidth]{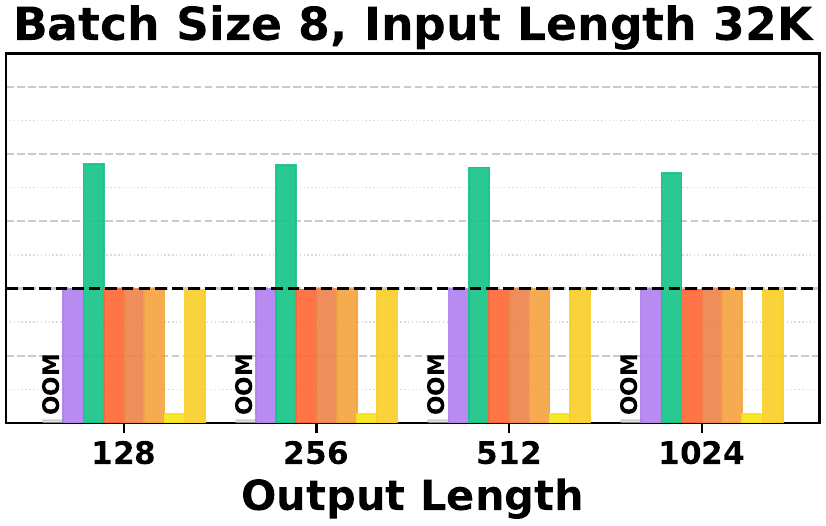}\hfill
\includegraphics[width=0.23449856\textwidth]{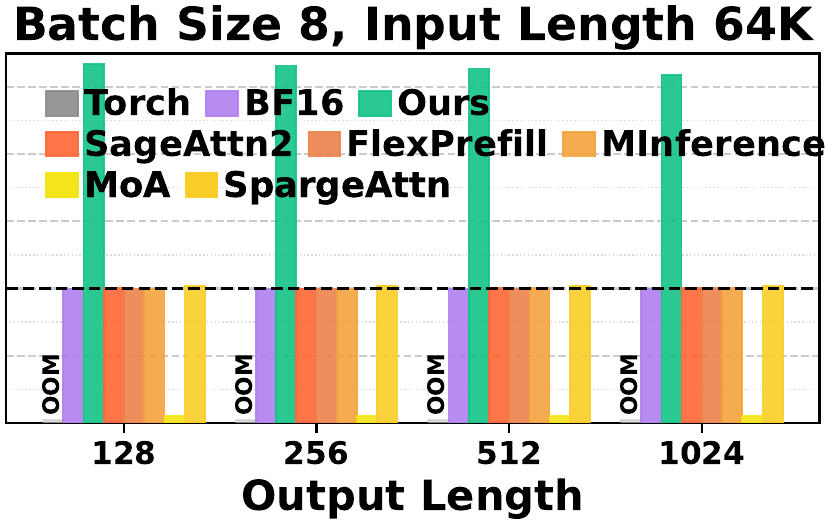}
\vspace{-8pt}
\end{minipage}
\end{adjustbox}
\caption{\textbf{TPOT speedup on H100 NVL.} Qwen2-7B-Instruct (prompt group 2 of 2). Rows use batch sizes 1, 2, 4, and 8; columns use 128, 16,384, 32,768, 65,536 input tokens. Each panel varies output length over 128, 256, 512, and 1024 tokens. BF16 is normalized to $1\times$; higher is better. The 128-token column repeats the first group as a shared reference. Method colors and order are fixed across panels.}
\label{fig:system-h100-qwen2_7b-tpot-2}
\end{figure*}

\begin{figure*}[!tp]
\centering
\begin{adjustbox}{max width=\textwidth,max totalheight=0.40\textheight}
\begin{minipage}{\textwidth}
\centering
\includegraphics[width=0.27650431\textwidth]{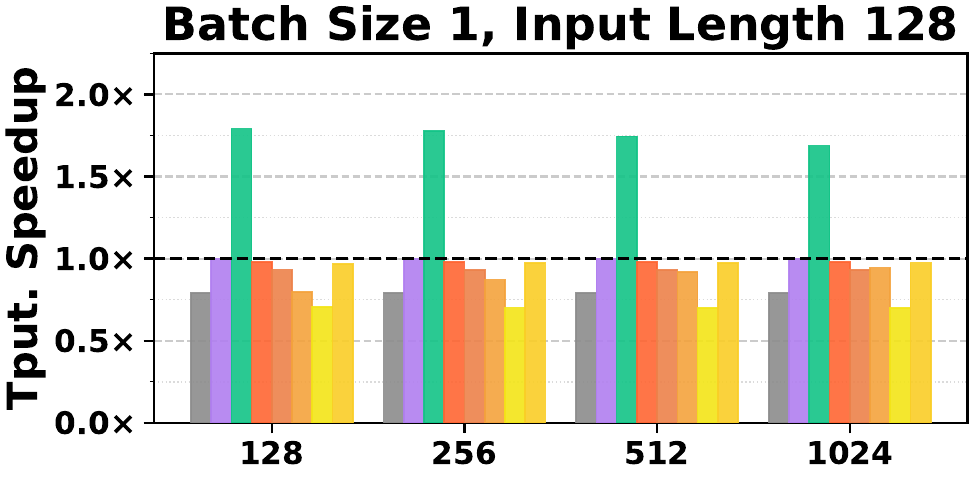}\hfill
\includegraphics[width=0.23449856\textwidth]{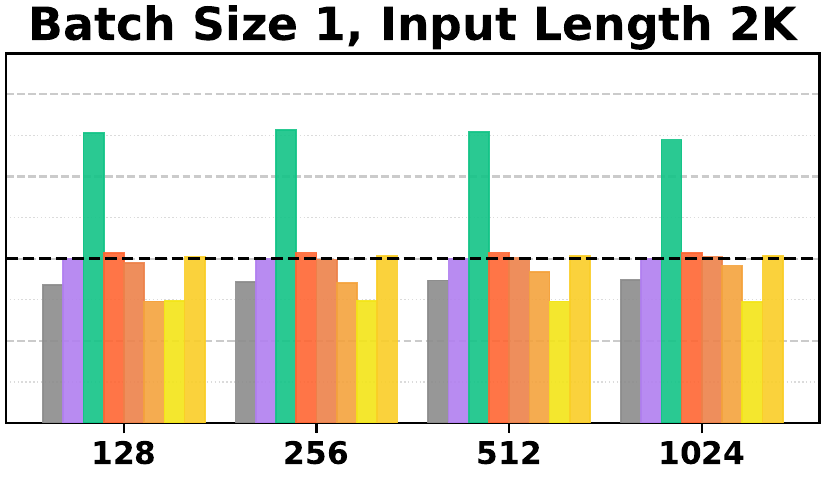}\hfill
\includegraphics[width=0.23449856\textwidth]{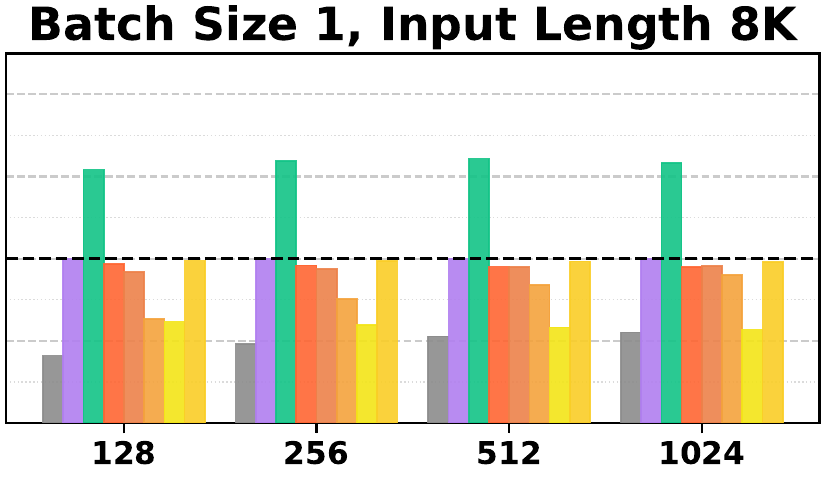}\hfill
\includegraphics[width=0.23449856\textwidth]{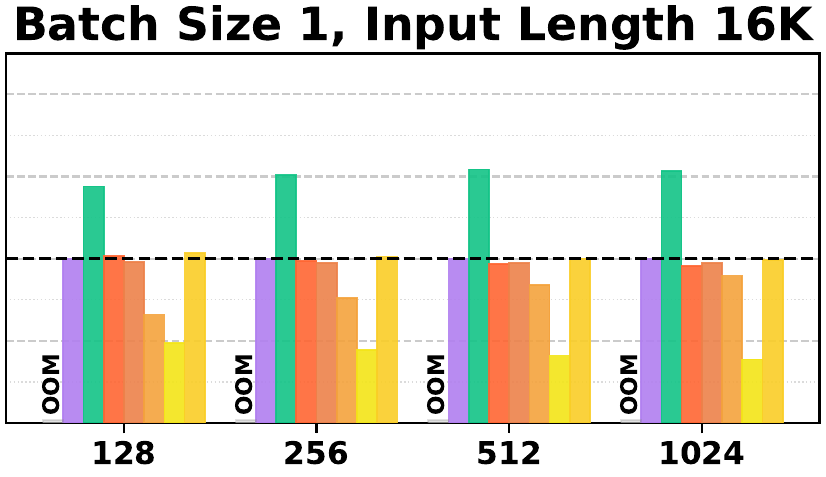}\\[-3pt]
\includegraphics[width=0.27650431\textwidth]{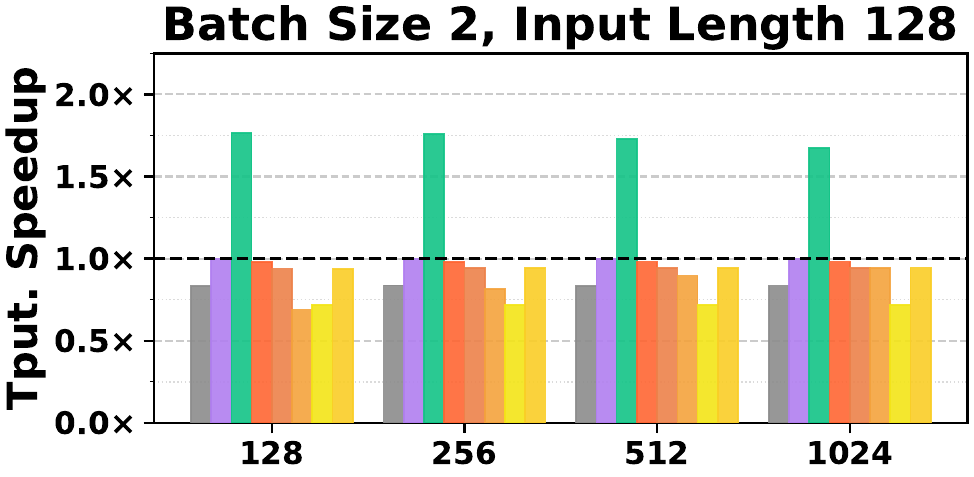}\hfill
\includegraphics[width=0.23449856\textwidth]{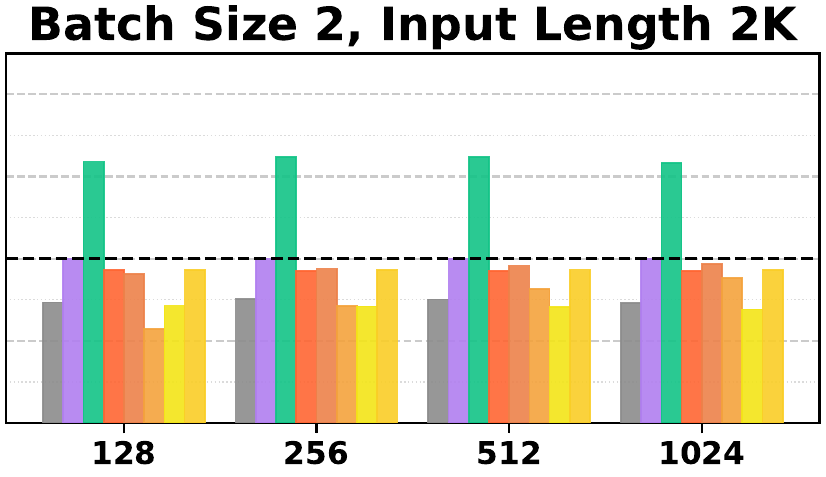}\hfill
\includegraphics[width=0.23449856\textwidth]{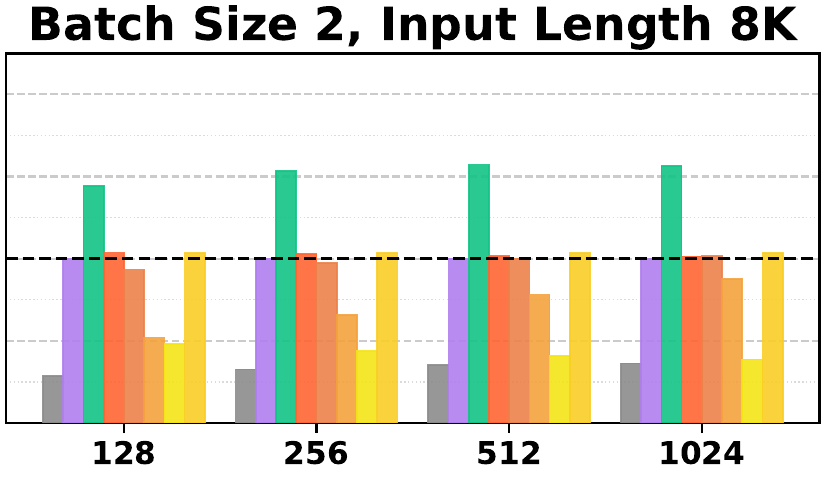}\hfill
\includegraphics[width=0.23449856\textwidth]{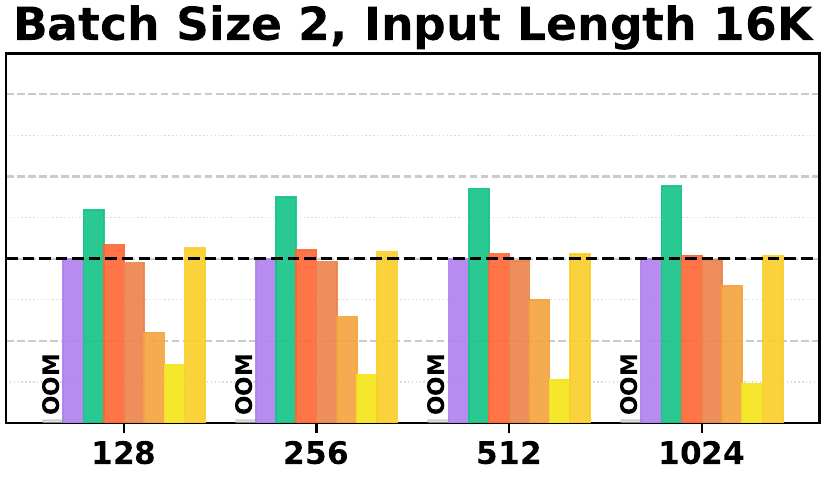}\\[-3pt]
\includegraphics[width=0.27650431\textwidth]{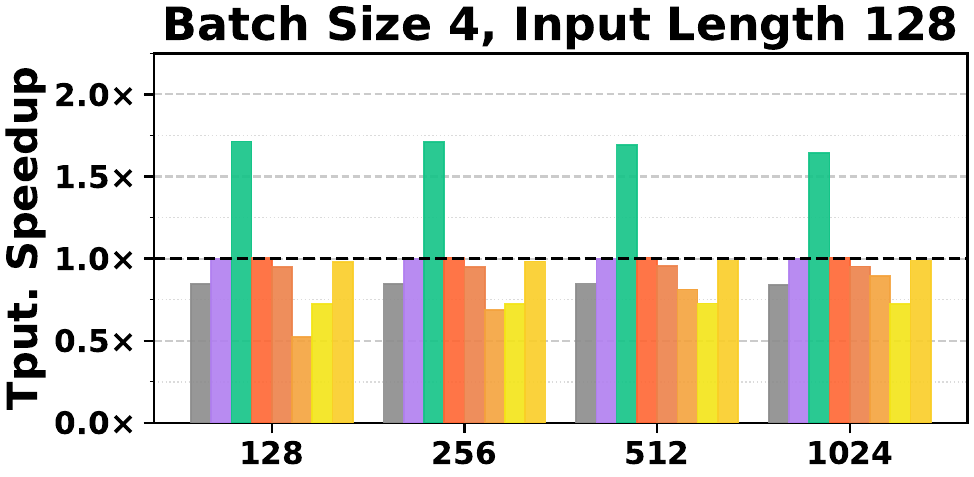}\hfill
\includegraphics[width=0.23449856\textwidth]{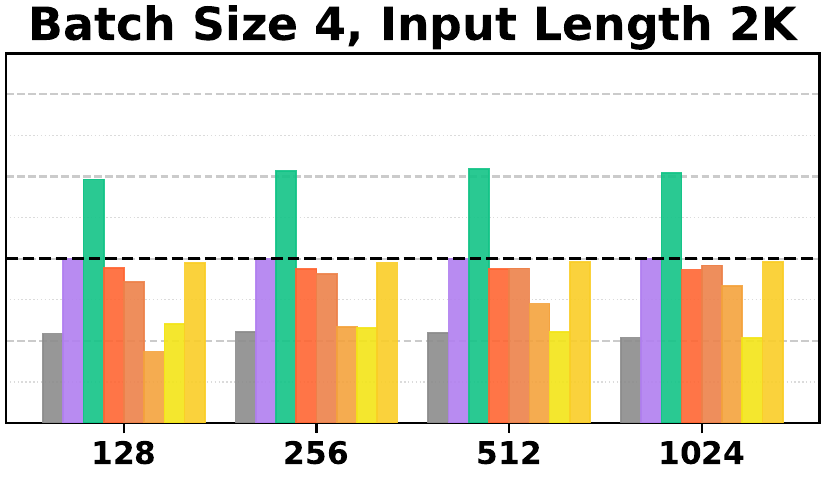}\hfill
\includegraphics[width=0.23449856\textwidth]{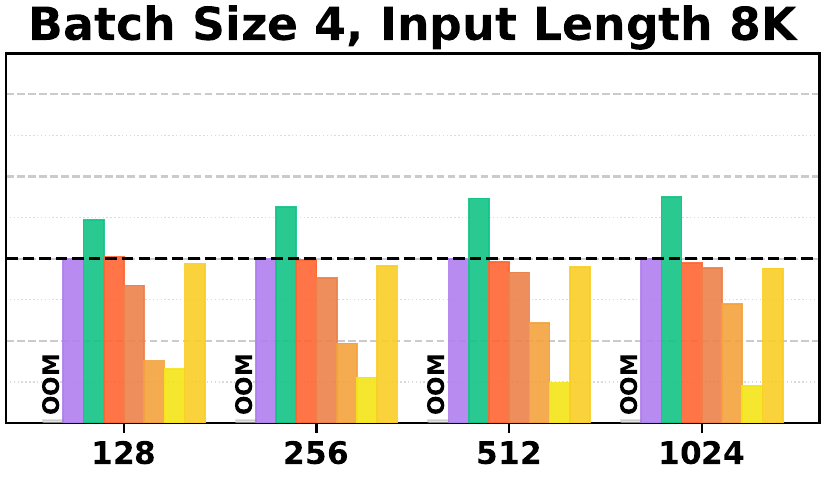}\hfill
\includegraphics[width=0.23449856\textwidth]{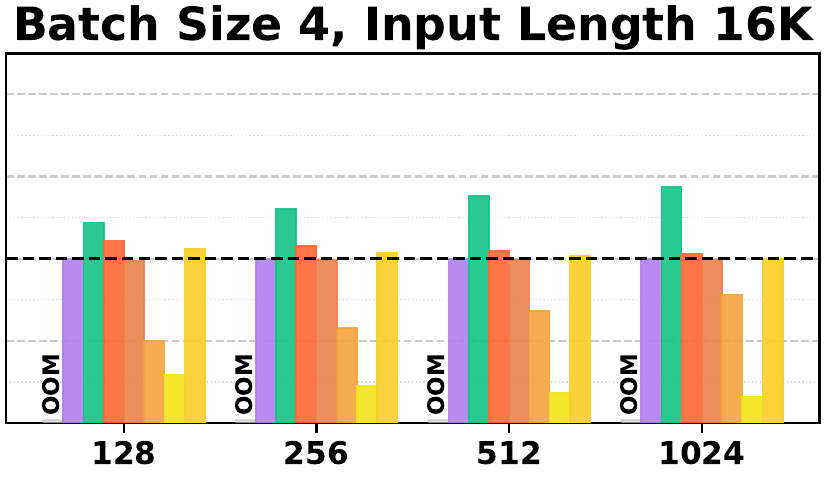}\\[-3pt]
\includegraphics[width=0.27650431\textwidth]{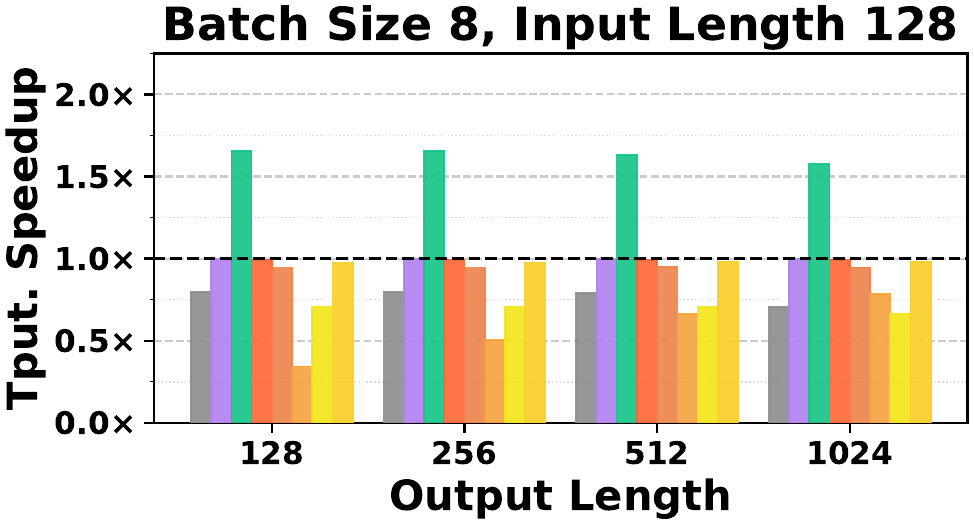}\hfill
\includegraphics[width=0.23449856\textwidth]{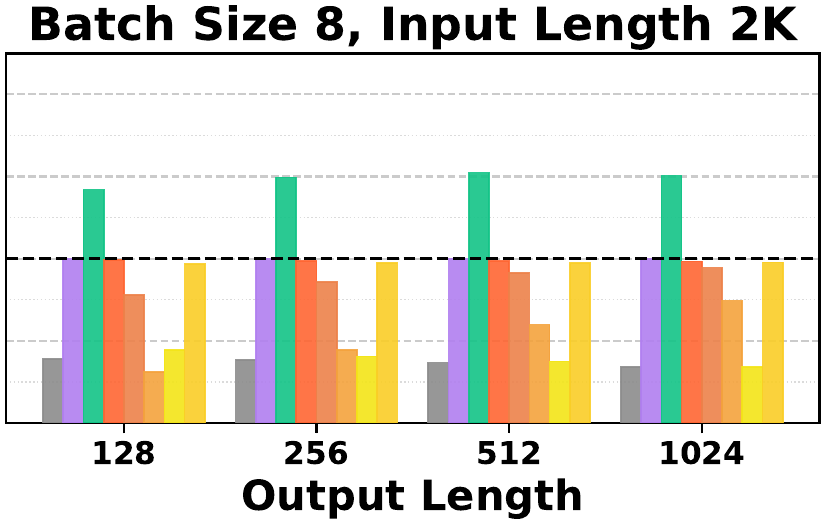}\hfill
\includegraphics[width=0.23449856\textwidth]{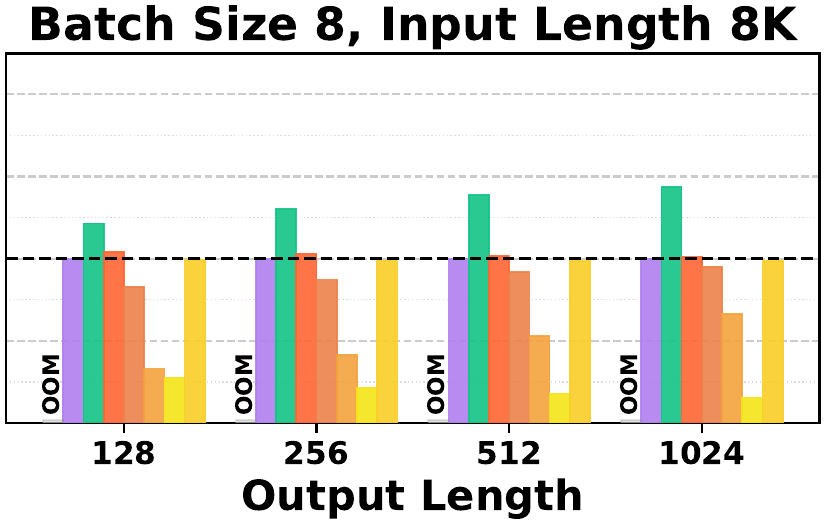}\hfill
\includegraphics[width=0.23449856\textwidth]{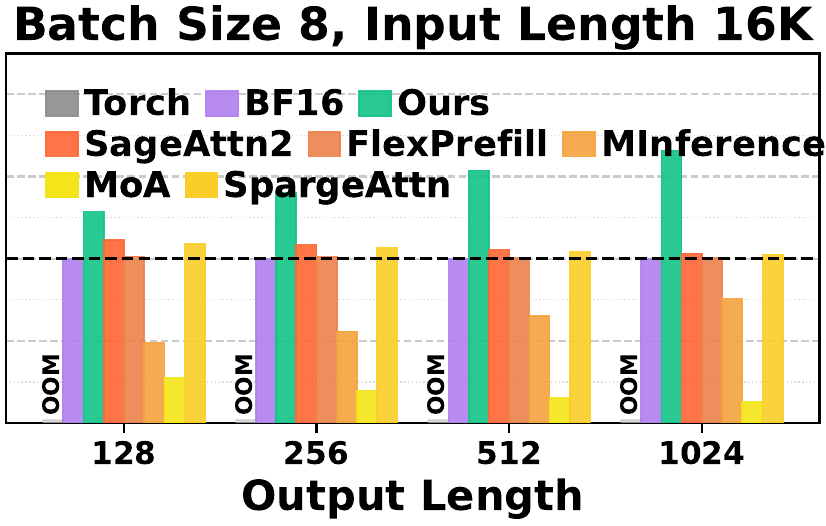}
\vspace{-8pt}
\end{minipage}
\end{adjustbox}
\caption{\textbf{Generation-throughput ratio on H100 NVL.} Ministral-3-8B-Instruct-2512-BF16. Rows use batch sizes 1, 2, 4, and 8; columns use 128, 2,048, 8,192, 16,384 input tokens. Each panel varies output length over 128, 256, 512, and 1024 tokens. BF16 is normalized to $1\times$; higher is better. Method colors and order are fixed across panels.}
\label{fig:system-h100-ministral3_8b-throughput}
\par\vspace{8pt}
\begin{adjustbox}{max width=\textwidth,max totalheight=0.40\textheight}
\begin{minipage}{\textwidth}
\centering
\includegraphics[width=0.27650431\textwidth]{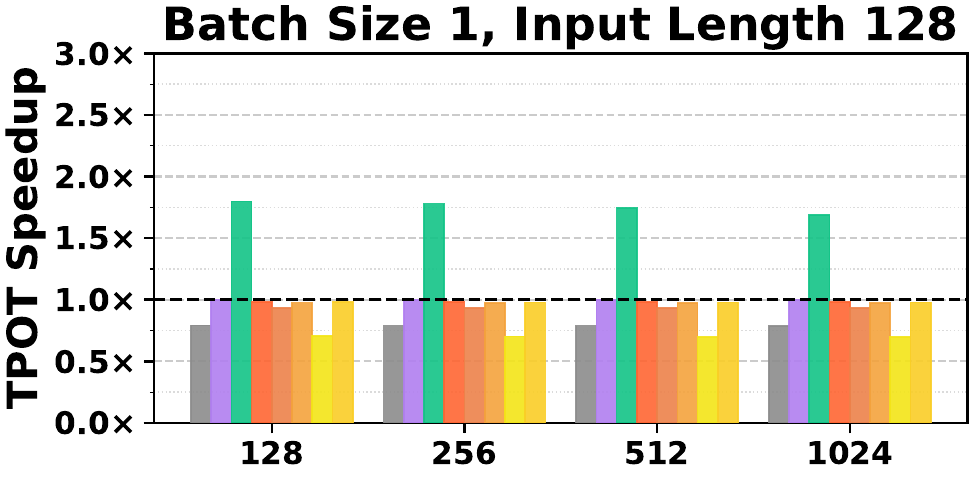}\hfill
\includegraphics[width=0.23449856\textwidth]{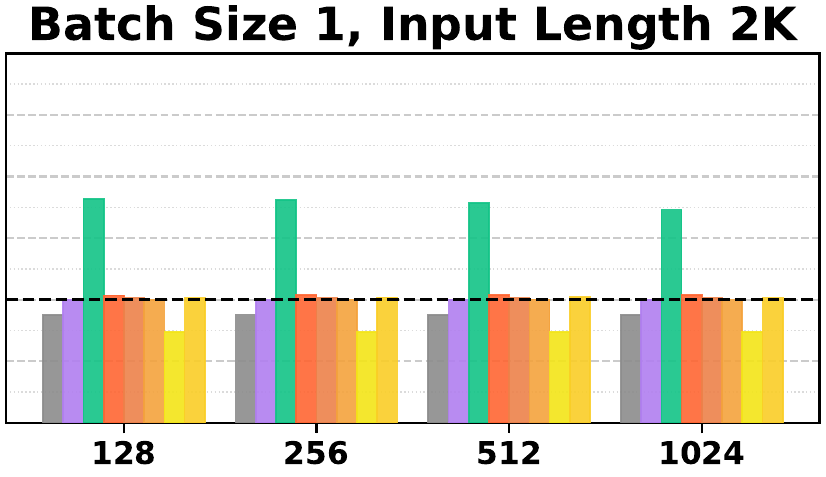}\hfill
\includegraphics[width=0.23449856\textwidth]{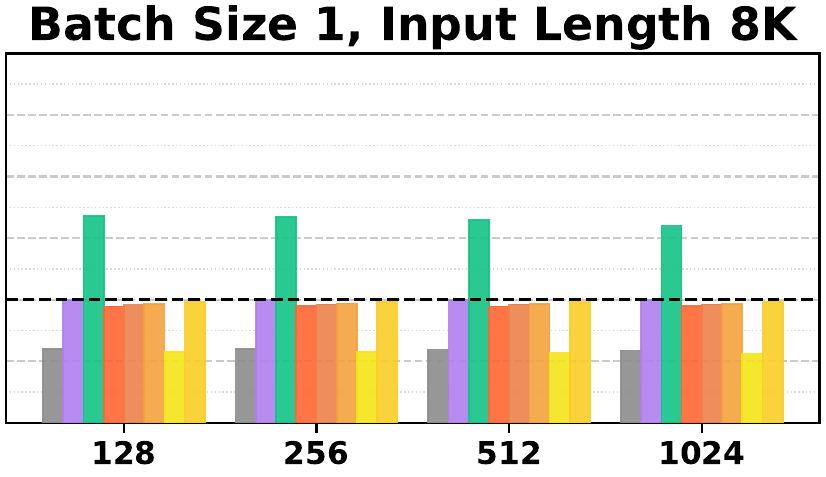}\hfill
\includegraphics[width=0.23449856\textwidth]{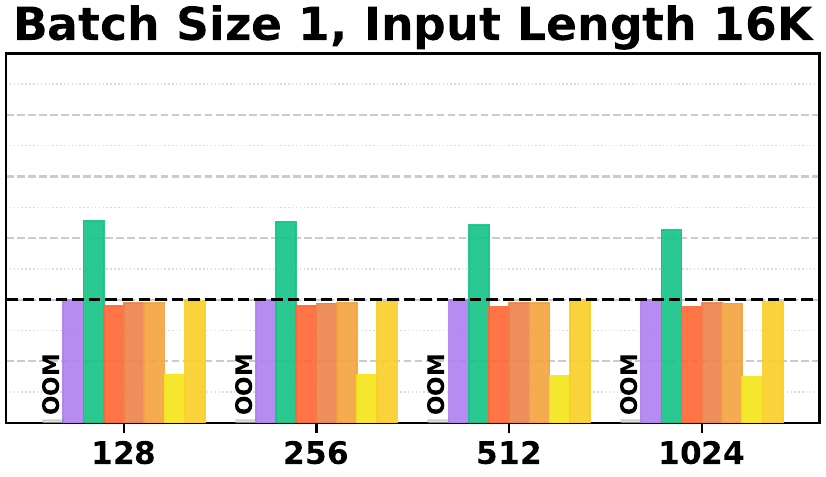}\\[-3pt]
\includegraphics[width=0.27650431\textwidth]{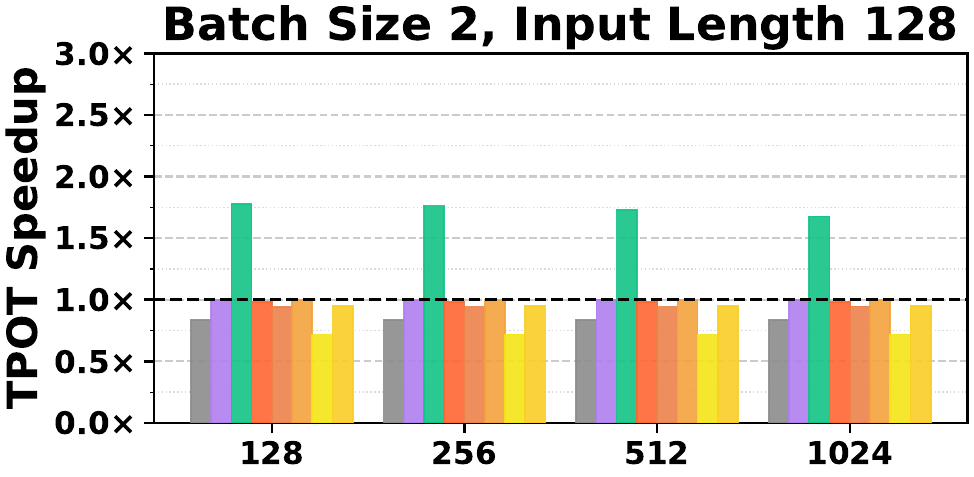}\hfill
\includegraphics[width=0.23449856\textwidth]{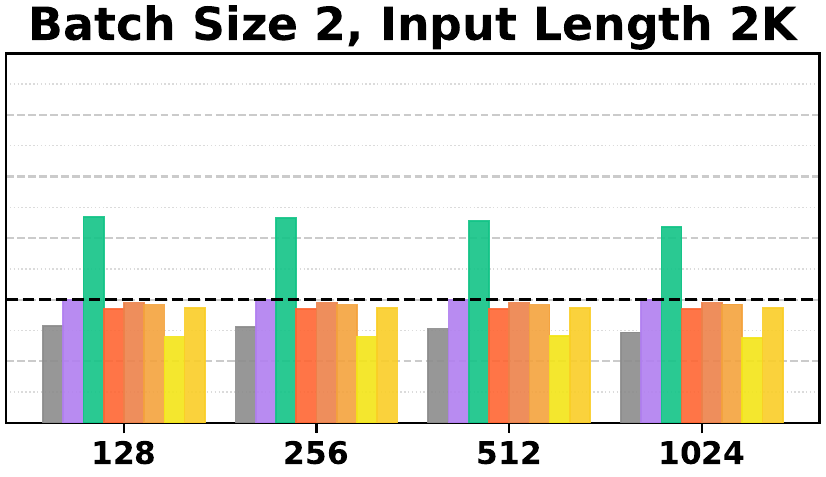}\hfill
\includegraphics[width=0.23449856\textwidth]{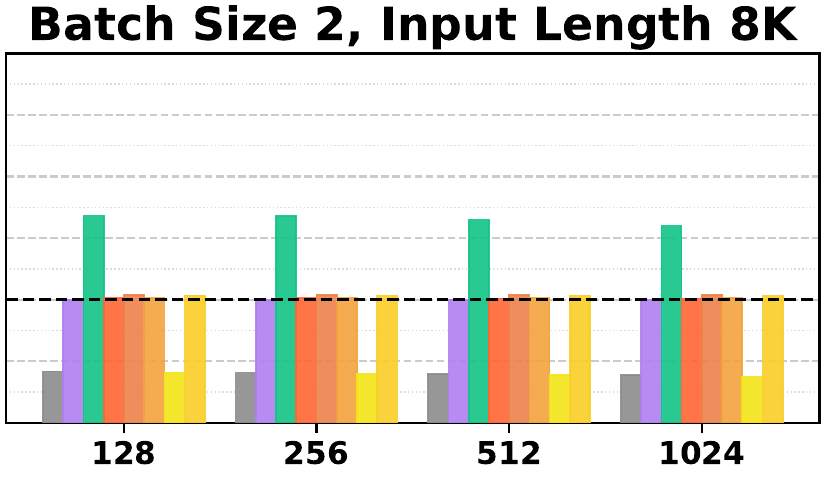}\hfill
\includegraphics[width=0.23449856\textwidth]{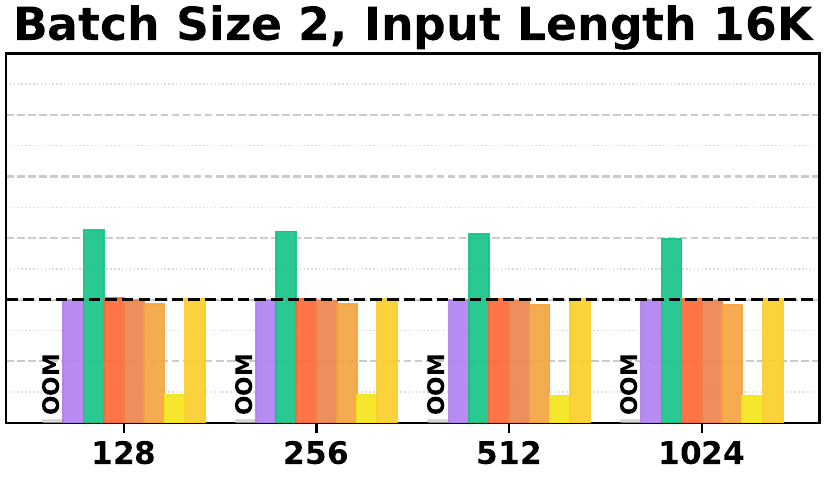}\\[-3pt]
\includegraphics[width=0.27650431\textwidth]{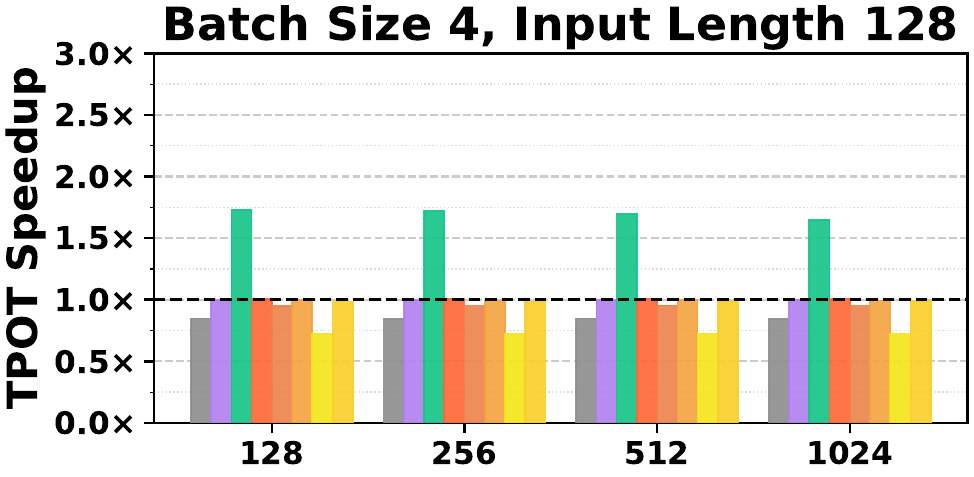}\hfill
\includegraphics[width=0.23449856\textwidth]{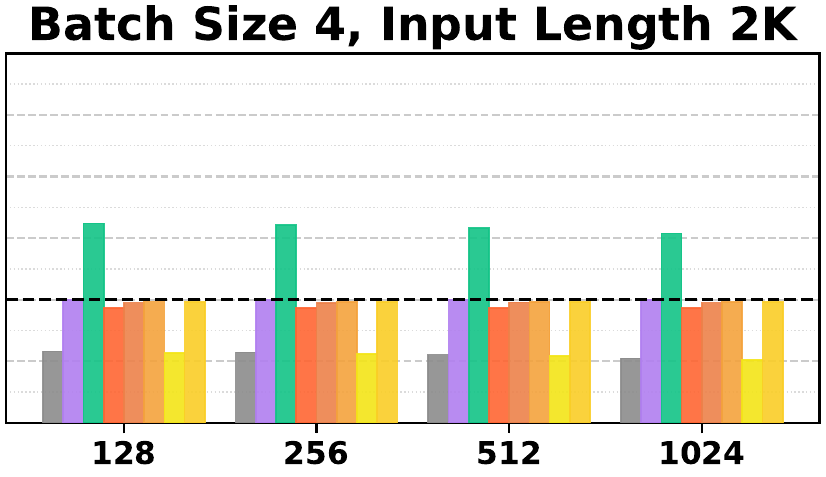}\hfill
\includegraphics[width=0.23449856\textwidth]{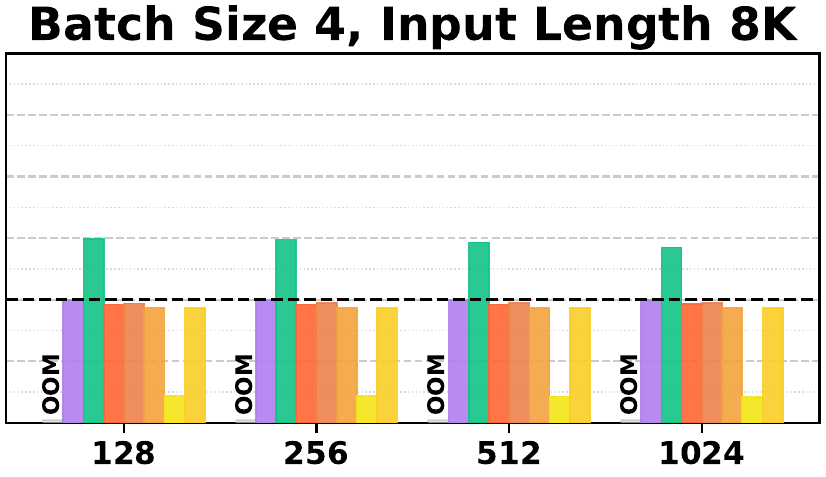}\hfill
\includegraphics[width=0.23449856\textwidth]{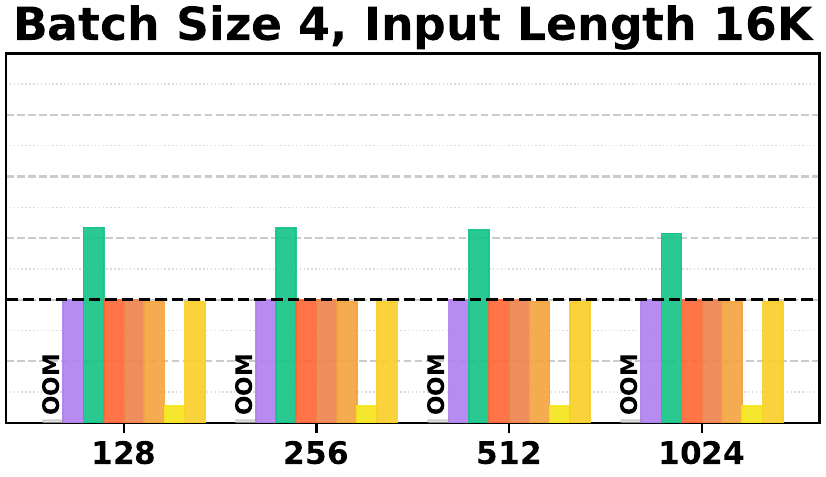}\\[-3pt]
\includegraphics[width=0.27650431\textwidth]{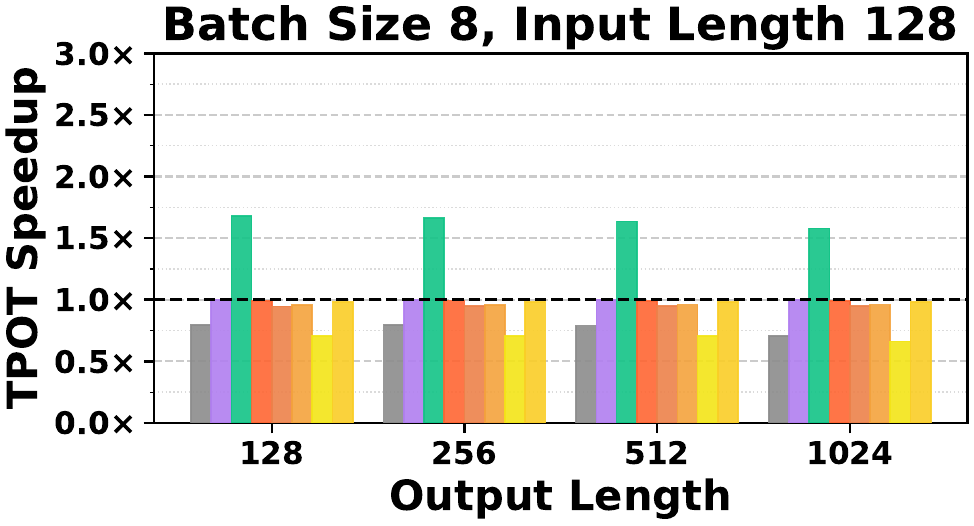}\hfill
\includegraphics[width=0.23449856\textwidth]{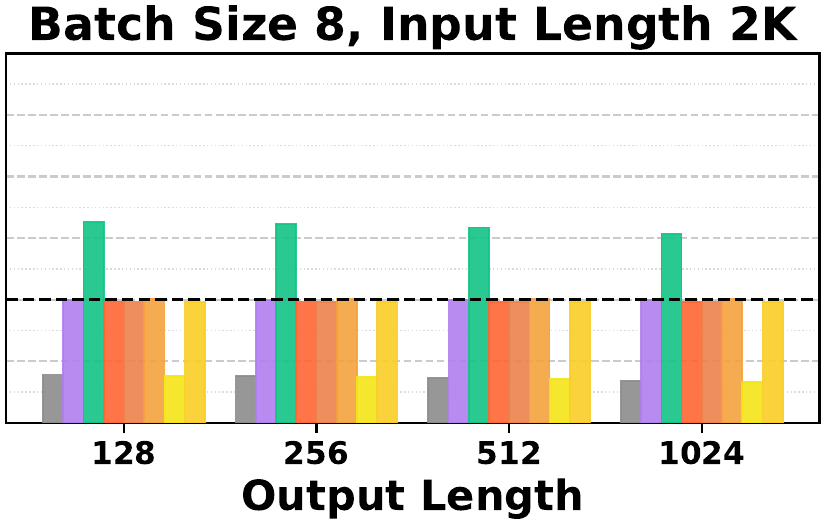}\hfill
\includegraphics[width=0.23449856\textwidth]{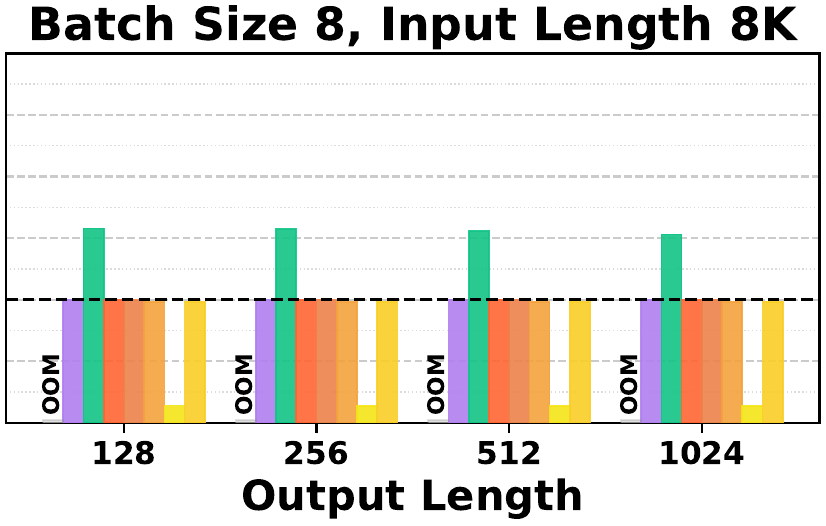}\hfill
\includegraphics[width=0.23449856\textwidth]{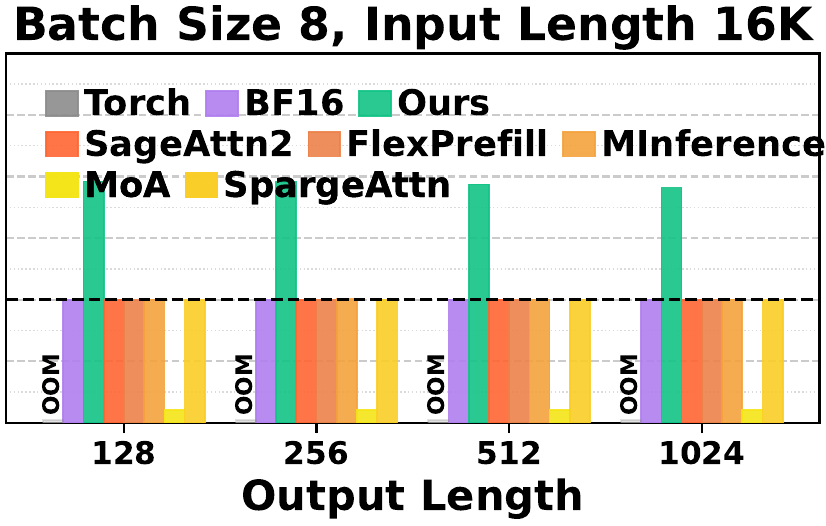}
\vspace{-8pt}
\end{minipage}
\end{adjustbox}
\caption{\textbf{TPOT speedup on H100 NVL.} Ministral-3-8B-Instruct-2512-BF16. Rows use batch sizes 1, 2, 4, and 8; columns use 128, 2,048, 8,192, 16,384 input tokens. Each panel varies output length over 128, 256, 512, and 1024 tokens. BF16 is normalized to $1\times$; higher is better. Method colors and order are fixed across panels.}
\label{fig:system-h100-ministral3_8b-tpot}
\end{figure*}

\begin{figure*}[!p]
\centering
\begin{minipage}[t]{\dimexpr(\textwidth-\columnsep)/2\relax}
\vspace{0pt}
\begin{minipage}{\linewidth}
\centering
\includegraphics[width=0.53027926\linewidth]{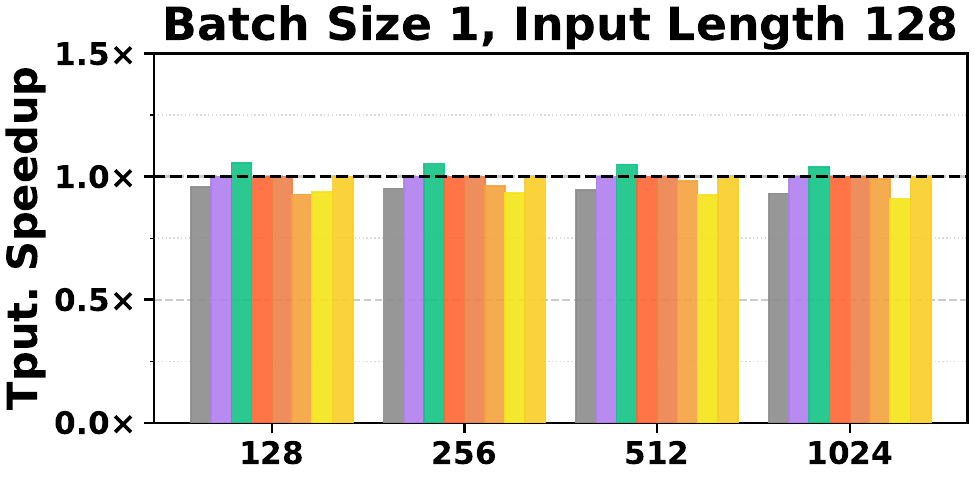}\hfill
\includegraphics[width=0.44972074\linewidth]{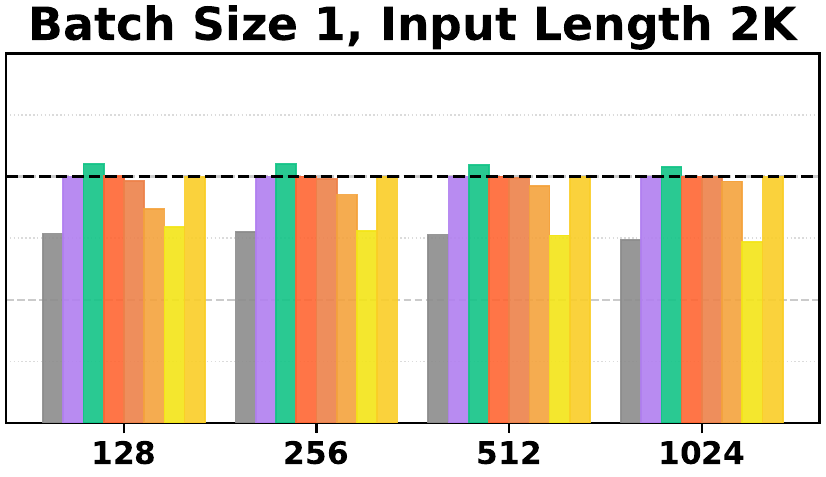}\\[-3pt]
\includegraphics[width=0.53027926\linewidth]{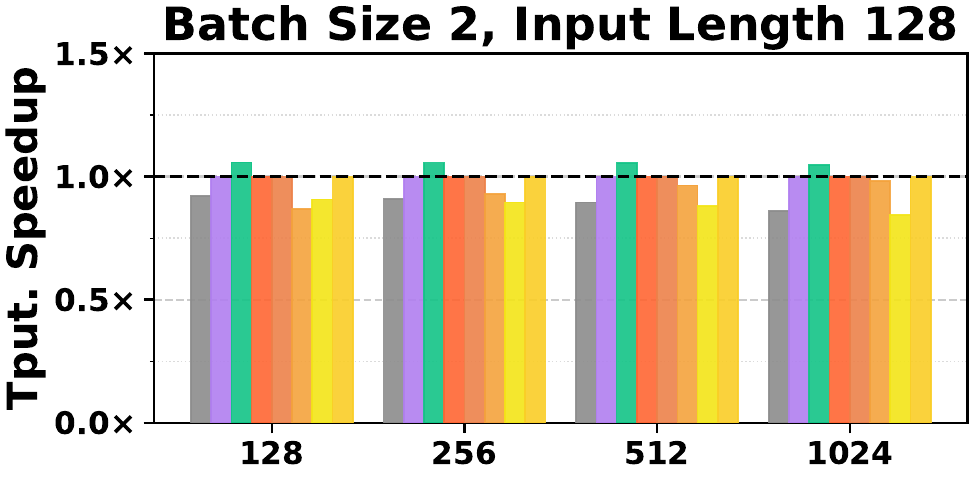}\hfill
\includegraphics[width=0.44972074\linewidth]{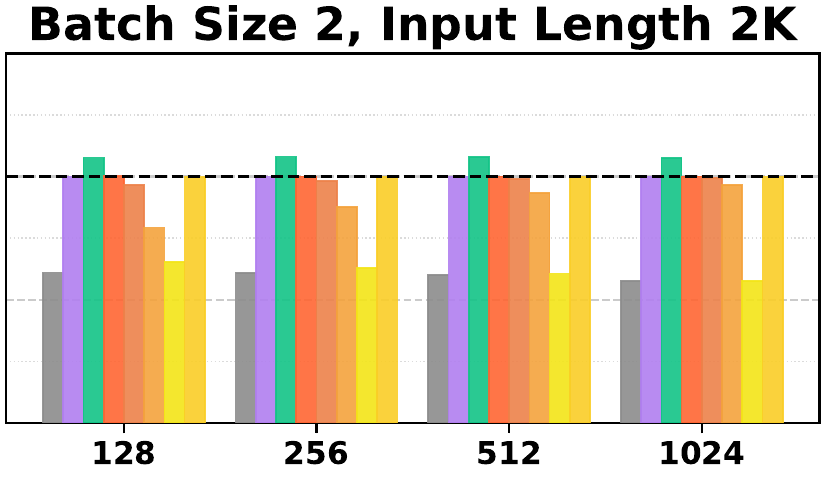}\\[-3pt]
\includegraphics[width=0.53027926\linewidth]{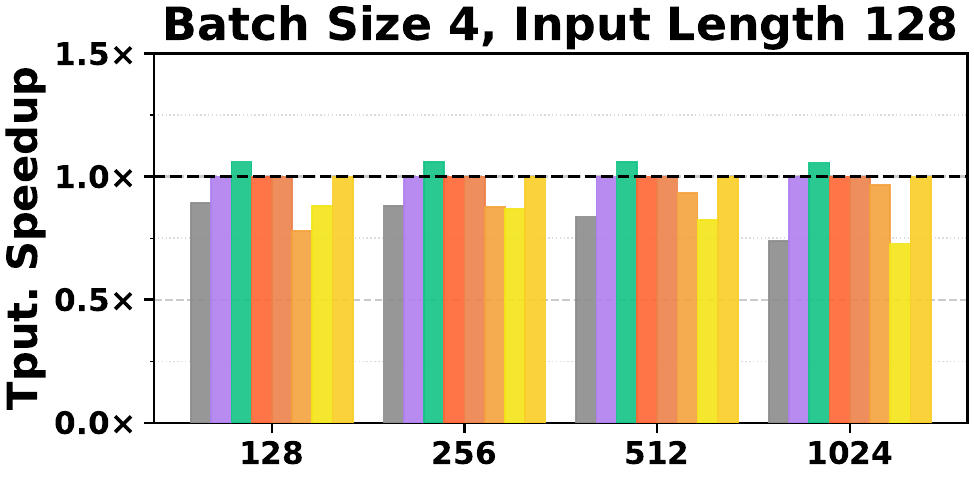}\hfill
\includegraphics[width=0.44972074\linewidth]{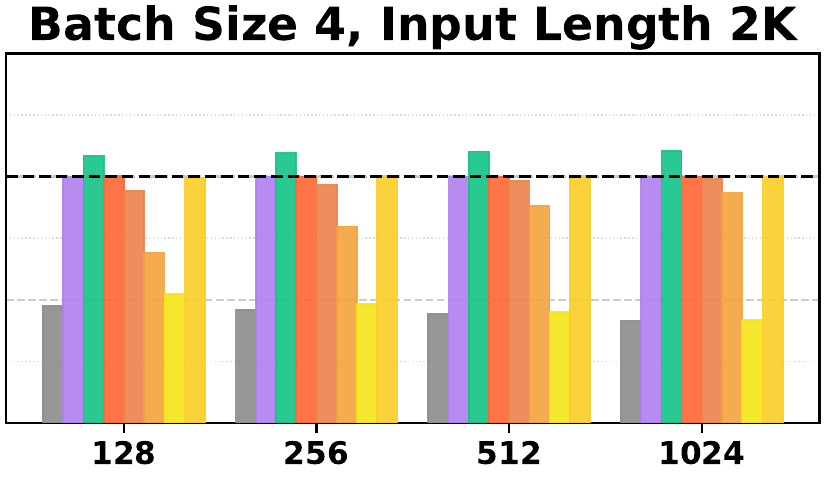}\\[-3pt]
\includegraphics[width=0.53027926\linewidth]{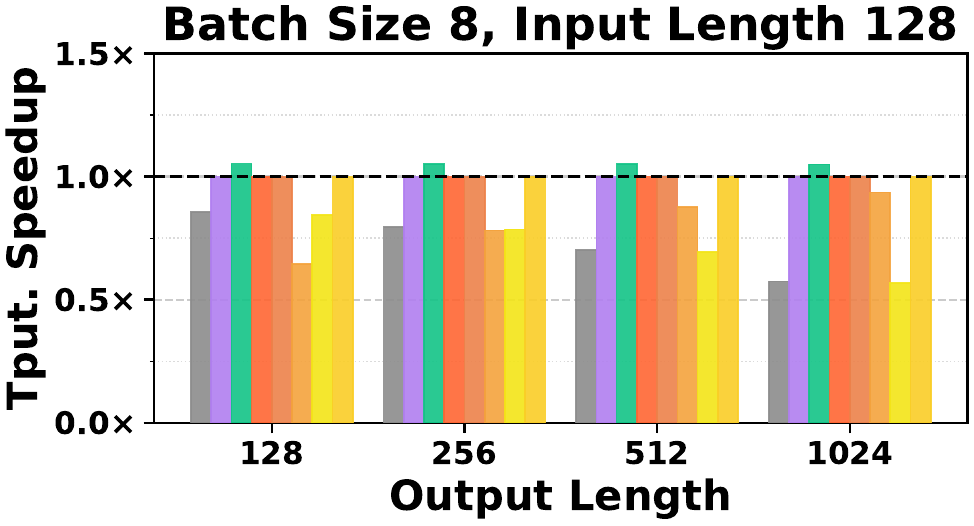}\hfill
\includegraphics[width=0.44972074\linewidth]{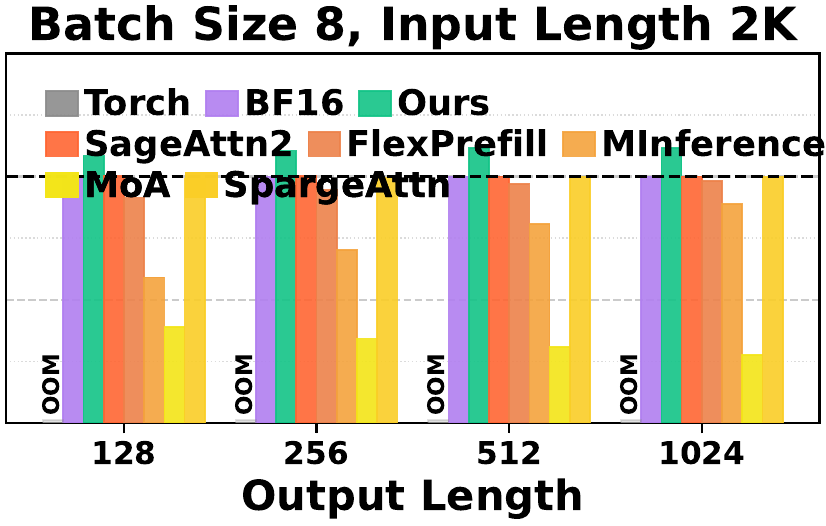}
\vspace{-8pt}
\captionof{figure}{\textbf{Generation-throughput ratio on RTX 4090.} Qwen2-7B-Instruct. Rows use batch sizes 1, 2, 4, and 8; columns use 128, 2,048 input tokens. Each panel varies output length over 128, 256, 512, and 1024 tokens. BF16 is normalized to $1\times$; higher is better. Method colors and order are fixed across panels.}
\label{fig:system-rtx4090-qwen2_7b-throughput}
\end{minipage}
\par\vspace{8pt}
\begin{minipage}{\linewidth}
\centering
\includegraphics[width=0.53027926\linewidth]{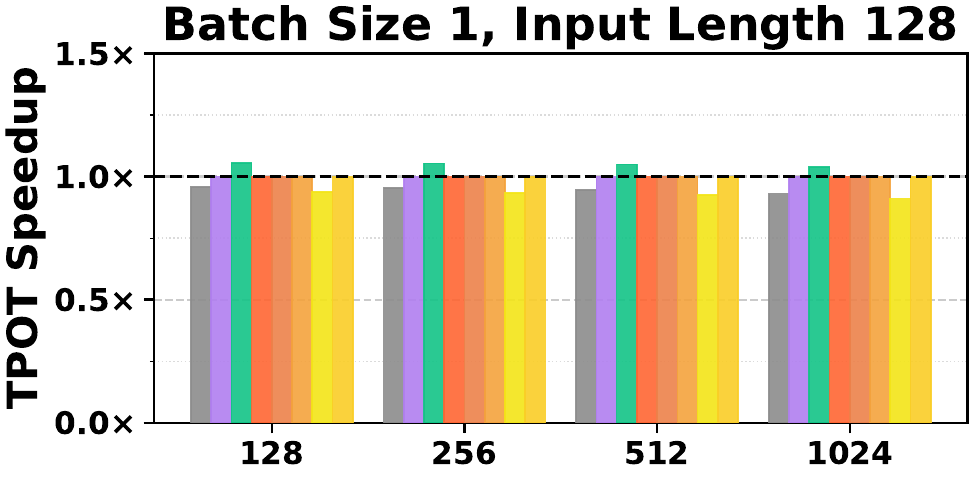}\hfill
\includegraphics[width=0.44972074\linewidth]{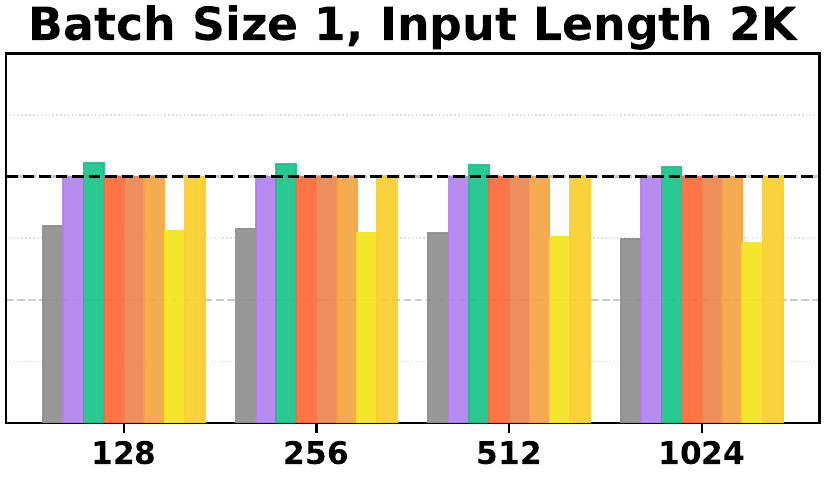}\\[-3pt]
\includegraphics[width=0.53027926\linewidth]{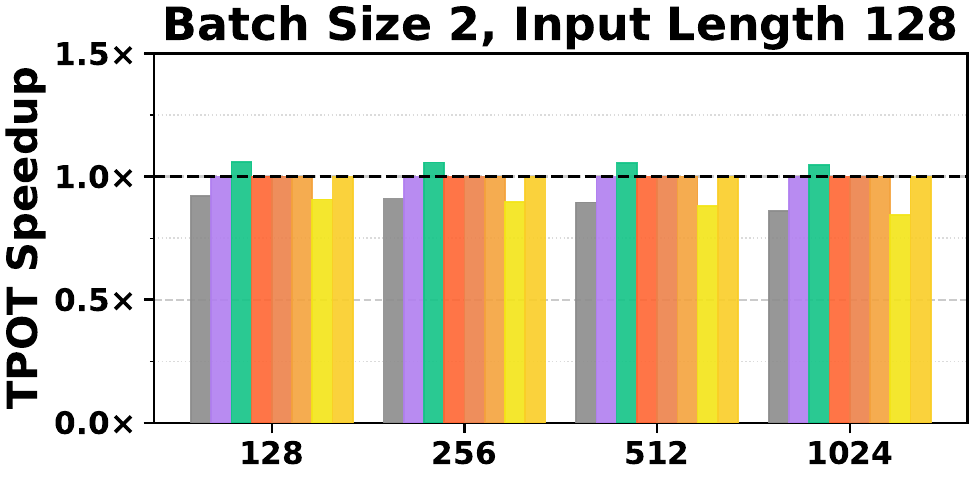}\hfill
\includegraphics[width=0.44972074\linewidth]{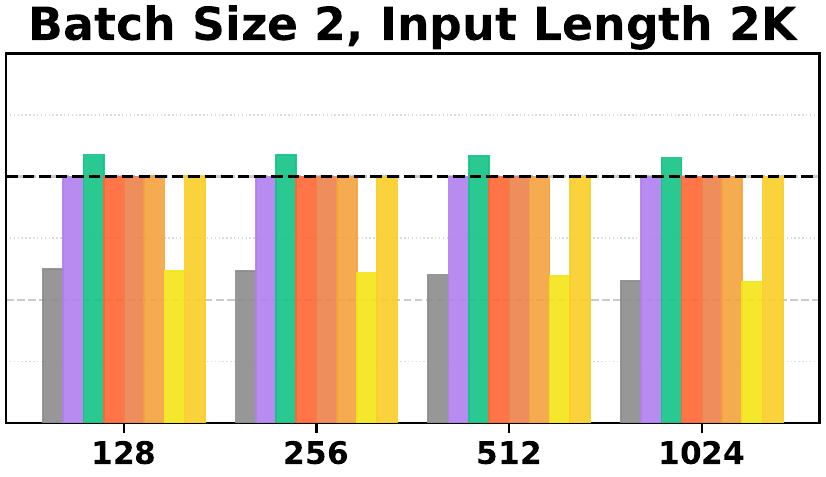}\\[-3pt]
\includegraphics[width=0.53027926\linewidth]{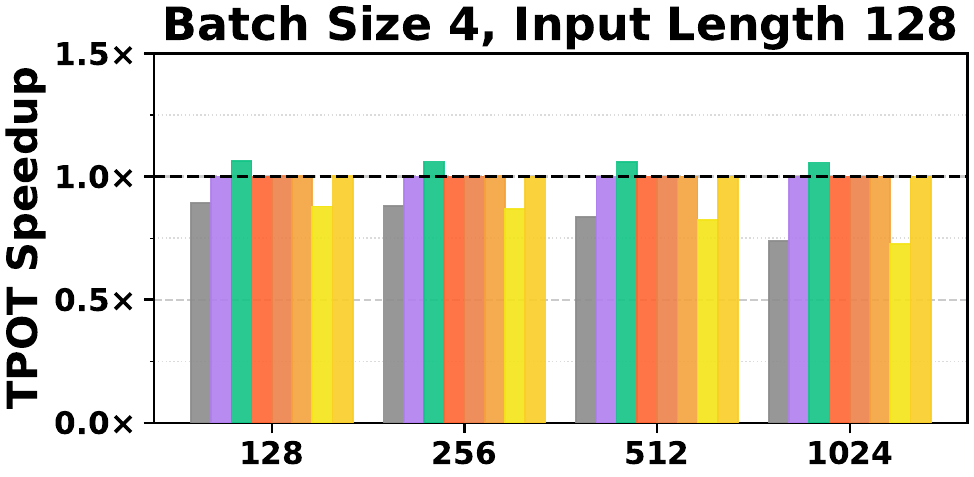}\hfill
\includegraphics[width=0.44972074\linewidth]{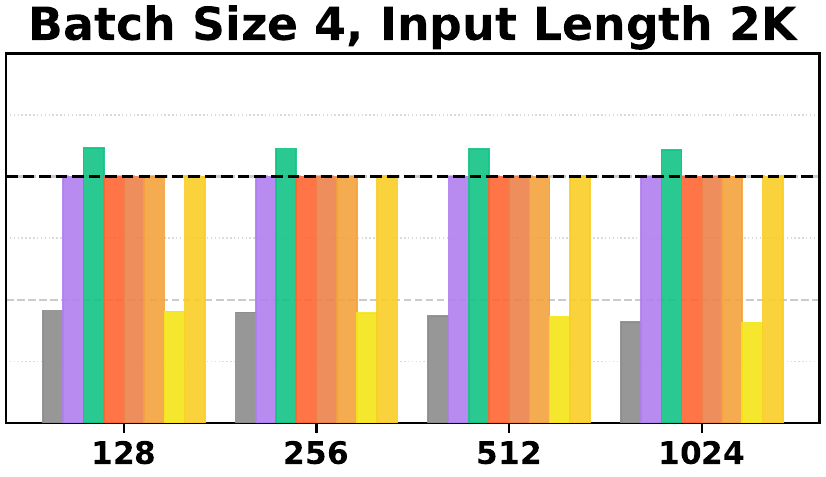}\\[-3pt]
\includegraphics[width=0.53027926\linewidth]{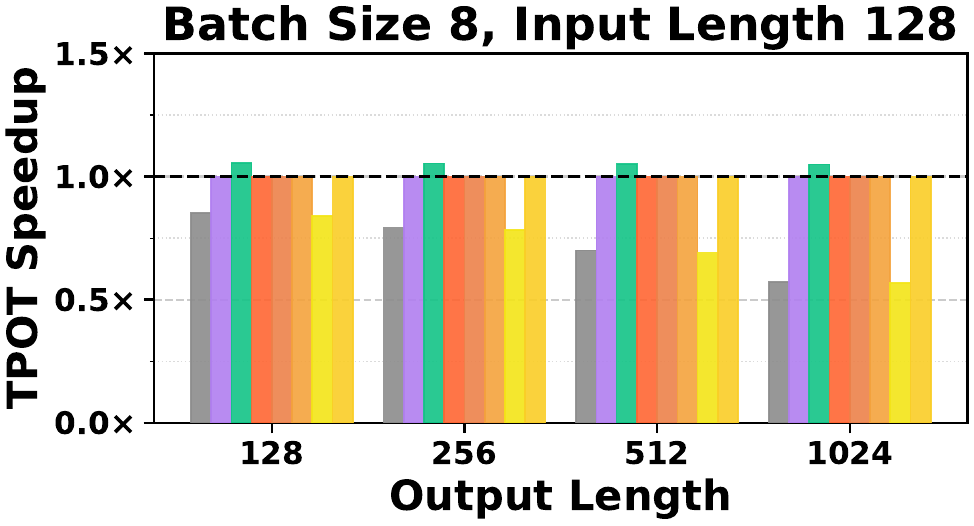}\hfill
\includegraphics[width=0.44972074\linewidth]{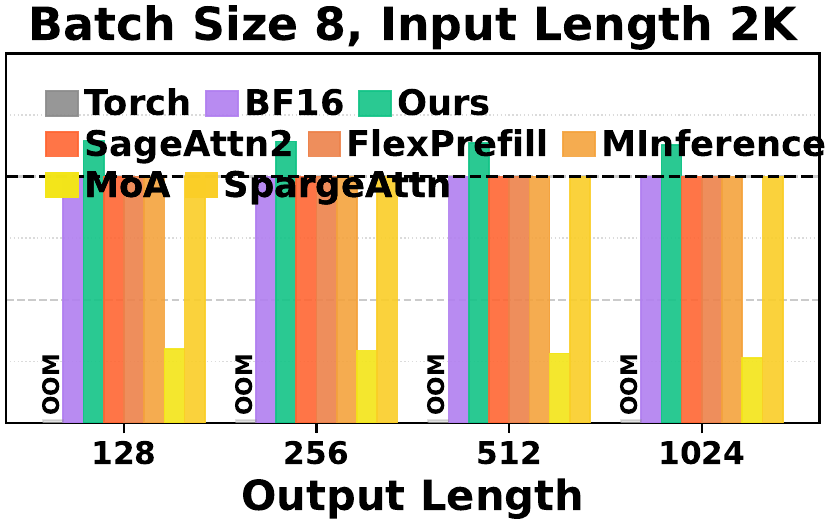}
\vspace{-8pt}
\captionof{figure}{\textbf{TPOT speedup on RTX 4090.} Qwen2-7B-Instruct. Rows use batch sizes 1, 2, 4, and 8; columns use 128, 2,048 input tokens. Each panel varies output length over 128, 256, 512, and 1024 tokens. BF16 is normalized to $1\times$; higher is better. Method colors and order are fixed across panels.}
\label{fig:system-rtx4090-qwen2_7b-tpot}
\end{minipage}
\end{minipage}\hfill
\begin{minipage}[t]{\dimexpr(\textwidth-\columnsep)/2\relax}
\vspace{0pt}
\begin{minipage}{\linewidth}
\centering
\includegraphics[width=0.53027926\linewidth]{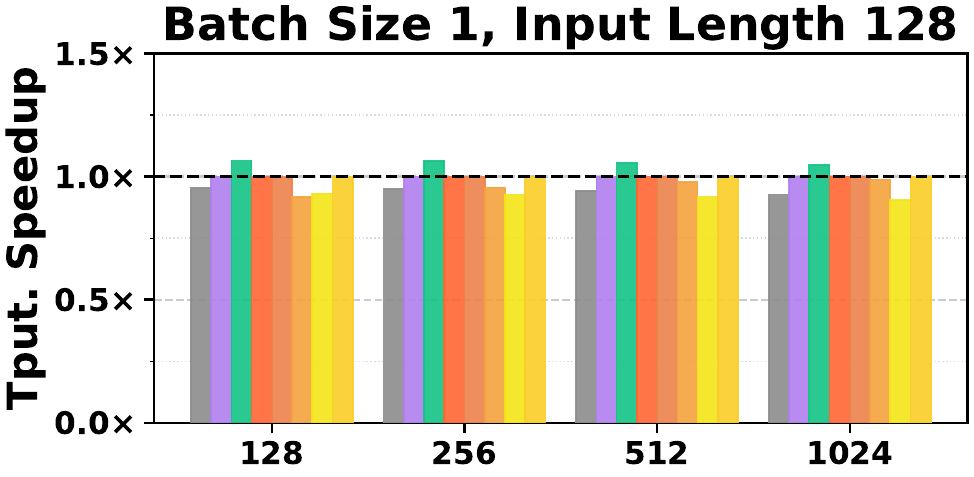}\hfill
\includegraphics[width=0.44972074\linewidth]{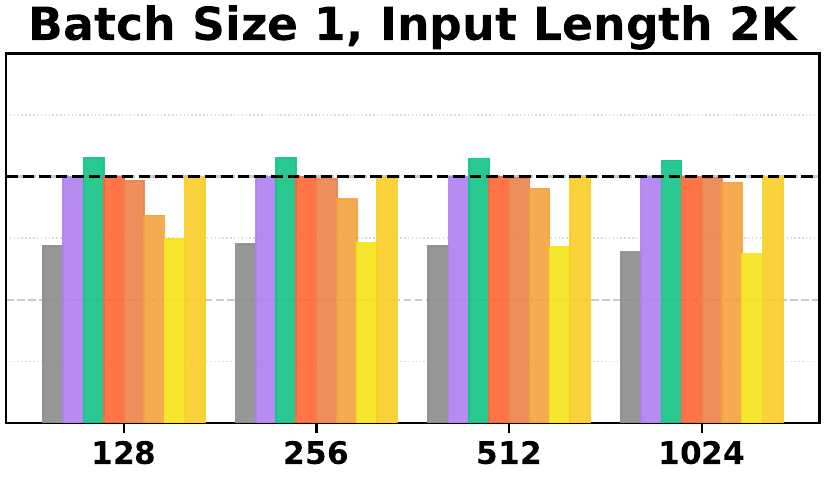}\\[-3pt]
\includegraphics[width=0.53027926\linewidth]{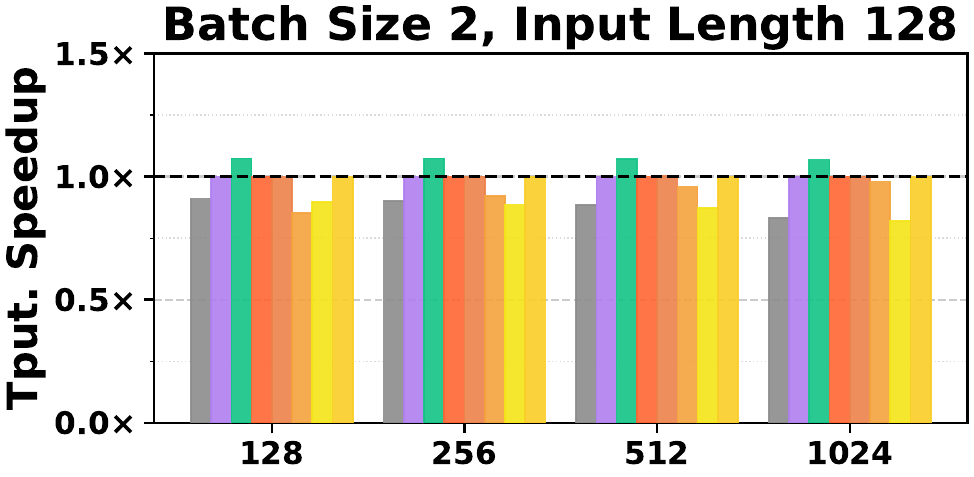}\hfill
\includegraphics[width=0.44972074\linewidth]{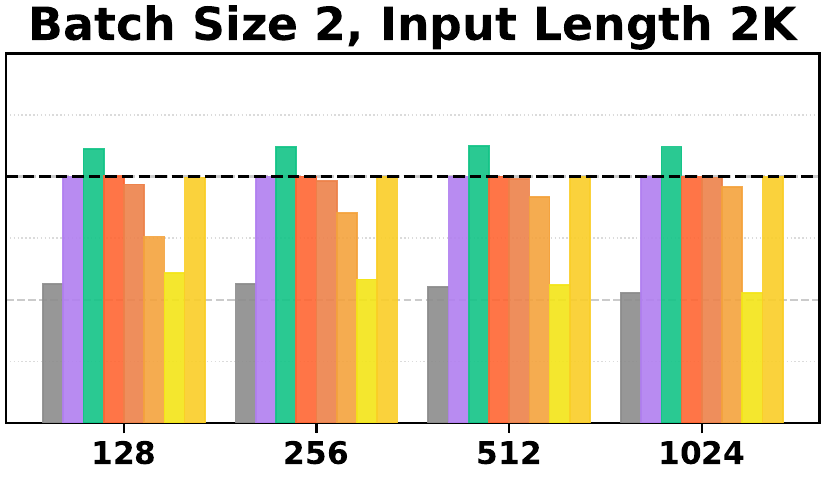}\\[-3pt]
\includegraphics[width=0.53027926\linewidth]{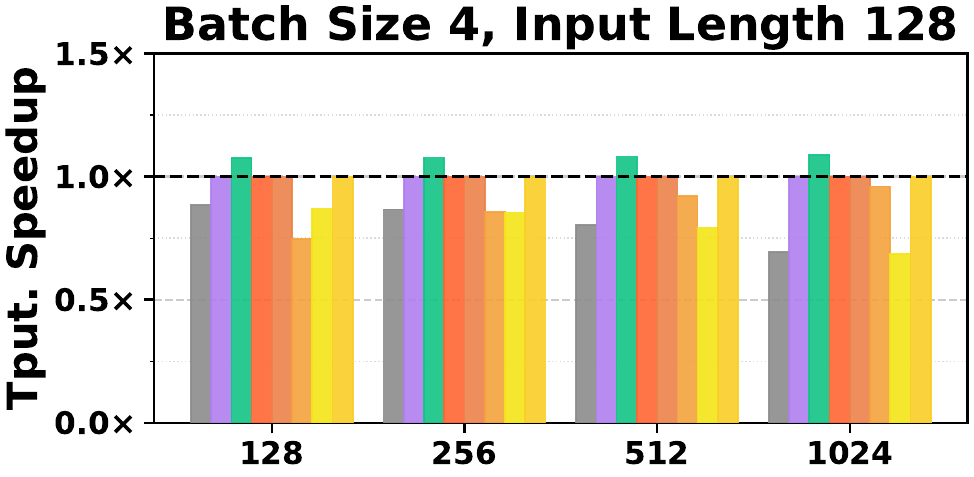}\hfill
\includegraphics[width=0.44972074\linewidth]{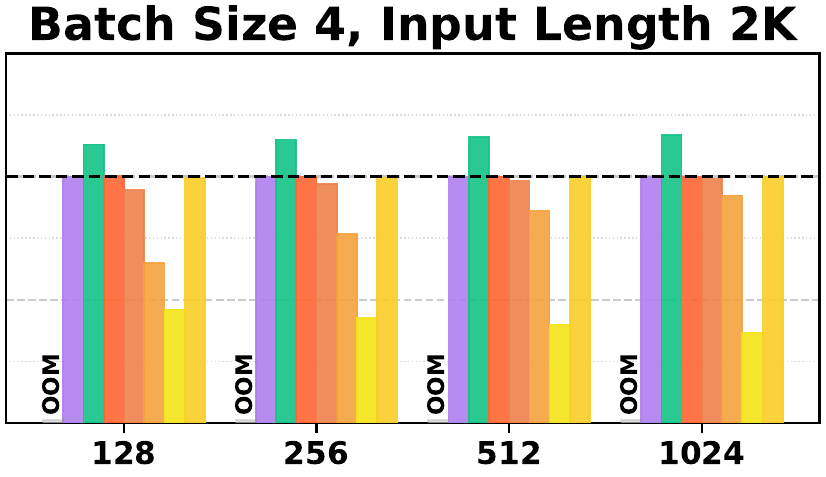}\\[-3pt]
\includegraphics[width=0.53027926\linewidth]{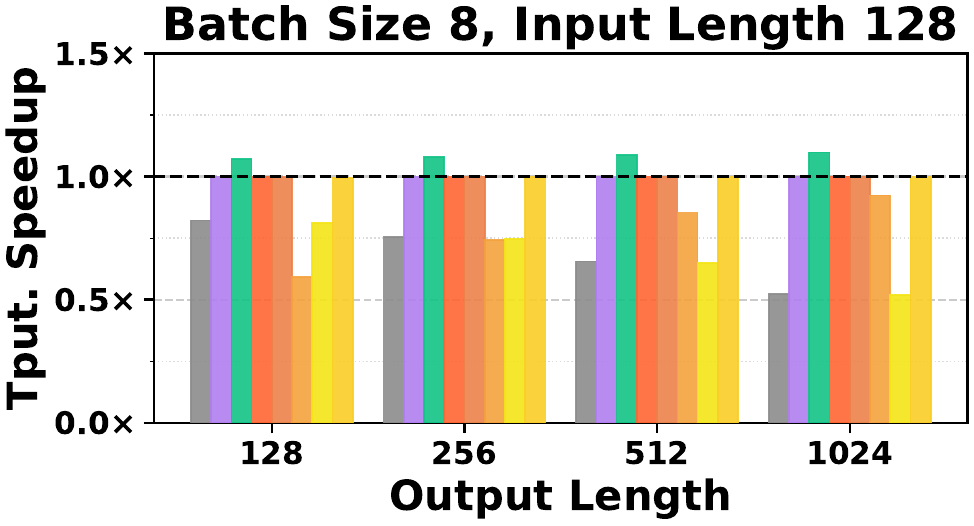}\hfill
\includegraphics[width=0.44972074\linewidth]{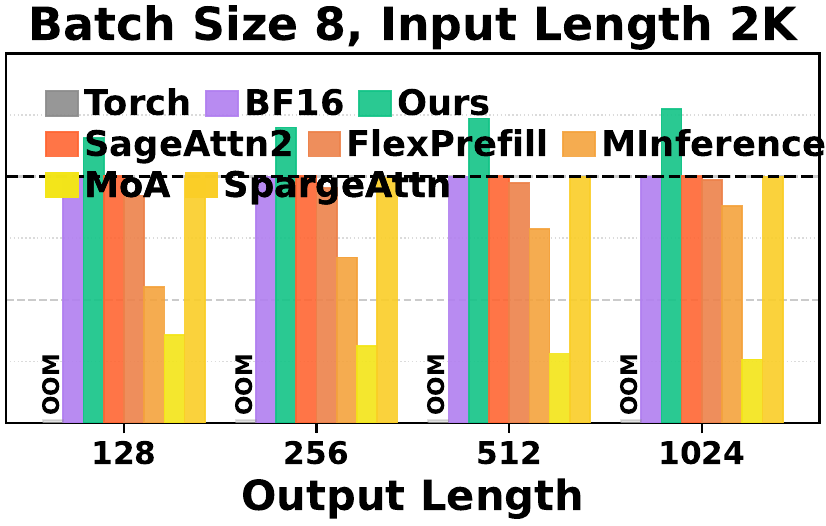}
\vspace{-8pt}
\captionof{figure}{\textbf{Generation-throughput ratio on RTX 4090.} Ministral-3-8B-Instruct-2512-BF16. Rows use batch sizes 1, 2, 4, and 8; columns use 128, 2,048 input tokens. Each panel varies output length over 128, 256, 512, and 1024 tokens. BF16 is normalized to $1\times$; higher is better. Panels share colors and order.}
\label{fig:system-rtx4090-ministral3_8b-throughput}
\end{minipage}
\par\vspace{8pt}
\begin{minipage}{\linewidth}
\centering
\includegraphics[width=0.53027926\linewidth]{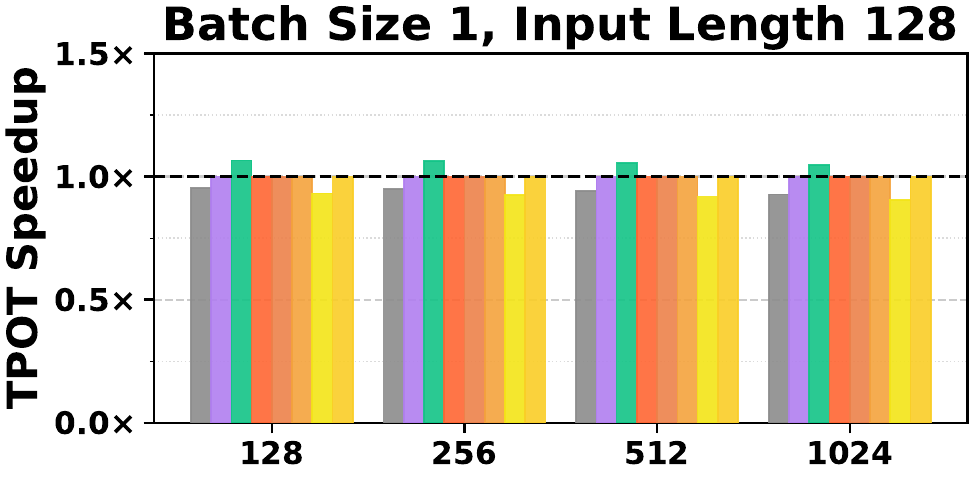}\hfill
\includegraphics[width=0.44972074\linewidth]{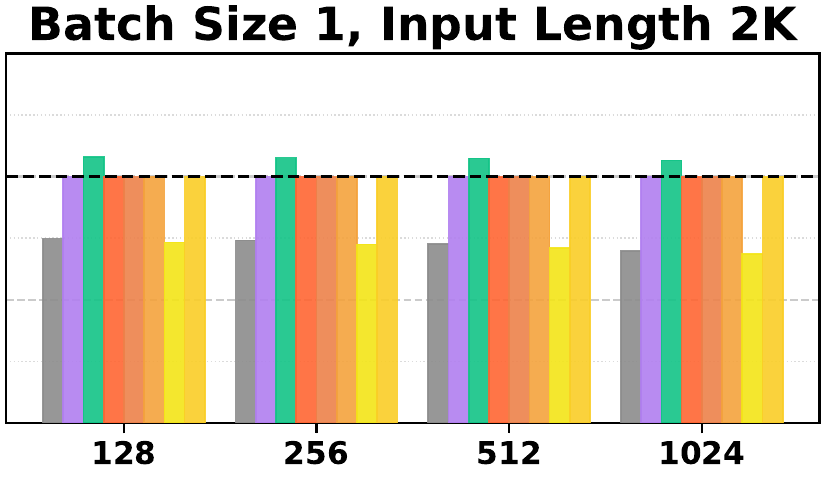}\\[-3pt]
\includegraphics[width=0.53027926\linewidth]{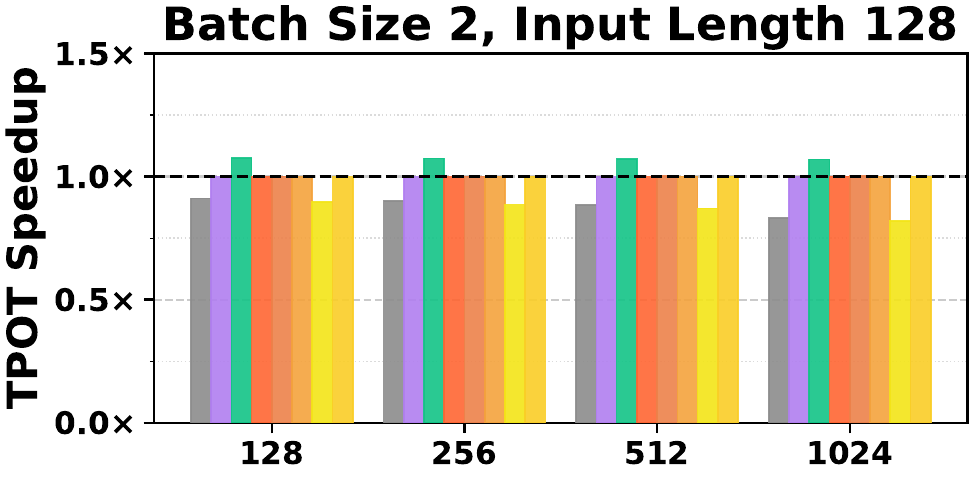}\hfill
\includegraphics[width=0.44972074\linewidth]{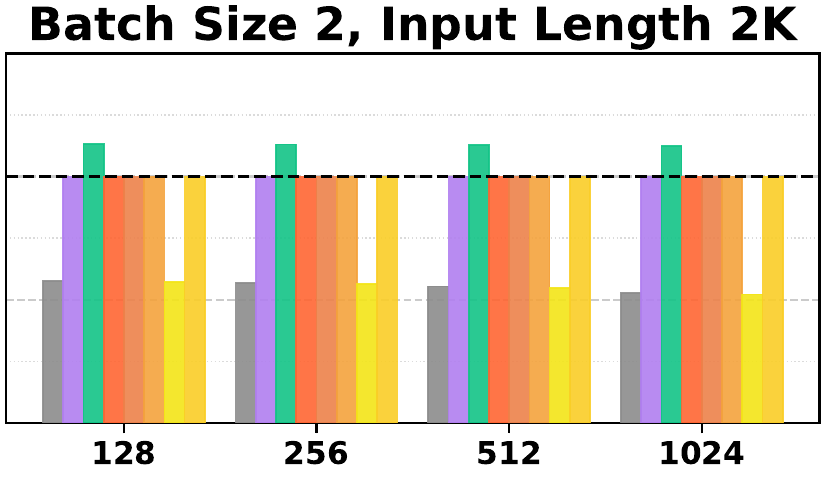}\\[-3pt]
\includegraphics[width=0.53027926\linewidth]{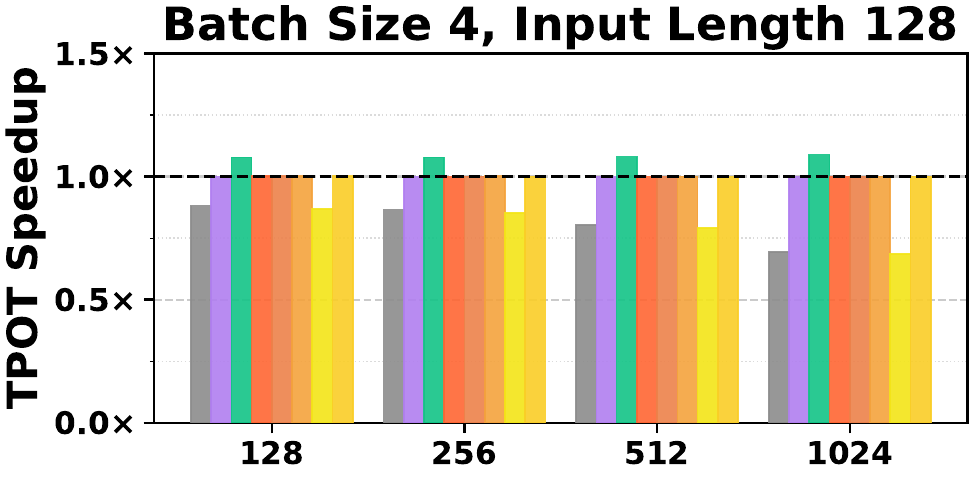}\hfill
\includegraphics[width=0.44972074\linewidth]{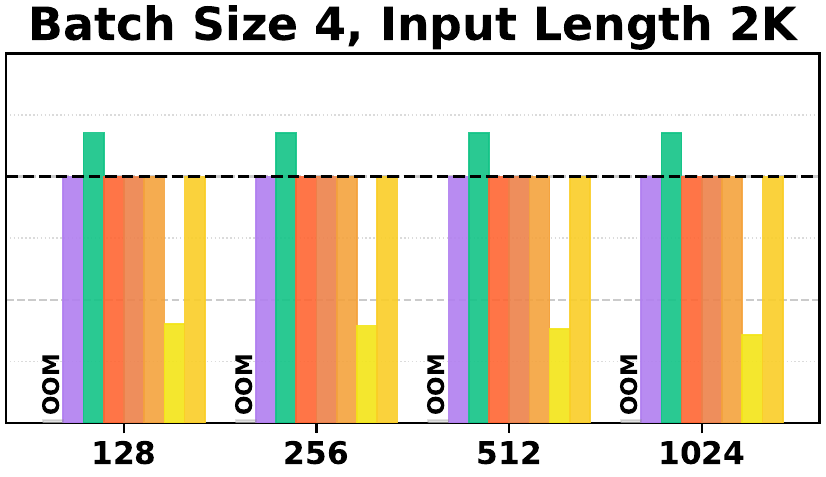}\\[-3pt]
\includegraphics[width=0.53027926\linewidth]{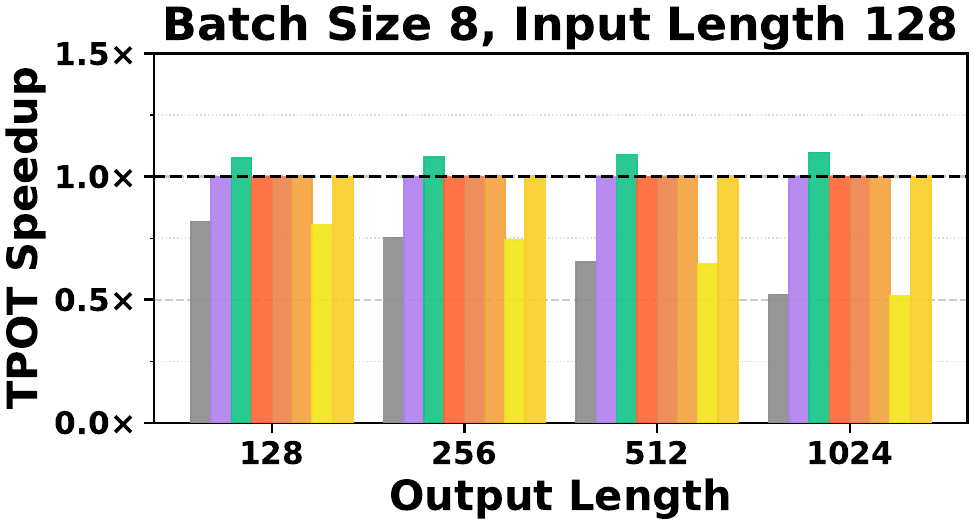}\hfill
\includegraphics[width=0.44972074\linewidth]{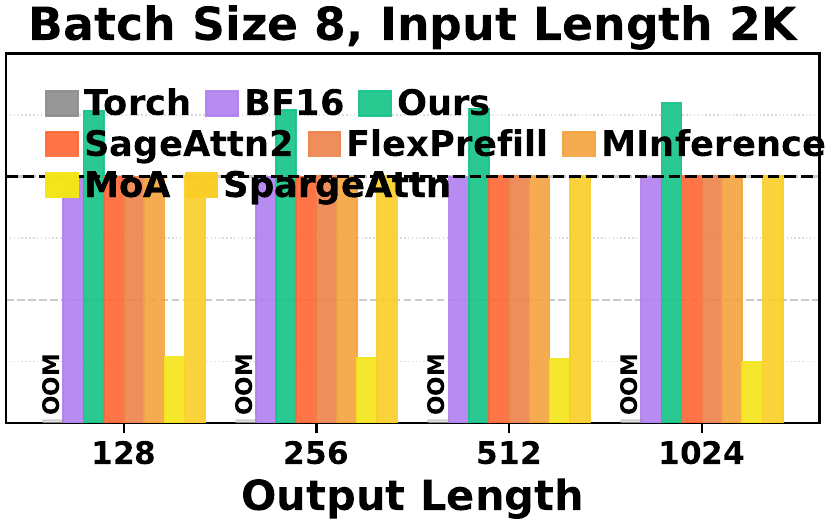}
\vspace{-8pt}
\captionof{figure}{\textbf{TPOT speedup on RTX 4090.} Ministral-3-8B-Instruct-2512-BF16. Rows use batch sizes 1, 2, 4, and 8; columns use 128, 2,048 input tokens. Each panel varies output length over 128, 256, 512, and 1024 tokens. BF16 is normalized to $1\times$; higher is better. Method colors and order are fixed across panels.}
\label{fig:system-rtx4090-ministral3_8b-tpot}
\end{minipage}
\end{minipage}
\end{figure*}

\begin{figure*}[!p]
\centering
\begin{minipage}{\textwidth}
\subsection{General and Long-Context Quality}
\label{app:extended-quality}
\begin{minipage}{\textwidth}
\centering
\captionof{table}{General-task results for Qwen2-7B-Instruct. Accuracy is reported in percent (higher is better); WikiText-2 uses word perplexity (lower is better). BF16 is the upper reference; \texttt{all\_int4} is the lower reference. Colors compare with BF16: green denotes equal or better results, blue worse results, and gray BF16. Bold marks the best result among Ours and Others; displayed ties share the mark.}
\label{tab:general-qwen}
\small
\setlength{\tabcolsep}{3pt}
\begin{adjustbox}{max width=\textwidth}
\renewcommand{\arraystretch}{1.08}
\begin{tabular*}{\textwidth}{@{\extracolsep{\fill}}llrrrrrrrr@{}}
\toprule
\multirow{2}{*}{\textbf{Benchmark}} & \multirow{2}{*}{\textbf{Metric}} & \multicolumn{1}{c}{\textbf{Upper Ref.}} & \multicolumn{1}{c}{\textbf{Lower Ref.}} & \multicolumn{1}{c}{\textbf{Ours}} & \multicolumn{5}{c}{\textbf{Others}}\\
\cmidrule(lr){3-3}\cmidrule(lr){4-4}\cmidrule(lr){5-5}\cmidrule(lr){6-10}
 & & \textbf{BF16} & \textbf{\texttt{all\_int4}} & \textbf{\modelname} & \textbf{MInference} & \textbf{FlexPrefill} & \textbf{SageAttn} & \textbf{MoA} & \textbf{SpargeAttn}\\
\midrule
WikiText-2 & PPL $\downarrow$ & \cellcolor{heatneutral}9.68 & \cellcolor{belowblue!25}9.73 & \cellcolor{abovegreen!25}9.68 & \cellcolor{abovegreen!25}\textbf{9.67} & \cellcolor{belowblue!55}10.15 & \cellcolor{abovegreen!25}9.68 & \cellcolor{belowblue!25}9.75 & \cellcolor{belowblue!25}9.70\\
HellaSwag & Acc. $\uparrow$ & \cellcolor{heatneutral}80.62 & \cellcolor{abovegreen!25}80.62 & \cellcolor{abovegreen!55}\textbf{81.12} & \cellcolor{abovegreen!25}80.63 & \cellcolor{abovegreen!25}80.65 & \cellcolor{abovegreen!25}80.65 & \cellcolor{abovegreen!25}80.65 & \cellcolor{abovegreen!25}80.62\\
MMLU & Acc. $\uparrow$ & \cellcolor{heatneutral}71.62 & \cellcolor{belowblue!25}71.16 & \cellcolor{abovegreen!55}\textbf{72.07} & \cellcolor{belowblue!25}71.45 & \cellcolor{belowblue!25}71.60 & \cellcolor{belowblue!25}71.34 & \cellcolor{abovegreen!25}71.64 & \cellcolor{belowblue!25}71.51\\
ARC-Challenge & Acc. $\uparrow$ & \cellcolor{heatneutral}62.37 & \cellcolor{belowblue!25}62.29 & \cellcolor{abovegreen!70}\textbf{62.96} & \cellcolor{abovegreen!25}62.37 & \cellcolor{abovegreen!40}62.54 & \cellcolor{abovegreen!25}62.46 & \cellcolor{abovegreen!25}62.37 & \cellcolor{belowblue!25}62.29\\
LAMBADA & Acc. $\uparrow$ & \cellcolor{heatneutral}70.04 & \cellcolor{belowblue!25}69.94 & \cellcolor{abovegreen!55}\textbf{70.44} & \cellcolor{abovegreen!25}70.06 & \cellcolor{belowblue!25}69.84 & \cellcolor{belowblue!25}69.94 & \cellcolor{abovegreen!25}70.11 & \cellcolor{abovegreen!25}70.06\\
PIQA & Acc. $\uparrow$ & \cellcolor{heatneutral}80.52 & \cellcolor{abovegreen!25}80.52 & \cellcolor{abovegreen!55}\textbf{81.02} & \cellcolor{abovegreen!25}80.52 & \cellcolor{belowblue!25}80.47 & \cellcolor{abovegreen!55}80.79 & \cellcolor{abovegreen!25}80.52 & \cellcolor{abovegreen!25}80.52\\
WinoGrande & Acc. $\uparrow$ & \cellcolor{heatneutral}76.56 & \cellcolor{belowblue!25}76.32 & \cellcolor{abovegreen!55}76.82 & \cellcolor{belowblue!25}76.24 & \cellcolor{belowblue!25}76.48 & \cellcolor{abovegreen!55}76.80 & \cellcolor{belowblue!25}76.24 & \cellcolor{abovegreen!55}\textbf{76.95}\\
\bottomrule
\end{tabular*}
\end{adjustbox}
\end{minipage}
\vspace{2pt}
\begin{minipage}{\textwidth}
\centering
\captionof{table}{LV-Eval QA results for Qwen2-7B-Instruct. Dataset identifiers follow Section~\ref{sec:model-performance}. Scores use official task-specific metrics on a 0--100 scale; higher is better. Length labels follow the official dataset levels. BF16 is the upper reference; \texttt{all\_int4} is the lower reference. Colors compare with BF16: green denotes equal or better results, blue worse results, and gray BF16. Bold marks the best result among Ours, Others, and Ablations; displayed ties share the mark.}
\label{tab:quality-qwen}
\small
\setlength{\tabcolsep}{3pt}
\begin{adjustbox}{max width=\textwidth}
\renewcommand{\arraystretch}{1.08}
\begin{tabular}{@{}ccrrrrrrrrrrr@{}}
\toprule
\multirow{2}{*}{\textbf{Dataset}} & \multirow{2}{*}{\textbf{Level}} & \multicolumn{1}{c}{\textbf{Upper Ref.}} & \multicolumn{1}{c}{\textbf{Lower Ref.}} & \multicolumn{1}{c}{\textbf{Ours}} & \multicolumn{5}{c}{\textbf{Others}} & \multicolumn{3}{c}{\textbf{Ablations}}\\
\cmidrule(lr){3-3}\cmidrule(lr){4-4}\cmidrule(lr){5-5}\cmidrule(lr){6-10}\cmidrule(lr){11-13}
 & & \textbf{BF16} & \textbf{\texttt{all\_int4}} & \textbf{\modelname} & \textbf{MInference} & \textbf{FlexPrefill} & \textbf{SageAttn} & \textbf{MoA} & \textbf{SpargeAttn} & \textbf{Layer Same} & \textbf{All Same} & \textbf{Random}\\
\midrule
\multirow{3}{*}{\ds{1}} & 16K & \cellcolor{heatneutral}23.85 & \cellcolor{belowblue!40}22.22 & \cellcolor{belowblue!25}23.49 & \cellcolor{abovegreen!70}29.00 & \cellcolor{abovegreen!70}26.13 & \cellcolor{abovegreen!70}\textbf{32.60} & \cellcolor{belowblue!70}16.32 & \cellcolor{belowblue!70}20.85 & \cellcolor{belowblue!40}22.94 & \cellcolor{belowblue!55}21.78 & \cellcolor{belowblue!40}23.12\\
 & 32K & \cellcolor{heatneutral}24.74 & \cellcolor{belowblue!70}19.99 & \cellcolor{abovegreen!70}\textbf{25.65} & \cellcolor{belowblue!70}15.00 & \cellcolor{belowblue!70}11.89 & \cellcolor{belowblue!70}16.90 & \cellcolor{belowblue!70}16.55 & \cellcolor{belowblue!70}21.19 & \cellcolor{belowblue!55}23.01 & \cellcolor{belowblue!40}23.26 & \cellcolor{belowblue!70}20.49\\
 & 64K & \cellcolor{heatneutral}27.46 & \cellcolor{belowblue!70}20.68 & \cellcolor{abovegreen!70}\textbf{29.70} & \cellcolor{belowblue!70}6.40 & \cellcolor{belowblue!70}8.98 & \cellcolor{belowblue!70}9.10 & \cellcolor{belowblue!70}19.40 & \cellcolor{belowblue!40}26.15 & \cellcolor{belowblue!70}24.05 & \cellcolor{belowblue!70}17.07 & \cellcolor{belowblue!70}21.15\\
\midrule
\multirow{3}{*}{\ds{2}} & 16K & \cellcolor{heatneutral}16.00 & \cellcolor{belowblue!55}14.07 & \cellcolor{abovegreen!70}\textbf{16.62} & \cellcolor{belowblue!55}14.10 & \cellcolor{belowblue!25}15.74 & \cellcolor{belowblue!55}14.20 & \cellcolor{belowblue!40}14.86 & \cellcolor{belowblue!25}15.87 & \cellcolor{belowblue!40}14.56 & \cellcolor{belowblue!40}14.74 & \cellcolor{belowblue!55}13.59\\
 & 32K & \cellcolor{heatneutral}16.42 & \cellcolor{belowblue!40}14.94 & \cellcolor{abovegreen!55}\textbf{16.88} & \cellcolor{belowblue!70}11.50 & \cellcolor{belowblue!40}15.24 & \cellcolor{belowblue!70}13.50 & \cellcolor{belowblue!25}16.20 & \cellcolor{belowblue!40}15.61 & \cellcolor{belowblue!25}15.87 & \cellcolor{belowblue!40}15.60 & \cellcolor{belowblue!25}16.03\\
 & 64K & \cellcolor{heatneutral}13.70 & \cellcolor{belowblue!40}12.20 & \cellcolor{abovegreen!70}\textbf{14.45} & \cellcolor{belowblue!70}9.50 & \cellcolor{belowblue!40}12.65 & \cellcolor{belowblue!70}10.80 & \cellcolor{belowblue!25}13.49 & \cellcolor{belowblue!25}13.19 & \cellcolor{abovegreen!40}13.89 & \cellcolor{belowblue!55}11.83 & \cellcolor{belowblue!25}13.07\\
\midrule
\multirow{3}{*}{\ds{3}} & 16K & \cellcolor{heatneutral}12.43 & \cellcolor{abovegreen!70}16.64 & \cellcolor{abovegreen!70}13.36 & \cellcolor{abovegreen!70}\textbf{14.30} & \cellcolor{belowblue!25}12.21 & \cellcolor{belowblue!25}12.12 & \cellcolor{belowblue!25}11.91 & \cellcolor{abovegreen!55}12.74 & \cellcolor{belowblue!25}11.89 & \cellcolor{belowblue!25}12.27 & \cellcolor{abovegreen!55}12.76\\
 & 32K & \cellcolor{heatneutral}9.33 & \cellcolor{abovegreen!70}14.04 & \cellcolor{abovegreen!70}9.94 & \cellcolor{abovegreen!70}11.46 & \cellcolor{abovegreen!70}\textbf{15.16} & \cellcolor{abovegreen!55}9.76 & \cellcolor{abovegreen!70}10.50 & \cellcolor{belowblue!25}8.73 & \cellcolor{abovegreen!70}9.85 & \cellcolor{belowblue!40}7.98 & \cellcolor{abovegreen!70}10.51\\
 & 64K & \cellcolor{heatneutral}11.82 & \cellcolor{abovegreen!70}13.21 & \cellcolor{abovegreen!55}12.18 & \cellcolor{belowblue!25}11.15 & \cellcolor{abovegreen!70}\textbf{13.14} & \cellcolor{belowblue!40}10.84 & \cellcolor{belowblue!55}9.95 & \cellcolor{belowblue!40}10.97 & \cellcolor{belowblue!70}7.69 & \cellcolor{belowblue!40}10.24 & \cellcolor{belowblue!70}7.09\\
\midrule
\multirow{3}{*}{\ds{4}} & 16K & \cellcolor{heatneutral}14.89 & \cellcolor{belowblue!25}14.82 & \cellcolor{abovegreen!70}\textbf{16.87} & \cellcolor{abovegreen!70}15.79 & \cellcolor{abovegreen!70}15.47 & \cellcolor{abovegreen!70}15.65 & \cellcolor{belowblue!40}13.75 & \cellcolor{abovegreen!70}15.67 & \cellcolor{belowblue!25}14.62 & \cellcolor{abovegreen!40}15.07 & \cellcolor{abovegreen!55}15.37\\
 & 32K & \cellcolor{heatneutral}16.04 & \cellcolor{belowblue!70}12.53 & \cellcolor{belowblue!55}14.11 & \cellcolor{belowblue!40}\textbf{14.88} & \cellcolor{belowblue!55}14.17 & \cellcolor{belowblue!55}13.66 & \cellcolor{belowblue!70}10.44 & \cellcolor{belowblue!55}13.61 & \cellcolor{belowblue!70}11.95 & \cellcolor{belowblue!70}13.30 & \cellcolor{belowblue!70}12.15\\
 & 64K & \cellcolor{heatneutral}14.01 & \cellcolor{belowblue!40}12.91 & \cellcolor{abovegreen!70}\textbf{14.69} & \cellcolor{belowblue!40}12.78 & \cellcolor{belowblue!25}13.96 & \cellcolor{belowblue!25}13.70 & \cellcolor{belowblue!55}11.62 & \cellcolor{belowblue!40}12.94 & \cellcolor{belowblue!40}12.90 & \cellcolor{belowblue!70}9.90 & \cellcolor{belowblue!70}10.95\\
\midrule
\multirow{3}{*}{\ds{5}} & 16K & \cellcolor{heatneutral}18.24 & \cellcolor{abovegreen!70}18.94 & \cellcolor{abovegreen!70}\textbf{20.31} & \cellcolor{belowblue!70}14.90 & \cellcolor{belowblue!40}17.09 & \cellcolor{belowblue!55}15.80 & \cellcolor{belowblue!40}16.74 & \cellcolor{belowblue!40}17.14 & \cellcolor{belowblue!25}18.15 & \cellcolor{belowblue!70}15.30 & \cellcolor{belowblue!70}15.12\\
 & 32K & \cellcolor{heatneutral}18.04 & \cellcolor{abovegreen!40}18.15 & \cellcolor{abovegreen!70}\textbf{18.58} & \cellcolor{belowblue!70}12.60 & \cellcolor{belowblue!55}16.08 & \cellcolor{belowblue!70}13.80 & \cellcolor{belowblue!25}17.78 & \cellcolor{abovegreen!55}18.48 & \cellcolor{belowblue!55}15.37 & \cellcolor{belowblue!40}16.46 & \cellcolor{belowblue!40}16.37\\
 & 64K & \cellcolor{heatneutral}17.57 & \cellcolor{belowblue!40}16.01 & \cellcolor{abovegreen!40}\textbf{17.70} & \cellcolor{belowblue!70}9.00 & \cellcolor{belowblue!70}10.29 & \cellcolor{belowblue!70}11.40 & \cellcolor{belowblue!70}14.49 & \cellcolor{belowblue!70}14.75 & \cellcolor{belowblue!40}16.60 & \cellcolor{belowblue!70}12.88 & \cellcolor{belowblue!40}16.31\\
\midrule
\multirow{3}{*}{\ds{6}} & 16K & \cellcolor{heatneutral}17.33 & \cellcolor{belowblue!55}15.02 & \cellcolor{belowblue!25}17.27 & \cellcolor{belowblue!55}15.20 & \cellcolor{belowblue!70}13.65 & \cellcolor{belowblue!55}15.10 & \cellcolor{belowblue!70}14.08 & \cellcolor{belowblue!40}16.06 & \cellcolor{belowblue!40}15.93 & \cellcolor{belowblue!55}15.35 & \cellcolor{abovegreen!40}\textbf{17.47}\\
 & 32K & \cellcolor{heatneutral}15.41 & \cellcolor{belowblue!25}14.88 & \cellcolor{abovegreen!55}\textbf{15.91} & \cellcolor{belowblue!70}10.80 & \cellcolor{belowblue!70}12.26 & \cellcolor{belowblue!70}10.80 & \cellcolor{belowblue!40}13.99 & \cellcolor{abovegreen!25}15.44 & \cellcolor{belowblue!40}14.19 & \cellcolor{belowblue!70}11.76 & \cellcolor{abovegreen!40}15.55\\
 & 64K & \cellcolor{heatneutral}14.67 & \cellcolor{abovegreen!40}14.79 & \cellcolor{belowblue!25}\textbf{14.31} & \cellcolor{belowblue!70}8.10 & \cellcolor{belowblue!70}9.46 & \cellcolor{belowblue!70}9.50 & \cellcolor{belowblue!70}10.83 & \cellcolor{belowblue!40}13.37 & \cellcolor{belowblue!70}11.87 & \cellcolor{belowblue!70}8.73 & \cellcolor{belowblue!70}10.91\\
\midrule
\multirow{3}{*}{\ds{7}} & 16K & \cellcolor{heatneutral}38.49 & \cellcolor{belowblue!70}34.96 & \cellcolor{abovegreen!40}\textbf{38.65} & \cellcolor{belowblue!70}28.50 & \cellcolor{belowblue!70}29.07 & \cellcolor{belowblue!70}31.40 & \cellcolor{belowblue!70}30.63 & \cellcolor{belowblue!40}37.38 & \cellcolor{belowblue!25}37.94 & \cellcolor{belowblue!40}37.15 & \cellcolor{belowblue!40}37.40\\
 & 32K & \cellcolor{heatneutral}39.45 & \cellcolor{belowblue!70}33.05 & \cellcolor{belowblue!25}\textbf{39.03} & \cellcolor{belowblue!70}18.10 & \cellcolor{belowblue!70}20.71 & \cellcolor{belowblue!70}21.00 & \cellcolor{belowblue!70}29.02 & \cellcolor{belowblue!40}38.05 & \cellcolor{belowblue!70}36.28 & \cellcolor{belowblue!70}29.49 & \cellcolor{belowblue!70}34.35\\
 & 64K & \cellcolor{heatneutral}34.89 & \cellcolor{belowblue!70}32.02 & \cellcolor{belowblue!55}32.77 & \cellcolor{belowblue!70}11.40 & \cellcolor{belowblue!70}14.53 & \cellcolor{belowblue!70}13.70 & \cellcolor{belowblue!70}21.04 & \cellcolor{belowblue!55}33.13 & \cellcolor{belowblue!55}\textbf{33.17} & \cellcolor{belowblue!70}15.32 & \cellcolor{belowblue!70}24.09\\
\midrule
\multirow{3}{*}{\ds{8}} & 16K & \cellcolor{heatneutral}30.14 & \cellcolor{belowblue!40}29.08 & \cellcolor{abovegreen!70}\textbf{31.34} & \cellcolor{belowblue!70}21.20 & \cellcolor{belowblue!70}23.89 & \cellcolor{belowblue!70}22.60 & \cellcolor{belowblue!70}22.24 & \cellcolor{belowblue!25}29.65 & \cellcolor{belowblue!40}28.59 & \cellcolor{belowblue!40}28.73 & \cellcolor{belowblue!70}26.49\\
 & 32K & \cellcolor{heatneutral}25.83 & \cellcolor{belowblue!40}24.15 & \cellcolor{abovegreen!55}\textbf{26.31} & \cellcolor{belowblue!70}15.30 & \cellcolor{belowblue!70}17.67 & \cellcolor{belowblue!70}17.90 & \cellcolor{belowblue!70}22.73 & \cellcolor{belowblue!70}22.92 & \cellcolor{belowblue!40}24.62 & \cellcolor{belowblue!70}19.14 & \cellcolor{belowblue!70}23.06\\
 & 64K & \cellcolor{heatneutral}25.55 & \cellcolor{belowblue!40}24.17 & \cellcolor{abovegreen!70}\textbf{27.66} & \cellcolor{belowblue!70}12.80 & \cellcolor{belowblue!70}15.95 & \cellcolor{belowblue!70}15.90 & \cellcolor{belowblue!70}21.10 & \cellcolor{belowblue!40}24.37 & \cellcolor{abovegreen!70}27.40 & \cellcolor{belowblue!70}19.58 & \cellcolor{belowblue!55}23.33\\
\midrule
\multirow{3}{*}{\ds{9}} & 16K & \cellcolor{heatneutral}35.42 & \cellcolor{abovegreen!55}35.86 & \cellcolor{abovegreen!70}\textbf{36.81} & \cellcolor{belowblue!70}28.70 & \cellcolor{belowblue!70}22.74 & \cellcolor{belowblue!70}31.40 & \cellcolor{belowblue!70}28.45 & \cellcolor{belowblue!70}32.26 & \cellcolor{belowblue!40}34.41 & \cellcolor{belowblue!70}32.46 & \cellcolor{belowblue!70}32.15\\
 & 32K & \cellcolor{heatneutral}29.52 & \cellcolor{belowblue!55}27.28 & \cellcolor{belowblue!25}28.94 & \cellcolor{belowblue!70}19.30 & \cellcolor{belowblue!70}15.98 & \cellcolor{belowblue!70}23.50 & \cellcolor{belowblue!70}20.48 & \cellcolor{belowblue!25}\textbf{28.98} & \cellcolor{belowblue!70}25.29 & \cellcolor{belowblue!55}27.30 & \cellcolor{belowblue!70}26.35\\
 & 64K & \cellcolor{heatneutral}27.05 & \cellcolor{belowblue!70}22.55 & \cellcolor{belowblue!40}25.38 & \cellcolor{belowblue!70}13.00 & \cellcolor{belowblue!70}15.22 & \cellcolor{belowblue!70}16.10 & \cellcolor{belowblue!70}23.79 & \cellcolor{belowblue!40}\textbf{26.18} & \cellcolor{belowblue!70}23.74 & \cellcolor{belowblue!70}21.51 & \cellcolor{belowblue!70}23.52\\
\bottomrule
\end{tabular}
\end{adjustbox}
\end{minipage}
\setlength{\multicolsep}{5pt}
\begin{multicols}{2}
\textbf{General prediction across models.}
Each additional model reuses its calibrated boundary table across tasks. Qwen2-7B-Instruct and Ministral-3-8B-Instruct-2512-BF16 match BF16 word perplexity at the reported precision, at 9.68 and 9.73, respectively, and all six accuracy gaps are below 0.6 percentage points (Tables~\ref{tab:general-qwen} and~\ref{tab:general-ministral}). Near-reference predictions support model-level allocation across language modeling, reasoning, and understanding tasks.

\textbf{Long-context answer generation.}
LV-Eval additionally exercises the compressed prompt cache throughout generation. At 64K, Qwen2 scores 29.70 on CMRC, compared with 27.46 for BF16 and 26.15 for SpargeAttn; Ministral scores 20.71 on LooGLE-MIR, compared with 19.03 and 18.89, respectively (Tables~\ref{tab:quality-qwen} and~\ref{tab:quality-ministral}). The same model-specific boundaries therefore preserve useful context for generated answers as well as likelihood-based prediction.
\end{multicols}
\end{minipage}
\end{figure*}

\begin{figure*}[!p]
\centering
\begin{minipage}{\textwidth}

\begin{minipage}{\textwidth}
\centering
\captionof{table}{General-task results for Ministral-3-8B-Instruct-2512-BF16. Accuracy is reported in percent (higher is better); WikiText-2 uses word perplexity (lower is better). BF16 is the upper reference; \texttt{all\_int4} is the lower reference. Colors compare with BF16: green denotes equal or better results, blue worse results, and gray BF16. Bold marks the best of Ours and Others, including displayed ties.}
\label{tab:general-ministral}
\small
\setlength{\tabcolsep}{3pt}
\begin{adjustbox}{max width=\textwidth}
\renewcommand{\arraystretch}{0.90}
\begin{tabular*}{\textwidth}{@{\extracolsep{\fill}}llrrrrrrrr@{}}
\toprule
\multirow{2}{*}{\textbf{Benchmark}} & \multirow{2}{*}{\textbf{Metric}} & \multicolumn{1}{c}{\textbf{Upper Ref.}} & \multicolumn{1}{c}{\textbf{Lower Ref.}} & \multicolumn{1}{c}{\textbf{Ours}} & \multicolumn{5}{c}{\textbf{Others}}\\
\cmidrule(lr){3-3}\cmidrule(lr){4-4}\cmidrule(lr){5-5}\cmidrule(lr){6-10}
 & & \textbf{BF16} & \textbf{\texttt{all\_int4}} & \textbf{\modelname} & \textbf{MInference} & \textbf{FlexPrefill} & \textbf{SageAttn} & \textbf{MoA} & \textbf{SpargeAttn}\\
\midrule
WikiText-2 & PPL $\downarrow$ & \cellcolor{heatneutral}9.73 & \cellcolor{belowblue!25}9.75 & \cellcolor{abovegreen!25}9.73 & \cellcolor{abovegreen!25}\textbf{9.72} & \cellcolor{belowblue!40}9.84 & \cellcolor{abovegreen!25}9.73 & \cellcolor{belowblue!40}9.89 & \cellcolor{belowblue!25}9.75\\
HellaSwag & Acc. $\uparrow$ & \cellcolor{heatneutral}78.79 & \cellcolor{abovegreen!25}78.80 & \cellcolor{abovegreen!70}\textbf{79.30} & \cellcolor{abovegreen!25}78.84 & \cellcolor{abovegreen!25}78.85 & \cellcolor{abovegreen!40}78.95 & \cellcolor{abovegreen!25}78.87 & \cellcolor{abovegreen!25}78.85\\
MMLU & Acc. $\uparrow$ & \cellcolor{heatneutral}77.85 & \cellcolor{belowblue!25}77.68 & \cellcolor{abovegreen!55}\textbf{78.35} & \cellcolor{belowblue!25}77.82 & \cellcolor{abovegreen!25}77.92 & \cellcolor{abovegreen!25}77.94 & \cellcolor{belowblue!25}77.65 & \cellcolor{belowblue!25}77.60\\
ARC-Challenge & Acc. $\uparrow$ & \cellcolor{heatneutral}69.03 & \cellcolor{abovegreen!25}69.11 & \cellcolor{abovegreen!70}\textbf{69.61} & \cellcolor{abovegreen!40}69.20 & \cellcolor{abovegreen!25}69.03 & \cellcolor{abovegreen!25}69.03 & \cellcolor{belowblue!25}68.94 & \cellcolor{belowblue!25}68.94\\
LAMBADA & Acc. $\uparrow$ & \cellcolor{heatneutral}70.99 & \cellcolor{abovegreen!25}71.07 & \cellcolor{abovegreen!70}\textbf{71.53} & \cellcolor{abovegreen!40}71.12 & \cellcolor{abovegreen!25}71.05 & \cellcolor{abovegreen!55}71.22 & \cellcolor{abovegreen!55}71.43 & \cellcolor{abovegreen!40}71.12\\
PIQA & Acc. $\uparrow$ & \cellcolor{heatneutral}80.52 & \cellcolor{belowblue!25}80.36 & \cellcolor{abovegreen!55}\textbf{80.86} & \cellcolor{belowblue!25}80.30 & \cellcolor{belowblue!25}80.30 & \cellcolor{abovegreen!25}80.58 & \cellcolor{belowblue!25}80.36 & \cellcolor{belowblue!25}80.30\\
WinoGrande & Acc. $\uparrow$ & \cellcolor{heatneutral}77.98 & \cellcolor{belowblue!40}77.11 & \cellcolor{abovegreen!55}\textbf{78.40} & \cellcolor{belowblue!25}77.51 & \cellcolor{belowblue!40}77.19 & \cellcolor{belowblue!25}77.35 & \cellcolor{belowblue!25}77.58 & \cellcolor{belowblue!25}77.51\\
\bottomrule
\end{tabular*}
\end{adjustbox}
\end{minipage}
\vspace{2pt}
\begin{minipage}{\textwidth}
\centering
\captionof{table}{LV-Eval QA results for Ministral-3-8B-Instruct-2512-BF16. Dataset identifiers follow Section~\ref{sec:model-performance}. Scores use official task-specific metrics on a 0--100 scale; higher is better. Length labels follow the official dataset levels. BF16 is the upper reference; \texttt{all\_int4} is the lower reference. Colors compare with BF16: green denotes equal or better results, blue worse results, and gray BF16. Bold marks the best result among Ours, Others, and Ablations; displayed ties share the mark.}
\label{tab:quality-ministral}
\small
\setlength{\tabcolsep}{3pt}
\begin{adjustbox}{max width=\textwidth}
\renewcommand{\arraystretch}{0.90}
\begin{tabular}{@{}ccrrrrrrrrrrr@{}}
\toprule
\multirow{2}{*}{\textbf{Dataset}} & \multirow{2}{*}{\textbf{Level}} & \multicolumn{1}{c}{\textbf{Upper Ref.}} & \multicolumn{1}{c}{\textbf{Lower Ref.}} & \multicolumn{1}{c}{\textbf{Ours}} & \multicolumn{5}{c}{\textbf{Others}} & \multicolumn{3}{c}{\textbf{Ablations}}\\
\cmidrule(lr){3-3}\cmidrule(lr){4-4}\cmidrule(lr){5-5}\cmidrule(lr){6-10}\cmidrule(lr){11-13}
 & & \textbf{BF16} & \textbf{\texttt{all\_int4}} & \textbf{\modelname} & \textbf{MInference} & \textbf{FlexPrefill} & \textbf{SageAttn} & \textbf{MoA} & \textbf{SpargeAttn} & \textbf{Layer Same} & \textbf{All Same} & \textbf{Random}\\
\midrule
\multirow{3}{*}{\ds{1}} & 16K & \cellcolor{heatneutral}39.51 & \cellcolor{belowblue!25}39.14 & \cellcolor{abovegreen!70}\textbf{40.52} & \cellcolor{belowblue!25}39.30 & \cellcolor{belowblue!40}38.44 & \cellcolor{belowblue!40}38.22 & \cellcolor{belowblue!70}21.53 & \cellcolor{abovegreen!70}40.02 & \cellcolor{belowblue!25}39.03 & \cellcolor{belowblue!40}38.01 & \cellcolor{belowblue!70}31.35\\
 & 32K & \cellcolor{heatneutral}41.20 & \cellcolor{belowblue!55}39.42 & \cellcolor{belowblue!40}39.77 & \cellcolor{abovegreen!70}\textbf{42.35} & \cellcolor{belowblue!40}40.29 & \cellcolor{belowblue!55}39.48 & \cellcolor{belowblue!70}23.61 & \cellcolor{belowblue!55}39.23 & \cellcolor{belowblue!55}38.87 & \cellcolor{belowblue!70}35.72 & \cellcolor{belowblue!70}27.81\\
 & 64K & \cellcolor{heatneutral}40.11 & \cellcolor{belowblue!70}36.26 & \cellcolor{belowblue!25}39.53 & \cellcolor{abovegreen!70}\textbf{42.02} & \cellcolor{belowblue!70}36.72 & \cellcolor{belowblue!25}39.79 & \cellcolor{belowblue!70}22.37 & \cellcolor{belowblue!40}39.37 & \cellcolor{belowblue!25}39.68 & \cellcolor{belowblue!70}17.46 & \cellcolor{belowblue!70}20.14\\
\midrule
\multirow{3}{*}{\ds{2}} & 16K & \cellcolor{heatneutral}13.68 & \cellcolor{abovegreen!55}13.90 & \cellcolor{abovegreen!55}\textbf{14.11} & \cellcolor{belowblue!25}13.51 & \cellcolor{belowblue!40}12.78 & \cellcolor{belowblue!40}12.64 & \cellcolor{belowblue!25}13.23 & \cellcolor{belowblue!40}12.91 & \cellcolor{belowblue!25}13.05 & \cellcolor{belowblue!25}13.37 & \cellcolor{belowblue!25}13.04\\
 & 32K & \cellcolor{heatneutral}15.26 & \cellcolor{belowblue!40}13.93 & \cellcolor{abovegreen!70}\textbf{15.85} & \cellcolor{belowblue!25}14.84 & \cellcolor{abovegreen!55}15.62 & \cellcolor{belowblue!40}14.51 & \cellcolor{belowblue!40}13.84 & \cellcolor{belowblue!55}12.68 & \cellcolor{belowblue!40}14.42 & \cellcolor{belowblue!25}15.03 & \cellcolor{belowblue!55}12.82\\
 & 64K & \cellcolor{heatneutral}17.46 & \cellcolor{belowblue!55}15.42 & \cellcolor{belowblue!25}\textbf{16.91} & \cellcolor{belowblue!55}14.86 & \cellcolor{belowblue!40}16.44 & \cellcolor{belowblue!55}15.68 & \cellcolor{belowblue!55}14.78 & \cellcolor{belowblue!40}15.87 & \cellcolor{belowblue!55}15.57 & \cellcolor{belowblue!70}14.24 & \cellcolor{belowblue!70}14.22\\
\midrule
\multirow{3}{*}{\ds{3}} & 16K & \cellcolor{heatneutral}31.15 & \cellcolor{belowblue!40}29.48 & \cellcolor{abovegreen!70}\textbf{32.04} & \cellcolor{belowblue!25}30.48 & \cellcolor{belowblue!70}26.82 & \cellcolor{belowblue!25}30.69 & \cellcolor{abovegreen!55}31.40 & \cellcolor{belowblue!40}29.95 & \cellcolor{abovegreen!55}31.46 & \cellcolor{abovegreen!55}31.36 & \cellcolor{belowblue!40}29.52\\
 & 32K & \cellcolor{heatneutral}32.58 & \cellcolor{abovegreen!70}33.21 & \cellcolor{abovegreen!70}\textbf{34.34} & \cellcolor{belowblue!55}29.93 & \cellcolor{belowblue!70}25.34 & \cellcolor{belowblue!55}30.70 & \cellcolor{belowblue!70}26.33 & \cellcolor{belowblue!25}31.89 & \cellcolor{belowblue!40}31.40 & \cellcolor{belowblue!70}25.52 & \cellcolor{belowblue!70}29.03\\
 & 64K & \cellcolor{heatneutral}25.90 & \cellcolor{abovegreen!70}27.63 & \cellcolor{abovegreen!70}\textbf{27.81} & \cellcolor{belowblue!70}22.70 & \cellcolor{belowblue!55}23.26 & \cellcolor{abovegreen!55}26.11 & \cellcolor{belowblue!70}22.70 & \cellcolor{belowblue!25}25.35 & \cellcolor{belowblue!70}23.02 & \cellcolor{belowblue!70}15.46 & \cellcolor{belowblue!70}21.49\\
\midrule
\multirow{3}{*}{\ds{4}} & 16K & \cellcolor{heatneutral}21.65 & \cellcolor{belowblue!40}20.70 & \cellcolor{belowblue!25}\textbf{21.22} & \cellcolor{belowblue!55}19.63 & \cellcolor{belowblue!70}18.61 & \cellcolor{belowblue!55}19.93 & \cellcolor{belowblue!70}14.49 & \cellcolor{belowblue!40}20.14 & \cellcolor{belowblue!40}20.47 & \cellcolor{belowblue!40}20.37 & \cellcolor{belowblue!70}15.62\\
 & 32K & \cellcolor{heatneutral}21.21 & \cellcolor{belowblue!40}20.34 & \cellcolor{belowblue!25}\textbf{21.13} & \cellcolor{belowblue!40}19.98 & \cellcolor{belowblue!40}20.11 & \cellcolor{belowblue!25}20.52 & \cellcolor{belowblue!70}14.63 & \cellcolor{belowblue!40}20.09 & \cellcolor{belowblue!25}20.72 & \cellcolor{belowblue!70}17.38 & \cellcolor{belowblue!70}15.51\\
 & 64K & \cellcolor{heatneutral}20.77 & \cellcolor{belowblue!25}20.39 & \cellcolor{belowblue!40}20.04 & \cellcolor{belowblue!25}\textbf{20.27} & \cellcolor{belowblue!40}19.43 & \cellcolor{belowblue!40}19.58 & \cellcolor{belowblue!70}15.00 & \cellcolor{belowblue!40}19.46 & \cellcolor{belowblue!40}19.20 & \cellcolor{belowblue!70}10.37 & \cellcolor{belowblue!70}12.50\\
\midrule
\multirow{3}{*}{\ds{5}} & 16K & \cellcolor{heatneutral}27.62 & \cellcolor{abovegreen!70}28.52 & \cellcolor{abovegreen!70}\textbf{30.46} & \cellcolor{belowblue!40}26.91 & \cellcolor{belowblue!25}27.36 & \cellcolor{belowblue!40}26.79 & \cellcolor{abovegreen!70}28.87 & \cellcolor{abovegreen!70}28.54 & \cellcolor{abovegreen!70}28.29 & \cellcolor{abovegreen!70}28.56 & \cellcolor{belowblue!25}27.51\\
 & 32K & \cellcolor{heatneutral}26.46 & \cellcolor{abovegreen!70}26.98 & \cellcolor{abovegreen!70}\textbf{29.22} & \cellcolor{belowblue!25}25.97 & \cellcolor{belowblue!40}25.57 & \cellcolor{belowblue!40}25.35 & \cellcolor{belowblue!40}25.14 & \cellcolor{abovegreen!70}27.81 & \cellcolor{abovegreen!25}26.55 & \cellcolor{belowblue!70}22.67 & \cellcolor{belowblue!55}23.85\\
 & 64K & \cellcolor{heatneutral}26.93 & \cellcolor{belowblue!40}26.03 & \cellcolor{abovegreen!55}\textbf{27.34} & \cellcolor{belowblue!25}26.23 & \cellcolor{belowblue!40}25.71 & \cellcolor{belowblue!40}25.99 & \cellcolor{belowblue!70}20.80 & \cellcolor{belowblue!40}25.72 & \cellcolor{abovegreen!40}27.06 & \cellcolor{belowblue!70}15.79 & \cellcolor{belowblue!70}19.82\\
\midrule
\multirow{3}{*}{\ds{6}} & 16K & \cellcolor{heatneutral}21.07 & \cellcolor{belowblue!40}20.04 & \cellcolor{abovegreen!70}\textbf{21.64} & \cellcolor{belowblue!25}20.75 & \cellcolor{abovegreen!40}21.25 & \cellcolor{belowblue!40}19.62 & \cellcolor{belowblue!70}17.79 & \cellcolor{belowblue!40}19.48 & \cellcolor{belowblue!40}19.51 & \cellcolor{belowblue!55}18.63 & \cellcolor{belowblue!70}17.51\\
 & 32K & \cellcolor{heatneutral}19.49 & \cellcolor{abovegreen!55}19.76 & \cellcolor{abovegreen!70}\textbf{20.02} & \cellcolor{belowblue!25}19.23 & \cellcolor{abovegreen!55}19.96 & \cellcolor{belowblue!25}19.08 & \cellcolor{belowblue!70}16.32 & \cellcolor{belowblue!25}19.08 & \cellcolor{belowblue!40}18.42 & \cellcolor{belowblue!70}13.92 & \cellcolor{belowblue!70}15.93\\
 & 64K & \cellcolor{heatneutral}19.03 & \cellcolor{abovegreen!55}19.35 & \cellcolor{abovegreen!70}\textbf{20.71} & \cellcolor{belowblue!40}18.18 & \cellcolor{belowblue!40}17.87 & \cellcolor{belowblue!40}17.99 & \cellcolor{belowblue!70}13.44 & \cellcolor{belowblue!25}18.89 & \cellcolor{abovegreen!55}19.29 & \cellcolor{belowblue!70}8.27 & \cellcolor{belowblue!70}11.44\\
\midrule
\multirow{3}{*}{\ds{7}} & 16K & \cellcolor{heatneutral}57.89 & \cellcolor{belowblue!40}56.53 & \cellcolor{abovegreen!70}\textbf{60.26} & \cellcolor{abovegreen!25}57.92 & \cellcolor{belowblue!25}57.36 & \cellcolor{belowblue!55}55.76 & \cellcolor{belowblue!70}45.50 & \cellcolor{belowblue!40}57.08 & \cellcolor{abovegreen!70}58.90 & \cellcolor{belowblue!40}56.96 & \cellcolor{belowblue!70}53.15\\
 & 32K & \cellcolor{heatneutral}57.31 & \cellcolor{belowblue!55}55.04 & \cellcolor{abovegreen!55}\textbf{57.56} & \cellcolor{belowblue!40}55.92 & \cellcolor{belowblue!70}53.84 & \cellcolor{belowblue!40}56.54 & \cellcolor{belowblue!70}42.82 & \cellcolor{belowblue!40}56.53 & \cellcolor{belowblue!40}56.36 & \cellcolor{belowblue!70}44.55 & \cellcolor{belowblue!70}39.54\\
 & 64K & \cellcolor{heatneutral}55.03 & \cellcolor{belowblue!55}53.15 & \cellcolor{abovegreen!55}55.33 & \cellcolor{abovegreen!70}\textbf{56.59} & \cellcolor{belowblue!55}53.32 & \cellcolor{belowblue!25}54.98 & \cellcolor{belowblue!70}33.74 & \cellcolor{belowblue!40}54.00 & \cellcolor{abovegreen!70}55.75 & \cellcolor{belowblue!70}22.05 & \cellcolor{belowblue!70}24.72\\
\midrule
\multirow{3}{*}{\ds{8}} & 16K & \cellcolor{heatneutral}33.14 & \cellcolor{belowblue!70}30.05 & \cellcolor{abovegreen!70}\textbf{34.92} & \cellcolor{belowblue!55}31.41 & \cellcolor{belowblue!70}30.18 & \cellcolor{abovegreen!70}33.77 & \cellcolor{belowblue!70}23.81 & \cellcolor{belowblue!55}31.39 & \cellcolor{belowblue!55}31.04 & \cellcolor{belowblue!55}31.00 & \cellcolor{belowblue!70}26.39\\
 & 32K & \cellcolor{heatneutral}31.90 & \cellcolor{belowblue!55}30.06 & \cellcolor{belowblue!40}30.50 & \cellcolor{belowblue!70}27.60 & \cellcolor{belowblue!40}\textbf{31.12} & \cellcolor{belowblue!40}30.28 & \cellcolor{belowblue!70}25.88 & \cellcolor{belowblue!55}30.12 & \cellcolor{belowblue!55}29.37 & \cellcolor{belowblue!70}25.18 & \cellcolor{belowblue!70}25.17\\
 & 64K & \cellcolor{heatneutral}31.27 & \cellcolor{belowblue!55}29.05 & \cellcolor{abovegreen!70}\textbf{33.82} & \cellcolor{abovegreen!40}31.39 & \cellcolor{belowblue!55}28.99 & \cellcolor{belowblue!40}30.11 & \cellcolor{belowblue!70}25.42 & \cellcolor{belowblue!55}29.32 & \cellcolor{abovegreen!70}33.81 & \cellcolor{belowblue!70}19.64 & \cellcolor{belowblue!70}21.01\\
\midrule
\multirow{3}{*}{\ds{9}} & 16K & \cellcolor{heatneutral}41.97 & \cellcolor{abovegreen!70}44.18 & \cellcolor{abovegreen!70}43.95 & \cellcolor{belowblue!25}41.93 & \cellcolor{belowblue!40}40.49 & \cellcolor{belowblue!25}41.46 & \cellcolor{belowblue!70}28.79 & \cellcolor{abovegreen!70}\textbf{44.56} & \cellcolor{abovegreen!70}43.49 & \cellcolor{belowblue!25}41.87 & \cellcolor{belowblue!25}41.83\\
 & 32K & \cellcolor{heatneutral}41.76 & \cellcolor{abovegreen!70}43.10 & \cellcolor{abovegreen!70}\textbf{42.72} & \cellcolor{belowblue!55}39.41 & \cellcolor{belowblue!70}36.03 & \cellcolor{belowblue!40}40.18 & \cellcolor{belowblue!70}26.82 & \cellcolor{belowblue!40}40.91 & \cellcolor{belowblue!55}39.17 & \cellcolor{belowblue!55}39.25 & \cellcolor{belowblue!70}34.34\\
 & 64K & \cellcolor{heatneutral}44.43 & \cellcolor{belowblue!25}43.89 & \cellcolor{abovegreen!70}45.80 & \cellcolor{abovegreen!70}\textbf{45.88} & \cellcolor{belowblue!70}38.34 & \cellcolor{belowblue!70}40.69 & \cellcolor{belowblue!70}28.68 & \cellcolor{belowblue!70}40.62 & \cellcolor{abovegreen!25}44.43 & \cellcolor{belowblue!70}26.53 & \cellcolor{belowblue!70}22.25\\
\bottomrule
\end{tabular}
\end{adjustbox}
\end{minipage}
\setlength{\multicolsep}{5pt}
\begin{multicols}{2}
\textbf{Sensitivity-guided placement.}
Random preserves the selected boundary pairs and their aggregate modeled cost but changes which heads receive them. On LooGLE-SD at 64K, Qwen2 scores 32.77 with \modelname{} and 24.09 with Random, while Ministral scores 55.33 and 24.72. All Same, which applies one 32K local configuration throughout the model, scores 15.32 and 22.05. The randomized control establishes the value of sensitivity-guided placement at fixed modeled cost, while the shared-range control examines restricting all heads to one context range.

The separation from Random widens at longer dataset levels. On LooGLE-SD, the gap increases from 1.25 points at 16K to 8.68 at 64K for Qwen2, and from 7.11 to 30.61 points for Ministral. Because Random preserves the boundary-pair collection and aggregate modeled cost, this trend identifies configuration placement as increasingly consequential in these longer-context evaluations. The same amount of modeled computation supports stronger QA quality when its retained ranges and precisions follow head sensitivity.

\findingbox{%
At the same modeled computation cost, sensitivity-guided boundary placement preserves long-context QA quality better than random placement.
}
\end{multicols}
\end{minipage}
\end{figure*}

\end{document}